\documentclass{article}

\usepackage{microtype}
\usepackage{graphicx}
\usepackage{caption}
\usepackage{subcaption}
\usepackage{booktabs} 
\usepackage{enumitem}
\usepackage{xcolor}

\usepackage{hyperref}

\usepackage[accepted]{icml2025}

\usepackage{amsmath}
\usepackage{amssymb}
\usepackage{mathtools}
\usepackage{amsthm}

\usepackage[capitalize,noabbrev]{cleveref}

\theoremstyle{plain}

\theoremstyle{definition}

\theoremstyle{remark}

\usepackage[textsize=tiny]{todonotes}
\usepackage{bm}
\usepackage{nicefrac}
\usepackage{amssymb}

\let\emptyset\varnothing

\newcounter{daggerfootnote}

\icmltitlerunning{When Can You Get Away with Low Memory Adam?}

\begin{document}

\twocolumn[
\icmltitle{When Can You Get Away with Low Memory Adam?}

\icmlsetsymbol{equal}{*}

\begin{icmlauthorlist}
\icmlauthor{Dayal Singh Kalra}{cs}
\icmlauthor{John Kirchenbauer}{cs}
\icmlauthor{Maissam Barkeshli}{phys,jqi}
\icmlauthor{Tom Goldstein}{cs}

\end{icmlauthorlist}


\icmlaffiliation{cs}{Department of Computer Science, University of Maryland, College Park}
\icmlaffiliation{phys}{Department of Physics, University of Maryland, College Park}
\icmlaffiliation{jqi}{Joint Quantum Institute, University of Maryland, College Park}
\icmlcorrespondingauthor{Dayal Kalra}{dayal@umd.edu}

\vskip 0.3in
]


\printAffiliationsAndNotice{}  

\begin{abstract}
    Adam is the go-to optimizer for training modern machine learning models, but it requires additional memory to maintain the moving averages of the gradients and their squares. 
    While various low-memory optimizers have been proposed that sometimes match the performance of Adam, their lack of reliability has left Adam as the default choice.
    In this work, we apply a simple layer-wise Signal-to-Noise Ratio (SNR)
    analysis to quantify when second-moment tensors can be effectively replaced by their means across different dimensions. Our SNR analysis reveals how architecture, training hyperparameters, and dataset properties impact compressibility along Adam's trajectory, naturally leading to \emph{SlimAdam}, a memory-efficient Adam variant. \emph{SlimAdam} compresses the second moments along dimensions with high SNR when feasible, and leaves when compression would be detrimental. Through experiments across a diverse set of architectures and training scenarios, we show that \emph{SlimAdam} matches Adam's performance and stability while saving up to $98\%$ of total second moments. Code for \emph{SlimAdam} is available at \url{https://github.com/dayal-kalra/low-memory-adam}.
    \vspace{-0.1 in}

\end{abstract}

\section{Introduction}
\label{section:introduction}

Adam with weight decay \cite{adamwloshchilov2018} has become the standard optimizer choice in modern machine learning, consistently outperforming non-adaptive optimizers such as Stochastic Gradient Descent with momentum (SGD-M). Its success is typically attributed to adapting to the geometry of the landscape by estimating the ``effective learning rate'' for each parameter using a moving average of the squared gradients. An additional benefit of this adaptive mechanism is that the optimal learning rate is less sensitive to changes in the training recipe.

\begin{figure}[!t]
    \centering
    \includegraphics[width=0.7\linewidth]{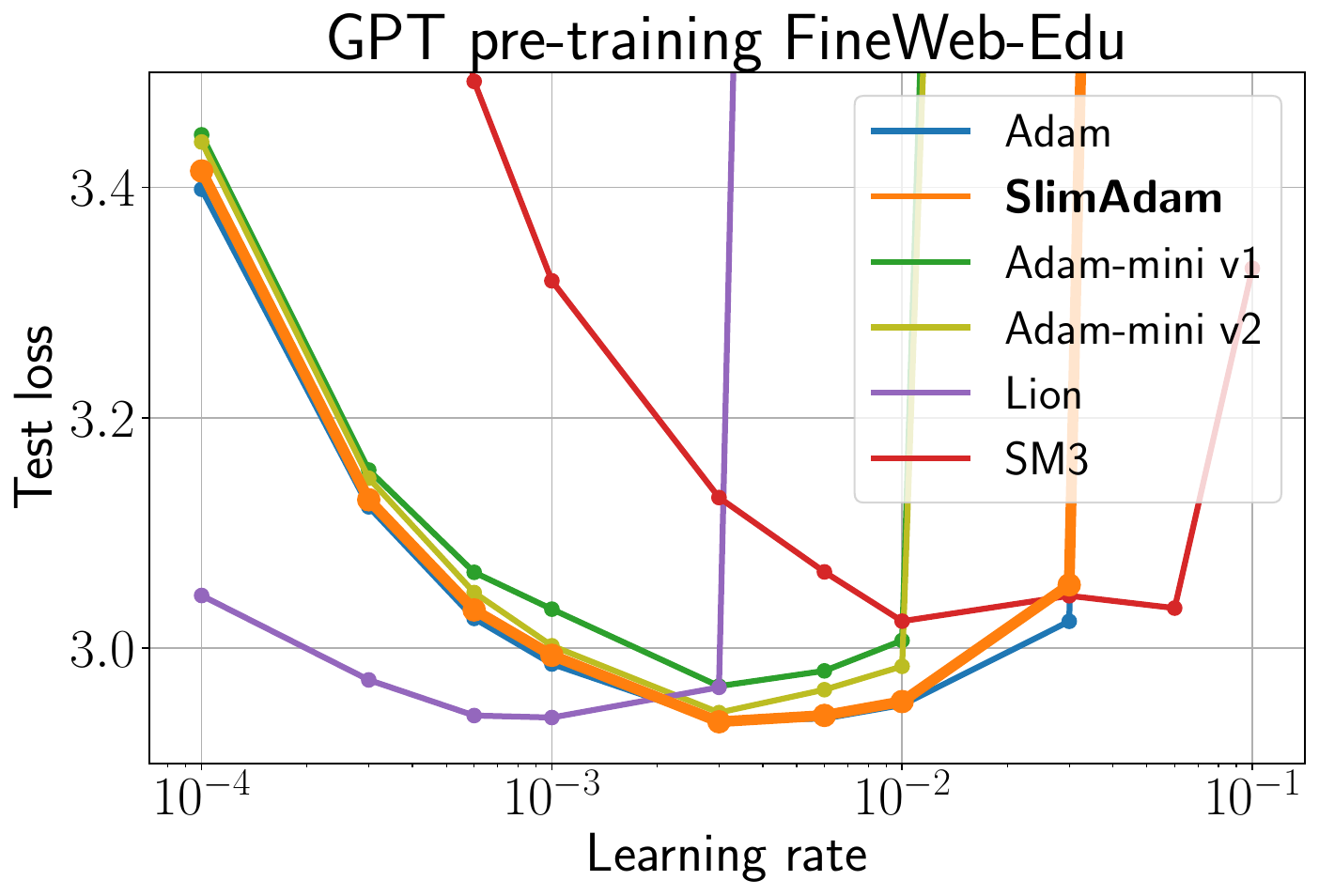}
    \caption{Comparison of common low-memory optimizers on GPT pre-training task using Fineweb-Edu dataset. 
    \emph{SlimAdam} matches Adam's performance with a nearly identical U-shaped loss curve. 
    }
    \label{fig:optim-comparison}
    \vspace{-0.2 in}
\end{figure}

While these factors conspire to make Adam the go-to optimizer for training language models, it requires additional memory beyond SGD-M. It requires storing moving averages of both first and second moments, doubling the optimizer's memory footprint.
This memory cost becomes particularly crucial in resource-limited settings, where the memory allocated to the optimizer states could otherwise be used for the model parameters or activations. 

To avoid the extra memory footprint of Adam, various low-memory optimizers have been proposed \cite{adafactor-shazeer18a,novograd2020,sm3,modoranu2024microadam}. These optimizers are a free lunch in some settings -- slashing memory usage with no detectable loss in performance \cite{zhao2024deconstructingmakesgoodoptimizer,zhang2024adamminiusefewerlearning} -- but they compromise training performance in others \cite{luo-etal-2023-came}. While the potential benefits of low-memory optimizers are clear, a lack of understanding as to when they will perform well is a major barrier to widespread adoption, as the expense of training modern generative models makes engineers unwilling to take risks such as modifying core components in the training recipe.

We argue that a practical low-memory alternative to Adam should exhibit the following properties. First and foremost, it must maintain optimization efficacy, showing no degradation in performance. Additionally, it should preserve Adam's robustness to minor changes in the training hyperparameters. Finally, the low-memory optimizer should immediately work with the same hyperparameter choices as Adam so that users can swap in a low-memory optimizer without major re-tuning.

\Cref{fig:optim-comparison} reveals a natural dichotomy in the space of low-memory optimizers: (1) those that yield learning rate sensitivity curves similar to Adam's, and (2) those that deviate substantially, exhibiting major shifts in optimal learning rates and expected training dynamics.
The first group comprises Adam-mini and our proposed \emph{SlimAdam} which are both constructed by replacing individual second moments with their means along specific dimensions, whereas the latter group composed of Lion, SM3, and Adafactor are all significantly different algorithms.
In this work, we focus on the first category of low-memory optimizers, as they can serve as a drop-in replacement for Adam. Our goal here is to develop a principled framework to help users understand and quantify when these low-memory variants of Adam are appropriate for their problem, thereby improving the reliability of low-memory optimizers and providing deeper insights into Adam's dynamics.

\textbf{Contributions:}
We propose and study a simple measure of the compressibility of Adam's second-moment memory. By examining the Signal-to-Noise Ratio (SNR) of the second moment tensor in each layer, we quantify when individual second moments can be effectively replaced by their means across different matrix/tensor dimensions (such as $\text{fan}_{\text{in}}$, $\text{fan}_{\text{out}}$, or both dimensions). 
Our SNR-based metrics reveal that layers exhibit varying degrees of compressibility along different dimensions, and this compressibility can depend strongly on the architecture, training hyperparameters, and dataset properties. For example, when training a transformer language model, an optimizer should compress key and query second moments in only the $\text{fan}_{\text{in}}$ dimension, as behaviors in the $\text{fan}_{\text{out}}$ dimensions are inconsistent across the multiple heads stacked in that dimension.
While some layer types show consistent compressibility patterns across training configurations, we also observe that some layer types show varying compression trends. 
These inconsistent patterns suggest that a one-size-fits-all approach to low-memory optimization is suboptimal.

To demonstrate the utility of our findings, we implement \emph{SlimAdam}, a memory-efficient variant of Adam that adaptively compresses the second moments along the most efficient dimensions, or selectively leaves layers uncompressed when needed to maintain stability. By taking an adaptive approach to compression, \emph{SlimAdam} preserves desirable properties of Adam while significantly reducing memory usage. For instance, it saves $98\%$ of second moments in $\sim 124M$ parameter GPT-style Transformer trained on language tasks. Further, we show that \emph{SlimAdam} matches Adam's performance as well as robustness to the choice of learning rate.

Our analysis also reveals a surprising property of Adam: it uses significantly more second moments at large learning rates than required for optimal performance. For instance, in GPT-style Transformers trained on language modeling, while the SNR analysis suggests that $\sim 35\%$ of second moments could be compressed at Adam's optimal learning rate, \emph{SlimAdam} actually achieves Adam's performance while compressing $98\%$ of them. This intriguing finding suggests that the majority of Adam's per-parameter adaptivity isn't necessarily required for optimal training.

\subsection{Related Work}
\label{section:related_works}

The superiority of Adam is observed primarily in language modeling, with SGD performing comparably to Adam in image classification settings \cite{whyadaptive2020}. 
This disparity has motivated several investigations into the unique challenges of language modeling landscapes, with studies identifying several explanations. \cite{whyadaptive2020,ahn2024linear} demonstrated that the heavy-tailed distribution of the stochastic gradient noise in language modeling cases causes SGD to perform worse than Adam. 
\cite{pan2022toward} attributed Adam's faster convergence to ``directional sharpness,'' which is the curvature along the update direction. 
Adding to these findings, \cite{zhang2024transformersneedadamhessian} illustrated that the Hessian spectrum across parameter blocks varies heavily and suggested that SGD performs worse because it applies a single learning rate to all blocks. 
Further insights come from \cite{kunstner2024heavytailedclassimbalanceadam}, who showed that, in settings with heavy-tailed class imbalance, SGD struggles to decrease loss in infrequent classes, while adaptive optimizers are less sensitive to this imbalance. 
\cite{zhao2024deconstructingmakesgoodoptimizer} argued that Adam's advantage over SGD in language modeling primarily stems from using per-parameter adaptive learning rates in two specific components---LayerNorm and the final layer---positing that for all other layers, a single shared second moment is sufficient.

\begin{figure*}[!htb]
\centering
\begin{minipage}[b]{0.245\textwidth}
    \centering
    \includegraphics[width=\textwidth]{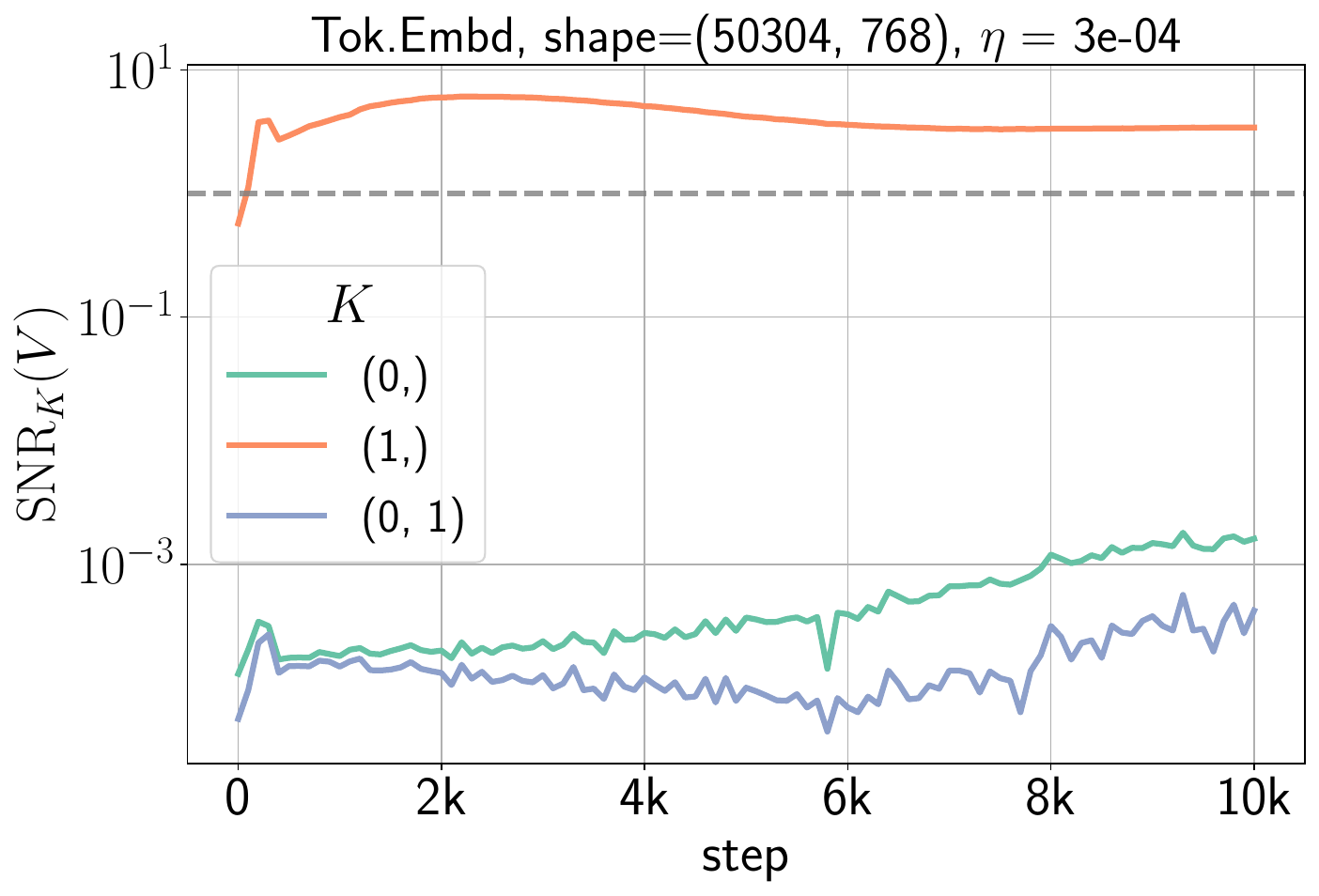}
\end{minipage}
\hfill
\begin{minipage}[b]{0.245\textwidth}
    \centering
    \includegraphics[width=\textwidth]{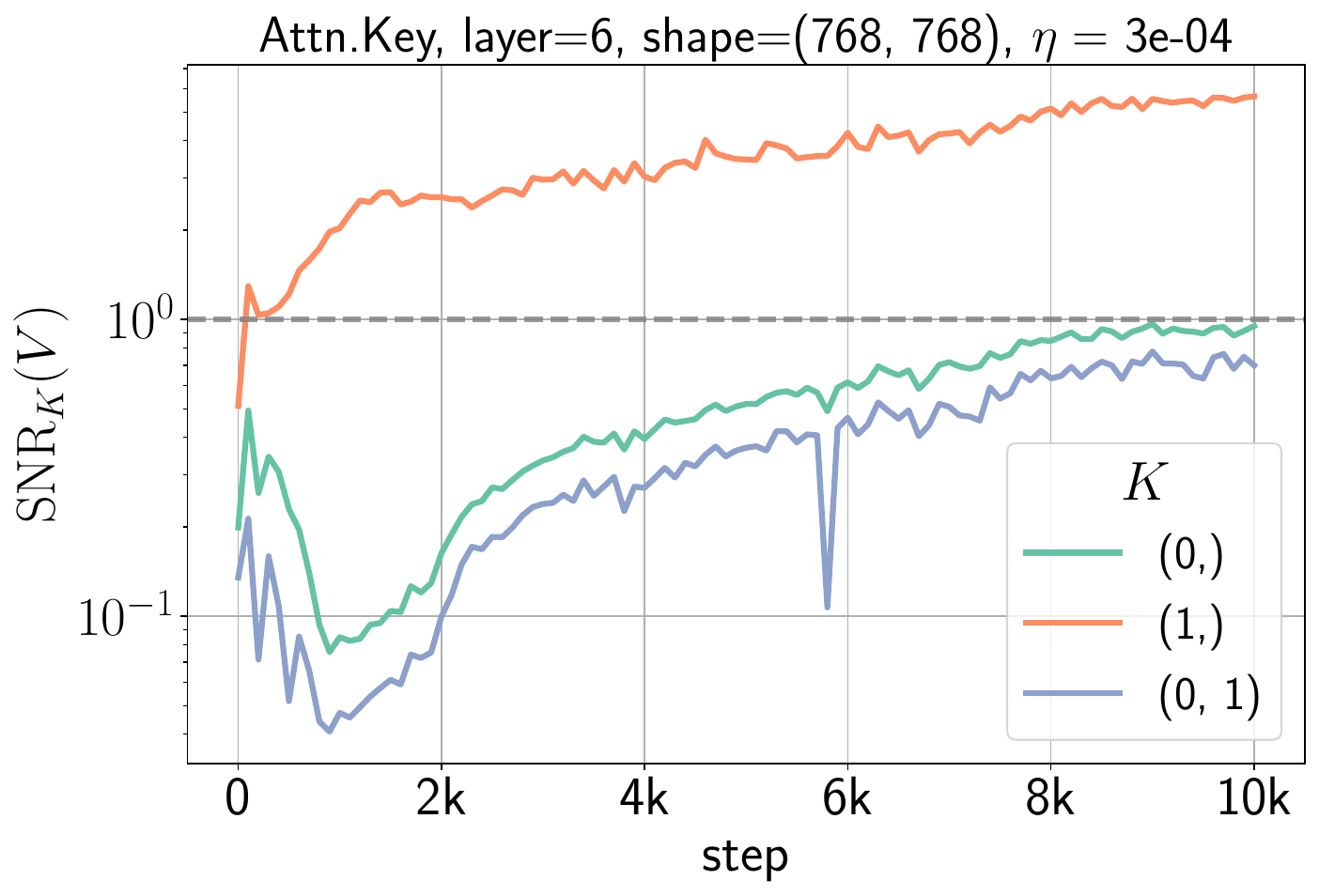}
\end{minipage}
\hfill
\begin{minipage}[b]{0.245\textwidth}
    \centering
    \includegraphics[width=\textwidth]{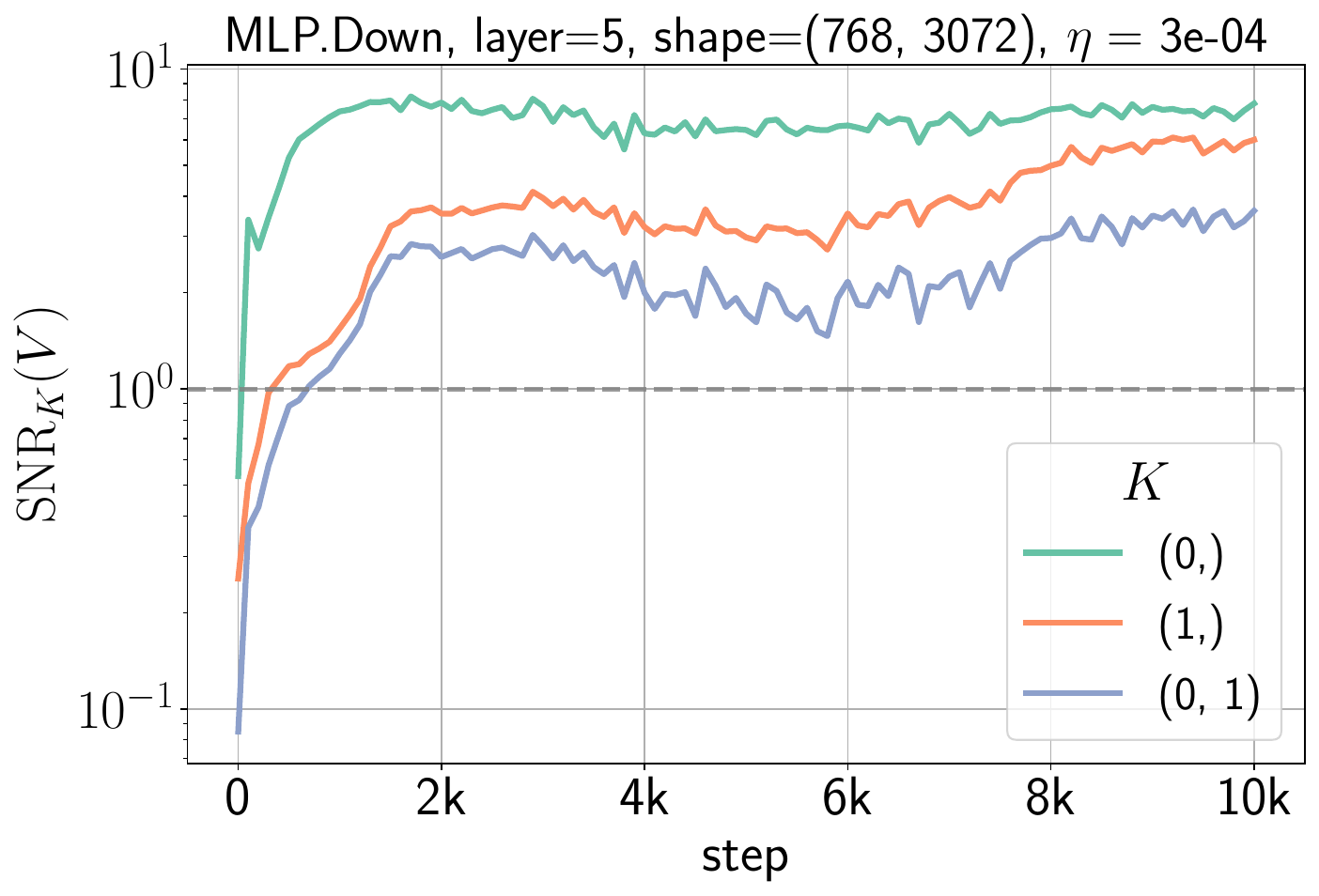}
\end{minipage}
\hfill
\begin{minipage}[b]{0.245\textwidth}
    \centering
    \includegraphics[width=\textwidth]{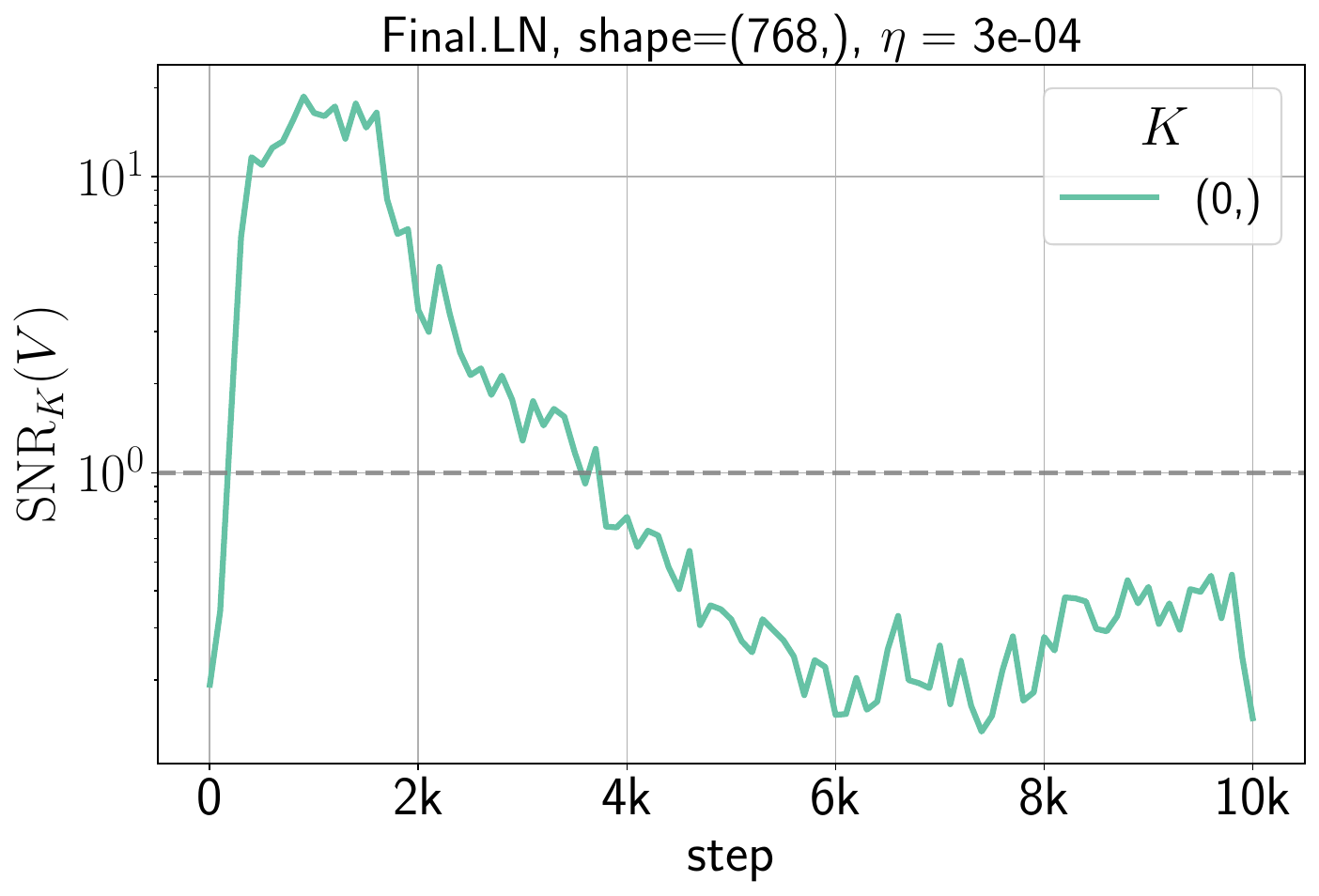}
\end{minipage}

\caption{SNR trajectories of selected second-moment blocks of GPT-small model trained on OpenWebText. Different compression dimensions are denoted as: $K = 0$ for $\text{fan}_{\text{out}}$, $K = 1$ for $\text{fan}_{\text{in}}$, and $K = (0, 1)$ for both dimensions. }
\label{fig:snr-curves-gpt-small-openweb}
\end{figure*}

\textbf{Low-memory optimizers:}
Several approaches have been proposed to reduce Adam's memory footprint in the past few years.
Adafactor \cite{adafactor-shazeer18a} approximates the second-moment matrix of a layer using a moving average of the row and column sums of the squared gradients.
SM3 \cite{sm3} groups parameters into sets based on similarity, such that each parameter can belong to multiple sets. Then, it maintains a moving average of the maximum of squared moments for each set and approximates a second-moment entry using the minimum value across different sets it belongs to.
Lion \cite{chen2023symbolic} is an algorithmically discovered optimizer that only tracks momentum and uses sign operation to estimate the update.
MicroAdam \cite{modoranu2024microadam} combines gradient sparsification, quantization, and error feedback to compress optimizer states. 
Adam-mini \cite{zhang2024adamminiusefewerlearning} assigns adaptive learning rates to block partitions based on the Hessian spectrum at initialization. 
In \Cref{appendix:related-works}, we further discuss closely related works in detail.

\section{Notations and Preliminaries}
\label{section:preliminaries}

\paragraph{Adam:} Consider a loss function $L(\bm{\theta})$ parameterized by parameters $\bm{\theta}$. For a weight matrix $W \in \mathbb{R}^{\text{fan}_{\text{out}} \times \text{fan}_{\text{in}}}$, let $G_t:= \nabla_W L(\bm{\theta}_t)$ denote its gradient at step $t$. Adam updates these weights using learning rate $\eta_t$ and the moving averages of the first two moments of gradients, denoted by $M_t$ and $V_t$, with coefficients $\beta_1$ and $\beta_2$, respectively. The equations governing the updates are:
\begin{align}
     & M_{t+1} = \beta_1 M_t + (1 - \beta_1) G_t \nonumber \\
     & V_{t+1} = \beta_2 V_t + (1 - \beta_2) G^2_t \nonumber \\
     & W_{t+1} = W_t - \eta_t \frac{\hat{M}_{t+1}}{\sqrt{\hat{V}_{t+1}} + \epsilon}.
     \label{equation:adam}
\end{align}
Here, $\hat{M}_t = \frac{M_t}{1 - \beta_1^t}$ and $\hat{V}_t = \frac{V_t}{1 - \beta_2^t}$ are the bias-corrected moments and $\epsilon$ is a small scalar used for numerical stability.

For our analysis, we generalize Adam to a family of low-memory variants parameterized by layer-specific sharing dimensions. For each layer, we compute an estimate of the second moments by averaging squared gradients across specified dimensions $K$ (${\text{fan}_{\text{in}}}$, ${\text{fan}_{\text{out}}}$, or both). 
The difference compared to Adam lies in the second moment update:
\begin{align}
    & V_{t+1} = \beta_2 V_t + (1 - \beta_2) \mathbb{E}_K \left[G^2_t \right],
    \label{equation:adashare}
\end{align}
where $\mathbb{E}_K[\cdot]$ denotes an average over dimensions $K$. 
Since Adam's second moment acts as a per-parameter ``effective'' learning rate, averaging these moments along dimensions $K$ is equivalent to sharing a common learning rate. 
The above optimizer coincides with Adam when $K = \emptyset$. Another notable limiting case is AdaLayer \cite{zhao2024deconstructingmakesgoodoptimizer}, which maintains one second moment per parameter block. 
In \Cref{section:slimadam}, we introduce \emph{SlimAdam}, a special member of the low memory Adam family, where the averaging dimensions $K$ are determined by our SNR analysis.

Throughout this work, we partition second moments using the default model parameter partitioning scheme that groups parameters at the granularity of layer components (e.g., weights, biases, and attention components), rather than fine-grained divisions such as per-attention-head partitioning as in \cite{zhang2024adamminiusefewerlearning}.
While more fine-grained partitioning could offer additional insights, using a simple partitioning scheme ensures applicability to a broad range of architectures without having to modify the analysis or optimizer code. Nevertheless, we still account for special dimensions such as attention heads while interpreting the results.
We use $K = 0$ for $\text{fan}_{\text{out}}$, $K = 1$ for $\text{fan}_{\text{in}}$ and $K = (0, 1)$ to denote sharing along both dimensions.

\begin{figure*}[!htb]
\centering
\begin{minipage}[b]{0.245\textwidth}
    \centering
    \includegraphics[width=\textwidth]{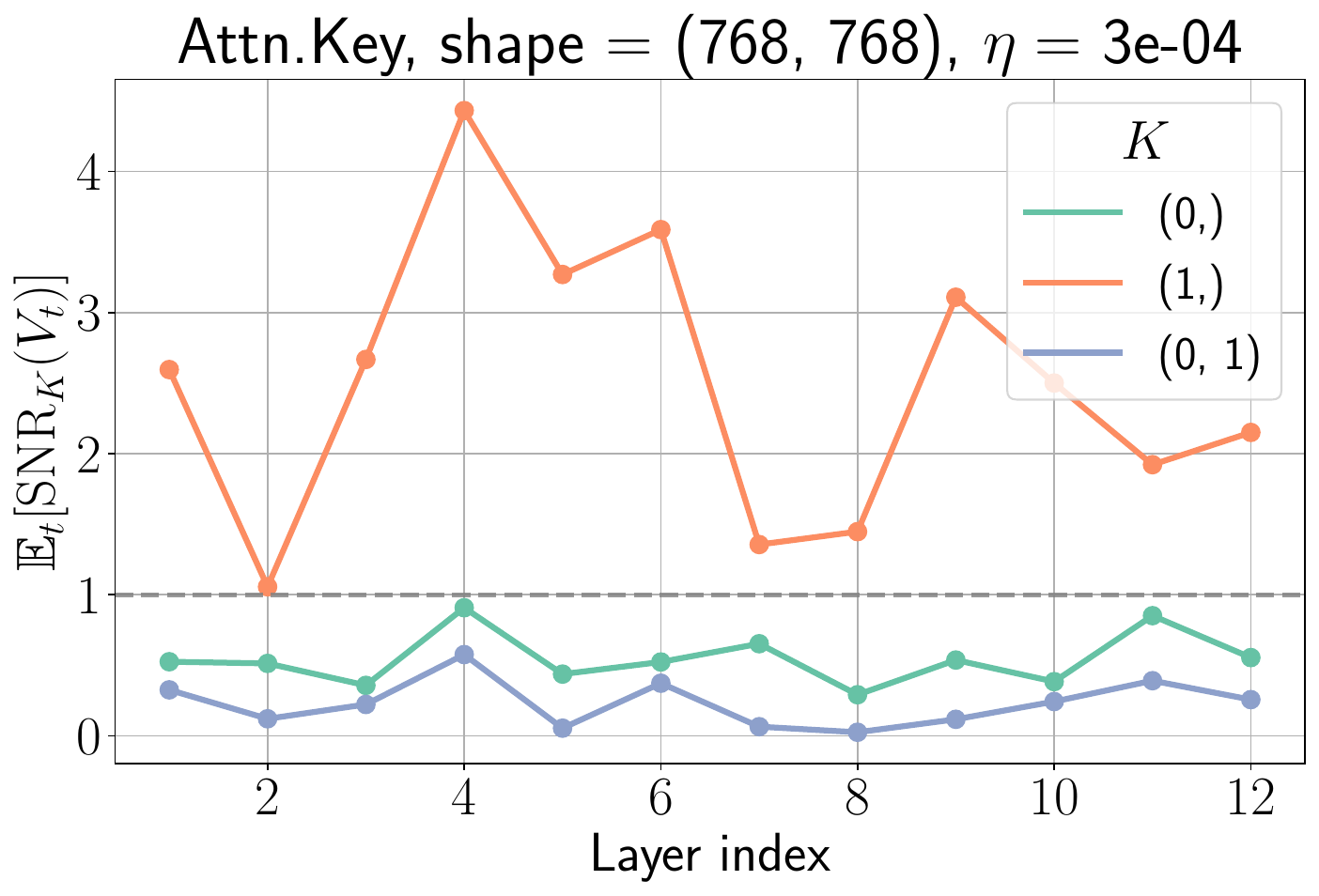}
\end{minipage}
\hfill
\begin{minipage}[b]{0.245\textwidth}
    \centering
    \includegraphics[width=\textwidth]{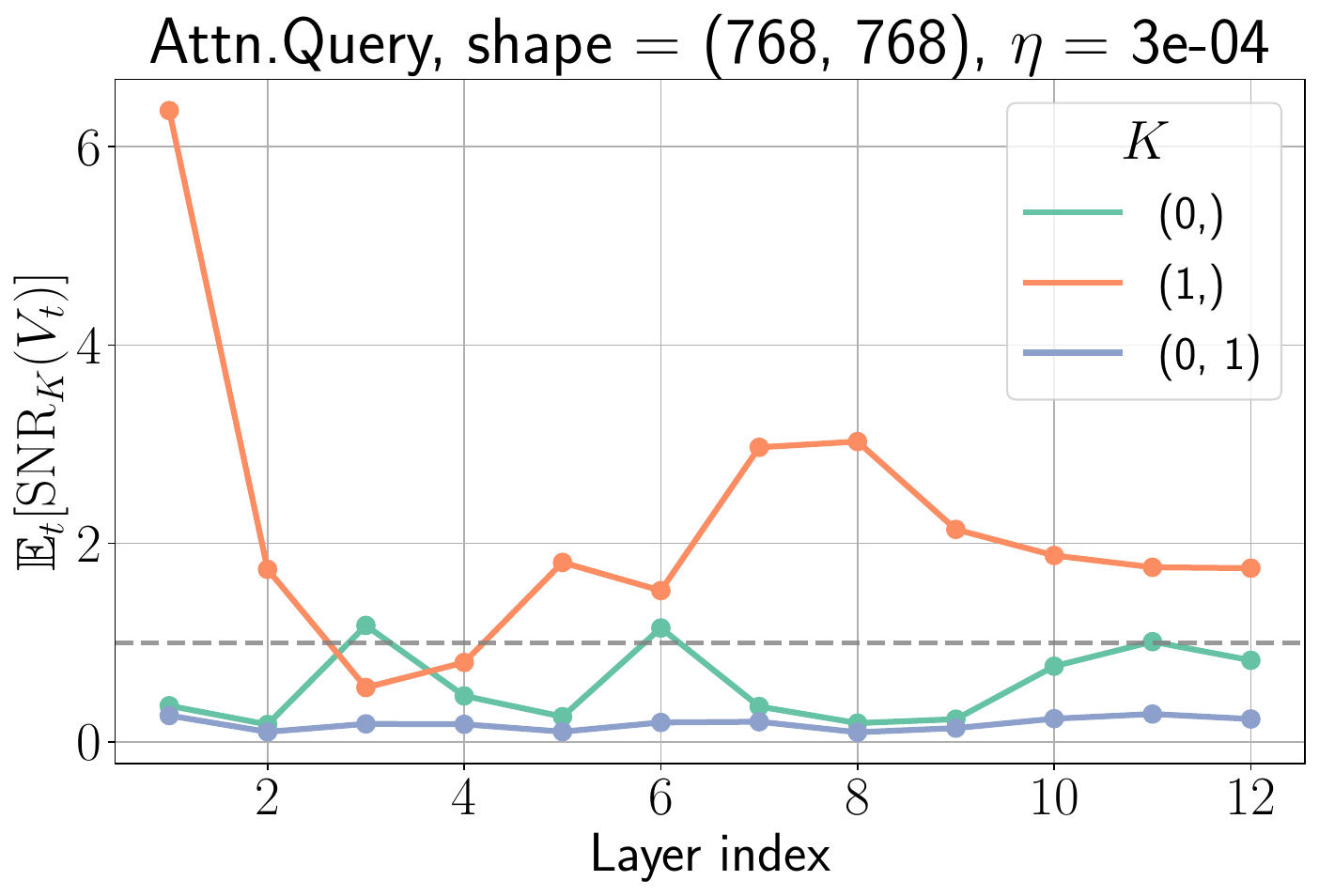}
\end{minipage}
\hfill
\begin{minipage}[b]{0.245\textwidth}
    \centering
    \includegraphics[width=\textwidth]{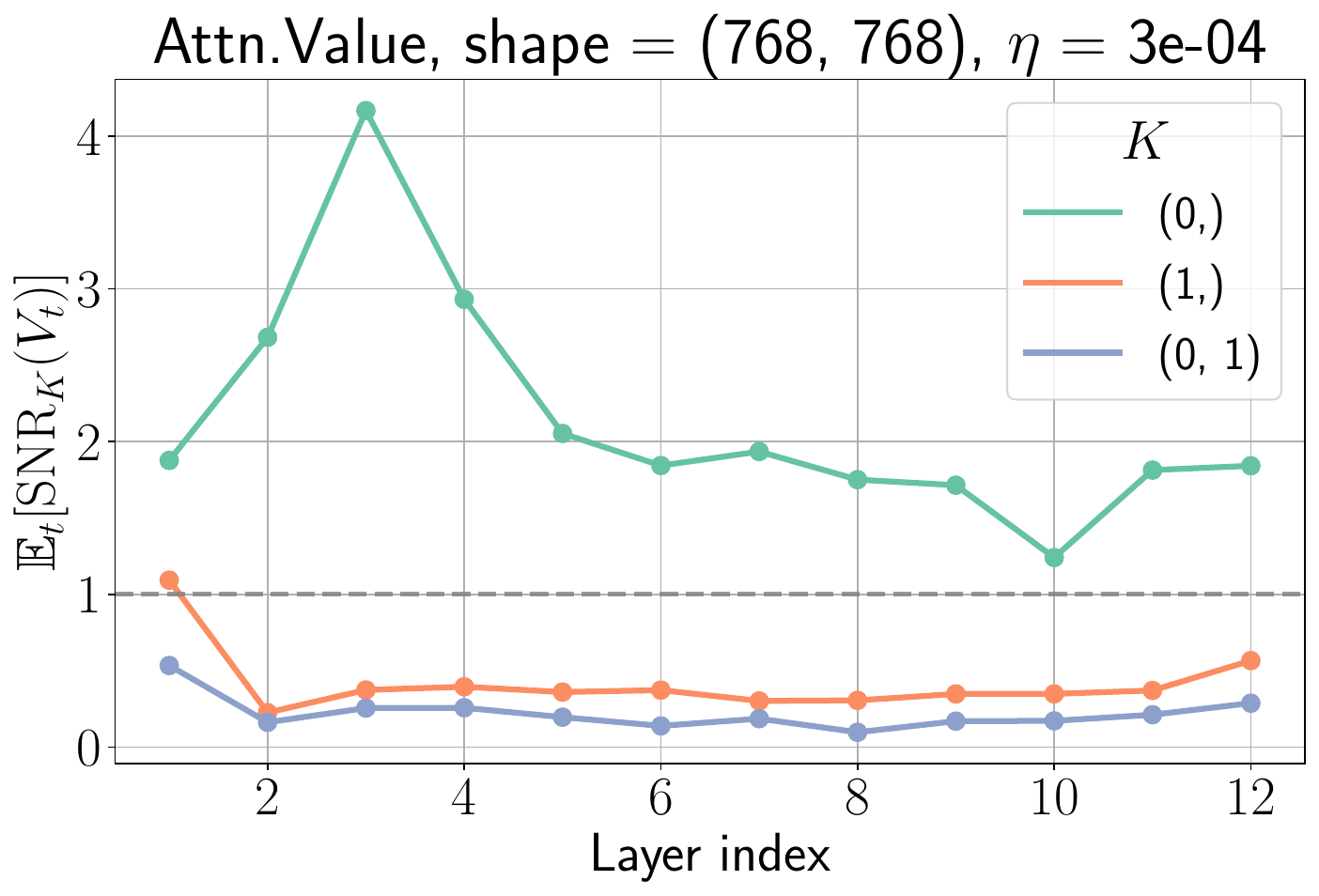}
\end{minipage}
\hfill
\begin{minipage}[b]{0.245\textwidth}
    \centering
    \includegraphics[width=\textwidth]{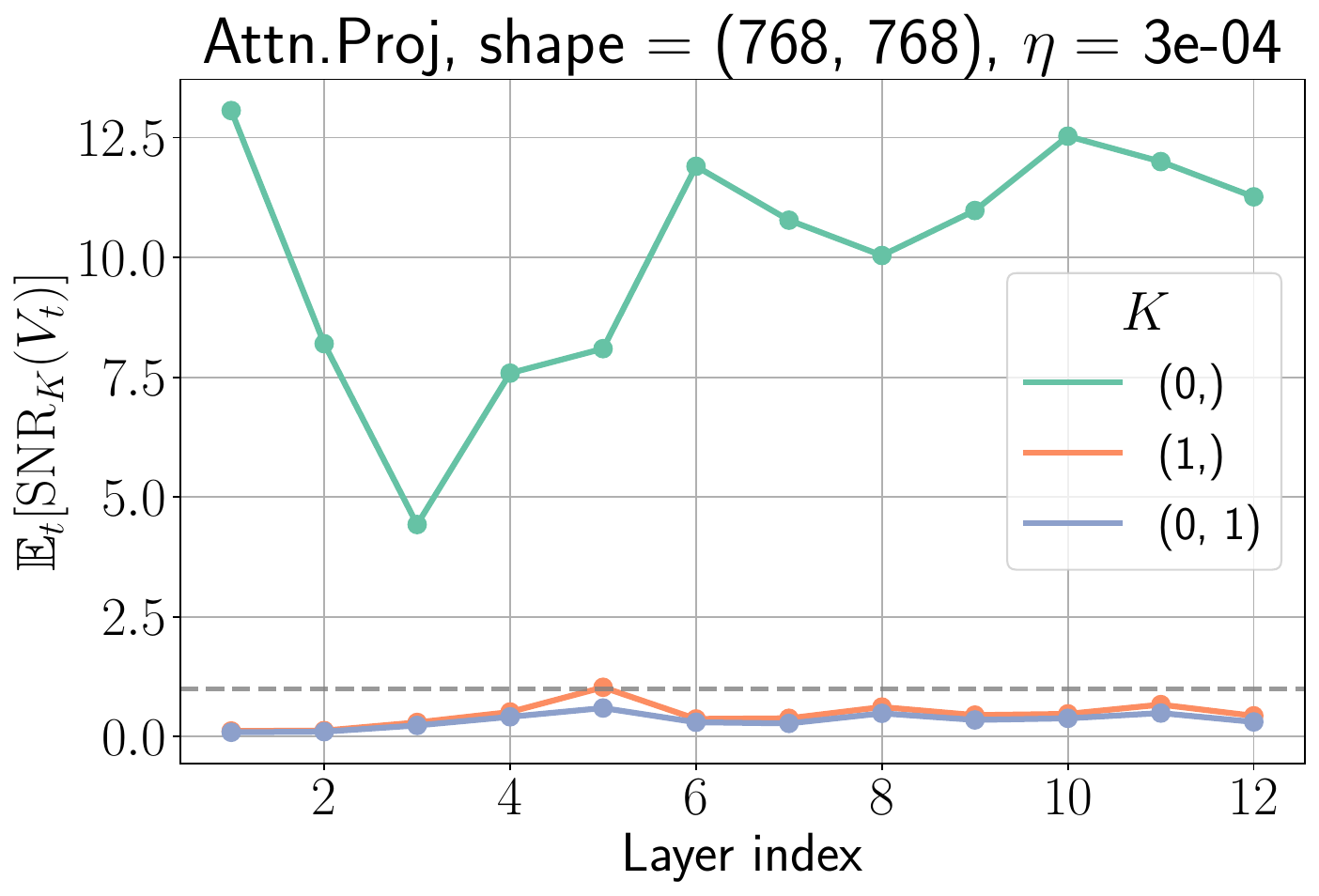}
\end{minipage}

\begin{minipage}[b]{0.245\textwidth}
    \centering
    \includegraphics[width=\textwidth]{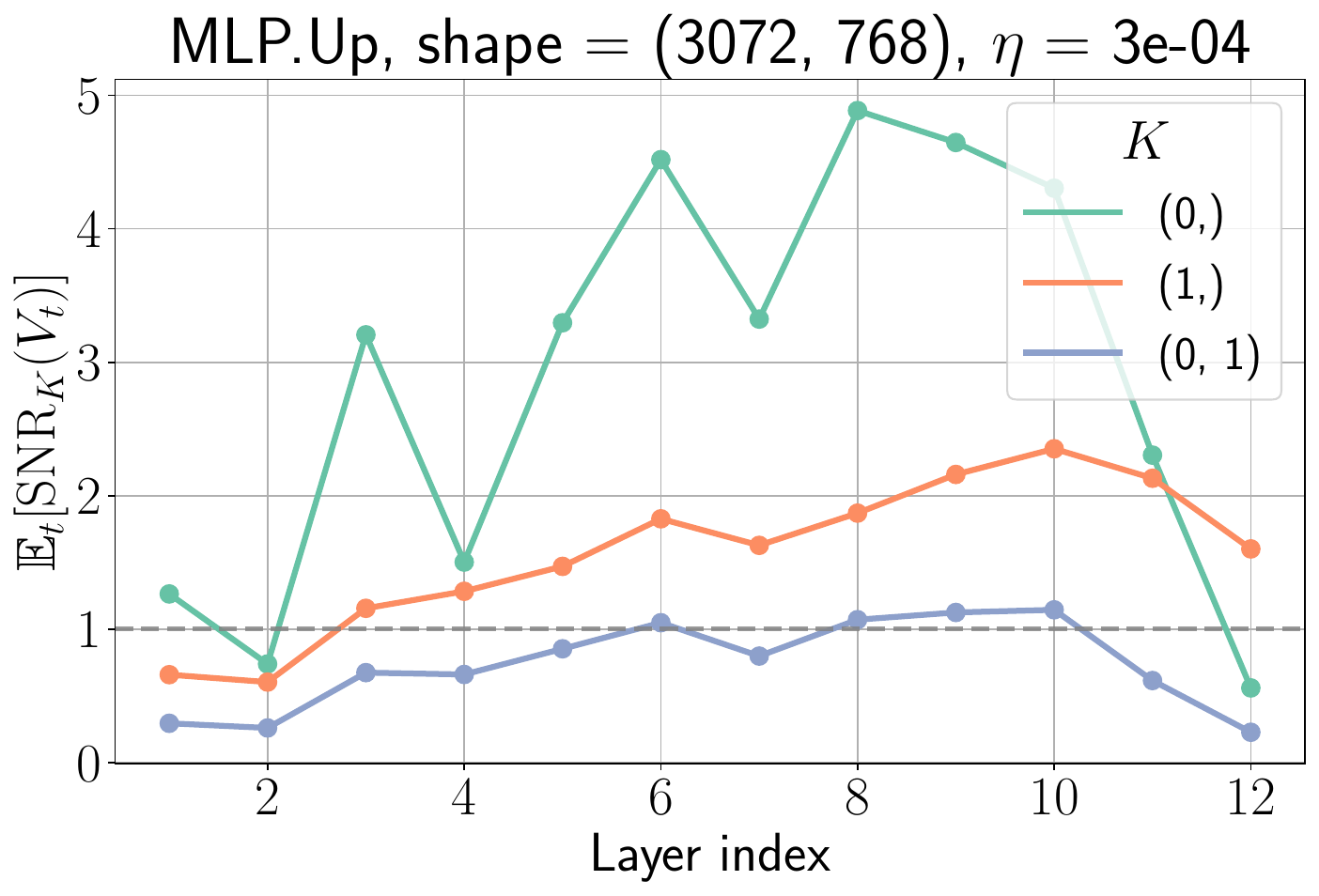}
\end{minipage}
\hfill
\begin{minipage}[b]{0.245\textwidth}
    \centering
    \includegraphics[width=\textwidth]{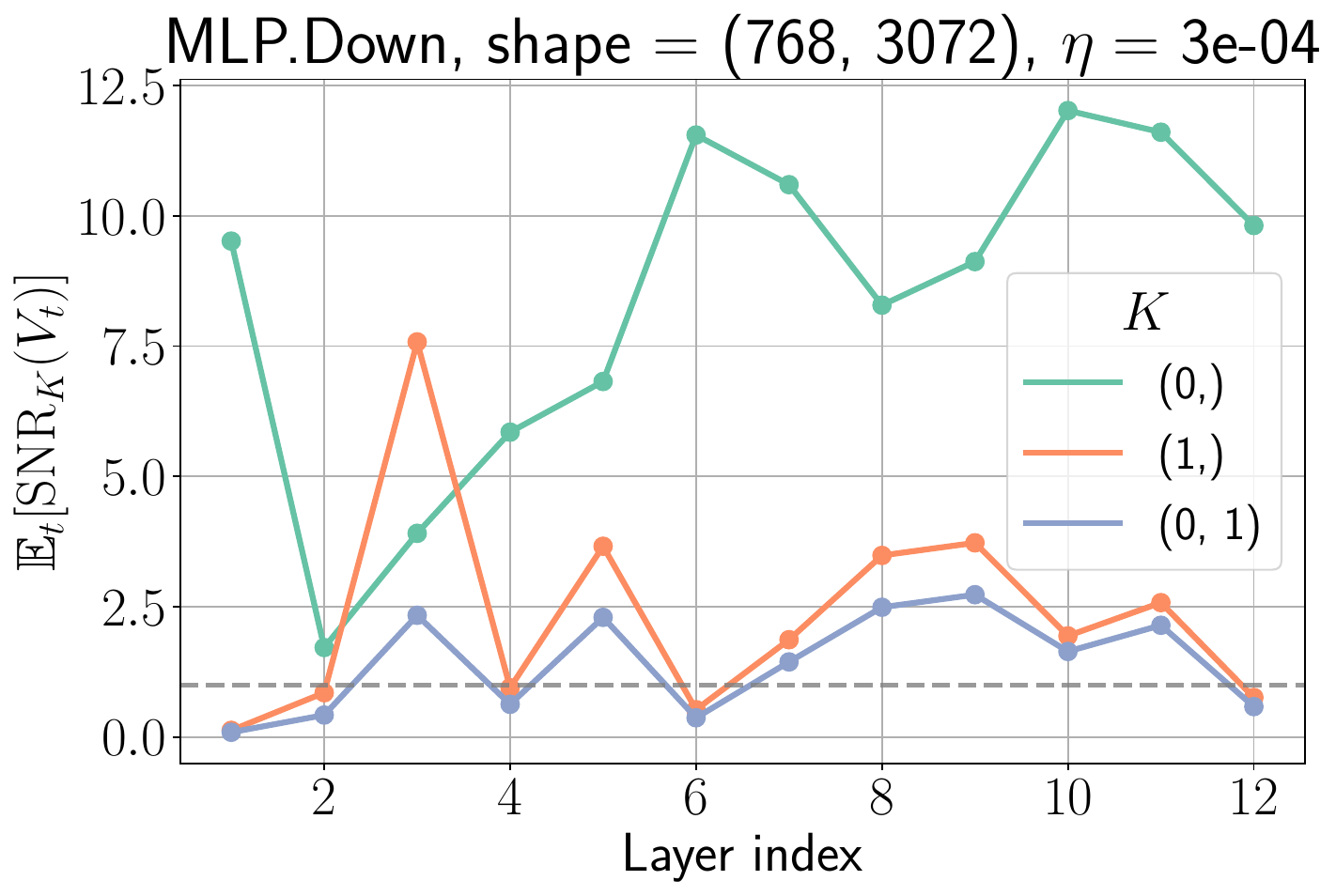}
\end{minipage}
\hfill
\begin{minipage}[b]{0.245\textwidth}
    \centering
    \includegraphics[width=\textwidth]{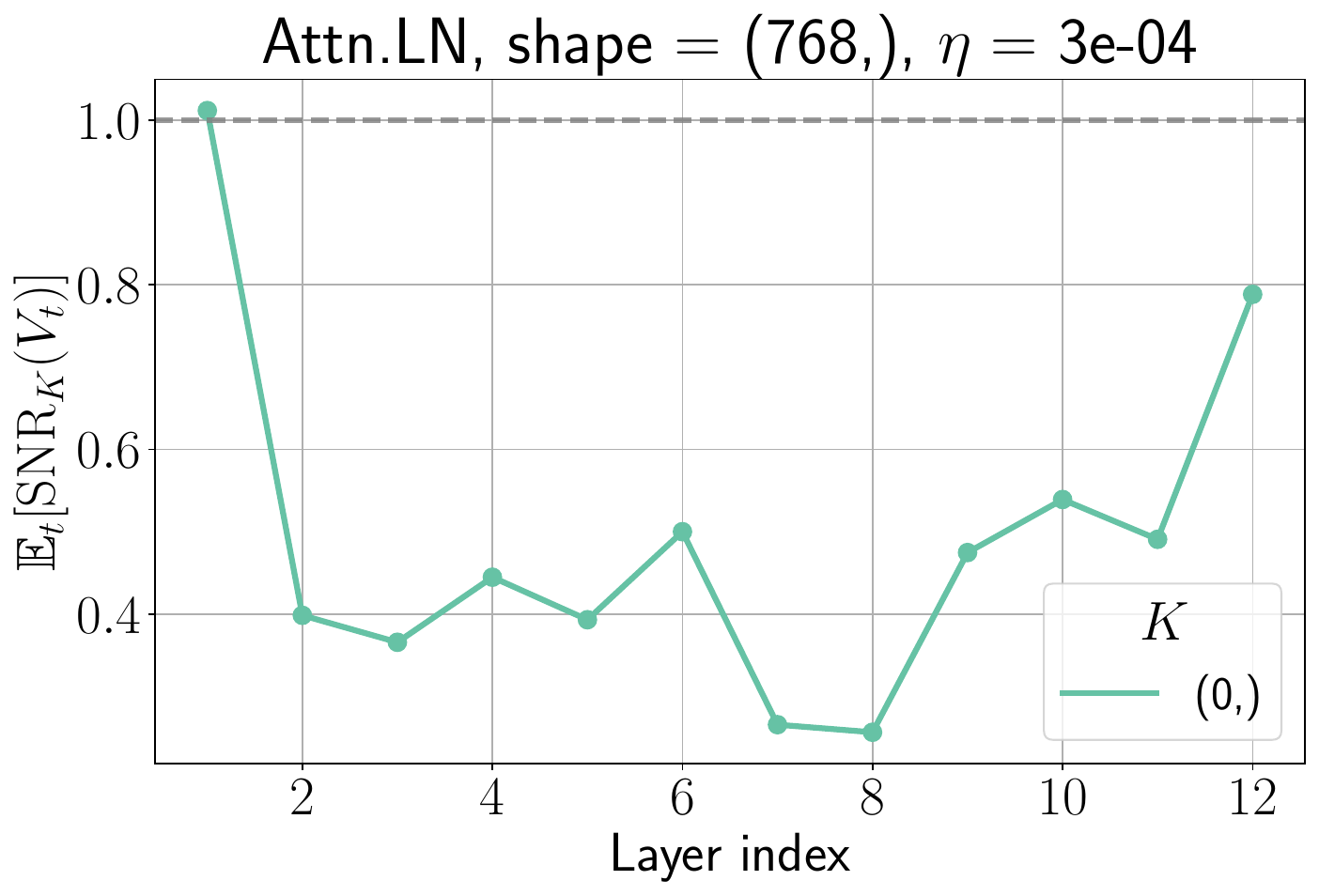}
\end{minipage}
\hfill
\begin{minipage}[b]{0.245\textwidth}
    \centering
    \includegraphics[width=\textwidth]{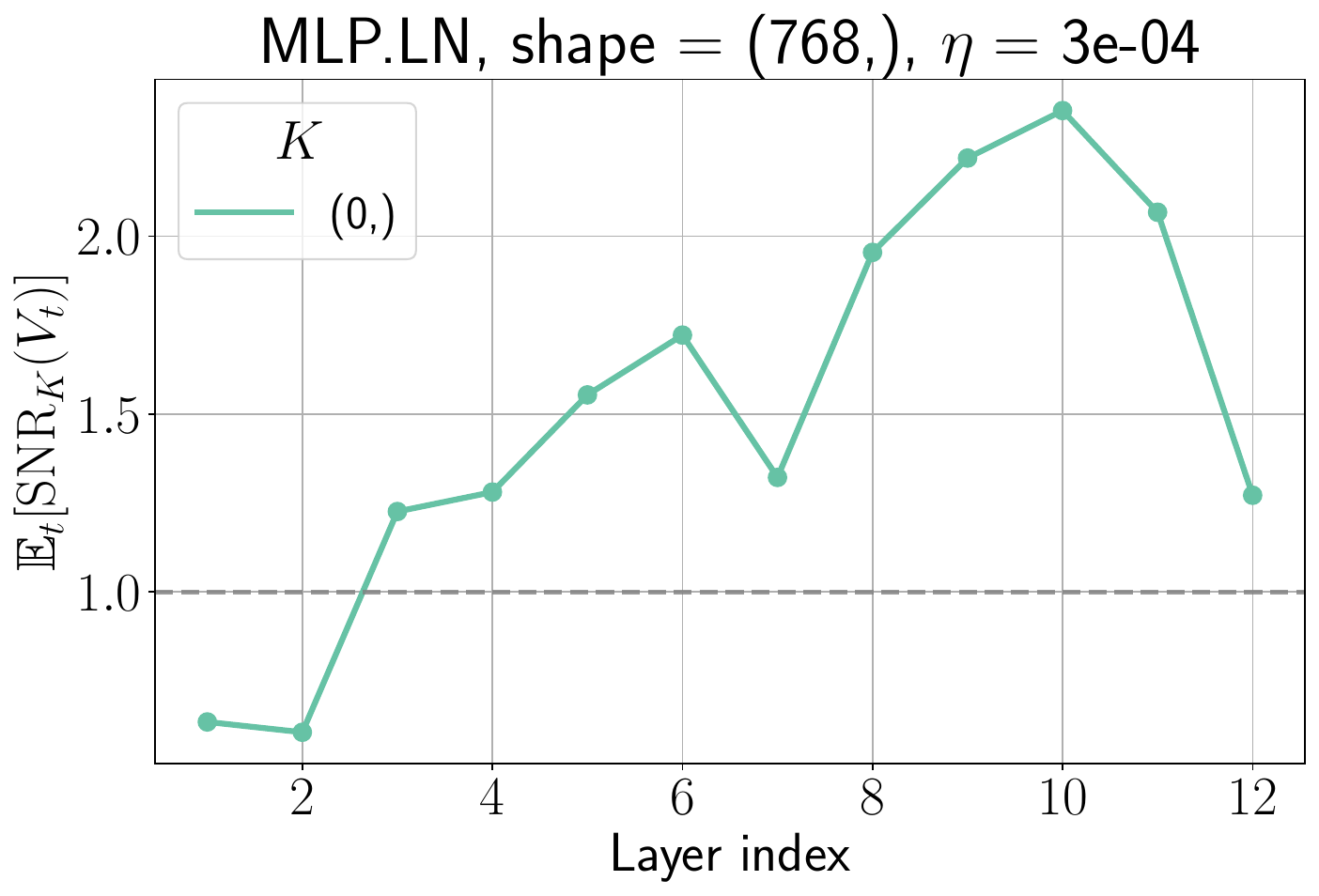}
\end{minipage}
\caption{Depth dependence of average SNR values for different second-moment blocks of the GPT-small model trained on OpenWebText. The experimental setup is the same as in \Cref{fig:snr-curves-gpt-small-openweb}.
}
\label{fig:snr-layer-gpt-small-openweb}
\end{figure*}

\section{SNR Analysis of Adam's Second Moments}
\label{section:snr-analysis}
This section analyzes how effectively Adam's per-parameter second moments can be replaced by their mean along different dimensions (such as $\text{fan}_{\text{in}}$, $\text{fan}_{\text{out}}$, or both) during training. The feasibility of such a compression depends on how tightly the entries are clustered around their mean value. If entries along a dimension exhibit low variance relative to their mean, they can be effectively represented by a single value. To quantify this concentration of values, we analyze the Signal-to-Noise Ratio (SNR) of the second moments during training. For a second moment matrix $V \in \mathbb{R}^{\text{fan}_{\text{out}} \times \text{fan}_{\text{in}}}$ and specified compression dimensions $K$, $\text{SNR}_{K}$ is defined as:
\vspace{-0.1 in}
\begin{align}
    \text{SNR}_K(V_t) = \mathbb{E}_{K'} \left[\frac{\left(\mathbb{E}_{K}[V_t]\right)^2}{\text{Var}_{K}[V_t]} \right]
\end{align}
where $\mathbb{E}_{K}[\cdot]$ and $\text{Var}_{K}[\cdot]$ compute the mean and variance along the specified dimensions $K$, while the outer expectation $\mathbb{E}_{K'} [\cdot]$ averages the ratio over the remaining dimensions to obtain a scalar.

$\text{SNR}_K$ quantifies the feasibility of compression along dimensions $K$ along an Adam trajectory. When $\text{SNR}_K \gtrsim 1$, the signal dominates the noise, indicating that entries can be effectively described by their mean, whereas $\text{SNR}_K \lesssim 1$ suggests that individual entries carry significant information that would be lost when the entries are replaced by their mean. 
This analysis not only suggests layers that can be compressed but also quantifies their relative compression feasibility.
As Adam adapts to the local geometry of the optimization landscape, SNR values also serve as a proxy for learning complexity during training, with lower SNR suggesting higher complexity and a need for per-parameter effective learning rates.

\subsection{Compressibility in Diverse Training Regimes}
\label{section:diverse-regimes}

We analyze the evolution of SNR across diverse training configurations (pre-training, fine-tuning, image classification) to uncover fundamental SNR trends. For each setup, experimental details and supplementary results are provided in \Cref{appendix:experimental-details} and \Cref{appendix:additional-snr-results}, respectively.

We introduce our methodology by examining a representative example. \Cref{fig:snr-curves-gpt-small-openweb} (left) shows SNR trajectories of the second-moment matrix for the Token Embedding layer of a GPT-small model trained on a language pre-training task. These SNR trajectories typically exhibit an early transient phase where their value quickly grows, followed by a late time phase where these values may consistently increase, decrease, or stabilize. We are interested in cases where it is feasible to replace the second moments by their mean throughout training. To this end, we define average SNR as:
\begin{align}
    \mathbb{E}_\tau\left[ \text{SNR}_K(V_\tau) \right] = \frac{1}{T} \sum_{\tau}^T \text{SNR}_K(V_\tau),
\end{align}
where $\tau$ indexes the training steps at which SNR is measured and $T$ is the total number of SNR measurements. The averaged SNR quantifies the feasibility of compression along dimensions $K$ throughout an Adam trajectory.

\subsubsection{Language Pre-training}
\label{section:pre-training}

We analyze GPT-style Transformers \cite{radford2019language} trained on two language modeling datasets: OpenWebText \cite{Gokaslan2019OpenWeb} and $10$B token subset of FineWeb-Edu \cite{penedo2024the}. 
\Cref{fig:snr-curves-gpt-small-openweb} shows SNR trajectories as a function of the optimization step for selected second-moment blocks of a GPT-small model trained on OpenWebText. \Cref{fig:snr-layer-gpt-small-openweb} presents the depth dependence of the averaged SNR values of different parameter types within a standard transformer block. 
The lack of consistency as to which compression dimension $K$ exhibits higher SNR across different layer types, suggests that optimal compression strategies must be customized for each parameter category rather than applying a uniform approach throughout the model. Below, we describe these trends in detail and discuss their implications. 

The Token Embedding and Language Modeling Head (LM Head\footnote{Unless otherwise mentioned, we use weight tying meaning that the Token Embedding and LM Head share the same underlying set of parameters and moments.}) second moments show a strong aversion to compressing along the token dimension (vocabulary dimension) while favoring compression along the embedding dimension. This pattern suggests that the subset of the parameter matrix corresponding to each individual token in the vocabulary evolves at its own pace during training, thereby requiring its own learning rate. 
This result aligns with recent studies \cite{zhang2024adamminiusefewerlearning,zhao2024deconstructingmakesgoodoptimizer} that suggest not compressing the token embedding and LM Head matrices in language modeling. Our SNR analysis extends their analysis by revealing that this aversion to compression is specific only to the token dimension.

\begin{figure*}[!htb]
\centering
\begin{minipage}[b]{0.245\textwidth}
    \centering
    \includegraphics[width=\textwidth]{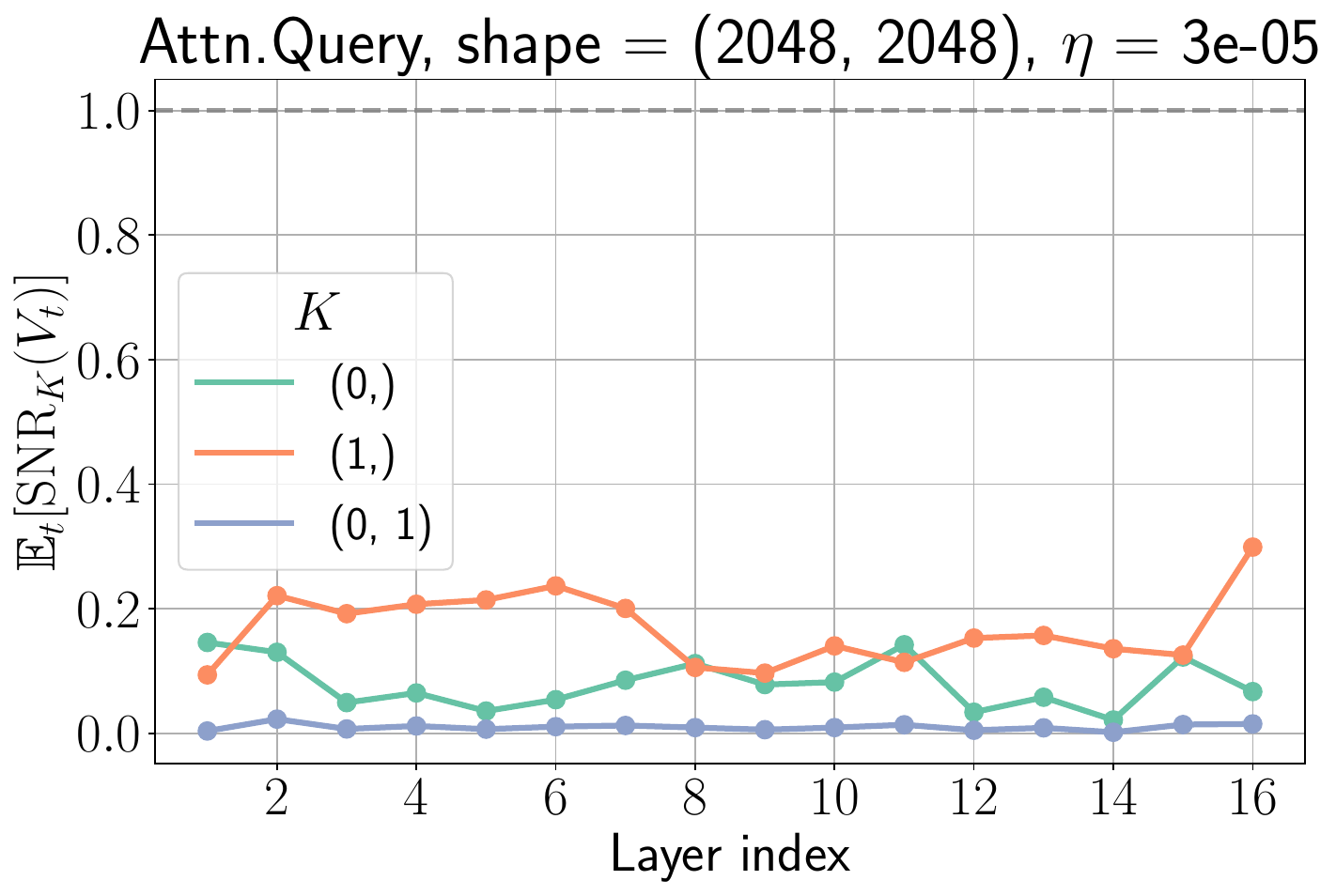}
\end{minipage}
\hfill
\begin{minipage}[b]{0.245\textwidth}
    \centering
    \includegraphics[width=\textwidth]{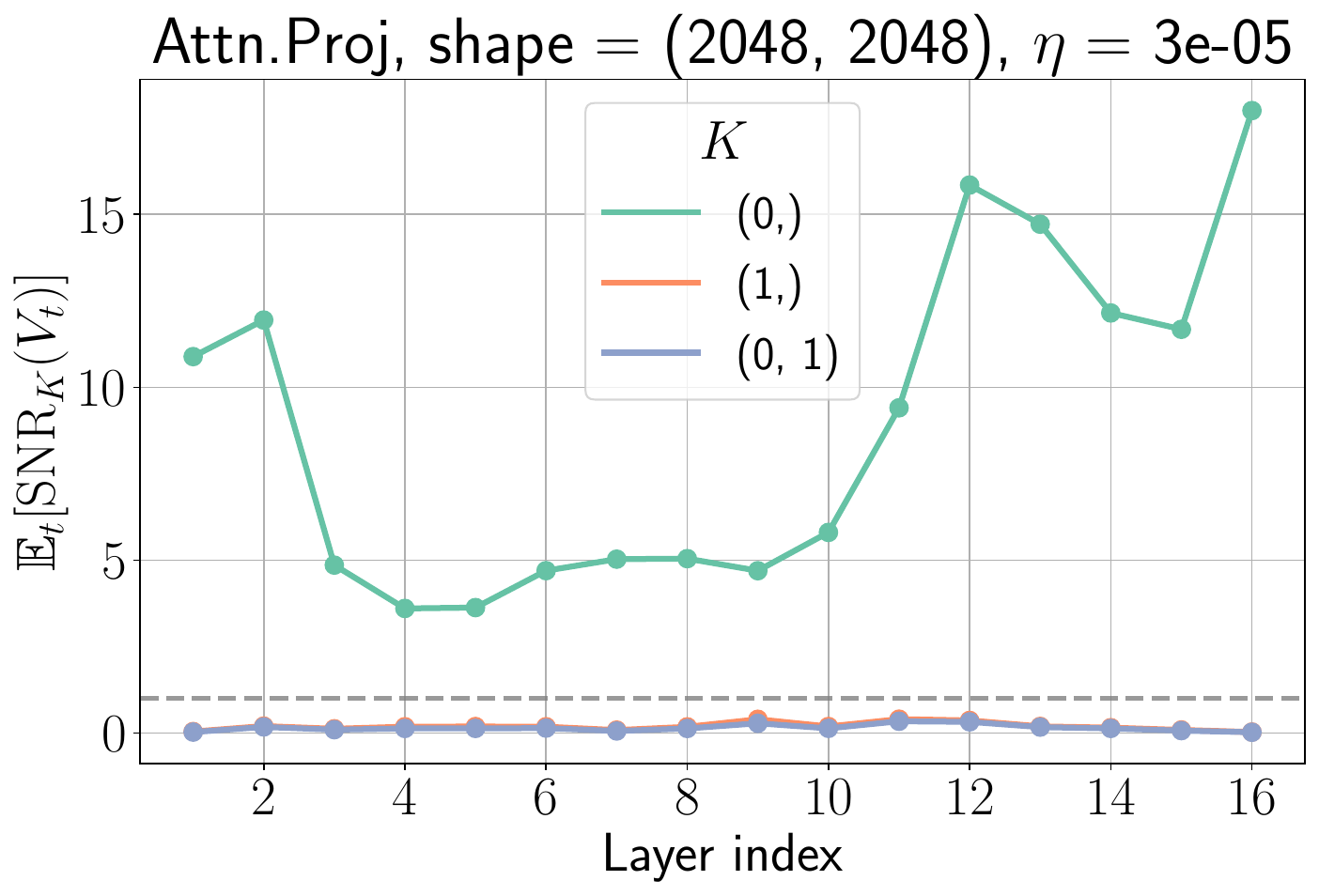}
\end{minipage}
\hfill
\begin{minipage}[b]{0.245\textwidth}
    \centering
    \includegraphics[width=\textwidth]{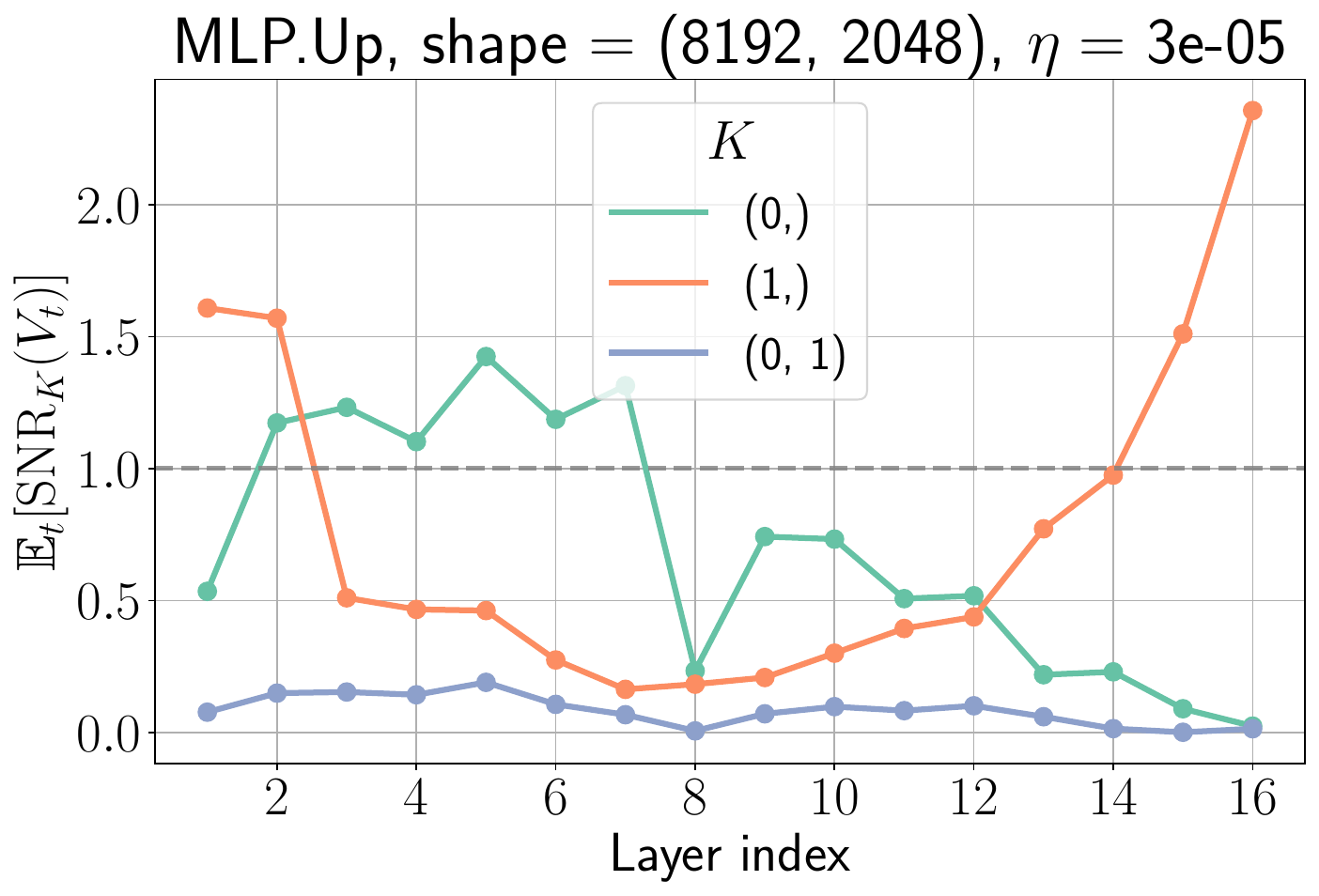}
\end{minipage}
\hfill
\begin{minipage}[b]{0.245\textwidth}
    \centering
    \includegraphics[width=\textwidth]{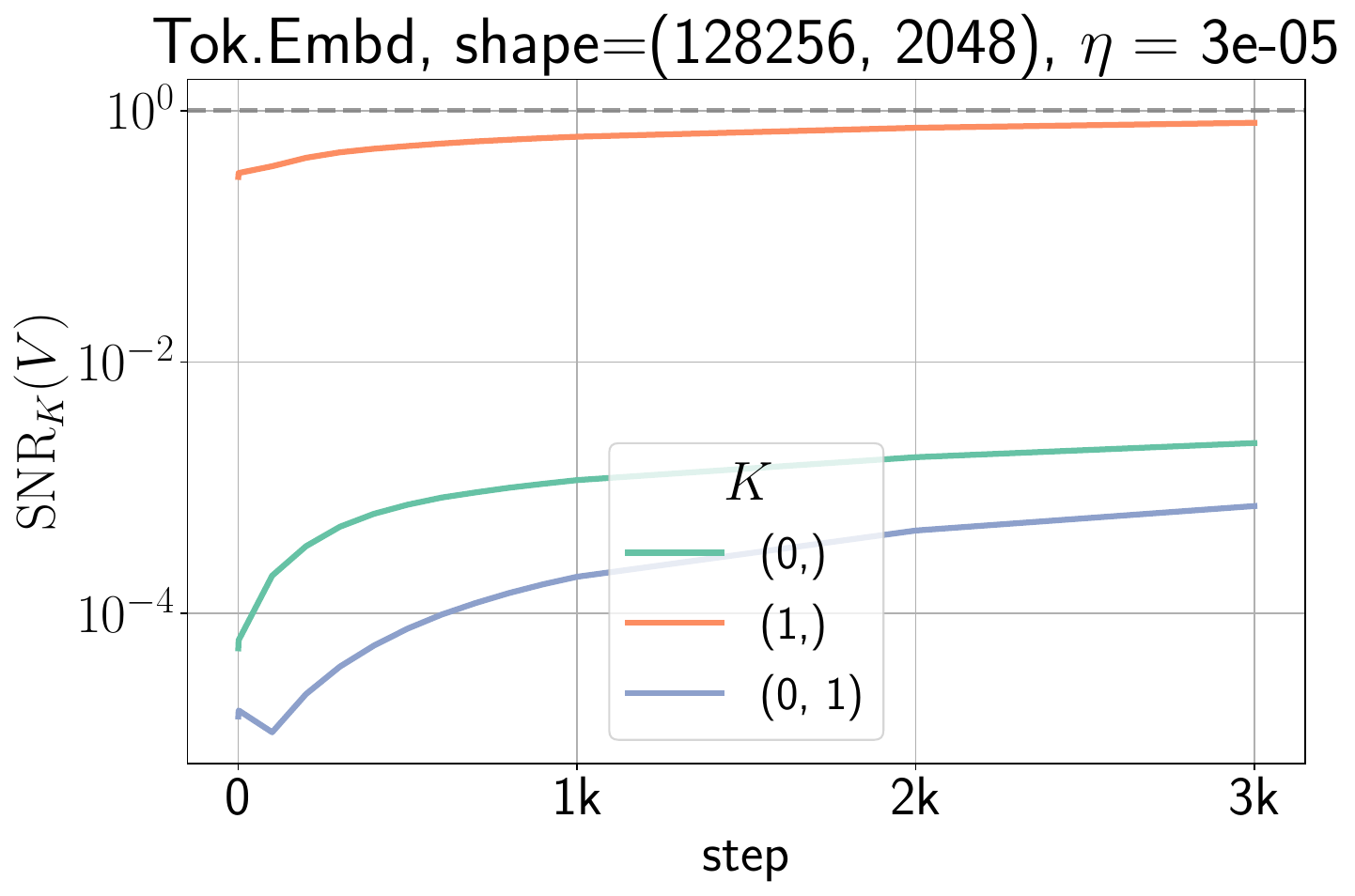}
\end{minipage}
\caption{SNR trends for selected layers of pre-trained Llama 3.2 1B fine-tuned on Alpaca dataset. For detailed results, see \Cref{appendix:snr-finetuning}.}
\label{fig:snr-layer-llama-1b-alpaca}
\vspace{-0.1 in}
\end{figure*}

The second moments of attention keys and queries consistently show aversion to compression along the $\text{fan}_{\text{out}}$ dimension, where multiple heads are stacked. This pattern suggests that each attention head requires its own effective learning rate. \cite{zhang2024adamminiusefewerlearning} reached similar conclusions through an independent Hessian-based analysis, corroborating our findings. On the other hand, the second moments of attention values and projections display a trend opposite to keys and queries as the moments for these layers are more compressible along the $\text{fan}_{\text{out}}$ dimension as compared to the $\text{fan}_{\text{in}}$ dimension. For the attention projection layer, aversion to compression along the $\text{fan}_{\text{in}}$ dimension (where heads are stacked) is intuitive, as the parameters corresponding to each attention head are intended to be able to evolve independently throughout training. However, for the same reason, the higher compressibility of second moments in the value layer along the head-stacked dimension is unexpected. Intuitions aside, from an absolute magnitude perspective, values and projection layers show higher averaged SNR values along the preferred dimension than keys and queries, indicating greater overall compressibility for the value and projection moments. 

Interestingly, by a similar magnitude argument, the MLP second moments exhibit greater compressibility than attention keys and queries. While in general MLP second moments prefer compression along the output dimension ($\text{fan}_{\text{out}}$), for some layer indices the second moment can also be compressed along the input dimension ($\text{fan}_{\text{in}}$) or even both dimensions simultaneously.

LayerNorm components show different SNR trends depending on their position in the network. The SNR values of the attention LayerNorms and final LayerNorm typically exhibit a sharp decline after an initial increase, suggesting incompressibility. In contrast, MLP LayerNorms maintain consistently high SNR values throughout training, indicating their second moments can be effectively compressed.

We validate the robustness of these results in \Cref{appendix:pretraining} by observing similar trends in a larger model (GPT-medium) and on a different dataset (FineWeb-edu).

\subsubsection{Language Fine-tuning}
\label{section:finetuning}

Next, we extend our SNR analysis to examine second-moment compressibility during fine-tuning, using Llama-3.2 \cite{grattafiori2024llama3herdmodels} on the Alpaca dataset \cite{alpaca}.
\Cref{fig:snr-layer-llama-1b-alpaca} shows the SNR trends of selected layers, which reveal layer-wise patterns with subtle distinctions from those observed for GPT pre-training (for complete results, see \Cref{fig:snr-layer-llama-1b-alpaca-full}, \Cref{appendix:snr-finetuning}).

We find lower SNR values across all layers during fine-tuning, suggesting an aversion to compressibility in general in this experimental setting. This is particularly pronounced in the attention mechanism, where key and query second moments exhibit SNR values well below $1.0$. While attention value and projection second moments maintain an SNR value above $1.0$ along $\text{fan}_{\text{out}}$ dimension, these values are notably smaller than those observed during GPT pre-training.

MLP layers display variable compressibility patterns. The first two MLP layers (MLP.Up and MLP.Gate) show sporadic compressibility ($\text{SNR} \gtrsim 1$) at certain depths, but without consistently favoring either input or output dimension compression. In comparison, the output MLP layer (MLP.Down), consistently maintains a high SNR value ($\text{SNR} \gtrsim 1$) across depths, favoring compression along the $\text{fan}_{\text{out}}$ dimension.

Attention and MLP RMSNorms show consistently low SNR values across layers, while the final RMSNorm's SNR gradually increases during training, eventually exceeding $1.0$. The token embeddings show reduced SNR values even along the embedding dimension, possibly due to a larger vocabulary relative to the embedding dimension for the Llama model than the GPT-small model.

\subsubsection{ResNet Image Classification}
\label{section:resnet}

\begin{figure}[!hb]
\centering
\begin{minipage}[b]{0.235\textwidth}
    \centering
    \includegraphics[width=\textwidth]{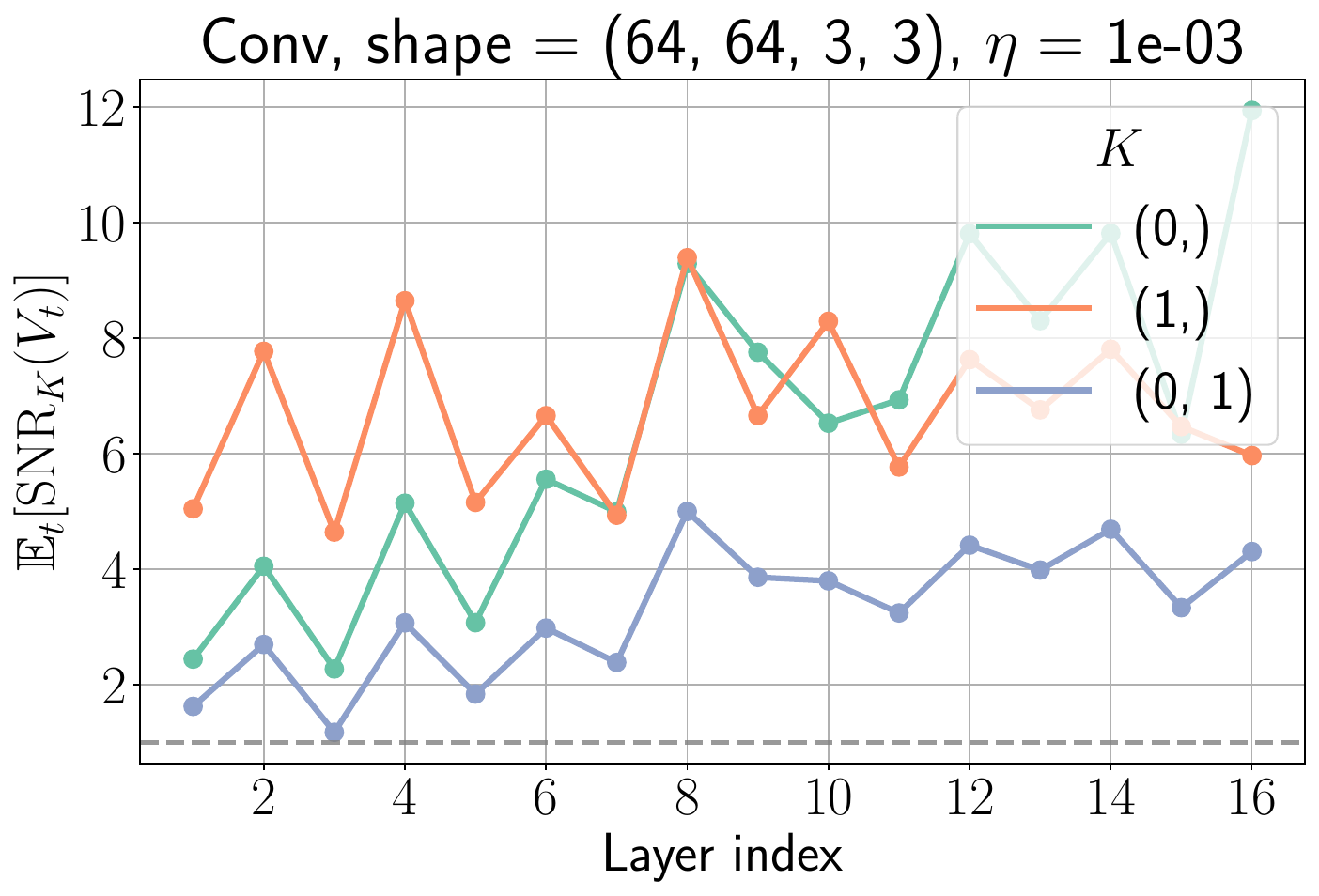}
\end{minipage}
\begin{minipage}[b]{0.235\textwidth}
    \centering
    \includegraphics[width=\textwidth]{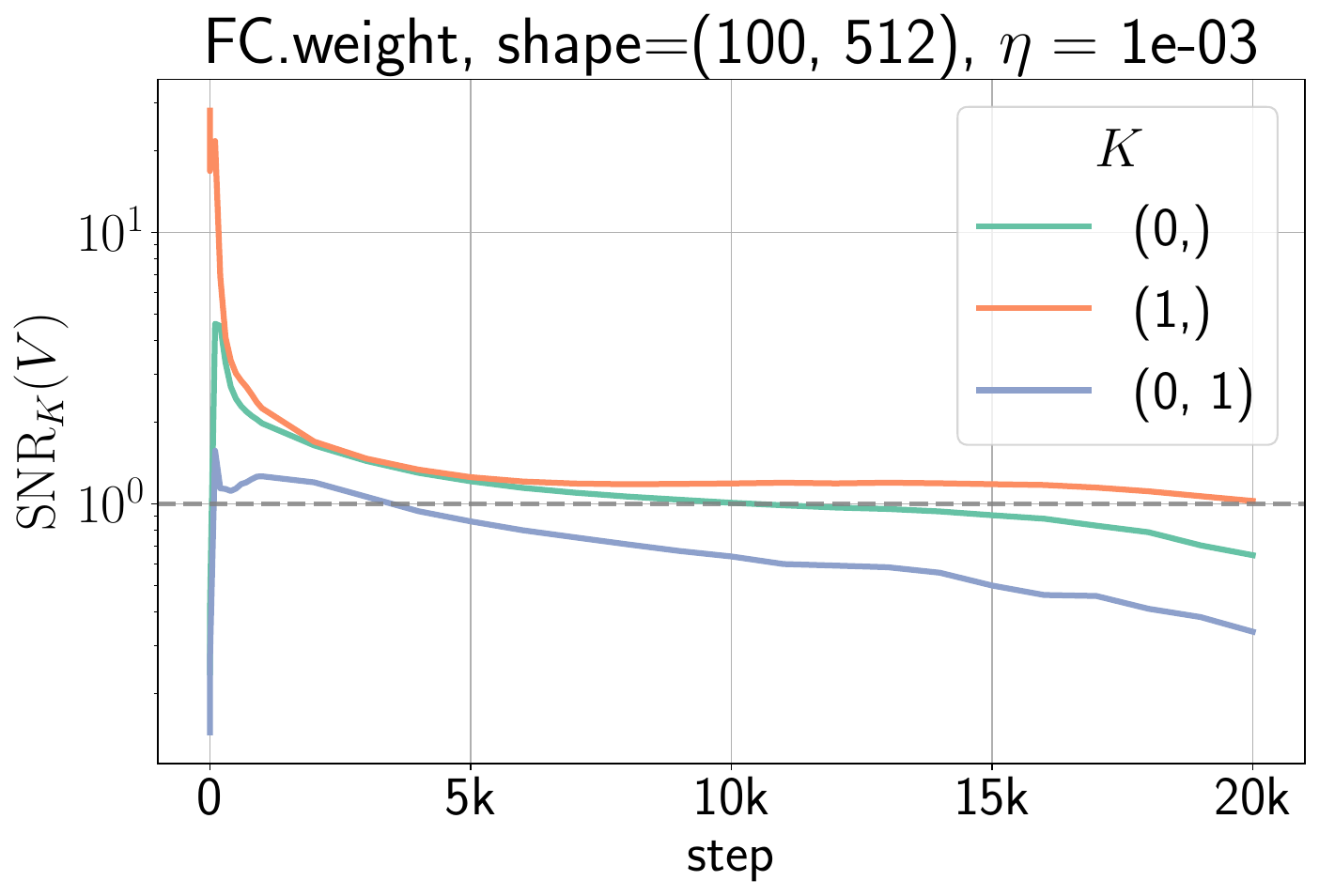}
\end{minipage}
\caption{SNR trends of ResNet-18 trained on CIFAR-100: (left) layer dependence of averaged SNR values on the intermediate convolutional layers, (right) SNR trajectories of the final layer.}
\label{fig:snr-resnet=cifar-100-main-text}
\end{figure}

\begin{figure*}[!htb]
\centering
\begin{minipage}[b]{0.245\textwidth}
    \centering
    \includegraphics[width=\textwidth]{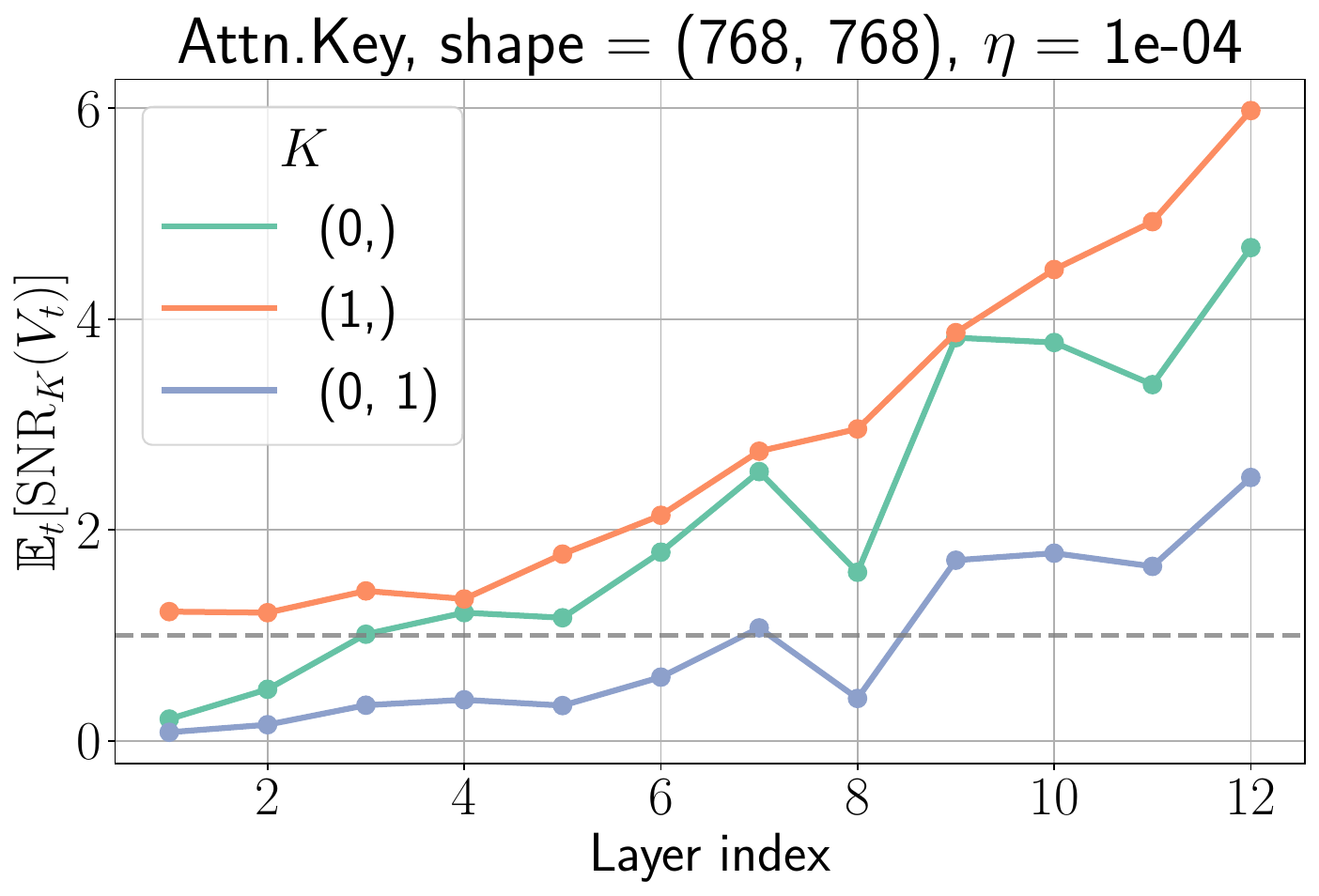}
\end{minipage}
\hfill
\begin{minipage}[b]{0.245\textwidth}
    \centering
    \includegraphics[width=\textwidth]{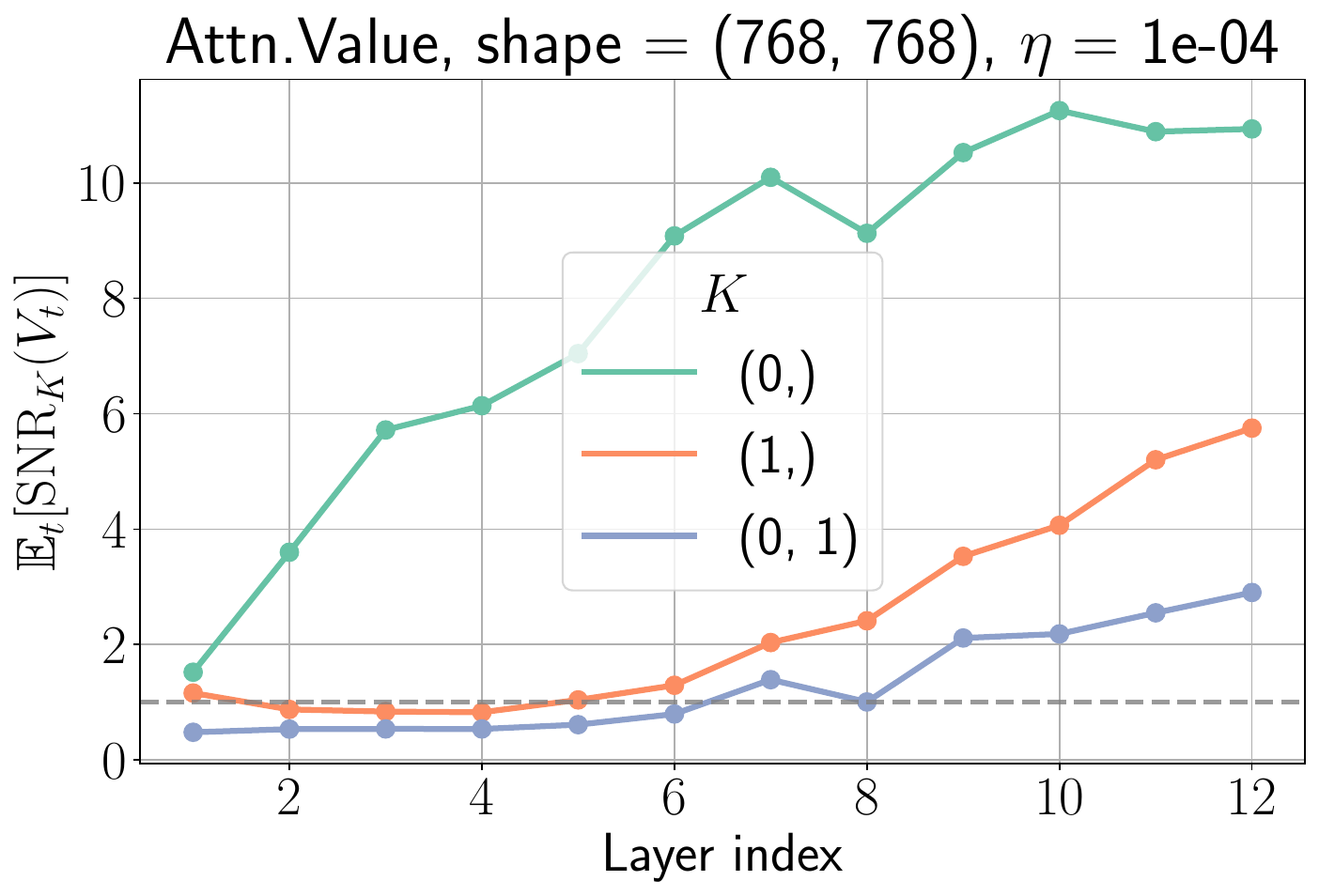}
\end{minipage}
\hfill
\begin{minipage}[b]{0.245\textwidth}
    \centering
    \includegraphics[width=\textwidth]{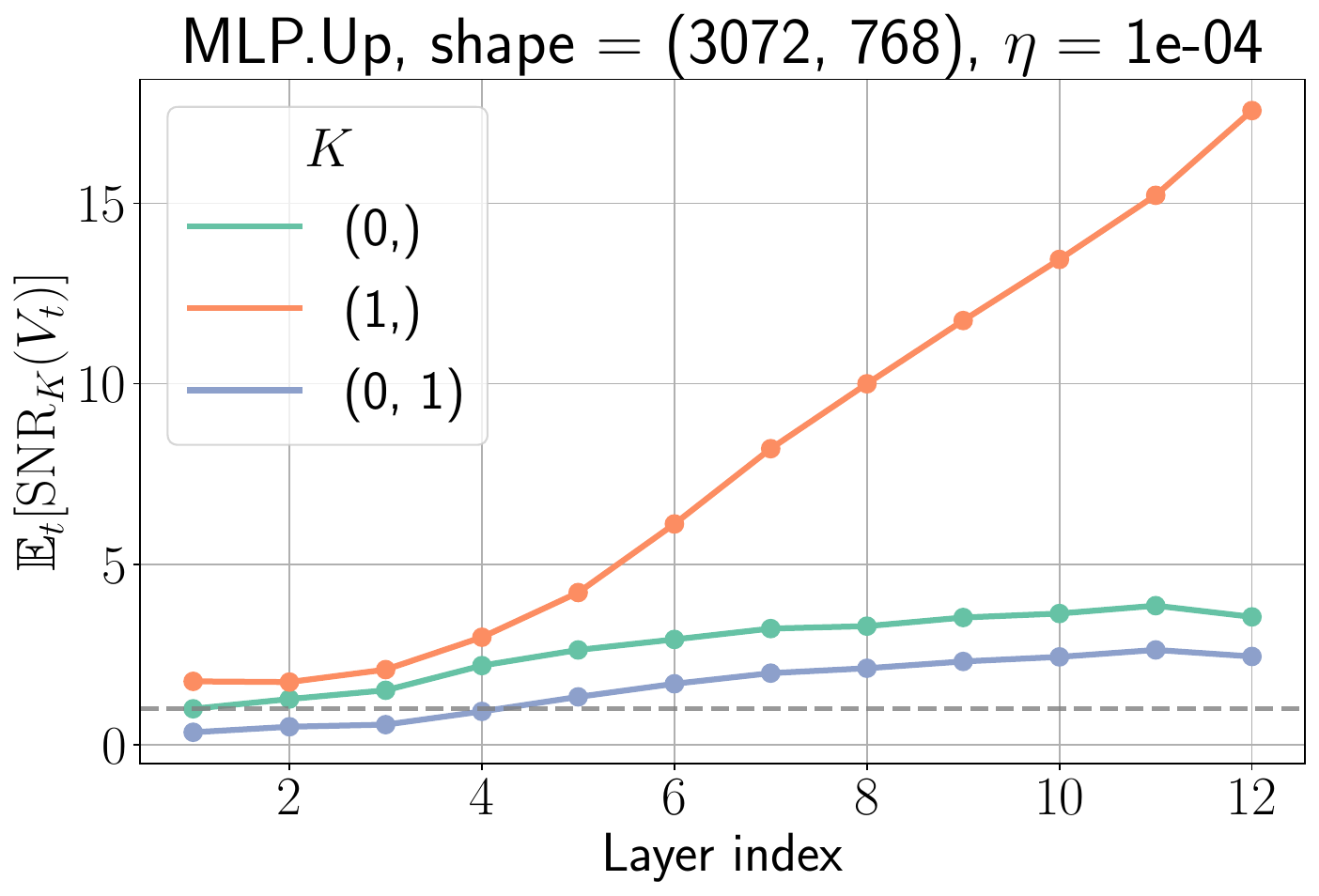}
\end{minipage}
\hfill
\begin{minipage}[b]{0.245\textwidth}
    \centering
    \includegraphics[width=\textwidth]{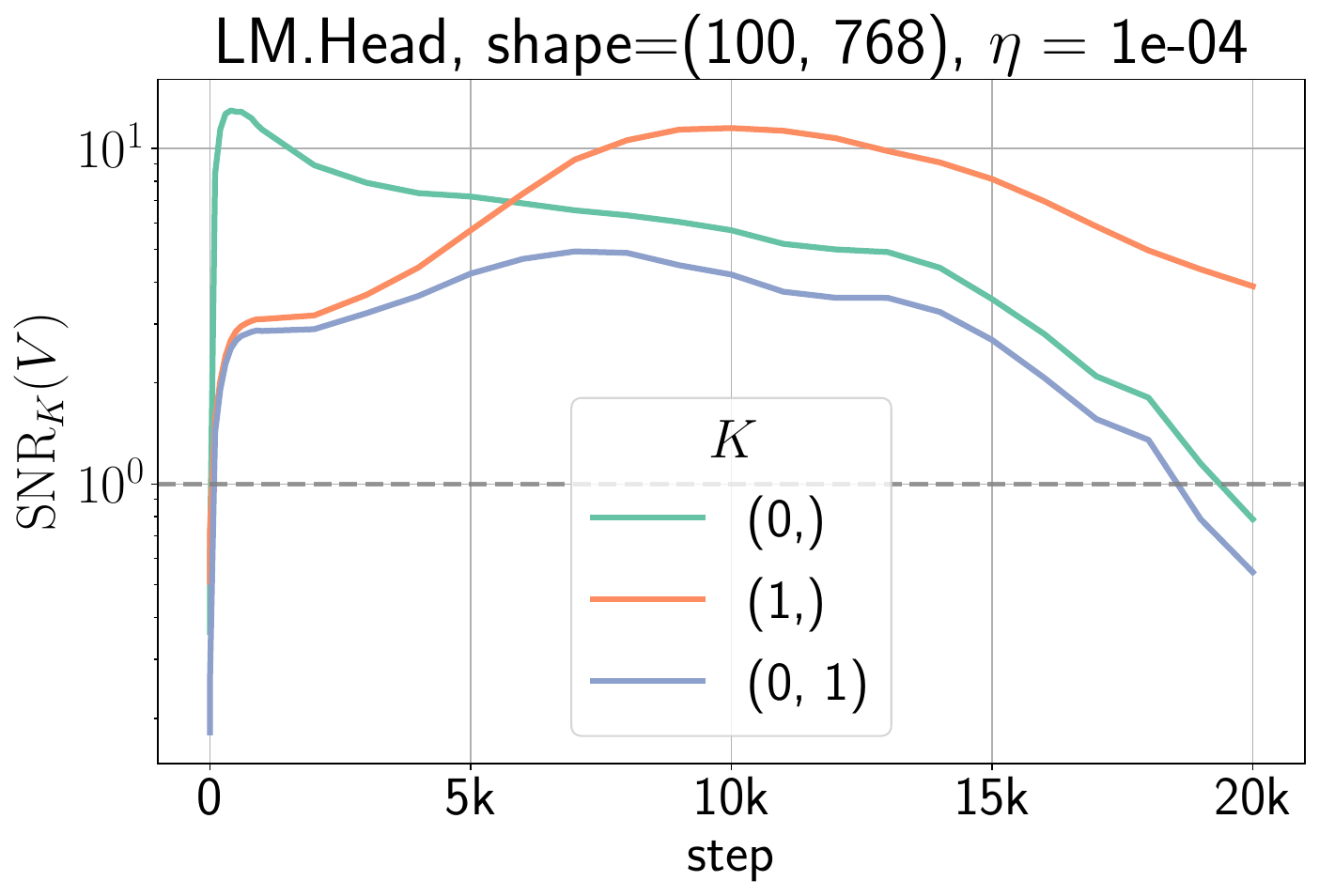}
\end{minipage}
\caption{SNR trends of selected layers of a $12$ layer ViT trained on CIFAR-100. For detailed results, see \Cref{fig:snr-vit-cifar-100-full}, \Cref{appendix:image-classification}.}
\label{fig:snr-vit-cifar-100-maintext}
\end{figure*}

Compared to language pre-training and fine-tuning settings, the second moments of ResNets trained on CIFAR-10 and CIFAR-100 (\Cref{fig:snr-resnet=cifar-100-main-text} and \Cref{appendix:image-classification}) exhibit high SNR values.
These SNR values suggest high second-moment compression feasibility across layers. 
In particular, the intermediate convolutional layers show exceptionally high SNR values across both $\text{fan}_{\text{in}}$ and $\text{fan}_{\text{out}}$ dimensions, with an increasing trend as a function of depth.
By comparison, the first and last layers behave differently. The first convolutional layer resists compression along the $\text{fan}_{\text{out}}$ dimension (shown in \Cref{fig:snr-resnet-cifar-100-full}, \Cref{appendix:image-classification}), while the final layer exhibits SNR values close to $1.0$ that decreases late in training. These results align with \cite{Hao17-loss_lanscape}, who demonstrated that ResNets exhibit remarkably smooth optimization landscapes.

\subsubsection{ViT Image Classification}
\label{section:vit}
\looseness -1
Next, we analyze Vision Transformers (ViTs) \cite{dosovitskiy2021an}, with GPT-2 Transformer adapted for image classification.
\Cref{fig:snr-vit-cifar-100-maintext} shows that ViTs trained on CIFAR-10 and CIFAR-100 exhibit SNR trends combining characteristics from both ResNet and GPT pre-training.

The attention moments maintain GPT-like SNR trends but with higher SNR values. The keys and query second moments favor $\text{fan}_{\text{in}}$ compression, while values and projections prefer $\text{fan}_{\text{out}}$ dimension. These attention components exhibit higher SNR values than GPT pre-training, with the averaged SNR increasing with depth for most layers.

Unlike GPT pre-training, the first MLP layer (MLP.Up) favors $\text{fan}_{\text{in}}$ compression instead of $\text{fan}_{\text{out}}$. This suggests that this layer type's compression behavior is training regime-dependent. 
By comparison, the second layer (MLP.Down) maintains GPT-like $\text{fan}_{\text{out}}$ preference and exhibits high SNR values along both dimensions. 

\looseness -1
Similar to ResNet's first convolution layer, ViT's patch embedding layer favors $\text{fan}_{\text{in}}$ compression. Meanwhile, the classification layer maintains SNR values close to $1.0$ without consistent preference toward a particular compression dimension. Notably, all LayerNorm components display surprisingly high SNR values, suggesting high compressibility.

\subsection{Compressibility Trends Across Training Regimes} 
\label{section:trends-across-regimes}

The SNR analysis in the previous section revealed several consistent compressibility trends and some regime-specific behaviors. Below, we summarize these findings.

\textbf{Attention:} The attention second moments exhibit consistent preferred compression dimensions, but with varying compressibility strengths across training regimes.
Key and query second moments consistently favor compression along $\text{fan}_{\text{in}}$ dimension while showing aversion to compression along $\text{fan}_{\text{out}}$ (head-stacked) dimension.
Values and projections display an opposite trend, favoring compressibility along $\text{fan}_{\text{out}}$ dimension. Value and projection layers generally exhibit higher SNR values than key and query layers, suggesting higher compressibility.
These trends persist across training regimes (GPT pre-training, Llama fine-tuning, and ViT image classification), suggesting these trends are intrinsic to the attention mechanism. However, the compressibility strength varies across training regimes, with ViT showing overall higher SNR values than GPT pre-training and fine-tuning exhibiting notably lower SNR values.

\textbf{MLPs:} 
Our GPT and ViT models share identical MLP blocks with two layers (MLP.Up and MLP.Down). The first layer shows task-dependent trends, with $\text{fan}_{\text{out}}$ preferred in the language pre-training and $\text{fan}_{\text{in}}$ favored in ViT image classification. The second layer (MLP.Down), consistently prefers $\text{fan}_{\text{out}}$ compression across both settings.
The pre-trained Llama model uses three layers in the MLP block (Up, Down, Gate). The first two layers (Up, Gate) show inconsistent compressibility trends, whereas the output layer (Down) favors $\text{fan}_{\text{out}}$ compression similar to the GPT setting.

\textbf{First and Last layer:} 
In language models, Token Embedding and LM Head show a strong aversion to compression along the token dimension, while allowing compression along the embedding dimension.
In image classification, the first layers exhibit a strong preference for $\text{fan}_{\text{in}}$ compression, while classification heads show inconsistent compression trends but maintain overall higher SNR values. 
Overall, image classification models exhibit substantially higher compressibility than language models.

\textbf{Normalization layers:} Normalization layers show domain-specific compressibility trends. Language models exhibit lower LayerNorm compressibility, while both BatchNorm and LayerNorm in vision models maintain higher compressibility throughout training.

\section{Factors Influencing Compressibility}
\label{section:understanding-compression-trends}
Our earlier analysis revealed various consistent SNR trends across training regimes. Here, we conduct experiments to analyze the effect of initialization, dataset properties, and optimization dynamics on these trends.

\subsection{Incompressibility under Heavy-Tailed Distributions}
\label{section:heavy-tailed-token-distribution}

\begin{figure}[!htb]
\centering
\begin{minipage}[b]{0.235\textwidth}
    \centering
    \includegraphics[width=\textwidth]{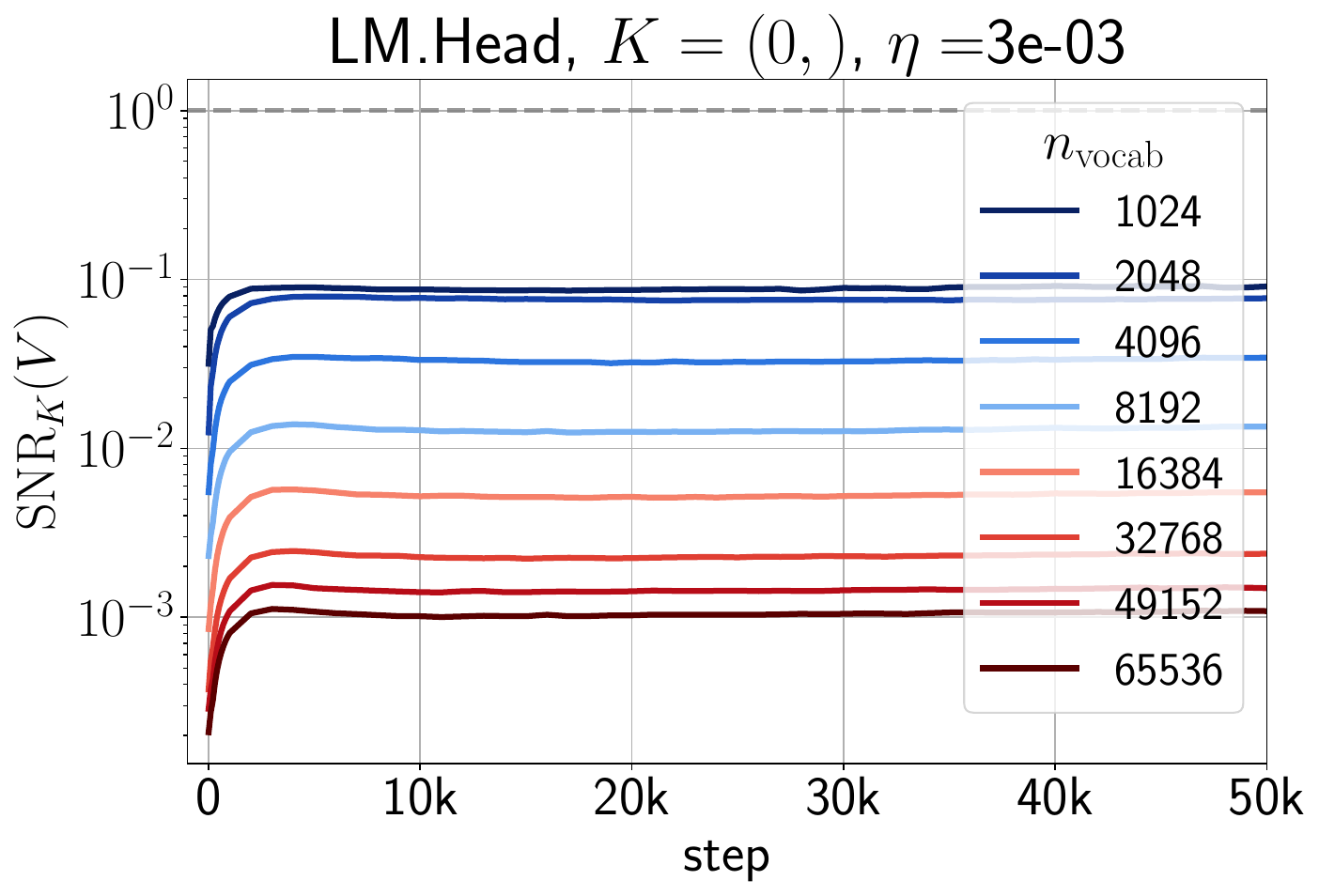}
\end{minipage}
\begin{minipage}[b]{0.235\textwidth}
    \centering
    \includegraphics[width=\textwidth]{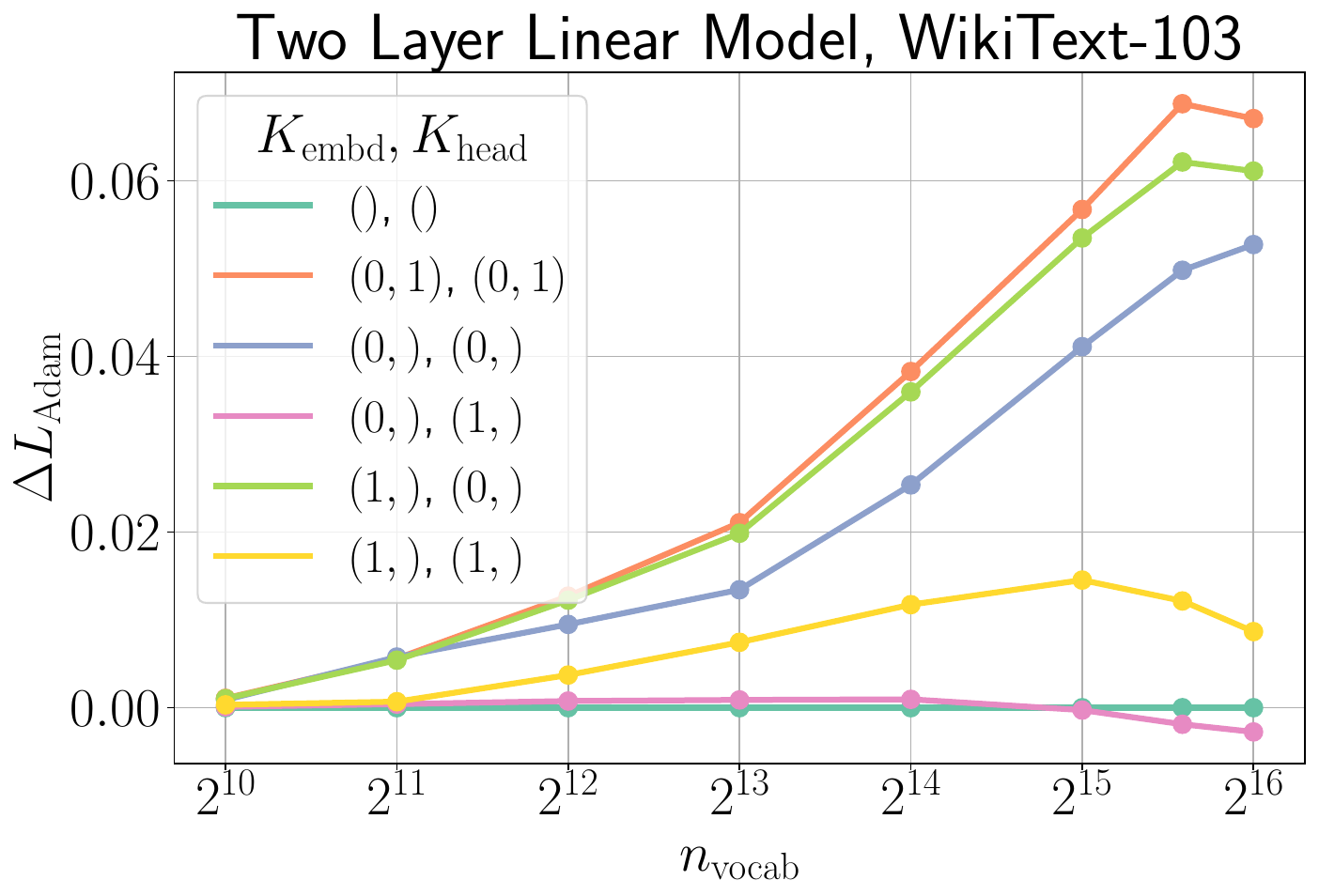}
\end{minipage}

\caption{(left) SNR trajectories of the linear head of the simplified two-layer model with varying vocabulary sizes. (right) Test loss gap $\Delta L_{\text{Adam}} = L_{(K_{\text{embd}}, K_{\text{head}})} - L_{\text{Adam}}$ of the linear model trained with Adam with shared second moments across dimensions $(K_{\text{embd}}, K_{\text{head}})$.}
\label{fig:vocab-experiment}
\end{figure}

In the previous section, we observed that language models strongly averse compression along the token dimension in the first and last layers. This resistance suggests that individual tokens require their own learning rates, as their gradients evolve at different paces. To better understand this phenomenon, we investigate how token frequency distribution influences compressibility.

We examine a simplified two-layer model, solely consisting of a token embedding matrix and a linear head. We train the model on the WikiText-103 dataset \cite{merity2017pointer} tokenized using BPE tokenizer \cite{BPE-Gage} with varying vocabulary sizes. By progressively reducing the vocabulary size, we systematically remove rare tokens to control the tail of the token distribution.
\Cref{fig:vocab-experiment} (left) shows that SNR values along the token dimension of the linear head decrease substantially at larger vocabularies, with similar trends observed for the token embedding matrix (see \Cref{appendix:vocab-experiment} for full results). This indicates that compression becomes increasingly challenging as vocabulary grows. 

We then analyze how vocabulary size affects model performance when trained with Adam with shared second moments (introduced in \Cref{equation:adashare}, \Cref{section:preliminaries}) along dimensions $(K_{\text{embd}}, K_{\text{head}})$.
\Cref{fig:vocab-experiment} (right) shows the loss gap between the above optimizer and standard Adam, defined as $\Delta L_{\text{Adam}} = L_{(K_{\text{embd}}, K_{\text{head}})} - L_{\text{Adam}}$. For large vocabularies, compression is only effective along embedding dimensions, while token-dimension compression degrades performance. In contrast, small vocabularies permit compression along both dimensions.

These findings extend the work of  \cite{kunstner2024heavytailedclassimbalanceadam}, which showed that Adam outperforms SGD on language tasks by making faster progress on rare tokens. Our analysis suggests that the apparent advantage of Adam in optimizing language models might stem in large part from the requirement that the Token Embedding and LM Head layers are allowed independent learning rates for each token in its vocabulary.

\subsection{Large Learning Rates reduce Compressibility}
\label{section:large-learning-rates}

\begin{figure}[!htb]
\centering

\begin{minipage}[b]{0.235\textwidth}
    \centering
    \includegraphics[width=\textwidth]{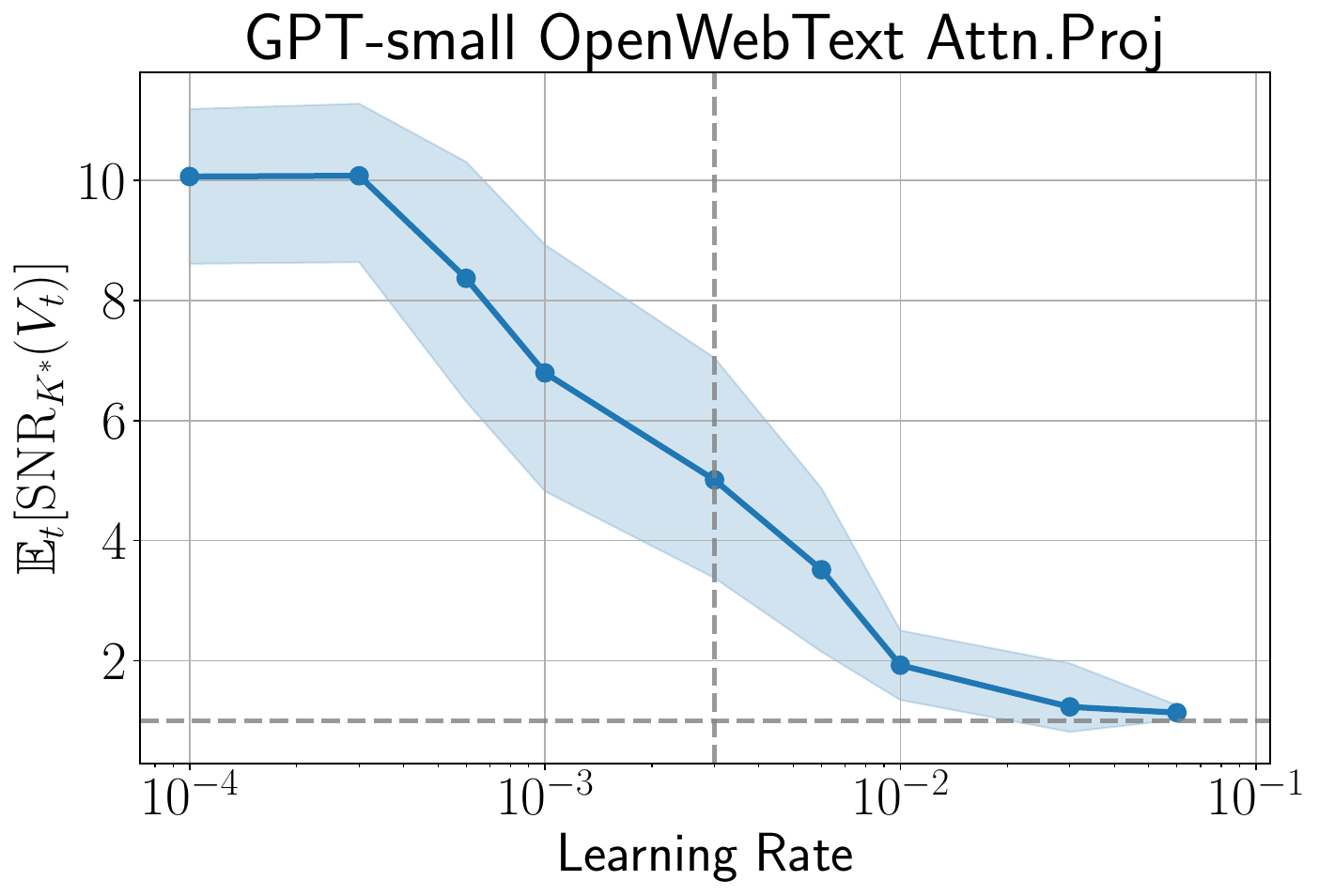}
\end{minipage}
\begin{minipage}[b]{0.235\textwidth}
    \centering
    \includegraphics[width=\textwidth]{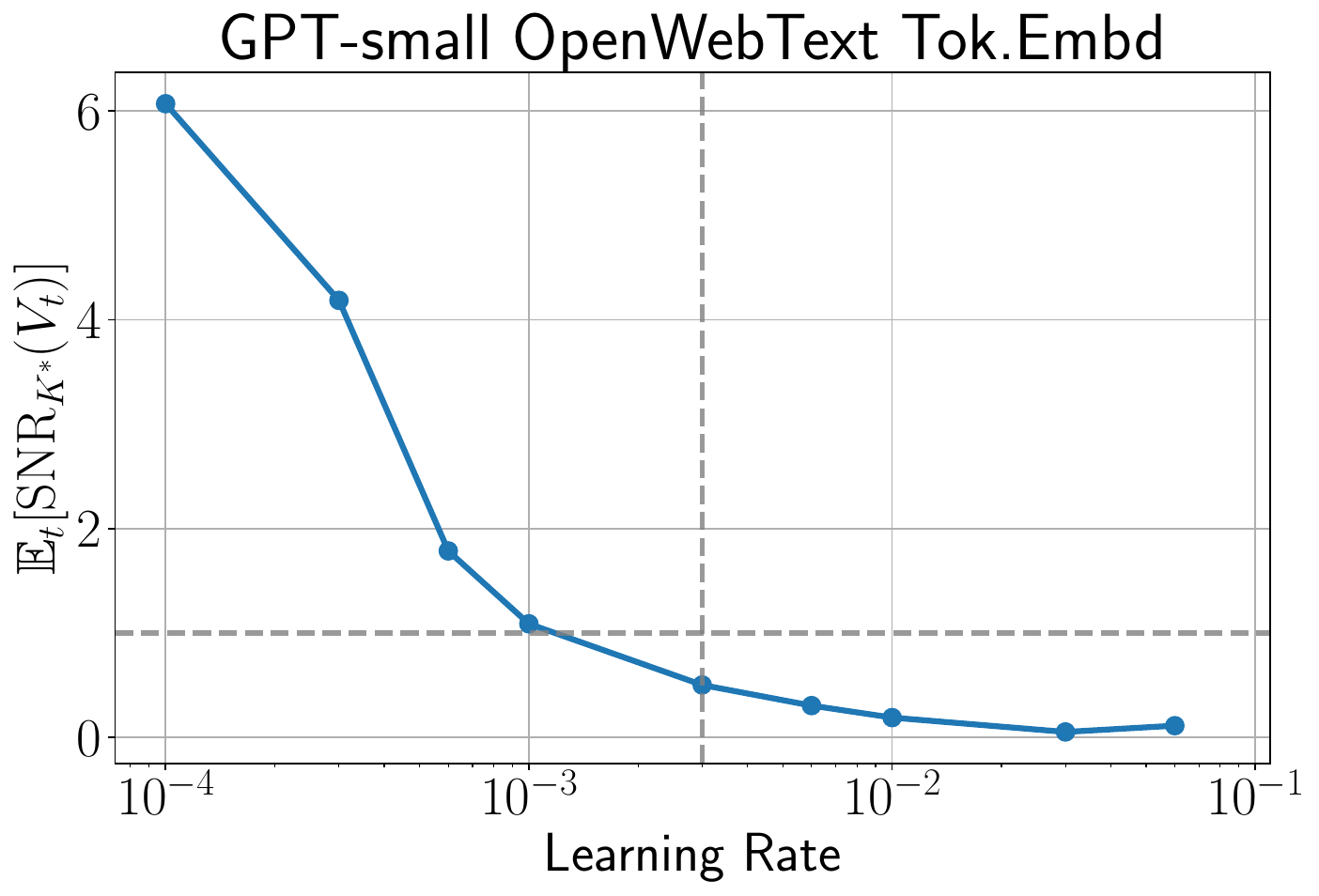}
\end{minipage}

\caption{The effect of learning rate on the averaged SNR values of selected layer types of a GPT-small model trained on OpenWebText. For each layer type, we select the compression dimension $K^*$ with the highest SNR. The shaded region around the mean trend shows the variation across depth.}
\label{fig:snr-trajectories-lr-gpt-small-openwebtext}
\end{figure}

In this section, we analyze how increasing the learning rate affects averaged SNR values and thereby compression feasibility. \Cref{fig:snr-trajectories-lr-gpt-small-openwebtext} demonstrates that increasing the learning rate consistently reduces SNR values across layers (see \Cref{appendix:large-learning-rates} for full results). For clarity, we focus on the preferred SNR compression dimension for each layer type. 
This decline in averaged SNR values suggests that higher learning rates cause training to explore regions of parameter space where the gradient distribution contains more outliers, thereby reducing compression feasibility.
Based on the effect of increasing the learning rate on SNR values, we classify layer types into two categories:

1. \emph{Layers that are compression-\textbf{averse} (SNR $\lesssim 1$) at the optimal learning rate:} Token Embedding/LM Head, LayerNorm, Attention keys, queries, first MLP layer (MLP.Up).

2.  \emph{Layers that are \textbf{amenable} to compression (SNR $\gtrsim 1$) at the optimal learning rate:} Attention values and projections and the second MLP layer (MLP.Down).

We observe similar trends for pre-trained Llama and ViT models, while ResNets remain compressible even at very high learning rates. In \Cref{section:slimadam}, we will quantify these architectural differences in compression feasibility.

\subsection{Effect of Initialization on Compressibility}
\label{section:initialization}

\begin{figure}[!ht]

    \centering
    \begin{minipage}[b]{0.235\textwidth}
    \centering
    \includegraphics[width=\textwidth]{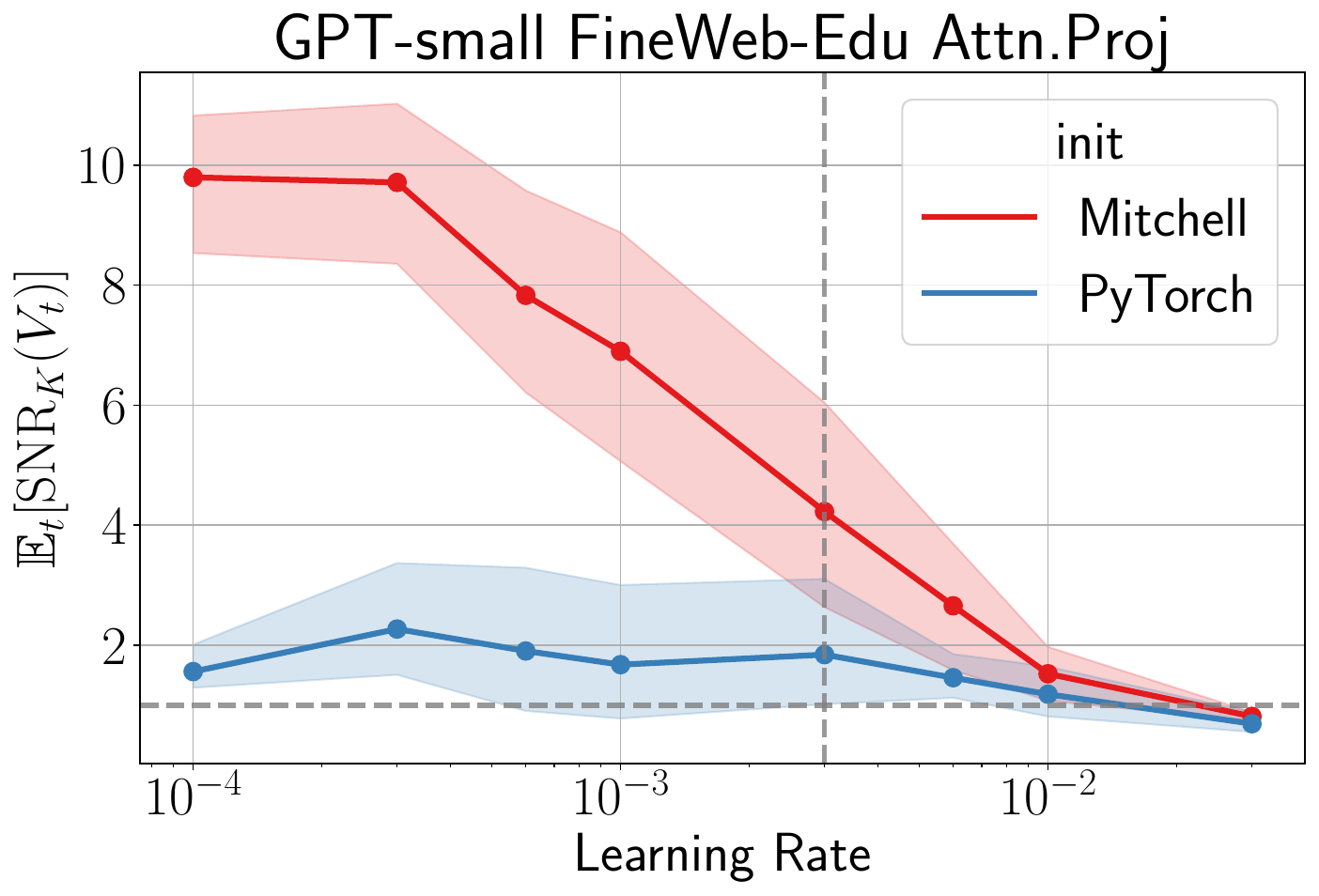}
\end{minipage}
\hfill
\begin{minipage}[b]{0.235\textwidth}
    \centering
    \includegraphics[width=\textwidth]{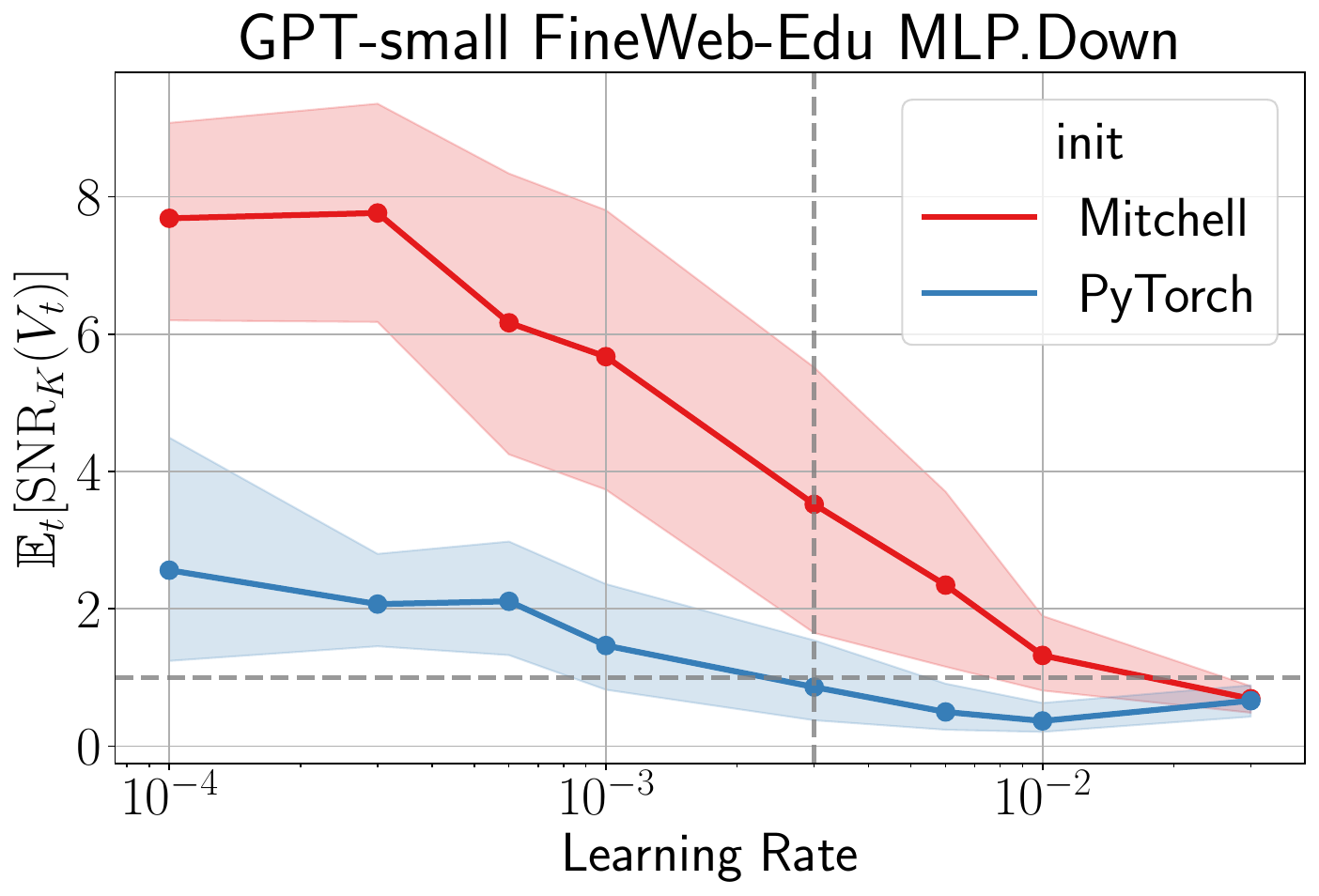}
\end{minipage}
\caption{The effect of initialization on SNR trends of GPT-small trained on FineWeb-Edu. For each layer type, we select the compression dimension $K^*$ with the highest SNR. The shaded region around the mean trend shows the variation across depth.}
\label{fig:snr-gpt-pytorch-init}
\end{figure}

\begin{figure*}[!htb]
\centering
\begin{minipage}[b]{0.245\textwidth}
    \centering
    \includegraphics[width=\textwidth]{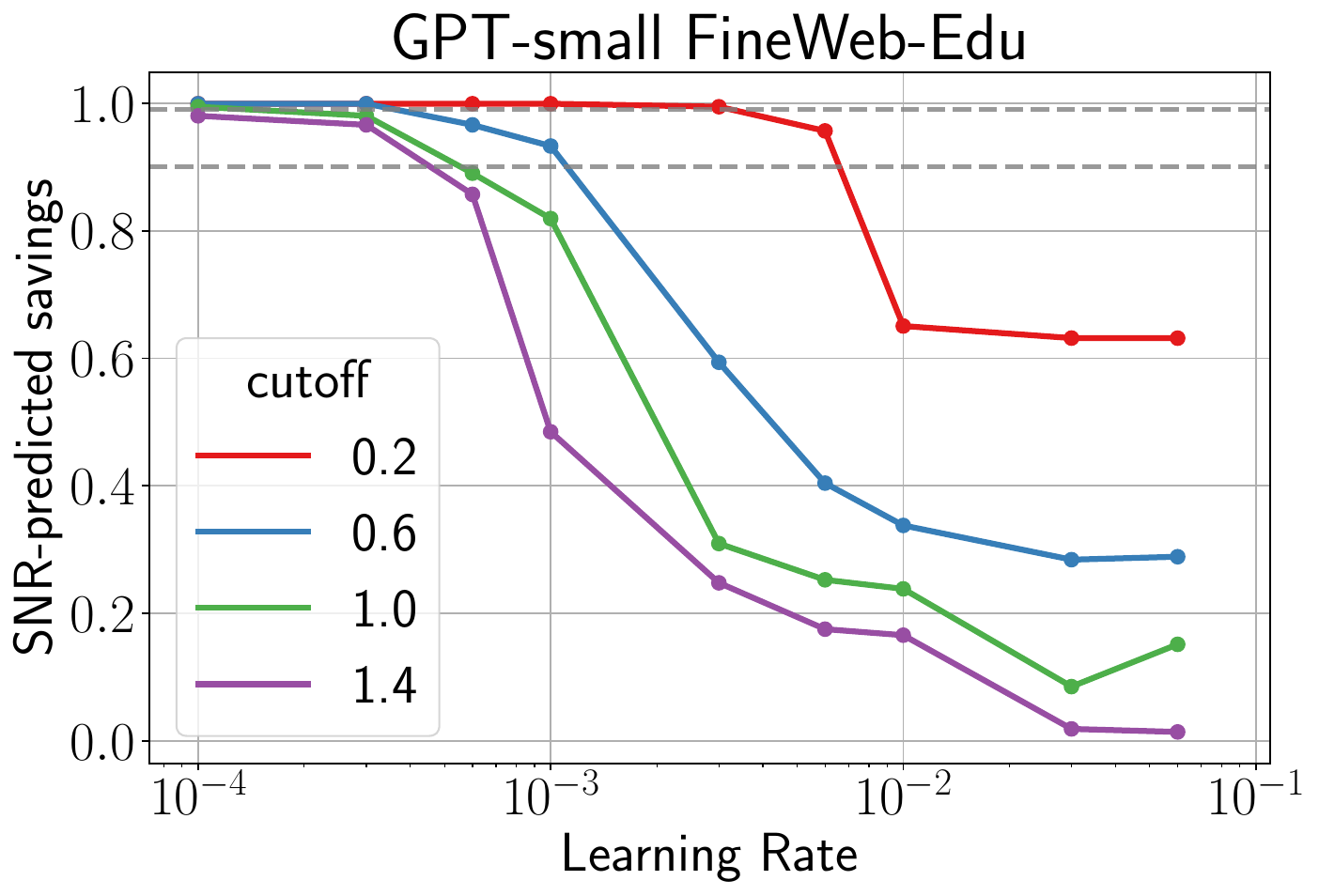}
\end{minipage}
\hfill
\begin{minipage}[b]{0.245\textwidth}
    \centering
    \includegraphics[width=\textwidth]{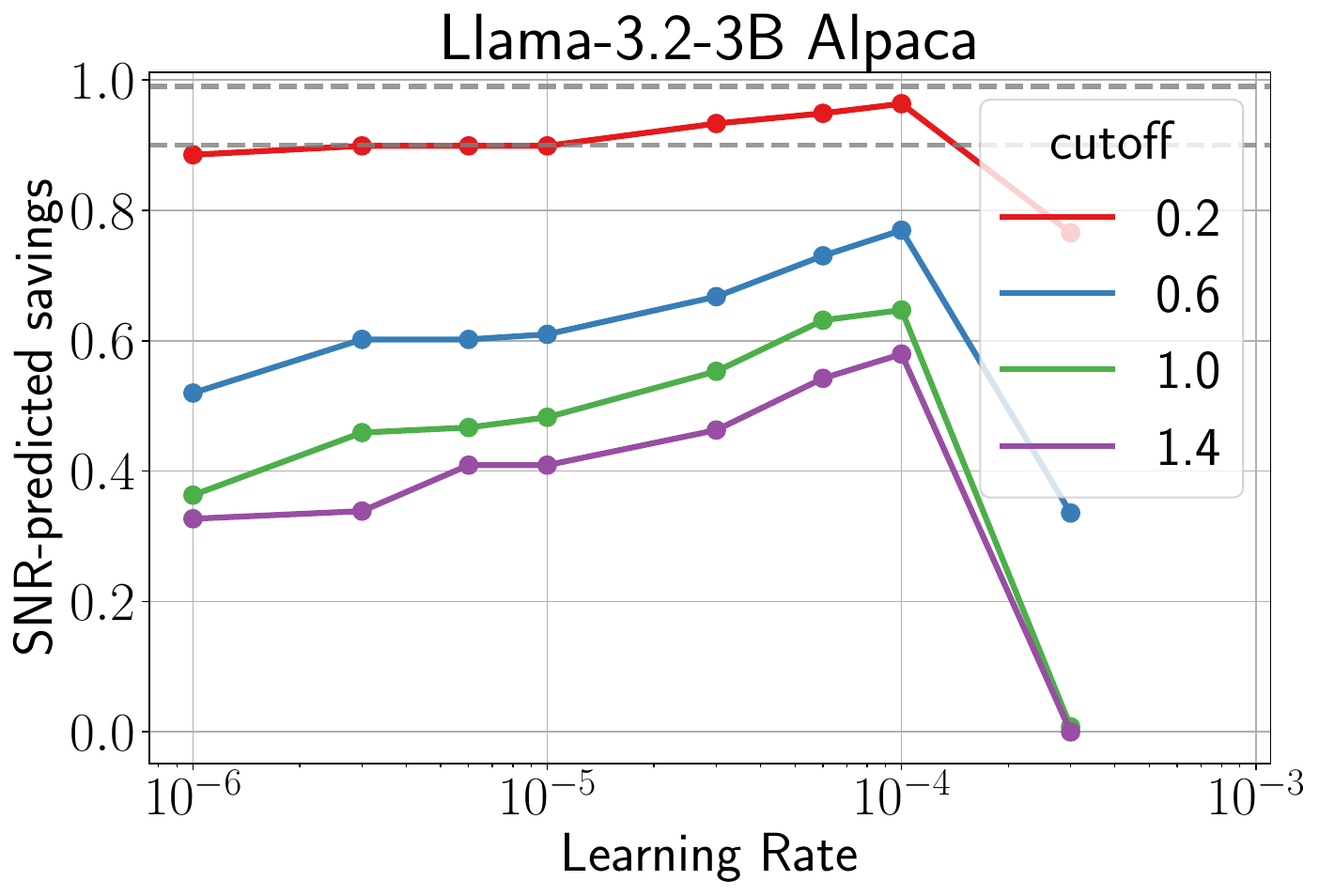}
\end{minipage}
\hfill
\begin{minipage}[b]{0.245\textwidth}
    \centering
    \includegraphics[width=\textwidth]{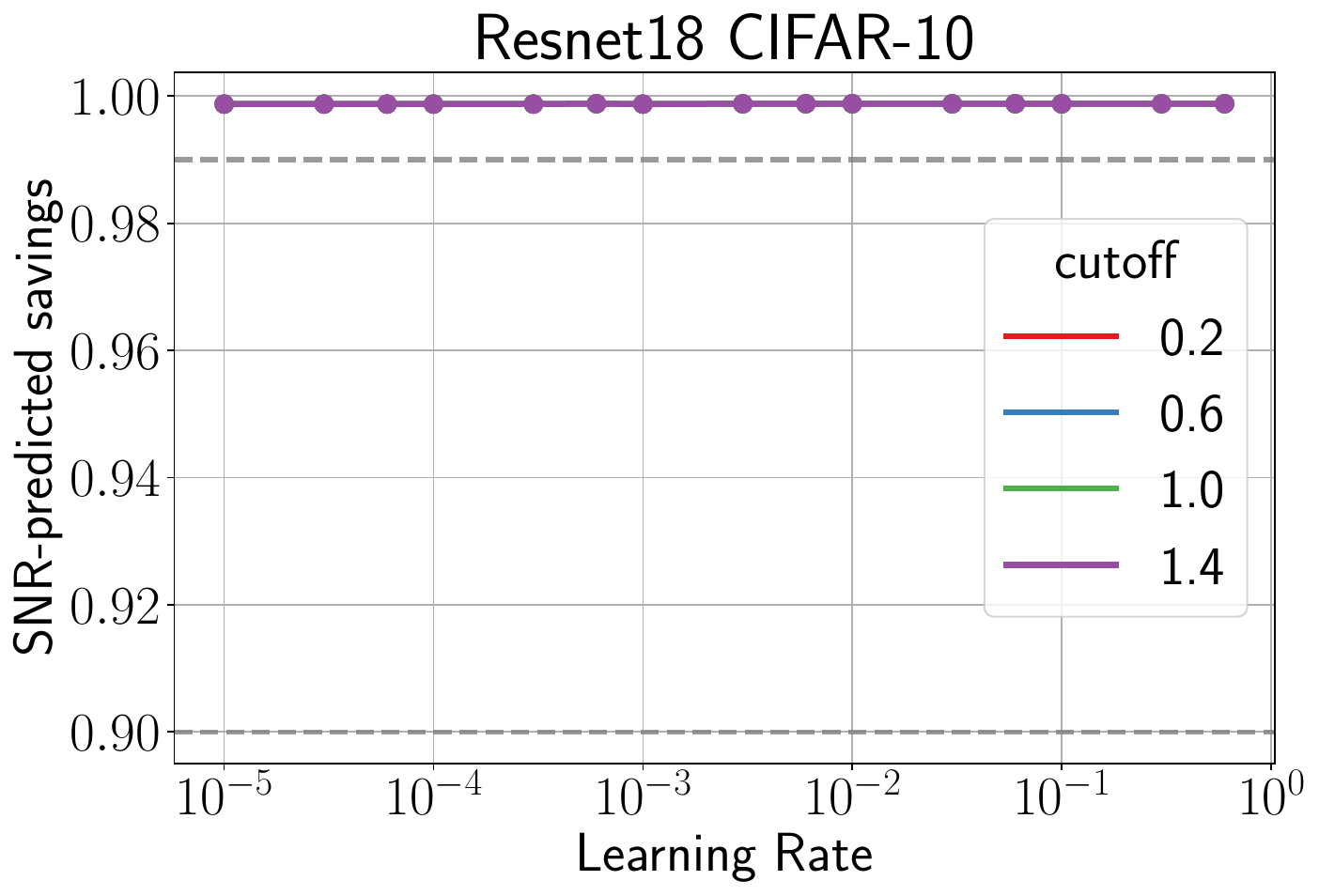}
\end{minipage}
\hfill
\begin{minipage}[b]{0.245\textwidth}
    \centering
    \includegraphics[width=\textwidth]{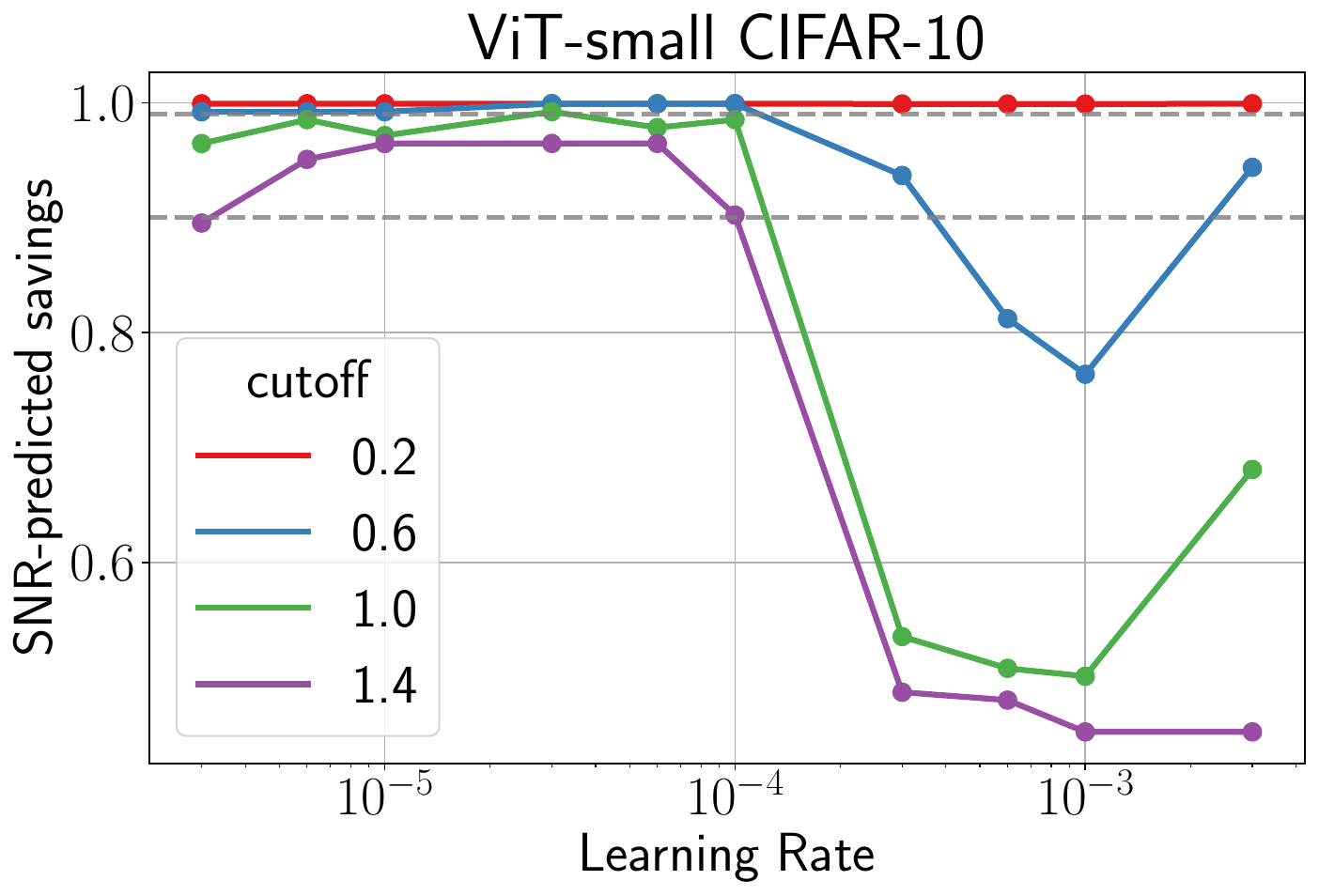}
\end{minipage}

\begin{minipage}[b]{0.245\textwidth}
    \centering
    \includegraphics[width=\textwidth]{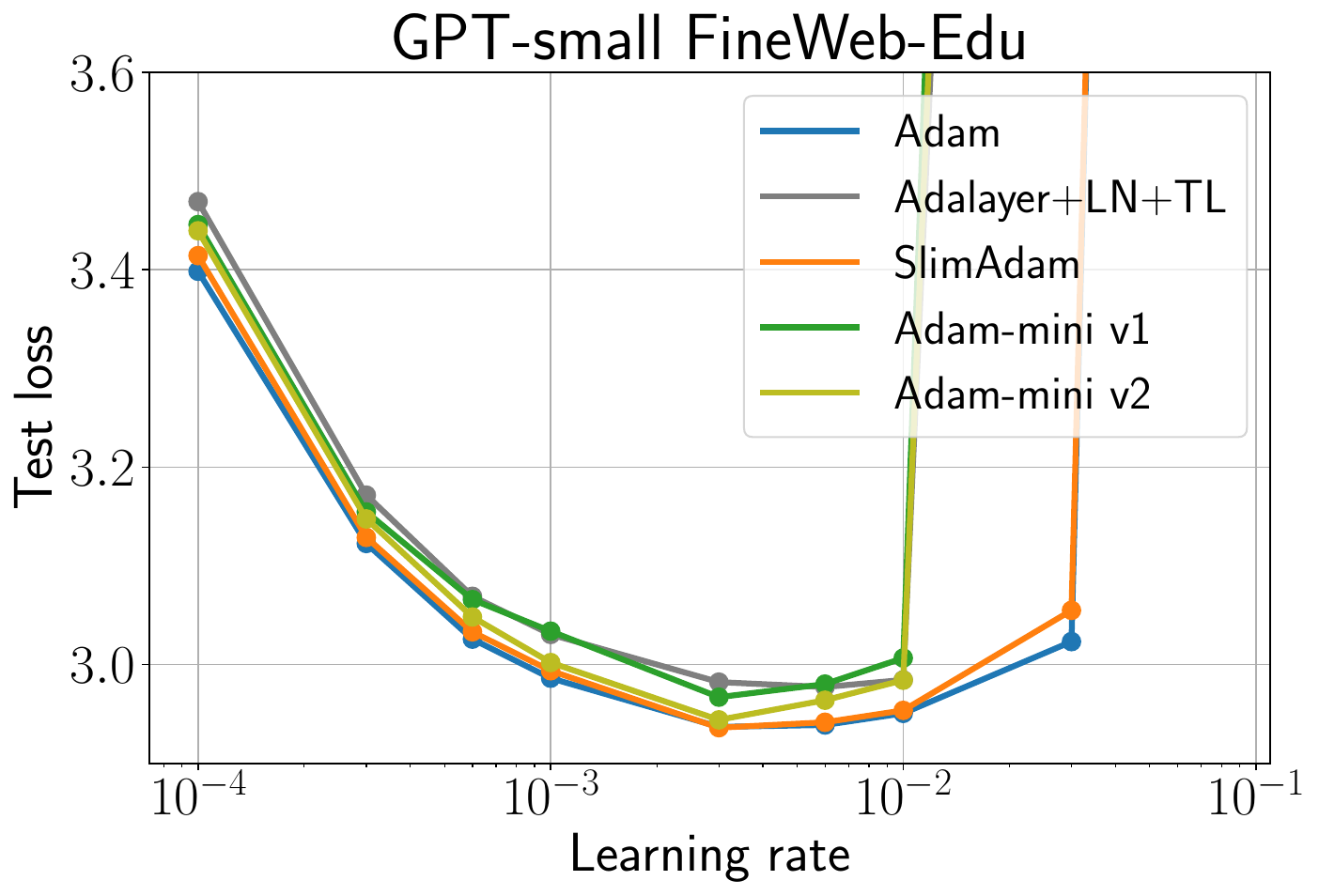}
\end{minipage}
\hfill
\begin{minipage}[b]{0.245\textwidth}
    \centering
    \includegraphics[width=\textwidth]{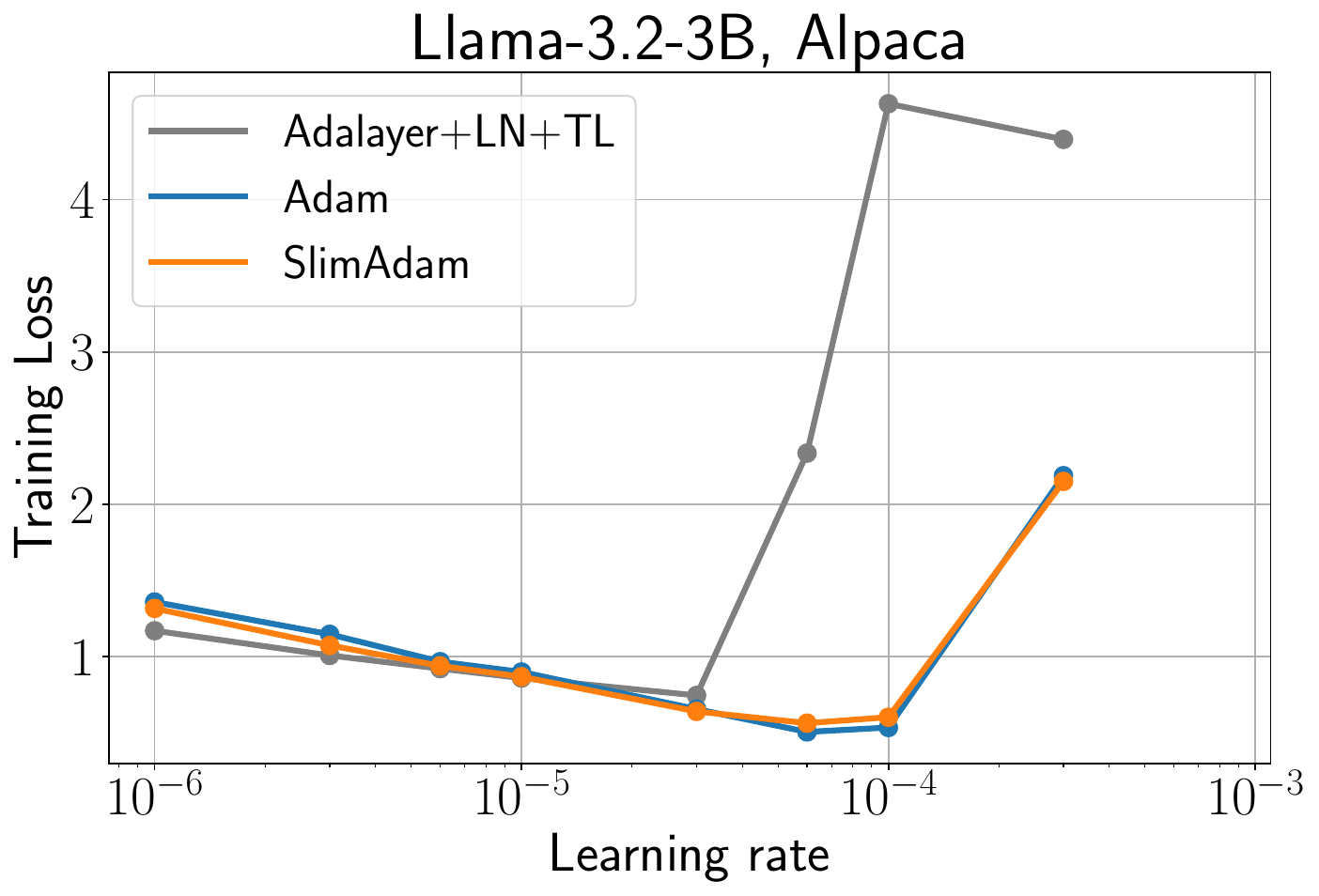}
\end{minipage}
\hfill
\begin{minipage}[b]{0.245\textwidth}
    \centering
    \includegraphics[width=\textwidth]{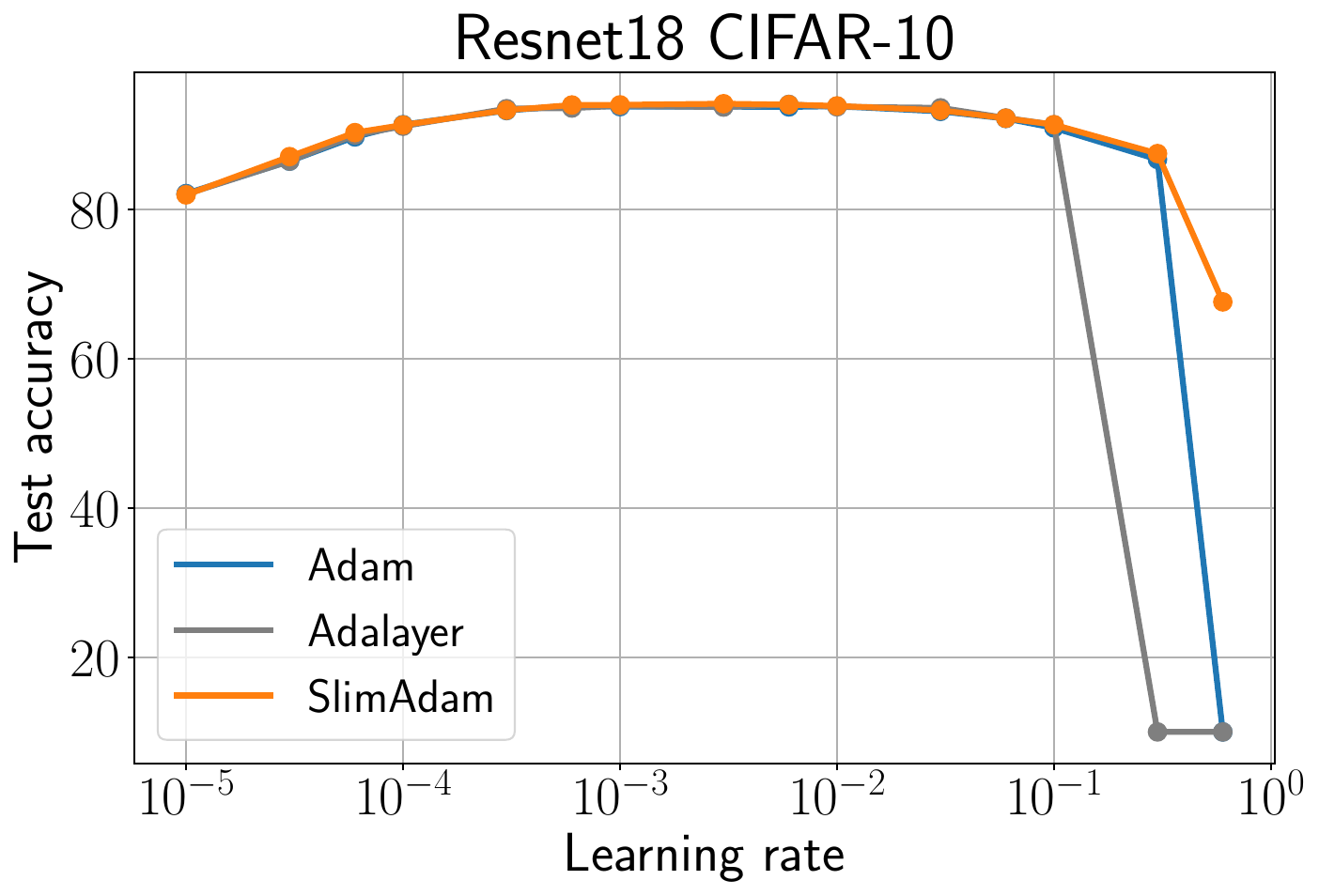}
\end{minipage}
\hfill
\begin{minipage}[b]{0.245\textwidth}
    \centering
    \includegraphics[width=\textwidth]{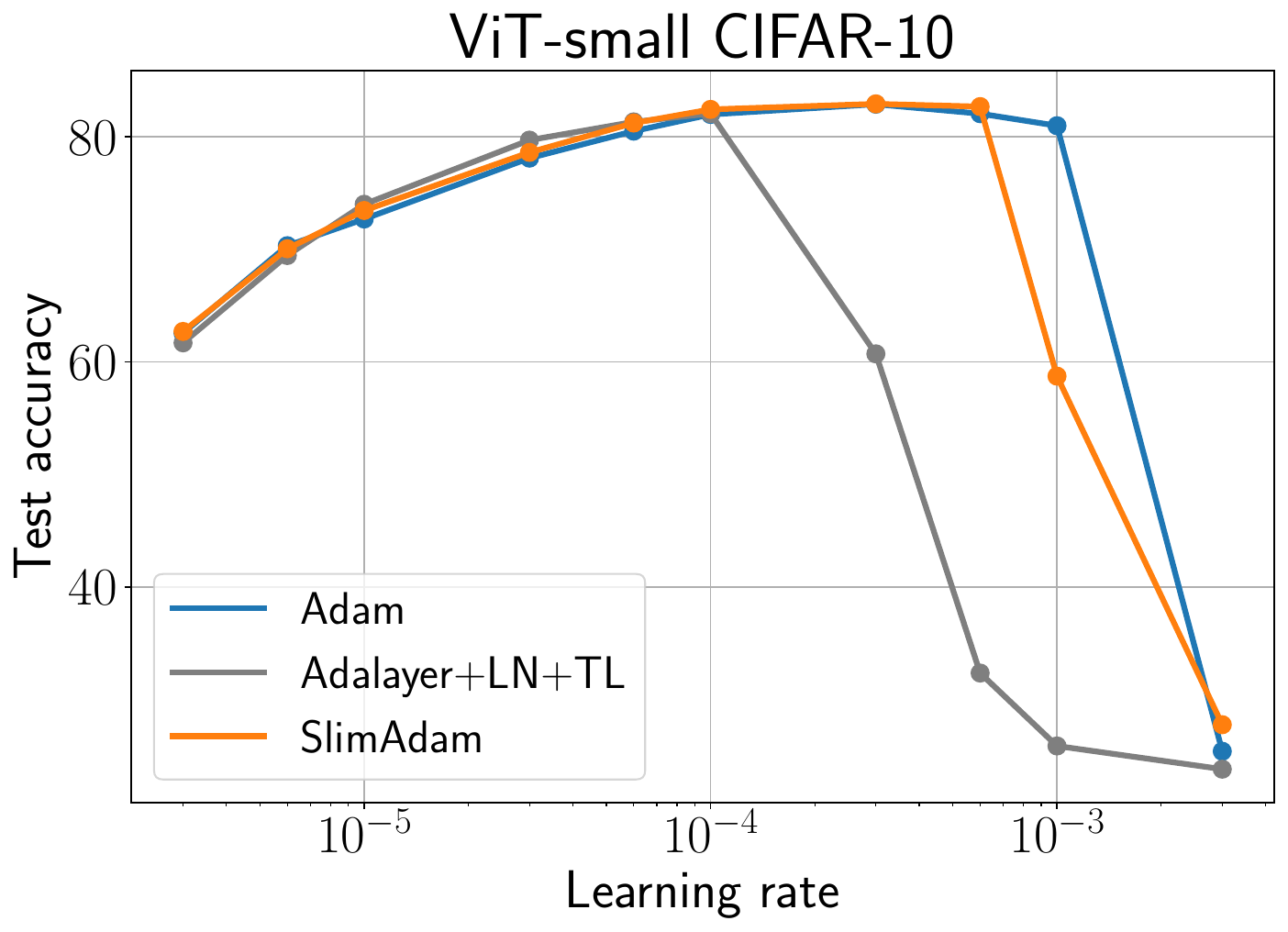}
\end{minipage}

\caption{(Top) Fraction of second moments potentially reducible (relative to Adam) as a function of learning rate and SNR cutoff across training configuration, as predicted by SNR analysis.
(Bottom) Performance comparison across learning rates between SlimAdam (with rules derived at learning rate $\eta = \text{3e-04}$) and baselines: Adam, AdaLayer (one second moment per block), AdaLayer+LN+TL (AdaLayer with uncompressed LayerNorm and LM head) \cite{zhao2024deconstructingmakesgoodoptimizer}, and Adam-mini versions v1 and v2 \cite{zhang2024adamminiusefewerlearning}. SlimAdam achieves Adam-level performance and stability while significantly reducing memory usage across all configurations. In \Cref{appendix:related-works}, we provide details about other optimizers. 
}
\label{fig:slim-memory-performance}
\vspace{-0.1in}
\end{figure*}

In this section, we examine the effect of initialization schemes on SNR trends and compressibility. While our earlier experiments in \Cref{section:snr-analysis} showed robust SNR patterns across model scales and datasets, we show that initialization significantly affects these trends. We compare Mitchell initialization\footnote{Mitchell initialization is implemented in the OLMo training code to achieve hyperparameter transfer across model scales, but followup work shows it may cause instability later in training \cite{olmo20242}.} \cite{groeneveld-etal-2024-olmo} used in \Cref{section:snr-analysis} against PyTorch's default initialization scheme.  A key feature of Mitchell initialization is that it scales the variance of layers that add to the residual stream (Attn.Proj and MLP.Down) with a factor of $1/$depth. 

\looseness -1
\Cref{fig:snr-gpt-pytorch-init} and \Cref{fig:snr-layer-gpt-dinit-small-finweb-full} in \Cref{appendix:initialization} show that Mitchell initialization leads to higher SNR values compared to the default PyTorch initialization across layers of the GPT-small model. In particular, Attn.Proj and MLP.Down layers show significantly higher SNR values. These exceptionally high SNR values provide empirical support for the $1/$depth scaling in Mitchell initialization.
As Adam's second moments adapt to the landscape geometry, these findings indicate that SNR analysis can serve as a proxy for evaluating initialization schemes by determining ones with higher SNR values.

\section{DIY: Build Your Own Low-Memory Adam}
\label{section:slimadam}

In the previous sections, we demonstrated that SNR trends vary across architectures, initialization schemes, dataset properties, and learning rates. We now test whether these SNR trends correctly identify when compression can be performed without sacrificing performance.
To put this to the test, we introduce \emph{SlimAdam}, a memory-efficient Adam variant that preserves Adam's performance and stability through SNR-guided compression. Given the averaged SNR trends, \emph{SlimAdam} (1) compresses matrix-like second moments along the dimension with the highest SNR if it exceeds a cutoff and (2) leaves vector-like second moments uncompressed due to their high variability and minimal effect on the overall memory.

\textbf{Memory Savings in Practice with SlimAdam:}
The SNR-predicted compressibility primarily depends on the learning rate and the SNR cutoff, with distinct patterns across architectures, as shown in the top panel of \Cref{fig:slim-memory-performance}. These results suggest that GPT and ViT models exhibit high compressibility ($\sim 98\%$) at small learning rates, though these savings reduce to $\sim 35\%$ at large learning rates. In contrast, Llama fine-tuning exhibits consistently low compressibility, indicating a more complex optimization landscape. By comparison, ResNets maintain high compressibility regardless of learning rate and cutoff value, suggesting an extremely smooth landscape.

\textbf{Implicit Bias of standard Adam towards low Compressibility:}
In theory, we would perform the SNR analysis at the optimal learning rate to determine compression rules. For Transformer-based models (GPT, Llama, and ViT), this approach will only save up to $35\%$ of second moments. 
Surprisingly, we find that a more aggressive compression is possible using compression rules derived at small learning rates. 
The bottom panel of \Cref{fig:slim-memory-performance} shows that \emph{SlimAdam} achieves Adam-level performance and stability using compression rules derived at learning rates $10 \times$ smaller than optimal.
For GPT, ViT, and ResNets, this saves $\sim 98\%$ second moments while matching Adam's performance and stability. 
Even in the more challenging landscape of fine-tuning tasks, it still achieves substantial memory savings of approximately $\sim 40\%$. 
The success of \emph{SlimAdam} in this setting suggests a previously unreported implicit bias in Adam \textemdash it uses significantly more second moments at large learning rates than required for optimal performance. This ``overparameterization'' in Adam's second moments may contribute to large magnitude weights and activations observed in language models
\cite{sun2024massive,oh2025housecardsmassiveweights}.
These results suggest that SNR analysis at small learning rates captures fundamental compression rules while avoiding artifacts that emerge when training Adam at large learning rates.

\begin{figure}[!t]
\centering
    \begin{minipage}[b]{0.235\textwidth}
    \centering
    \includegraphics[width=\textwidth]{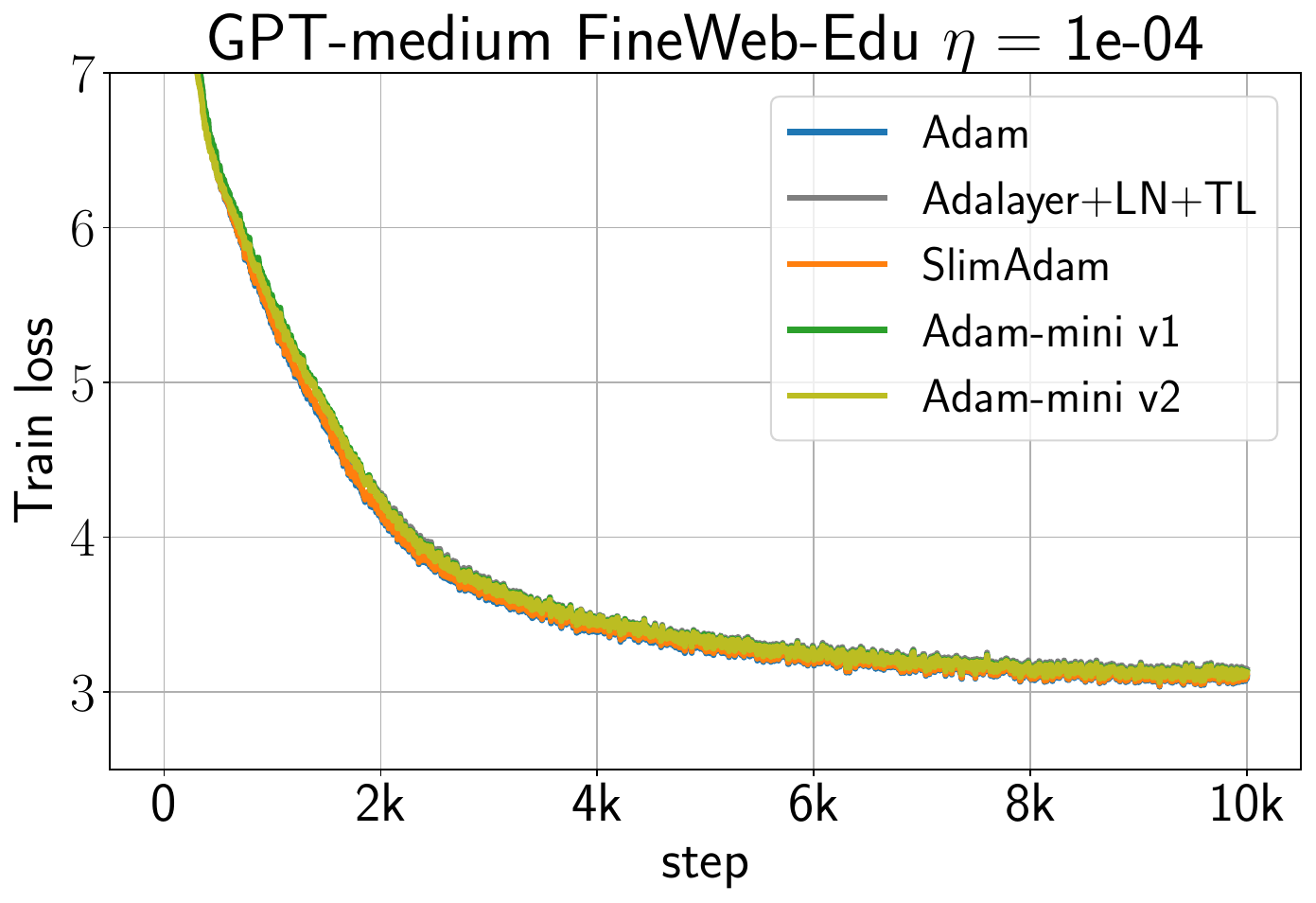}
\end{minipage}
\hfill
\begin{minipage}[b]{0.235\textwidth}
    \centering
    \includegraphics[width=\textwidth]{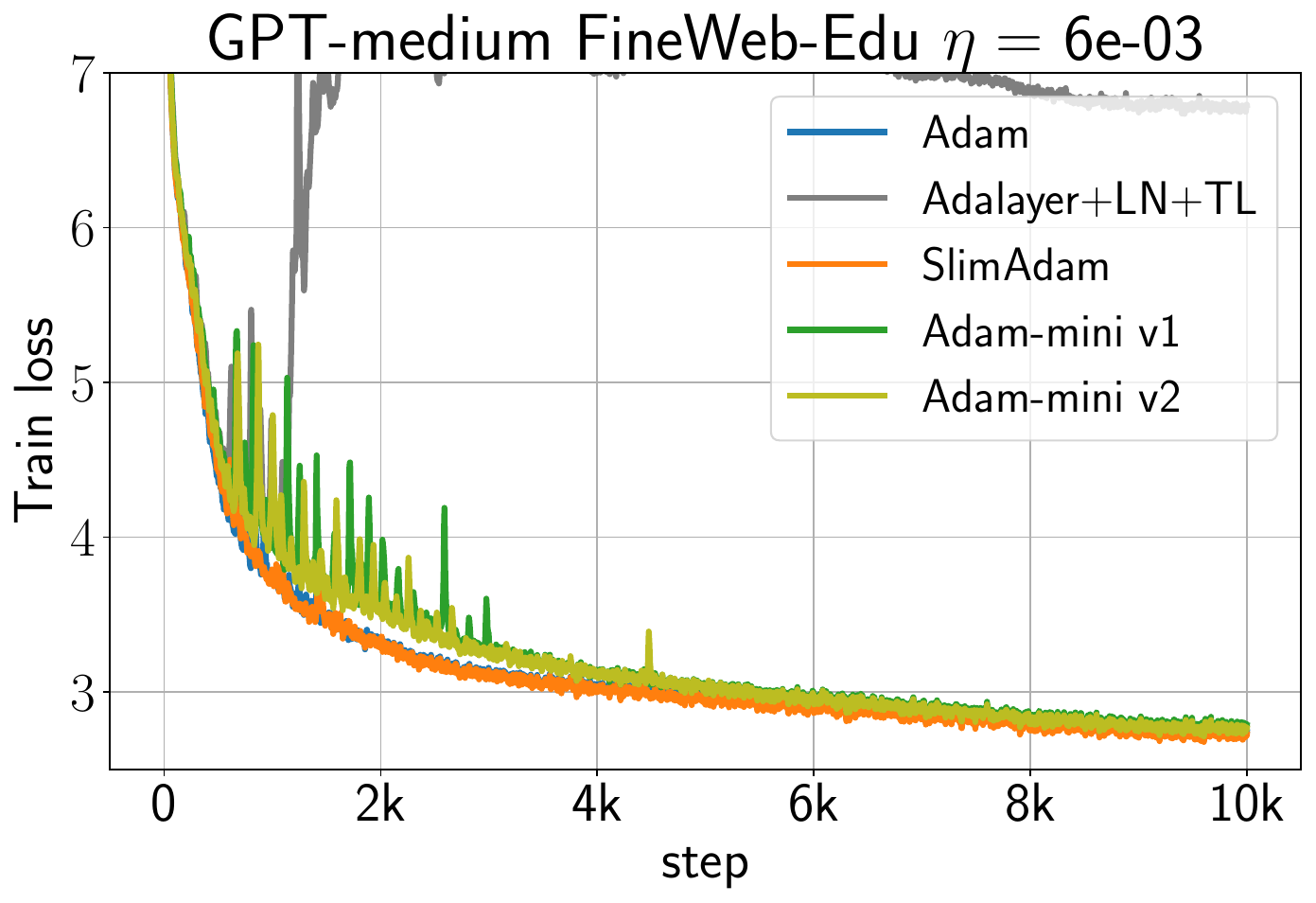}
\end{minipage}
    \caption{Training trajectories  (moving averages of 10 steps) of GPT-medium trained on the Finweb-Edu dataset. (left) all low-memory optimizers exhibit nearly identical curves at small learning rates, (right) at large learning rates, \emph{SlimAdam} exhibits nearly the same training dynamics as Adam, while other low-memory Adam variants experience training instabilities. 
    }
    \label{fig:train-loss-fineweb-lr}
    \vspace{-0.2 in}
\end{figure}

\textbf{Superior Stability of SlimAdam at High Learning Rates:}
\Cref{fig:train-loss-fineweb-lr} shows that \emph{SlimAdam} exhibits more stable training dynamics at large learning rates as compared to other low-memory Adam variants, such as AdaLayer \cite{zhao2024deconstructingmakesgoodoptimizer} and Adam-mini \cite{zhang2024adamminiusefewerlearning}. 
While the other low-memory Adam variants exhibit large training instabilities at Adam's optimal learning rate, \emph{SlimAdam} exhibits nearly the same training dynamics as Adam. 
This difference in stability is expected, as for Adam variants, the pre-conditioner $P^{-1} = \frac{1}{\sqrt{V}}$ directly influences the local instability threshold\footnote{The local instability threshold is the critical learning rate above which the loss increases in the next training step.} \cite{cohen2022adaptive,kalra2024why}. These results suggest that compressing the ``correct'' dimensions as guided by our SNR analysis is crucial for maintaining both stability and performance at large learning rates. In contrast, all low-memory Adam variants perform equally well at small learning rates.

We also analyze the robustness of \emph{SlimAdam}'s compression rules across different datasets and model sizes in \Cref{appendix:rule-transfer}. When switching from OpenWebText to FineWeb-Edu, we observe that compression rules remain consistent for most layers, with variations in only five matrices \textemdash with four being early MLP layers. Similarly, compression rules remain consistent across different model widths, with variations in only $12$ matrices ($8$ from early MLPs, $4$ from attention components). 
These results are intuitive since MLP layers exhibit high variability of compression dimensions. 
We find that these variations can be eliminated by using depth-averaged SNR for each layer type, resulting in more consistent trends. \Cref{fig:snr-mean-rules} in \Cref{appendix:rule-transfer} shows that rules derived from depth-averaged SNR produce identical results to per-layer compression rules.
This robustness has implications for efficient deployment in practice \textemdash compression rules can be identified using smaller models during preliminary experiments and then transferred to large ones.

\section{Discussion}

\looseness -1
Our computationally efficient SNR analysis independently confirms and extends several findings from prior work while overcoming their limitations. \cite{zhang2024adamminiusefewerlearning} used Hessian-based analysis of small models to construct a low-memory optimizer and then applied these rules to larger models, assuming transferability. A primary advantage of our approach is that we can directly analyze models of any scale without requiring expensive Hessian computations or assumptions about transferability between model sizes.
Similarly, \cite{zhao2024deconstructingmakesgoodoptimizer}'s extensive ablation studies showed that Adam's advantage over SGD in language modeling primarily stems from maintaining per-parameter second moments for two components: LM Head and LayerNorm. Our SNR analysis naturally uncovers these same trends and shows that for LM Head and Token Embedding, this aversion to compression is specific only to the token dimension.

\looseness -1
Beyond optimizer design, our SNR analysis also serves as a diagnostic tool.
The SNR values of the gradient's second-moment function as a proxy for learning complexity within each layer, with lower SNR indicating higher complexity. This insight naturally reveals regions of model architecture that could benefit from improvements. For instance, the low SNR values observed in token embeddings or language model heads suggest these components might benefit from more sophisticated layer designs. SNR analysis also enables a quantitative evaluation of the effectiveness of initialization schemes for different layers. \Cref{section:initialization} demonstrates how PyTorch's default initialization yields consistently lower SNR values compared to Mitchell initialization, indicating the scheme's suboptimality. 

In conclusion, we present a principled SNR framework to analyze when second moments can be effectively replaced with their means, naturally leading to \emph{SlimAdam}, a practical low-memory Adam variant which maintains its performance and stability while saving up to $98\%$ second moments. We hope our work furthers the communities' understanding of when low memory optimizers are safe to use in practice while deepening our fundamental understanding of how architecture, training regime, and optimizer design interact.

\section*{Acknowledgements}
This work is supported by NSF DMR-2345644 (D.S.K., and M.B.) and Office of Naval Research MURI program, DARPA TIAMAT, the National Science Foundation IIS-2212182, and the NSF TRAILS Institute 2229885 (J.K. and T.G.).
Commercial support was provided by Capital One Bank, the Amazon Research Award program, and Open Philanthropy (J.K. and T.G.).
The authors acknowledge the University of Maryland supercomputing resources (\url{http://hpcc.umd.edu}) made available for conducting the research reported in this paper. 
The authors would like to thank Jonas Geiping, Aditya Tomar, Siddharth Singh, and Abhinav Bhatele for discussions during early phases of the research and D.S.K. would like to thank Tianyu He for helpful discussions and comments on the final manuscript.

\bibliography{ICML2025/references}
\bibliographystyle{icml2025}

\newpage
\appendix
\onecolumn

\section{Detailed Comparison with Other Low-memory Optimizers}
\label{appendix:related-works}

\begin{figure*}[!htb]
\centering
\begin{minipage}[b]{0.245\textwidth}
    \centering
    \includegraphics[width=\textwidth]{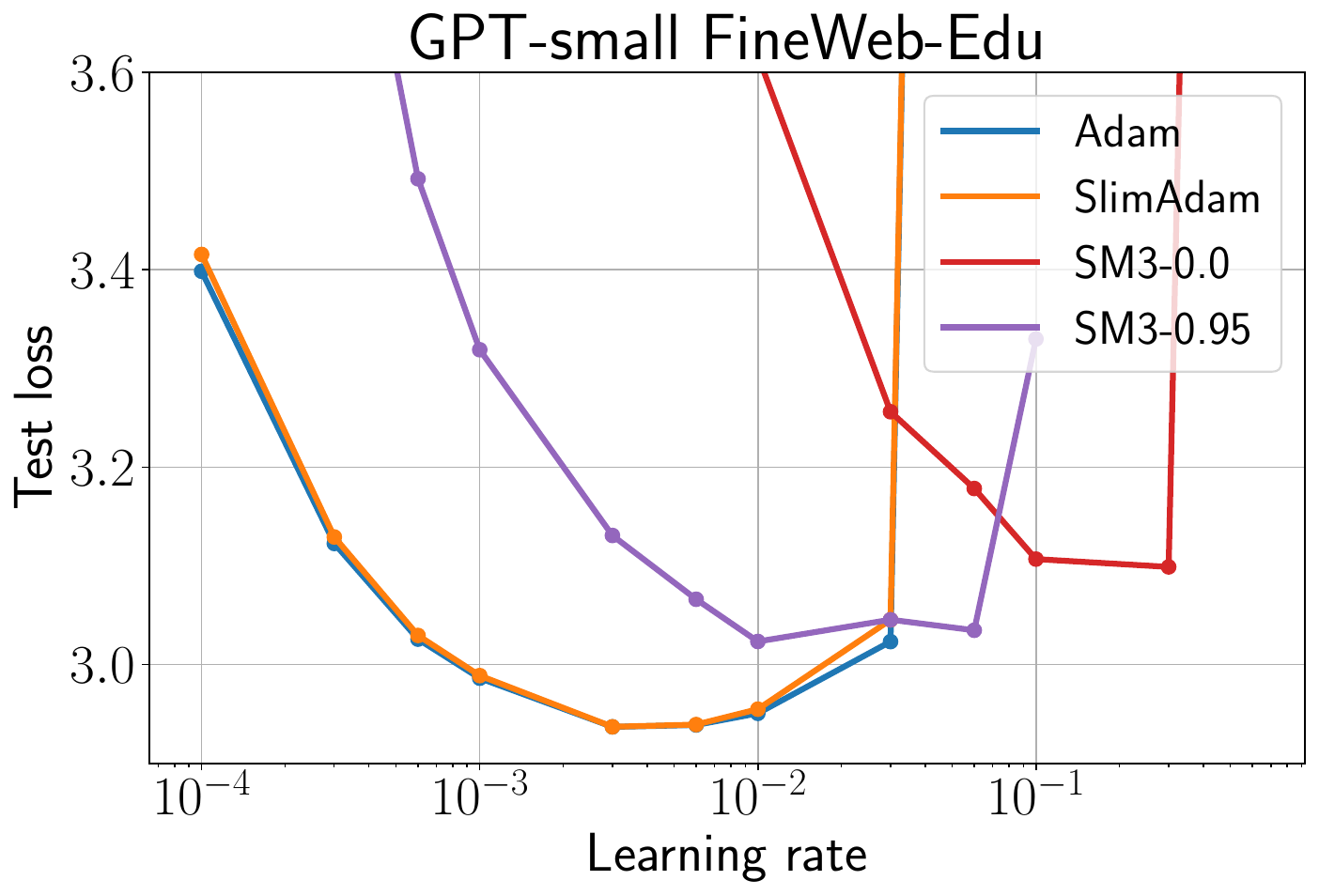}
    \subcaption{}
\end{minipage}
\begin{minipage}[b]{0.245\textwidth}
    \centering
    \includegraphics[width=\textwidth]{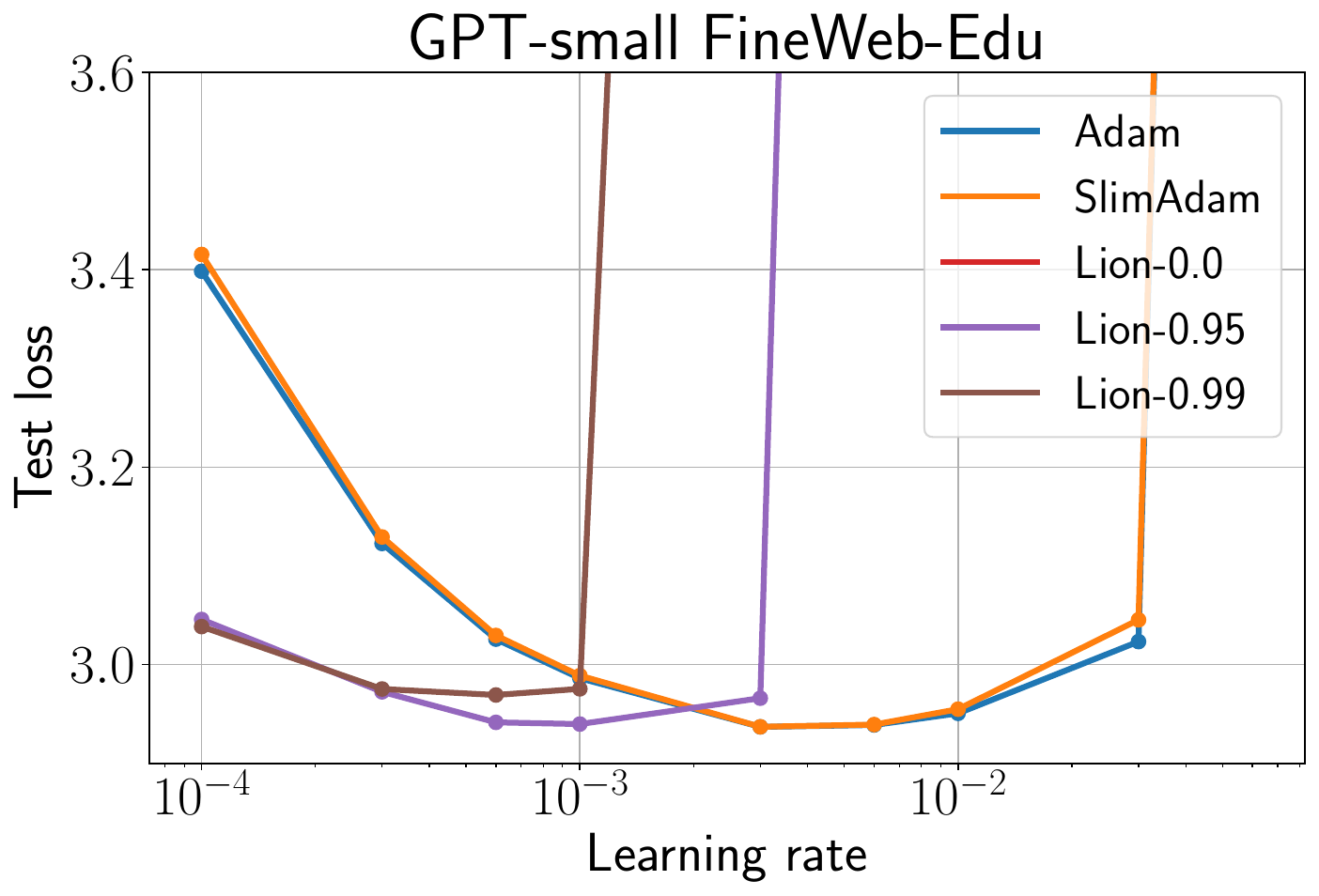}
    \subcaption{}
\end{minipage}
\begin{minipage}[b]{0.245\textwidth}
    \centering
    \includegraphics[width=\textwidth]{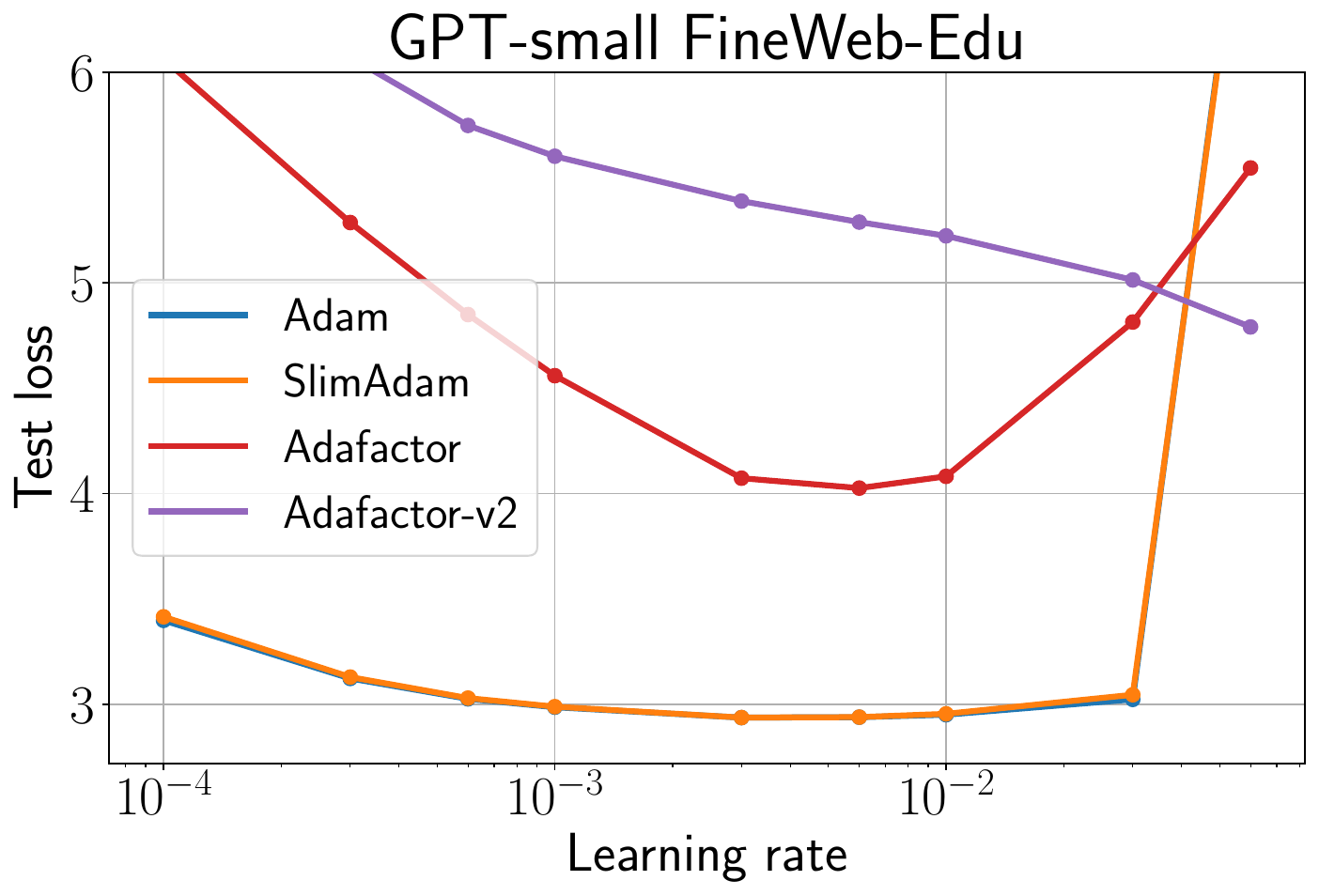}
    \subcaption{}
\end{minipage}
\caption{Comparison of \emph{SlimAdam} with different optimizers on GPT pre-training using Fineweb-Edu dataset.}
\label{fig:optim-comparisons-additional}
\end{figure*}

\textbf{Adam-mini}:
\cite{zhang2024adamminiusefewerlearning} introduced Adam-mini, which assigns adaptive learning rates to block partitions based on the Hessian spectrum at initialization. The initial release, Adam-mini v1.0.4 (referred to as Adam-mini v1), uses PyTorch's default block partitioning with two key modifications: (1) individual second moments are assigned to each parameter in the Token Embedding and LM Head, and (2) individual second moments are assigned to each key and query attention head. In a recent update, Adam-mini v1.1.1 (referred to as Adam-mini v2) revises this approach by assigning one second moment per output neuron in each layer, with two exceptions: (1) each key and query attention head receives its own second moment, and (2) each token dimension in the Token Embedding and LM Head receives its own second moment. LayerNorms are always compressed. 

Our SNR analysis identifies similar compression rules to Adam-mini but with two key differences. First, Adam-mini assigns one second moment to every output neuron of attention values, projection, and MLPs. In our convention, it amounts to $\text{fan}_{\text{in}}$ compression.
In comparison, our SNR analysis suggests that $\text{fan}_{\text{out}}$ compression is more appropriate for these layers. 
The second difference relates to LayerNorm parameters. While Adam-mini compresses these by default, our SNR analysis indicates that LayerNorm second moments show aversion to compression. We attribute \emph{SlimAdam}'s superior learning rate stability to its identification of these more appropriate compression dimensions.

\textbf{AdaLayer:} \cite{zhao2024deconstructingmakesgoodoptimizer} found that Adam's superior performance over SGD in language modeling primarily comes from using per-parameter adaptive learning rates in just two components: LayerNorm and the LM Head. All other layers can be trained with SGD. Following their naming convention, we use AdaLayer to refer to Adam with one second moment per weight/bias, and `AdaLayer+LN+TN' to denote AdaLayer with per-parameter second moments for LayerNorm and final layer parameters. 

While our SNR analysis supports their findings about Token Embedding/LM Head and LayerNorm second moments, we find that AdaLayer+LN+TN underperforms Adam and \emph{SlimAdam} using $2\%$ of Adam's second moments closely matches Adam's performance and stability.

\paragraph{SM3:} 
SM3 \cite{sm3}  groups parameters into sets based on similarity, such that each parameter can belong to multiple sets. Then, it maintains a moving average of the maximum of squared moments for each set and approximates a second-moment entry using the minimum value across different sets it belongs to.
We use the implementation from \cite{pytorch-sm3} with momentum $= 0.9$ and $\beta \in \{0.0, 0.95\}$. \Cref{fig:optim-comparisons-additional}(a) compares SM3 performance with different $\beta$ values on the GPT pre-training task. 
We observe that $\beta = 0.95$ performs better for GPT pre-training. We use this optimal $\beta$ value in the comparisons shown in \Cref{fig:optim-comparison}.

\paragraph{Lion:}
Lion \cite{chen2023symbolic} is an algorithmically discovered optimizer that only tracks momentum and uses the sign operation to determine update directions.
For the GPT-small experiment, we found that $\beta_2 = 0.95$ performs best when keeping $\beta_1 = 0.9$ fixed, as shown in \Cref{fig:optim-comparisons-additional}(b). Similar to other optimizers, we use a weight decay strength of $\lambda = 0.1$ and a gradient clipping threshold of $1.0$.

\paragraph{Adafactor:} 
\cite{adafactor-shazeer18a} approximates the second-moment matrix of a layer using a moving average of the row and column sums of the squared gradients. We evaluate two implementations: (1) the PyTorch implementation, which does not use a moving average of updates (referred to as Adafactor) and (2) the implementation by \cite{fairseq_adafactor}, which incorporates the moving average of updates (referred to as Adafactor v2). For both variants, we maintain the same learning rate schedule used in our default experiments. For Adafactor v2, this requires setting \texttt{relative\_step=False}. As shown in \Cref{fig:optim-comparisons-additional}(c), both Adafactor variants perform significantly worse than Adam. Due to this performance gap, we exclude these results from \Cref{fig:optim-comparison}.

\section{Experimental Details}
\label{appendix:experimental-details}

\paragraph{SNR measurement:}
We measured SNR values at regular intervals throughout training: every $100$ step for the first $1000$ steps, then every $1000$ step thereafter.

\subsection{Language Pre-training}
\label{appendix:pretraining-details}

\paragraph{Model and Datasets:}
We train GPT-style models \cite{radford2019language} using a codebase based on NanoGPT \cite{karpathy2022nanogpt} on two language modeling datasets: OpenWebText \cite{Gokaslan2019OpenWeb} and $10$B token subset of FineWeb-Edu \cite{penedo2024the}. The datasets are tokenized using the GPT tokenizer with a vocabulary size $n_{\text{vocab}} = 50, 304$. The models are trained with a context length of $T_n = 1024$. We use $n_{\text{layers}}$ to denote the number of layers, $n_{\text{heads}}$ to denote the number of heads, and $d_{\text{model}}$ to denote the embedding dimension.

We consider two model configurations:

1. GPT-small ($n_{\text{layers}} = 12$, $n_{\text{heads}}  = 12$, $d_{\text{model}} = 768$)\\
2. GPT-medium ($n_{\text{layers}} = 24$, $n_{\text{heads}} = 16$, $d_{\text{model}} 
 = 1024$).
 
Both with an MLP upscaling factor of $4$, learnable positional embedding, and weight tying, without biases.

 \paragraph{Initialization:} Unless specified, we consider the Mitchell initialization \cite{groeneveld-etal-2024-olmo}: For standard layers, the weights are initialized using a normal distribution $\mathcal{N}(0, 0.02^2)$, while residual projection layers (attention and MLP projections) use a scaled normal distribution $\mathcal{N}(0, 
 \nicefrac{0.02^2}{2n_{\text{layers}}})$. In \Cref{section:initialization}, we use PyTorch's default uniform initialization: $\mathcal{U}(- \frac{1}{\sqrt{\text{fan}{\text{in}}}}, \frac{1}{\sqrt{\text{fan}{\text{in}}}})$.

\paragraph{Training:} The training uses a micro-batch size of $32$ with $40$ gradient accumulation steps, resulting in an effective batch size of $B = 1,280$. All models are trained for $10,000$ steps using different Adam variants with the following hyperparameters: $\beta_1 = 0.9$, $\beta_2 = 0.95$, $\epsilon = 10^{-8}$, and weight decay strength $\lambda = 0.1$. The learning rate is linearly increased from zero to a target learning rate $\eta$ in $T_{\text{wrm}} = 2048$ steps, followed by cosine decay to $\eta_{\text{min}} = \nicefrac{\eta}{10.0}$.
Gradients are clipped at a maximum norm of $1.0$.

\subsection{Linear Model trained on WikiText}

\paragraph{Model Architecture:}
We consider a two-layer linear model composed of an embedding layer followed by a language model head, trained on WikiText-103 \cite{merity2017pointer}. The dataset is tokenized using BPE tokenization \cite{BPE-Gage,sennrich-etal-2016-neural} with different vocabulary sizes $V \in \{1024, 2048, 4096, 8192, 16384, 32768, 49152, 65536\}$. The embedding dimension is set to $d_{\text{model}} = 768$ and a context length of $T_n = 128$ is considered.

\paragraph{Initialization:}
The embedding parameters are initialized using a truncated normal distribution $\mathcal{N}(0, 1)$, while the language model head uses a truncated normal distribution $\mathcal{N}(0, \nicefrac{1}{\text{fan}_{\text{in}}})$.

\paragraph{Training:}
The training consists of one epoch with a batch size $B = 16$. The model is trained using Adam variants with hyperparameters $\beta_1 = 0.9$, $\beta_2 = 0.999$, $\epsilon = 10^{-8}$, and weight decay strength $\lambda = 10^{-4}$. The learning rate follows a schedule with linear warmup from zero to $\eta$ over $T_{\text{wrm}} = 2048$ steps, followed by cosine decay to $\eta_{\text{min}} = \nicefrac{\eta}{10.0}$. The optimal target learning rate is found by scanning the set $\{ 1\text{e-}4, 3\text{e-}4, 6\text{e-}4, 1\text{e-}3, 3\text{e-}3 \}$.

\subsection{Language Fine-tuning}
\label{appendix:finetuning-details}

\paragraph{Model and Datasets:} We consider pre-trained Llama-3.2 models \cite{grattafiori2024llama3herdmodels} and fine-tune them on the Alpaca dataset \cite{alpaca} using the torchtune library \cite{torchtune}.

\paragraph{Fine-tuning:} We finetune the models for $3$ epochs using a batch size $B = 16$, optimizer hyperparameters $\beta_1 = 0.9$, $\beta_2 =0.999$, $\epsilon = 10^{-8}$ and weight decay strength $\lambda = 0.1$. 

\subsection{Image Classification}
\label{appendix:image-classification-details}

\paragraph{Model and Datasets:} We train ResNet \cite{He2015DeepRL} and ViT \cite{dosovitskiy2021an} models on CIFAR-10 and CIFAR-100 datasets \cite{cifar10} with random crop and horizontal flip augmentations. 

\textbf{ResNet:} We consider the standard ResNet-18 architecture with batch normalization.

\textbf{ViT}:  We consider Vision Transformers \cite{dosovitskiy2021an}, with GPT-like architecture adapted for image classification using patch embeddings and a special class token. We consider two model configurations:  ViT-mini ($n_{\text{layers}} =6$  layers, $n_{\text{heads}} = 12$ heads, embedding dimension $d_{\text{model}} = 768$) and ViT-small ($n_{\text{layers}} =12$  layers, $n_{\text{heads}} = 12$ heads, embedding dimension $d_{\text{model}} = 768$). Both models are initialized using Mitchell initialization, do not use biases, and use a learnable class token and a patch size of $2$.

\paragraph{Training:} We train these models with a batch size of $B = 128$ for $100,000$ steps with optimization hyperparamters:  $\beta_1 = 0.9$, $\beta_2 =0.999$, $\epsilon = 10^{-8}$ and weight decay strength $\lambda = 0.01$. The learning rate is linearly increased from zero to a target learning rate $\eta$ in $T_{\text{wrm}} = 2048$ steps, followed by cosine decay to $\eta_{\text{min}} = \nicefrac{\eta}{10.0}$.

\section{SNR Analysis of Diverse Training Regimes}
\label{appendix:additional-snr-results}

\begin{figure*}[!htb]
\centering
\begin{minipage}[b]{0.245\textwidth}
    \centering
    \includegraphics[width=\textwidth]{figures/snr-analysis/snr-trajectories/GPT-small/snr_Attn.Key_L5_gpt2_openwebtext_SlimAdamW_T10000_ga40_d12_h12_n768_lr0.0003_wd0.1_bs32_b0.9_b0.95.pdf}
\end{minipage}
\hfill
\begin{minipage}[b]{0.245\textwidth}
    \centering
    \includegraphics[width=\textwidth]{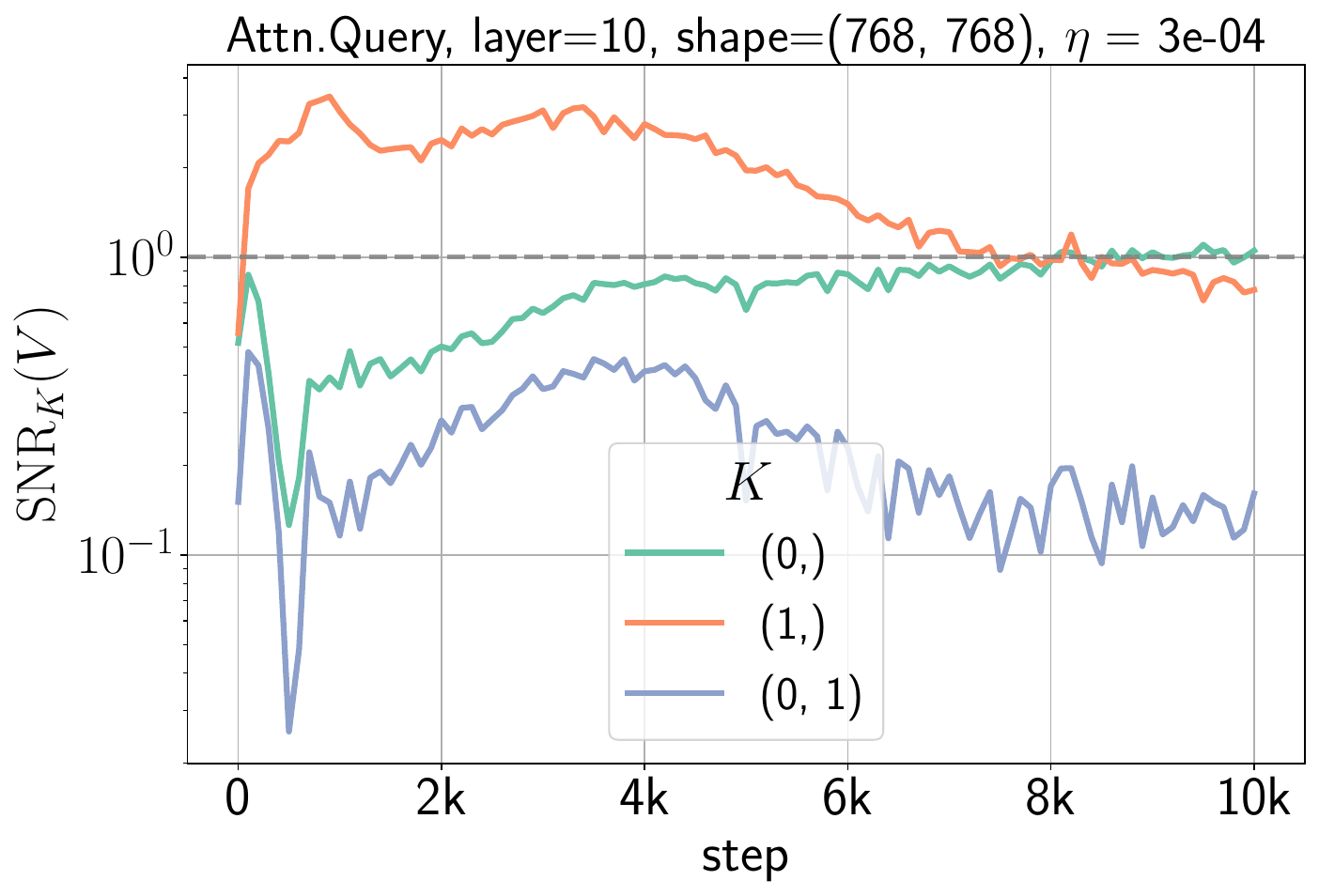}
\end{minipage}
\hfill
\begin{minipage}[b]{0.245\textwidth}
    \centering
    \includegraphics[width=\textwidth]{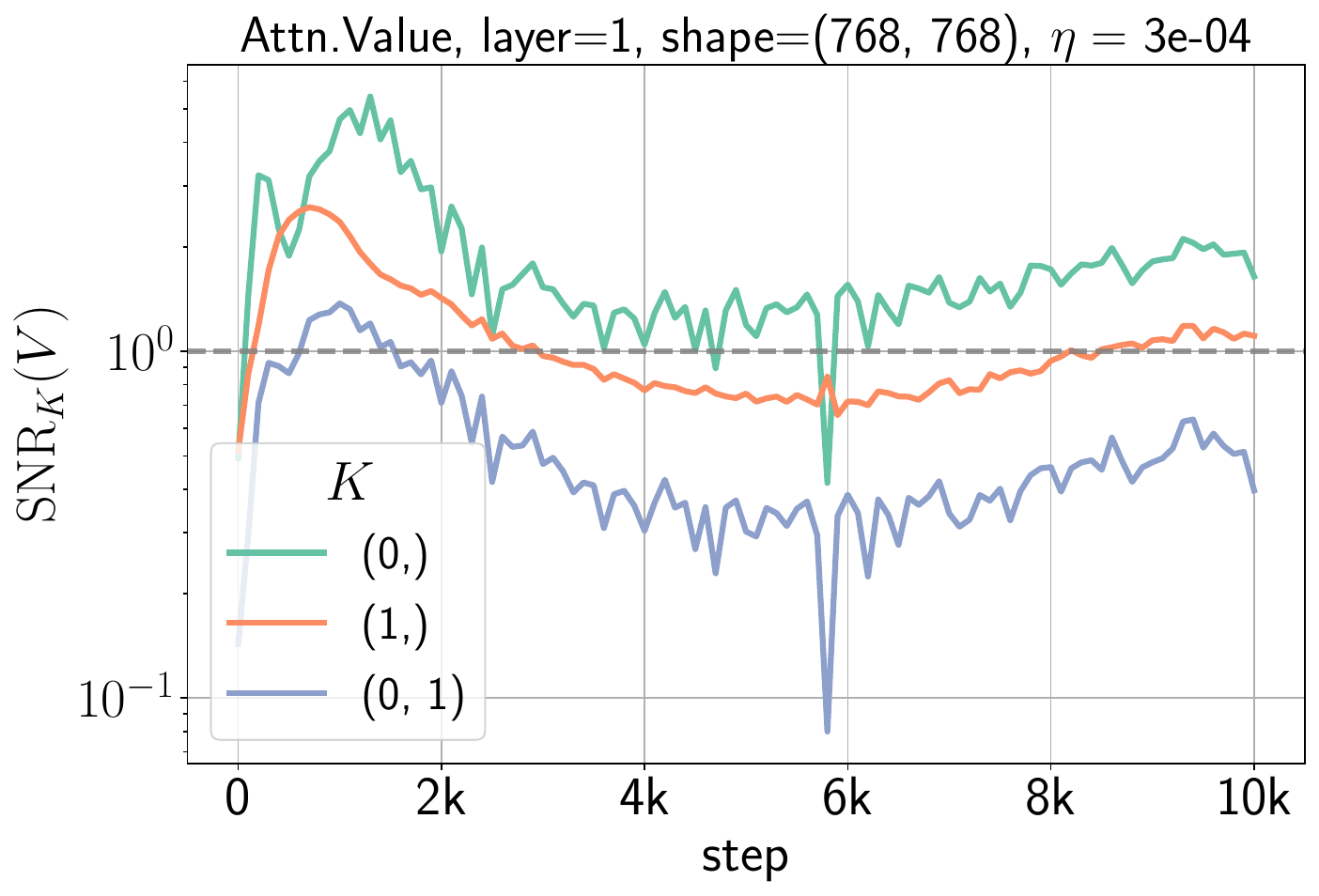}
\end{minipage}
\hfill
\begin{minipage}[b]{0.245\textwidth}
    \centering
    \includegraphics[width=\textwidth]{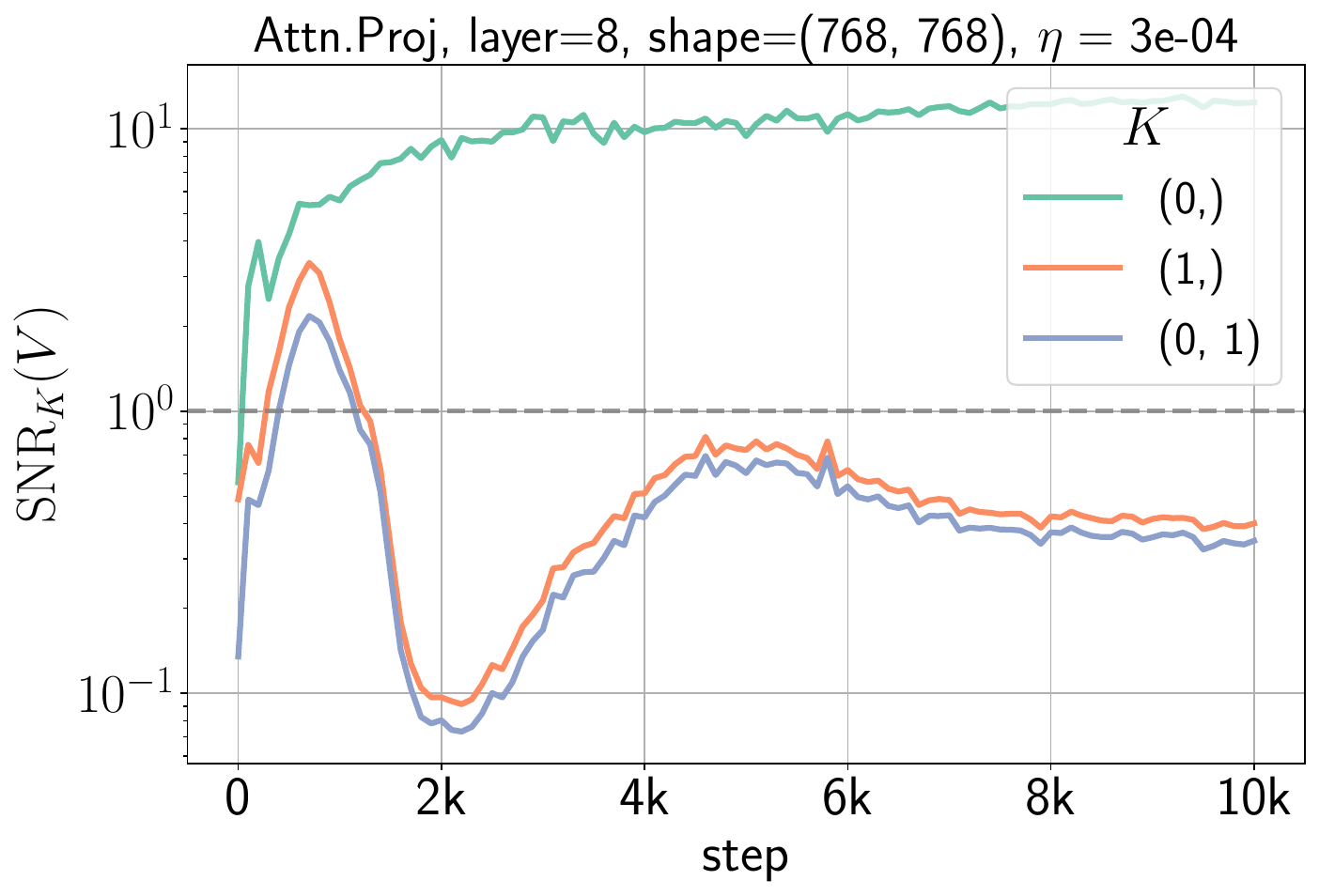}
\end{minipage}

\begin{minipage}[b]{0.245\textwidth}
    \centering
    \includegraphics[width=\textwidth]{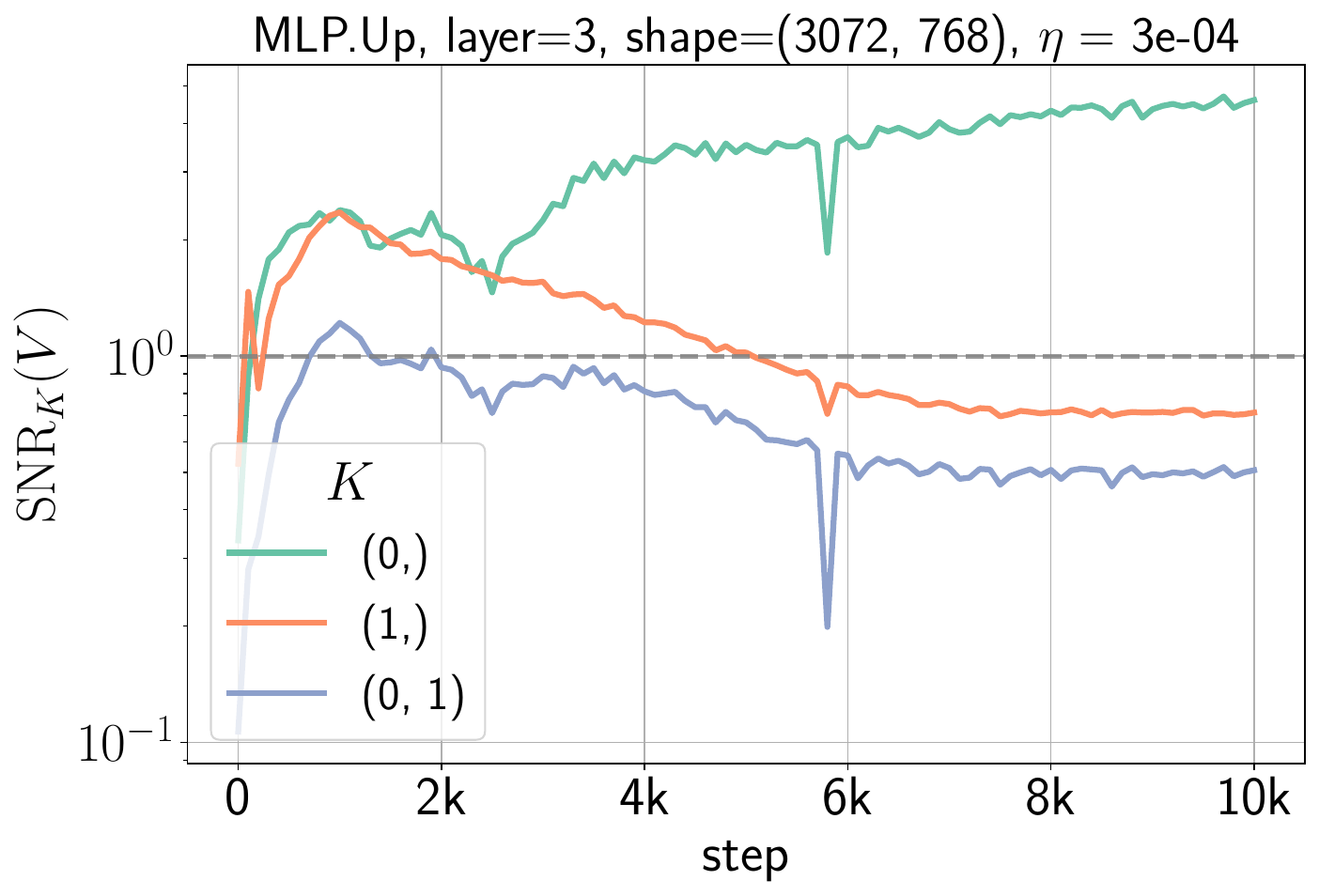}
\end{minipage}
\hfill
\begin{minipage}[b]{0.245\textwidth}
    \centering
    \includegraphics[width=\textwidth]{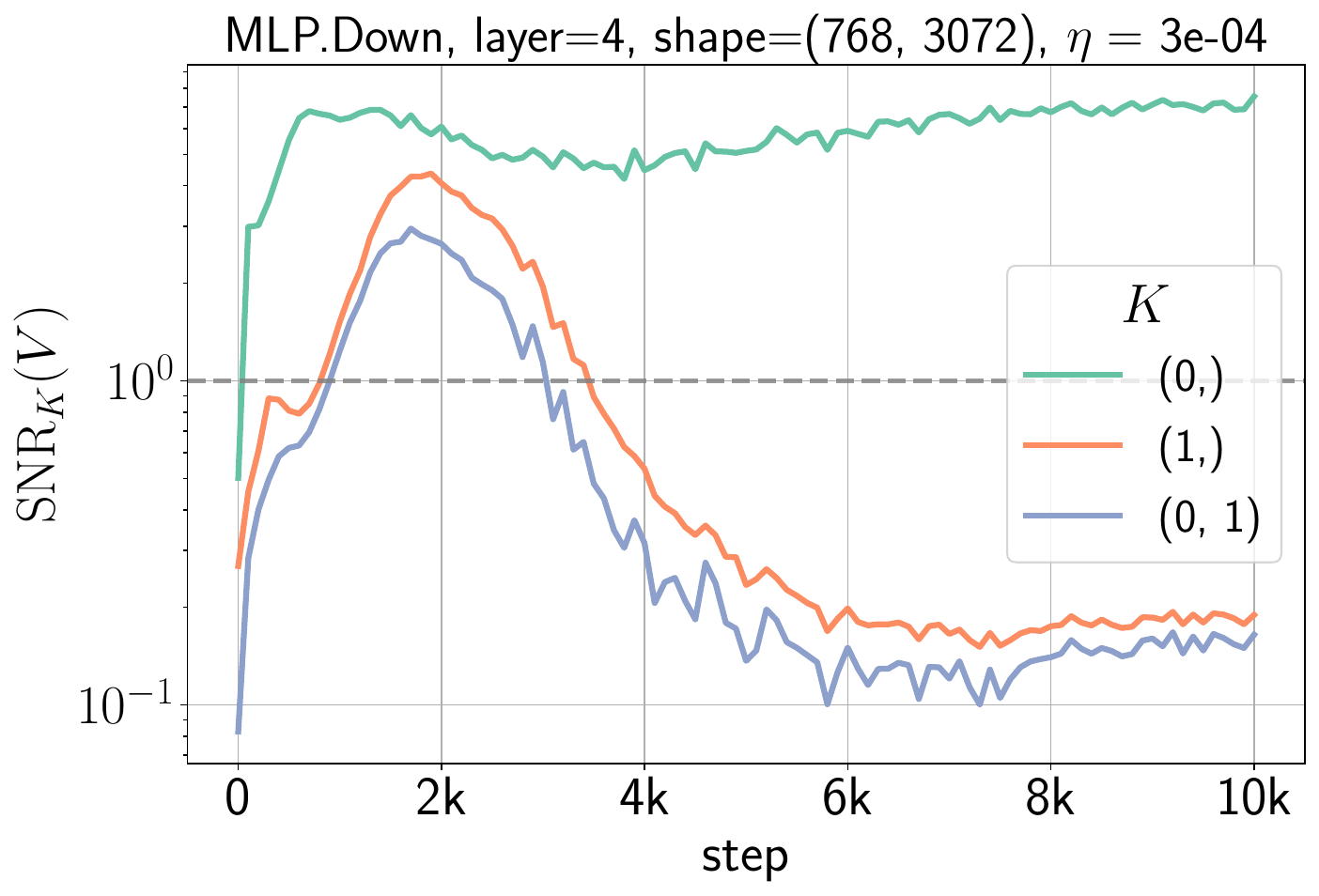}
\end{minipage}
\hfill
\begin{minipage}[b]{0.245\textwidth}
    \centering
    \includegraphics[width=\textwidth]{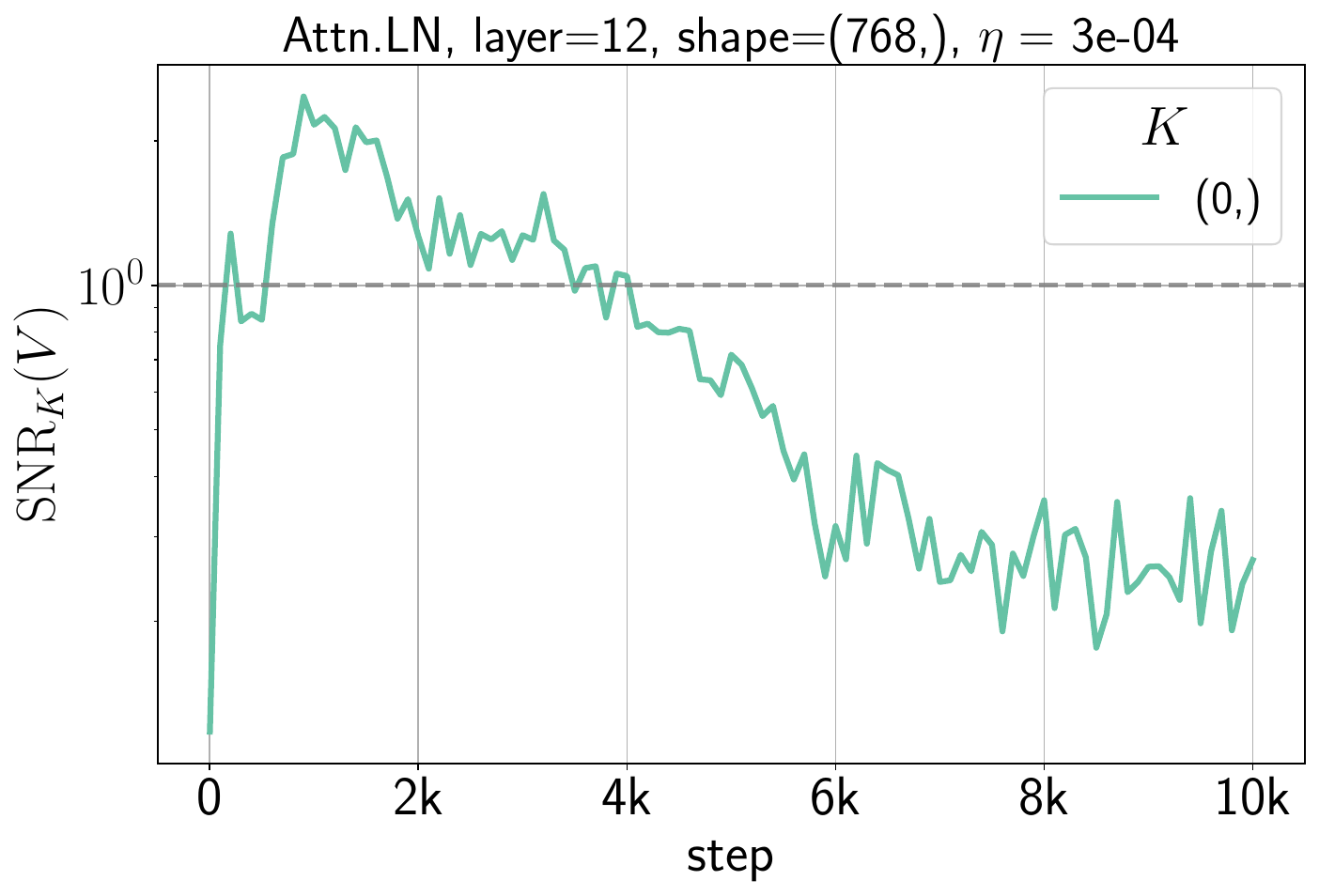}
\end{minipage}
\hfill
\begin{minipage}[b]{0.245\textwidth}
    \centering
    \includegraphics[width=\textwidth]{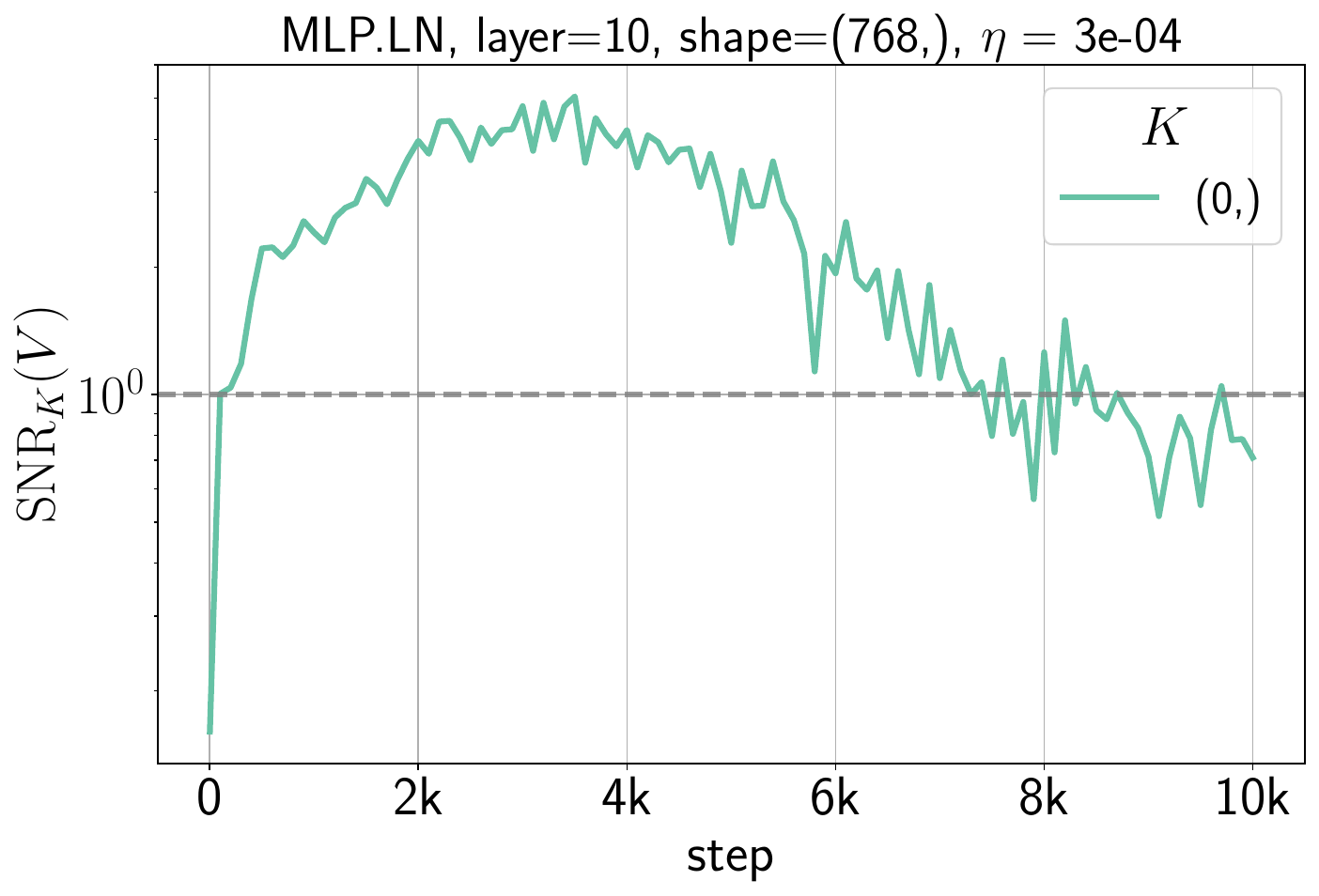}
\end{minipage}
\hfill
\begin{minipage}[b]{0.245\textwidth}
    \centering
    \includegraphics[width=\textwidth]{figures/snr-analysis/snr-trajectories/GPT-small/snr_Final.LN_LNone_gpt2_openwebtext_SlimAdamW_T10000_ga40_d12_h12_n768_lr0.0003_wd0.1_bs32_b0.9_b0.95.pdf}
\end{minipage}
\begin{minipage}[b]{0.245\textwidth}
    \centering
    \includegraphics[width=\textwidth]{figures/snr-analysis/snr-trajectories/GPT-small/snr_Tok.Embd_LNone_gpt2_openwebtext_SlimAdamW_T10000_ga40_d12_h12_n768_lr0.0003_wd0.1_bs32_b0.9_b0.95.pdf}
\end{minipage}
\begin{minipage}[b]{0.245\textwidth}
    \centering
    \includegraphics[width=\textwidth]{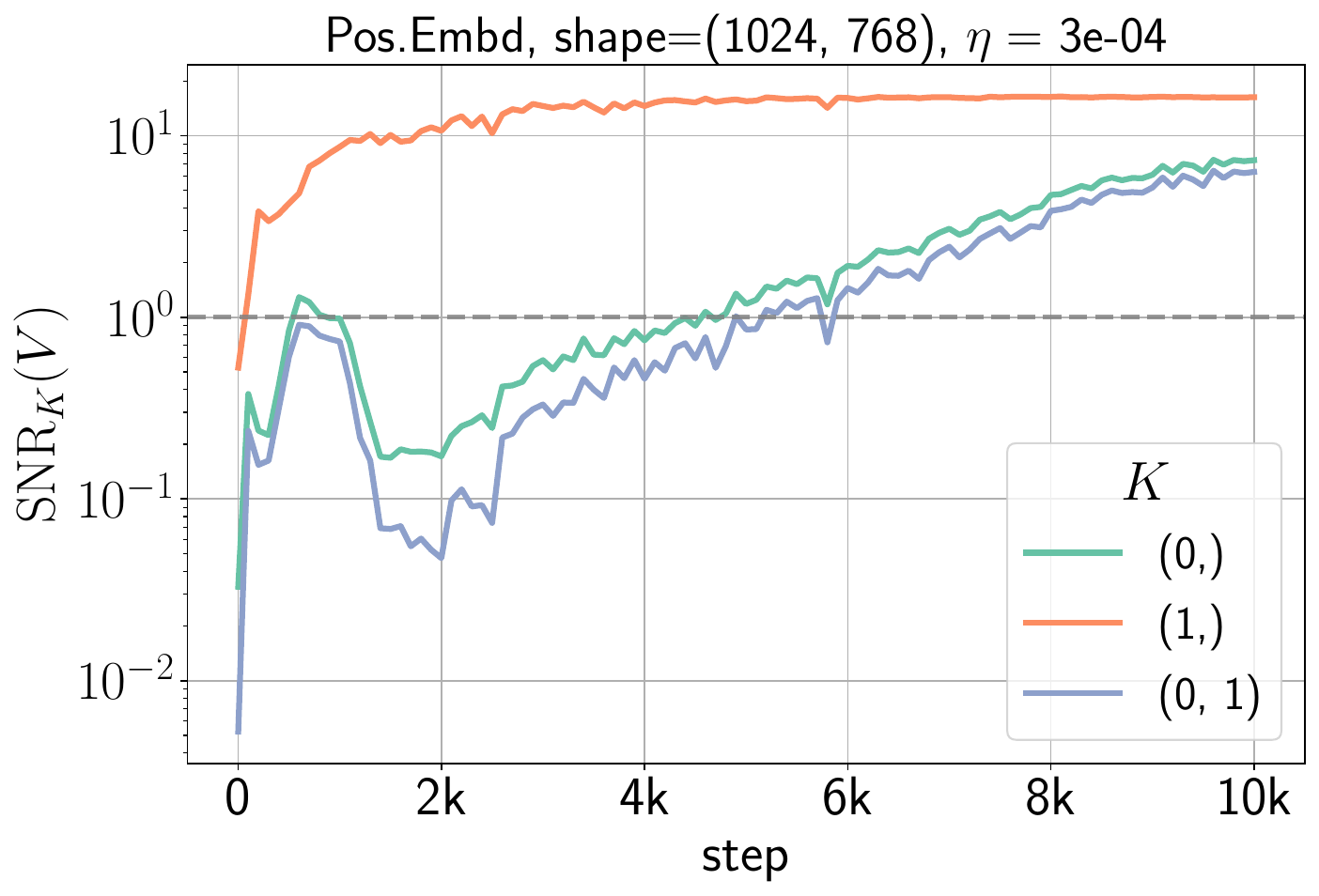}
\end{minipage}

\caption{SNR trajectories of GPT-small trained on OpenWebText. For each layer type, the layer number is selected at random.}
\label{fig:snr-curves-gpt-small-openweb-full}
\end{figure*}

\subsection{Language Pre-training}
\label{appendix:pretraining}

\begin{figure*}[!htb]
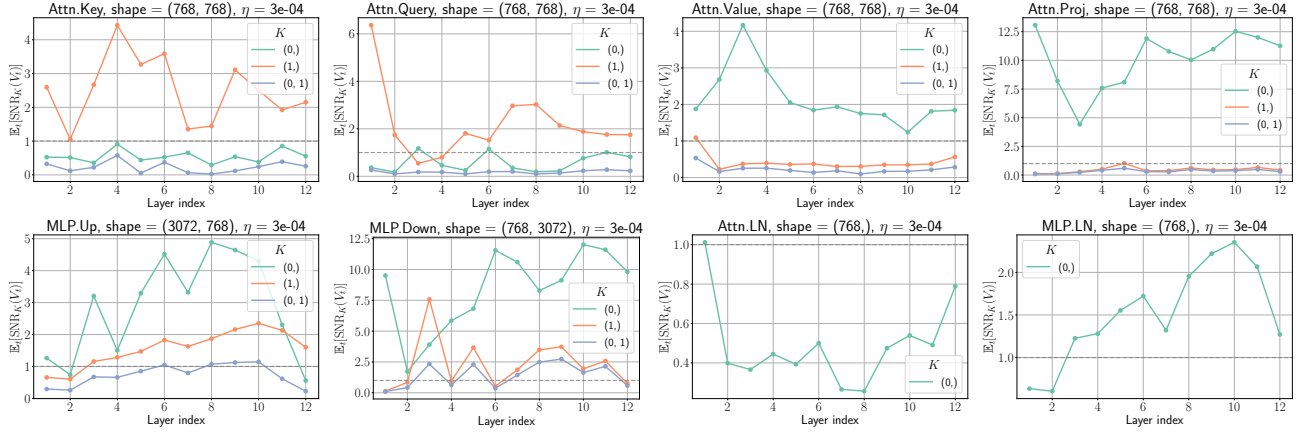

\centering
\begin{minipage}[b]{0.245\textwidth}
    \centering
    \includegraphics[width=\textwidth]{figures/snr-analysis/snr-layer/GPT-small/snr_layer_Attn.Key_gpt2_openwebtext_SlimAdamW_T10000_ga40_d12_h12_n768_lr0.0003_wd0.1_bs32_b0.9_b0.95.pdf}
\end{minipage}
\hfill
\begin{minipage}[b]{0.245\textwidth}
    \centering
    \includegraphics[width=\textwidth]{figures/snr-analysis/snr-layer/GPT-small/snr_layer_Attn.Query_gpt2_openwebtext_SlimAdamW_T10000_ga40_d12_h12_n768_lr0.0003_wd0.1_bs32_b0.9_b0.95.pdf}
\end{minipage}
\hfill
\begin{minipage}[b]{0.245\textwidth}
    \centering
    \includegraphics[width=\textwidth]{figures/snr-analysis/snr-layer/GPT-small/snr_layer_Attn.Value_gpt2_openwebtext_SlimAdamW_T10000_ga40_d12_h12_n768_lr0.0003_wd0.1_bs32_b0.9_b0.95.pdf}
\end{minipage}
\hfill
\begin{minipage}[b]{0.245\textwidth}
    \centering
    \includegraphics[width=\textwidth]{figures/snr-analysis/snr-layer/GPT-small/snr_layer_Attn.Proj_gpt2_openwebtext_SlimAdamW_T10000_ga40_d12_h12_n768_lr0.0003_wd0.1_bs32_b0.9_b0.95.pdf}
\end{minipage}

\begin{minipage}[b]{0.245\textwidth}
    \centering
    \includegraphics[width=\textwidth]{figures/snr-analysis/snr-layer/GPT-small/snr_layer_MLP.Up_gpt2_openwebtext_SlimAdamW_T10000_ga40_d12_h12_n768_lr0.0003_wd0.1_bs32_b0.9_b0.95.pdf}
\end{minipage}
\hfill
\begin{minipage}[b]{0.245\textwidth}
    \centering
    \includegraphics[width=\textwidth]{figures/snr-analysis/snr-layer/GPT-small/snr_layer_MLP.Down_gpt2_openwebtext_SlimAdamW_T10000_ga40_d12_h12_n768_lr0.0003_wd0.1_bs32_b0.9_b0.95.pdf}
\end{minipage}
\hfill
\begin{minipage}[b]{0.245\textwidth}
    \centering
    \includegraphics[width=\textwidth]{figures/snr-analysis/snr-layer/GPT-small/snr_layer_Attn.LN_gpt2_openwebtext_SlimAdamW_T10000_ga40_d12_h12_n768_lr0.0003_wd0.1_bs32_b0.9_b0.95.pdf}
\end{minipage}
\hfill
\begin{minipage}[b]{0.245\textwidth}
    \centering
    \includegraphics[width=\textwidth]{figures/snr-analysis/snr-layer/GPT-small/snr_layer_MLP.LN_gpt2_openwebtext_SlimAdamW_T10000_ga40_d12_h12_n768_lr0.0003_wd0.1_bs32_b0.9_b0.95.pdf}
\end{minipage}
\caption{Layer dependence of averaged SNR values of GPT-small trained on OpenWebText.}
\label{fig:snr-layer-gpt-small-openweb-full}
\end{figure*}

\begin{figure*}[!htb]
\centering
\begin{minipage}[b]{0.245\textwidth}
    \centering
    \includegraphics[width=\textwidth]{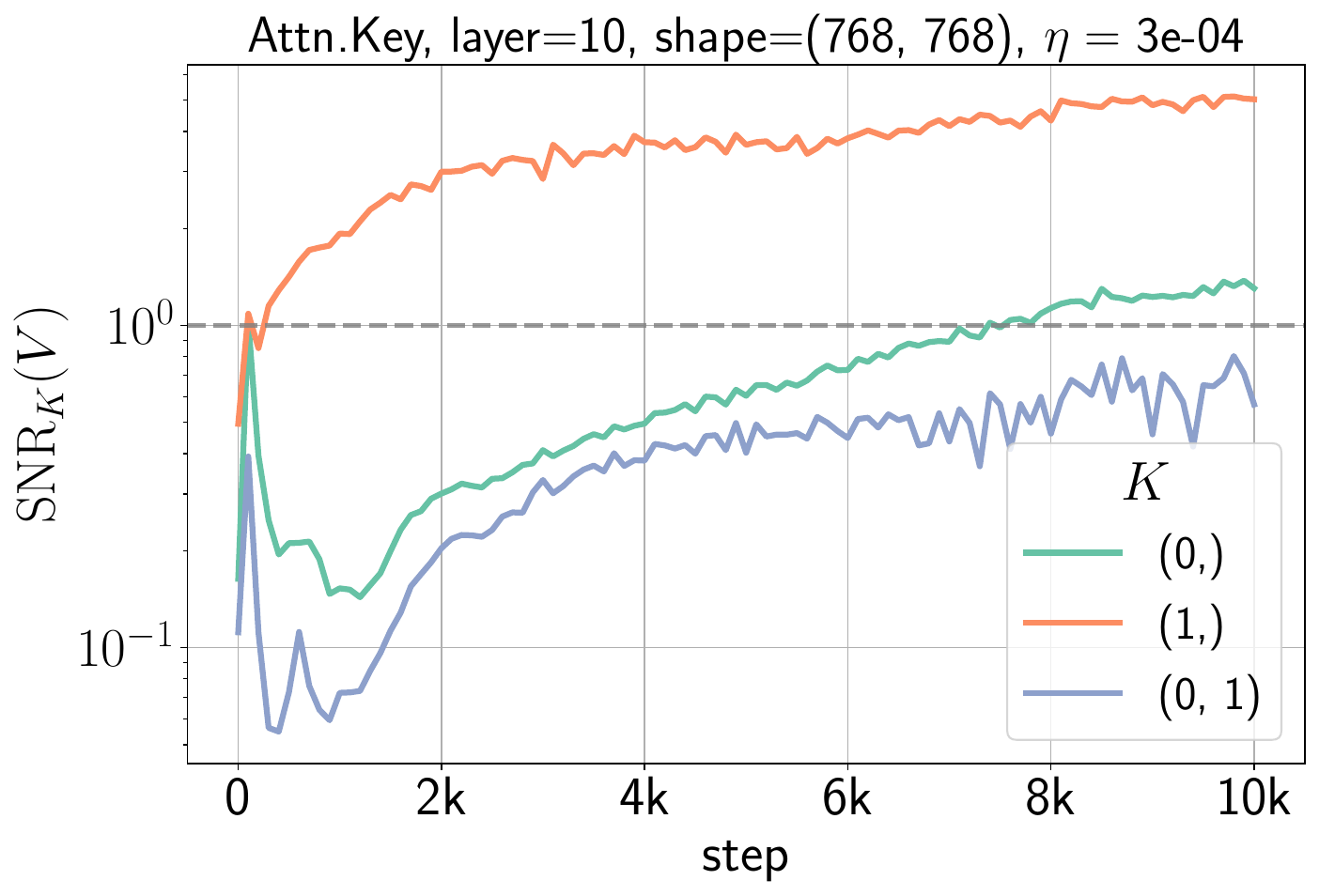}
\end{minipage}
\hfill
\begin{minipage}[b]{0.245\textwidth}
    \centering
    \includegraphics[width=\textwidth]{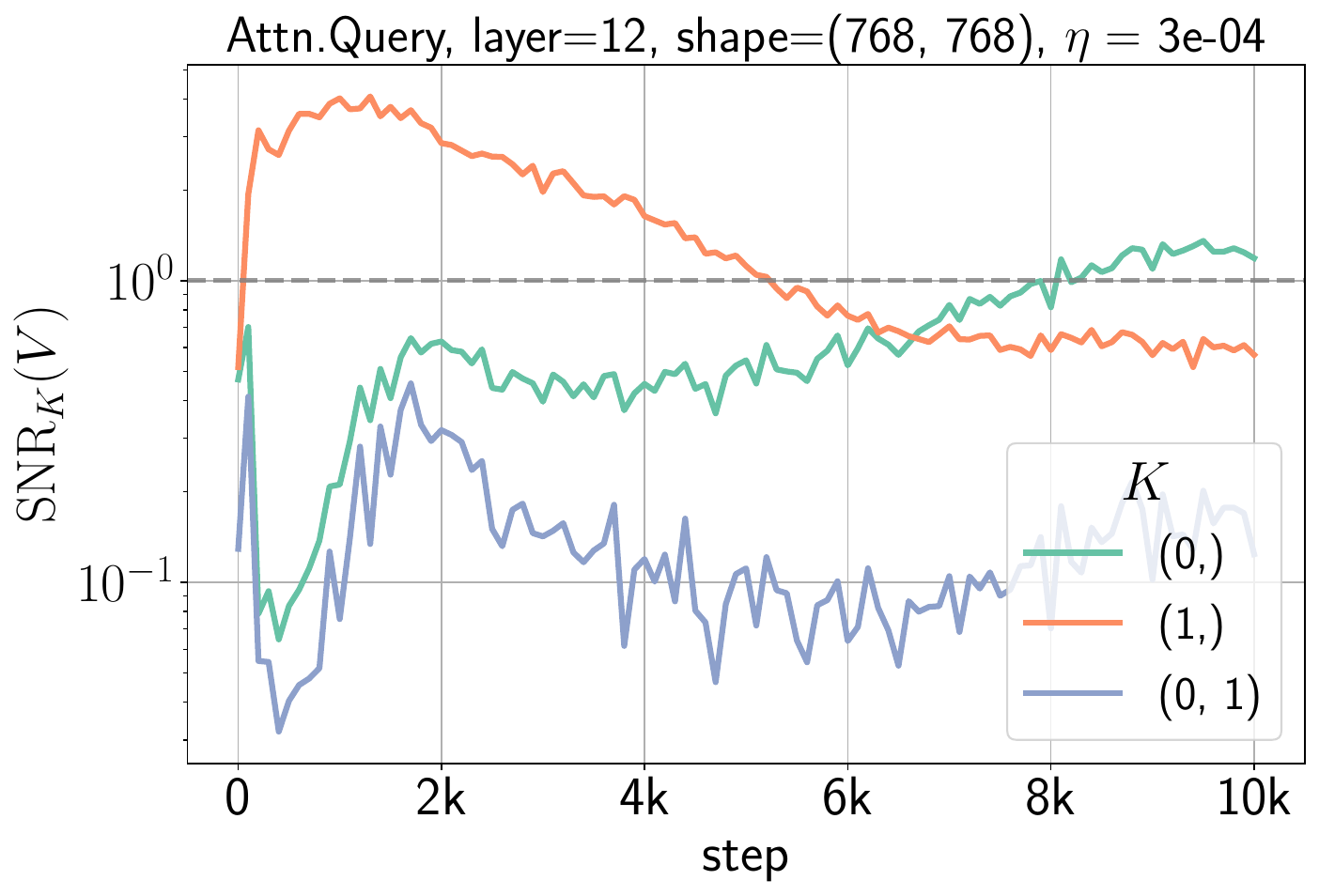}
\end{minipage}
\hfill
\begin{minipage}[b]{0.245\textwidth}
    \centering
    \includegraphics[width=\textwidth]{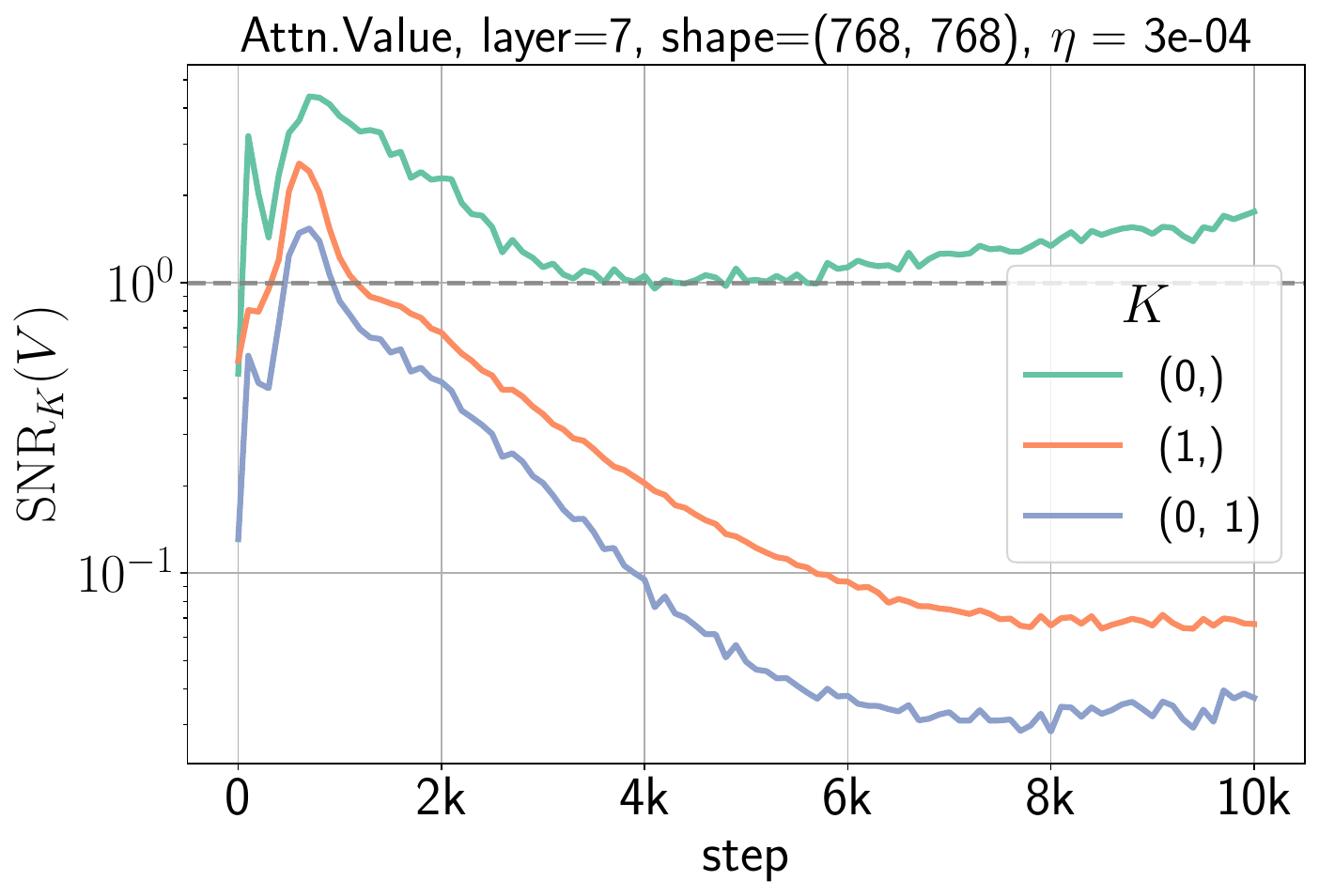}
\end{minipage}
\hfill
\begin{minipage}[b]{0.245\textwidth}
    \centering
    \includegraphics[width=\textwidth]{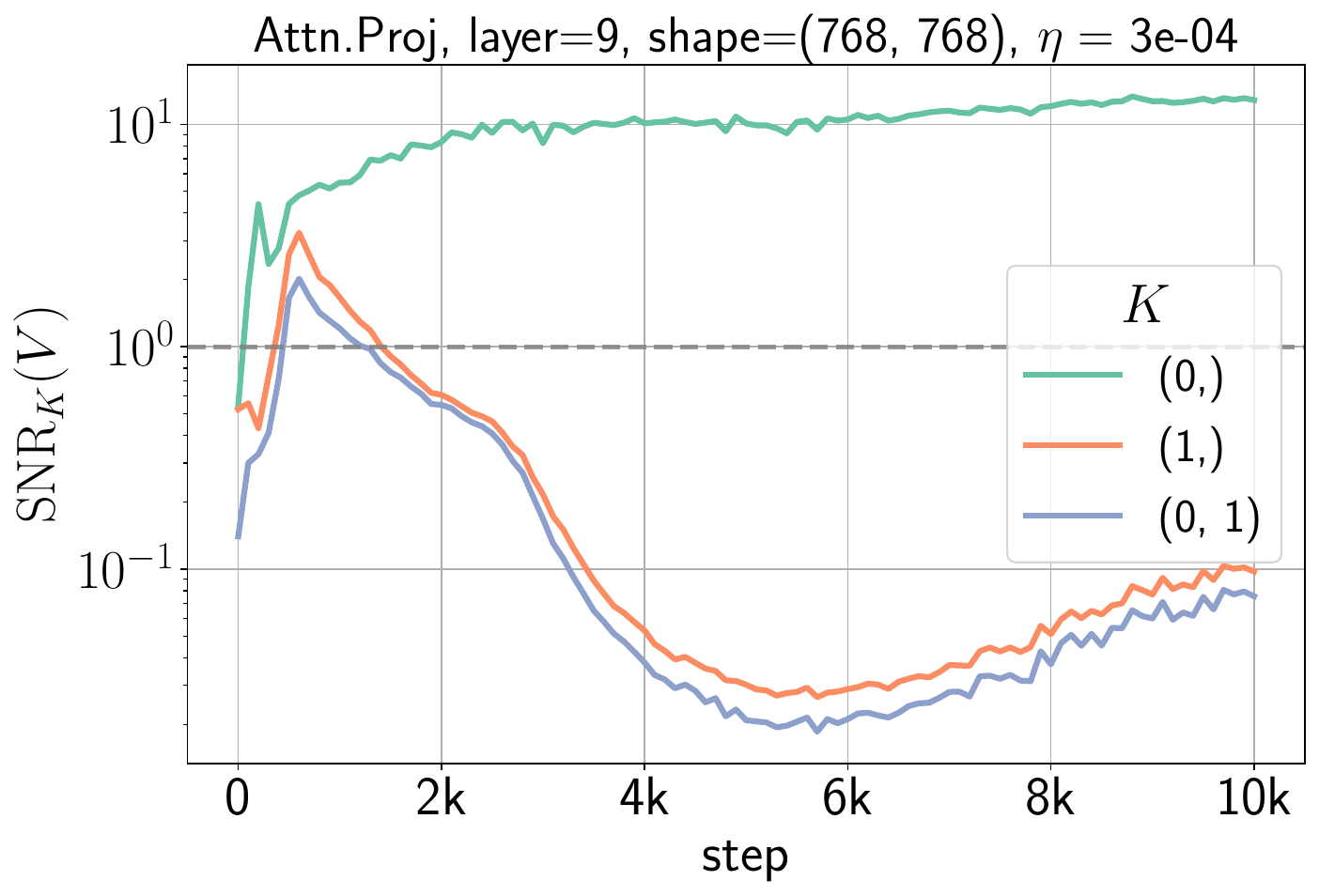}
\end{minipage}
\hfill
\begin{minipage}[b]{0.245\textwidth}
    \centering
    \includegraphics[width=\textwidth]{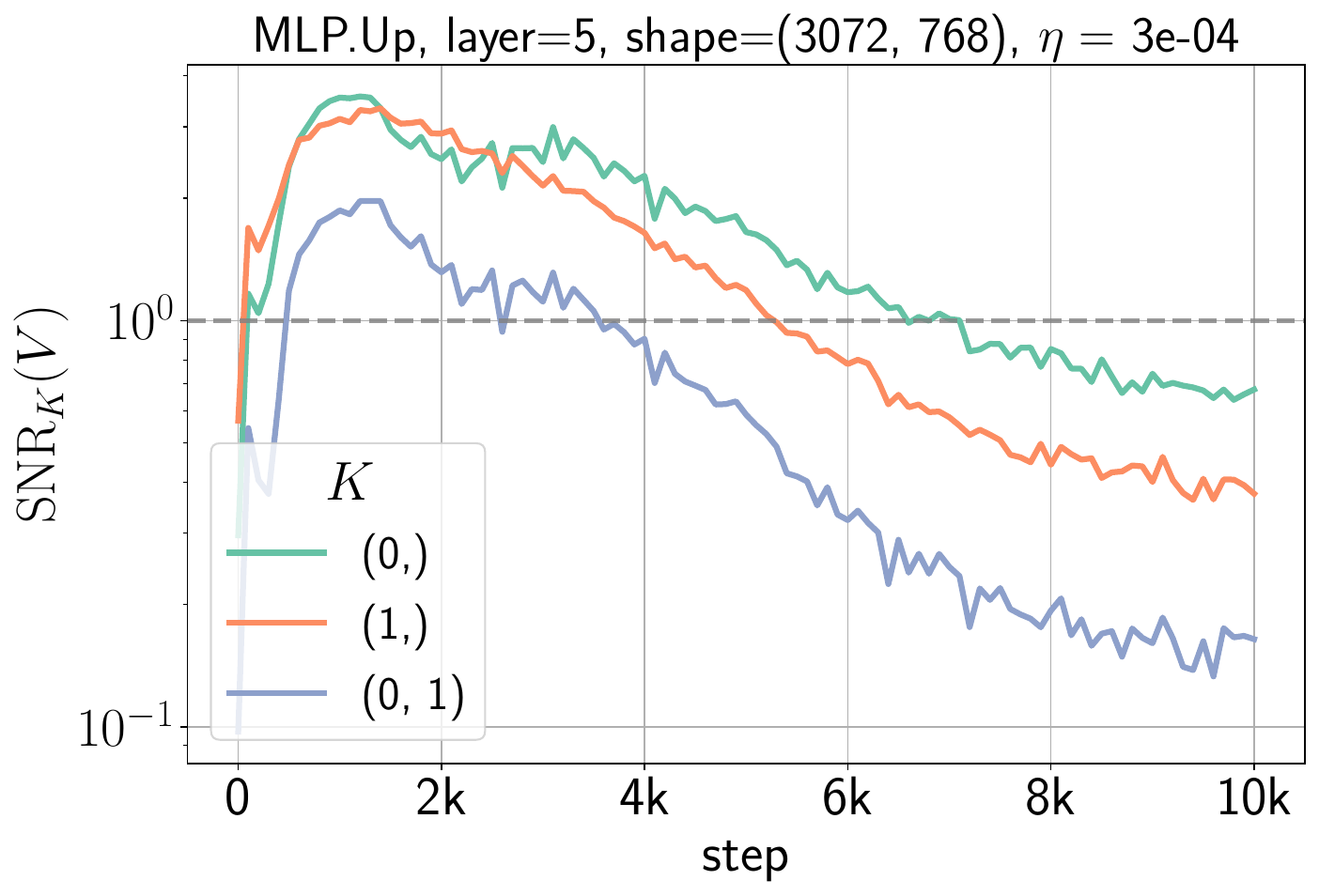}
\end{minipage}
\hfill
\begin{minipage}[b]{0.245\textwidth}
    \centering
    \includegraphics[width=\textwidth]{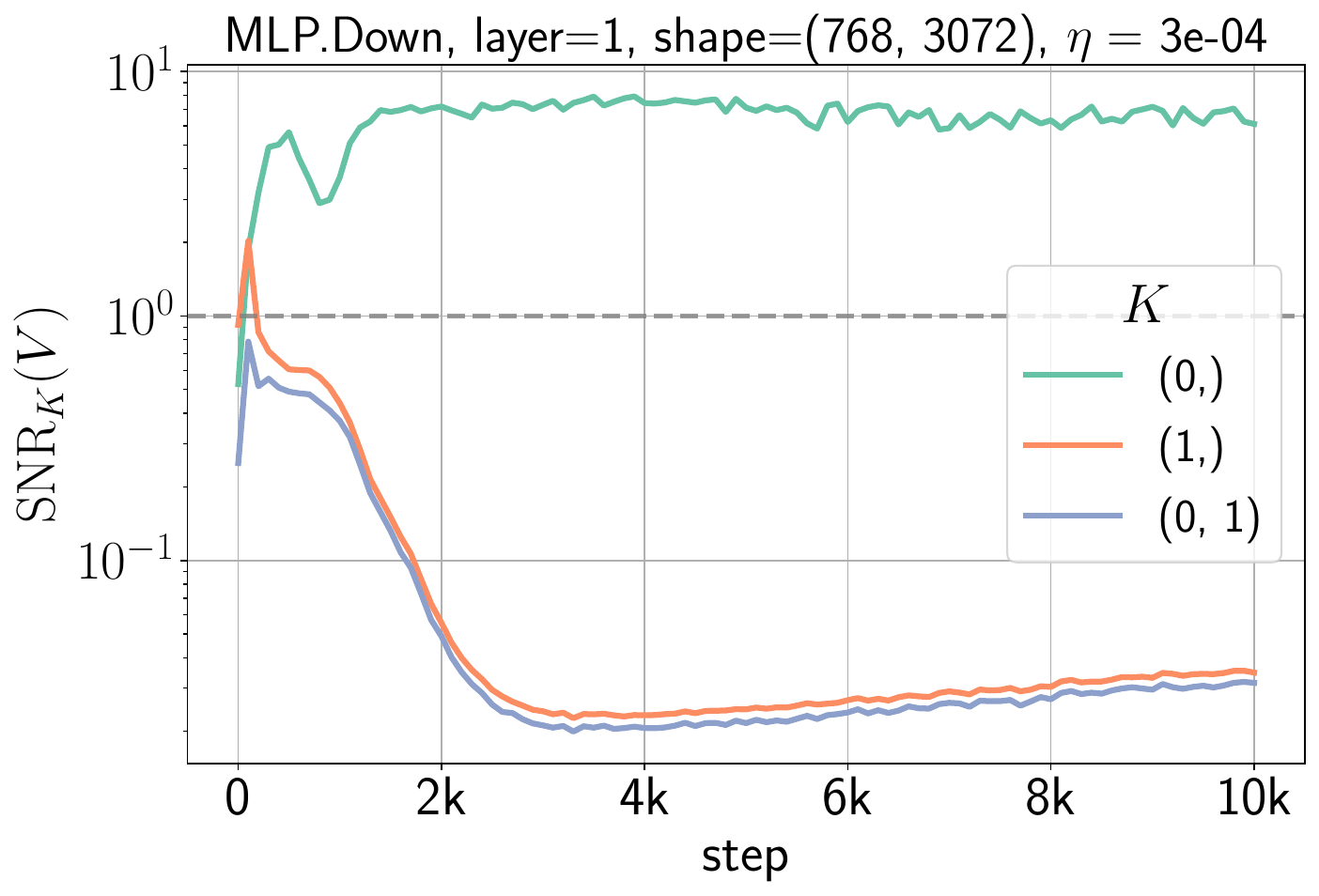}
\end{minipage}
\hfill
\begin{minipage}[b]{0.245\textwidth}
    \centering
    \includegraphics[width=\textwidth]{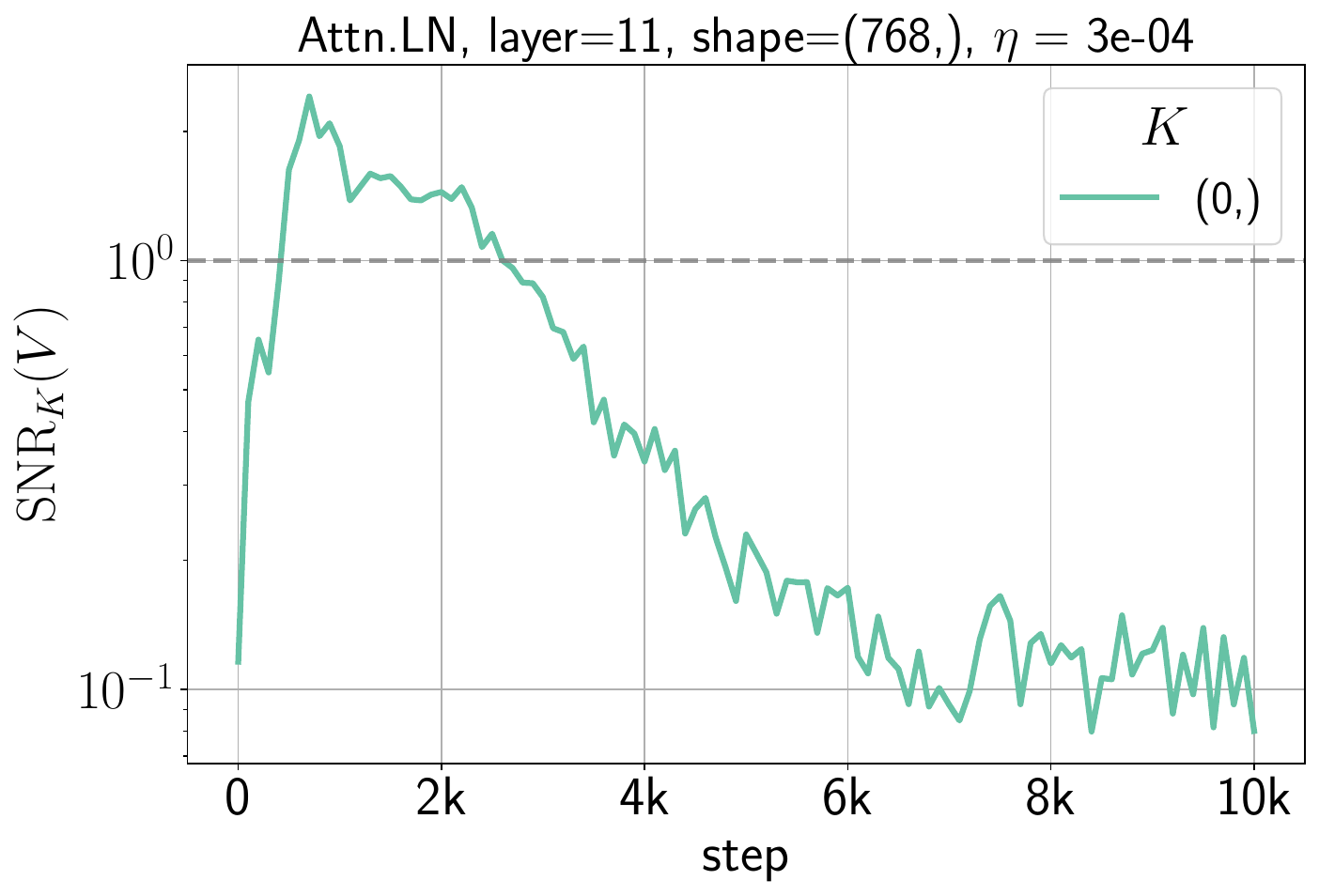}
\end{minipage}
\hfill
\begin{minipage}[b]{0.245\textwidth}
    \centering
    \includegraphics[width=\textwidth]{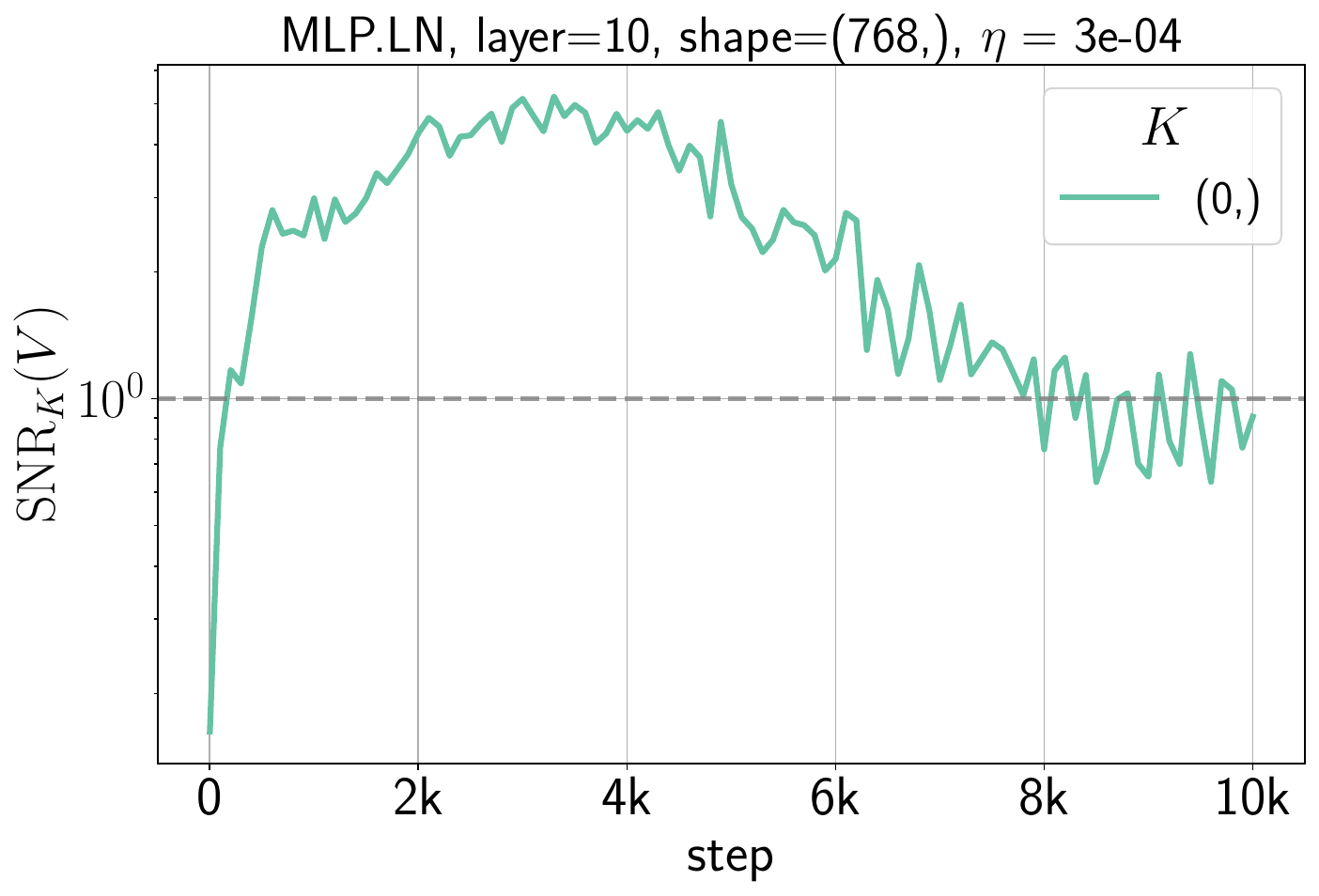}
\end{minipage}

\begin{minipage}[b]{0.245\textwidth}
    \centering
    \includegraphics[width=\textwidth]{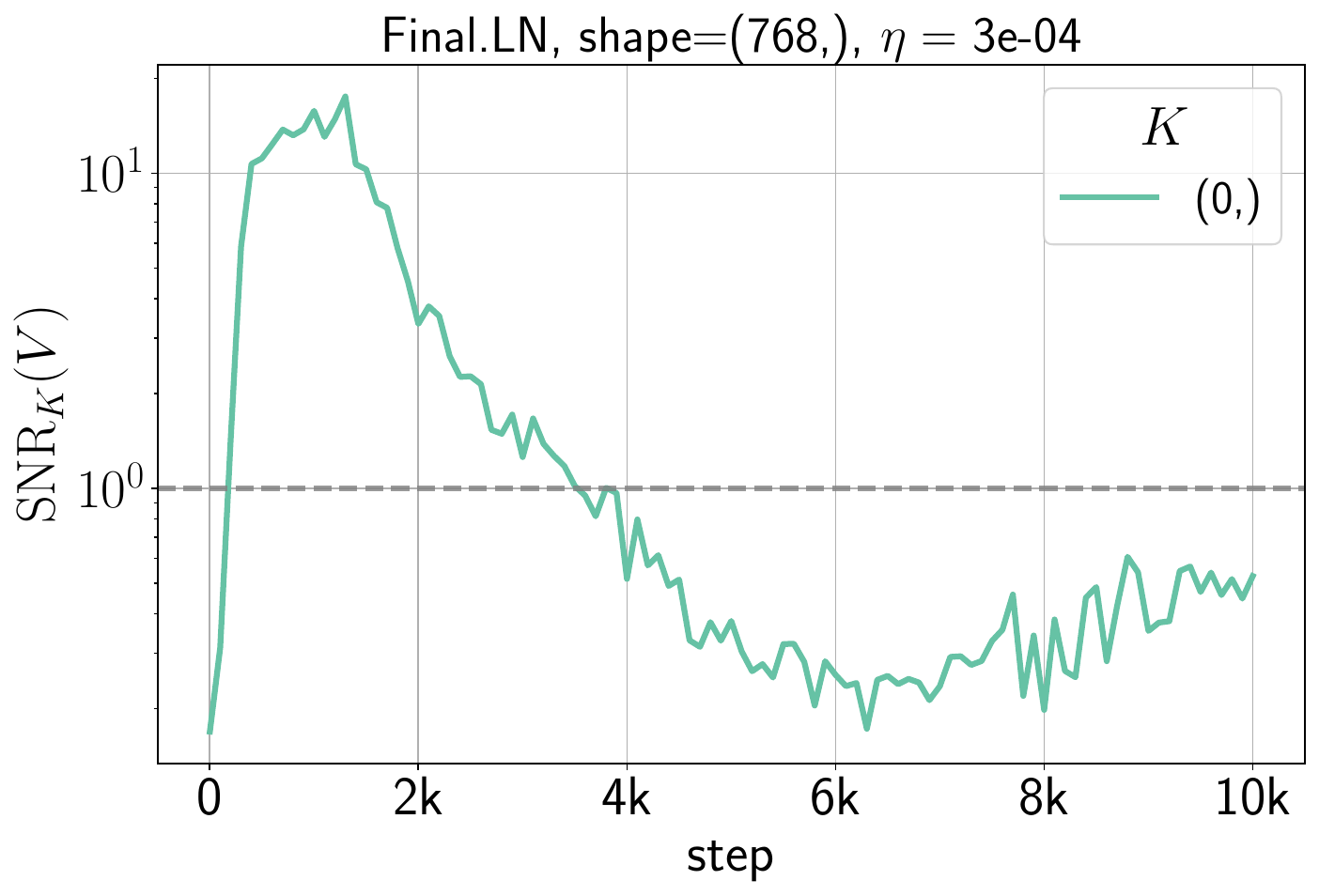}
\end{minipage}
\begin{minipage}[b]{0.245\textwidth}
    \centering
    \includegraphics[width=\textwidth]{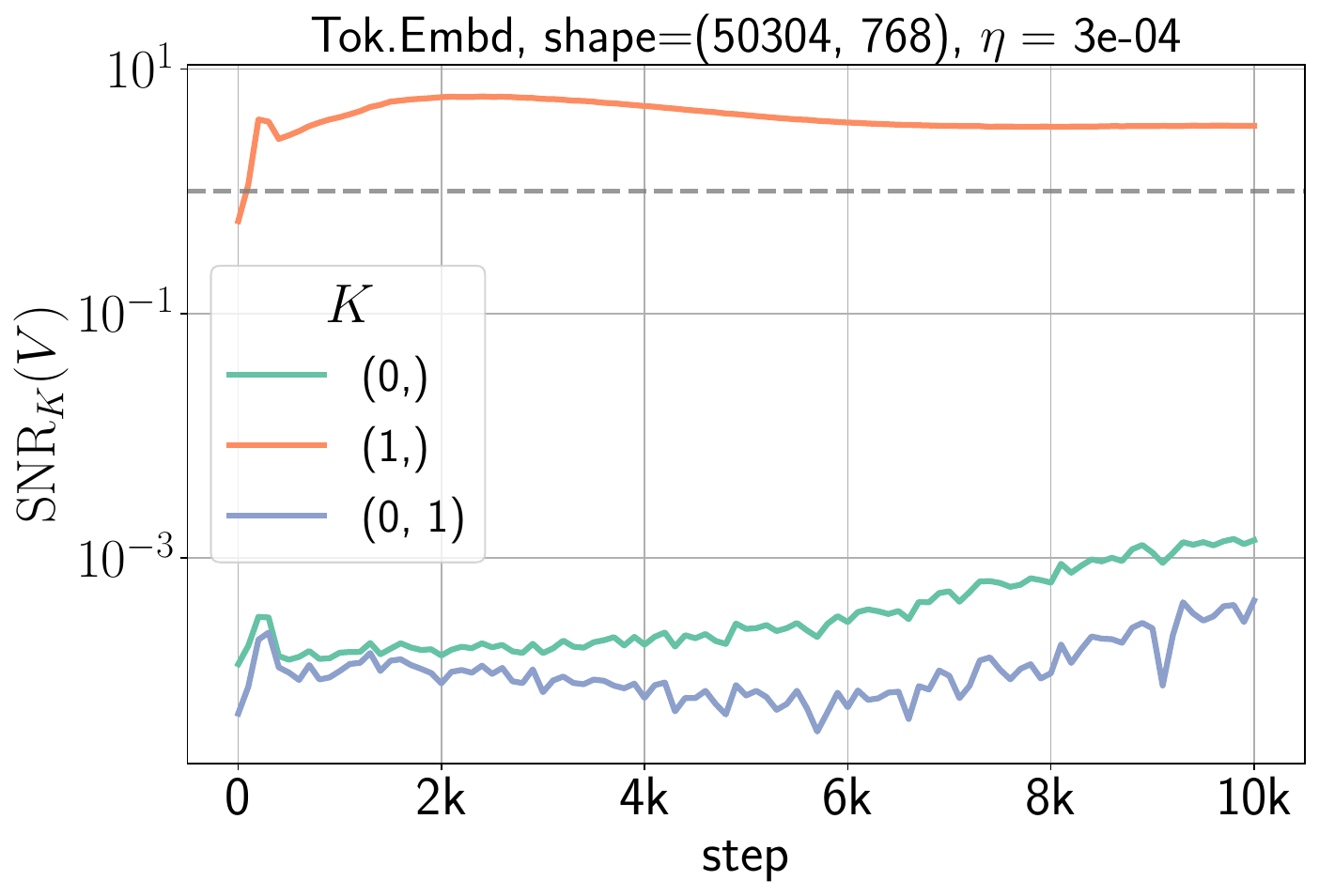}
\end{minipage}
\begin{minipage}[b]{0.245\textwidth}
    \centering
    \includegraphics[width=\textwidth]{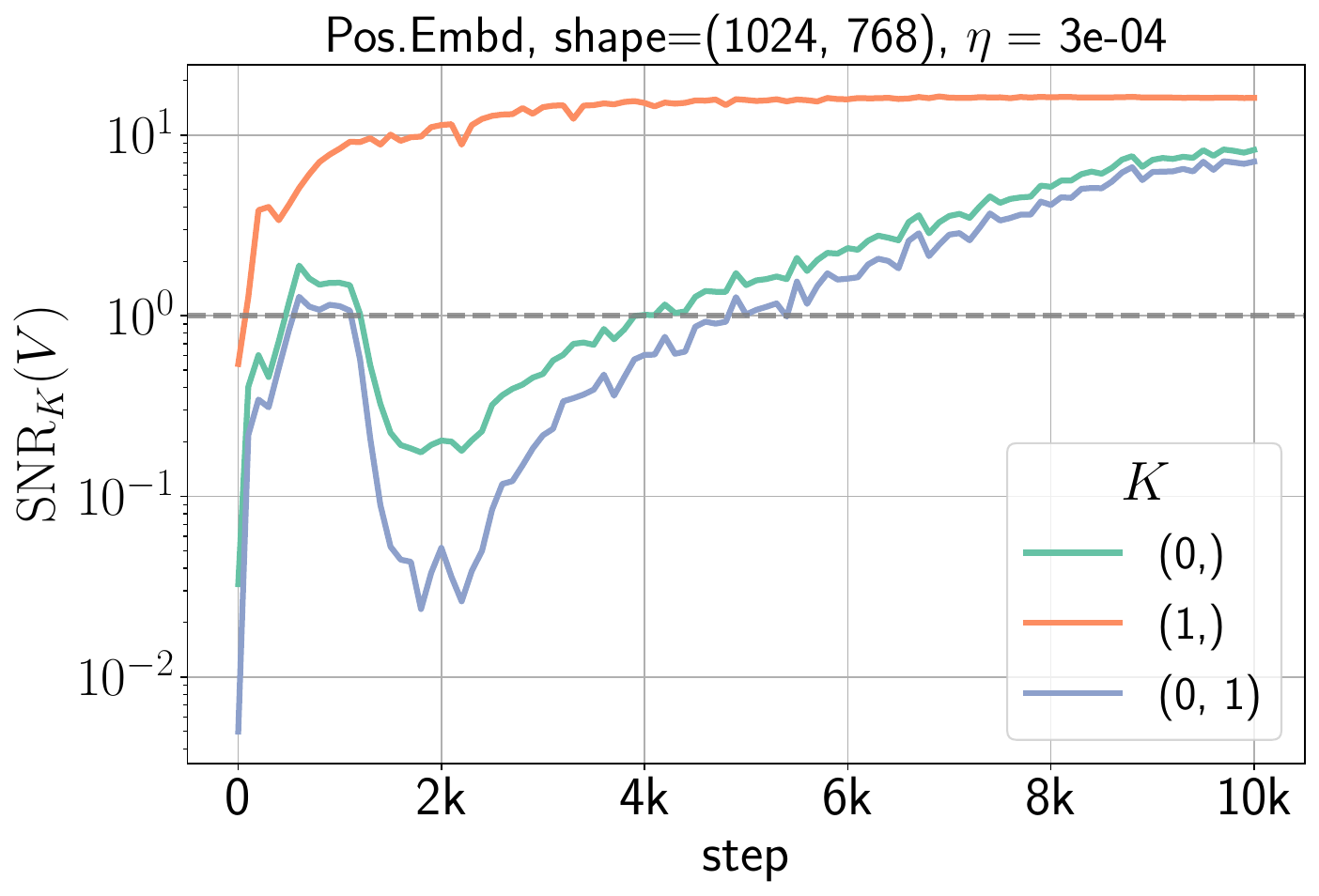}
\end{minipage}

\caption{SNR trajectories of GPT-small trained on $10$B subset of FineWeb-Edu. For each layer type, the layer number is selected at random.}
\label{fig:snr-curves-gpt-small-fineweb-full}
\end{figure*}

\begin{figure*}[!htb]
\centering
\begin{minipage}[b]{0.245\textwidth}
    \centering
    \includegraphics[width=\textwidth]{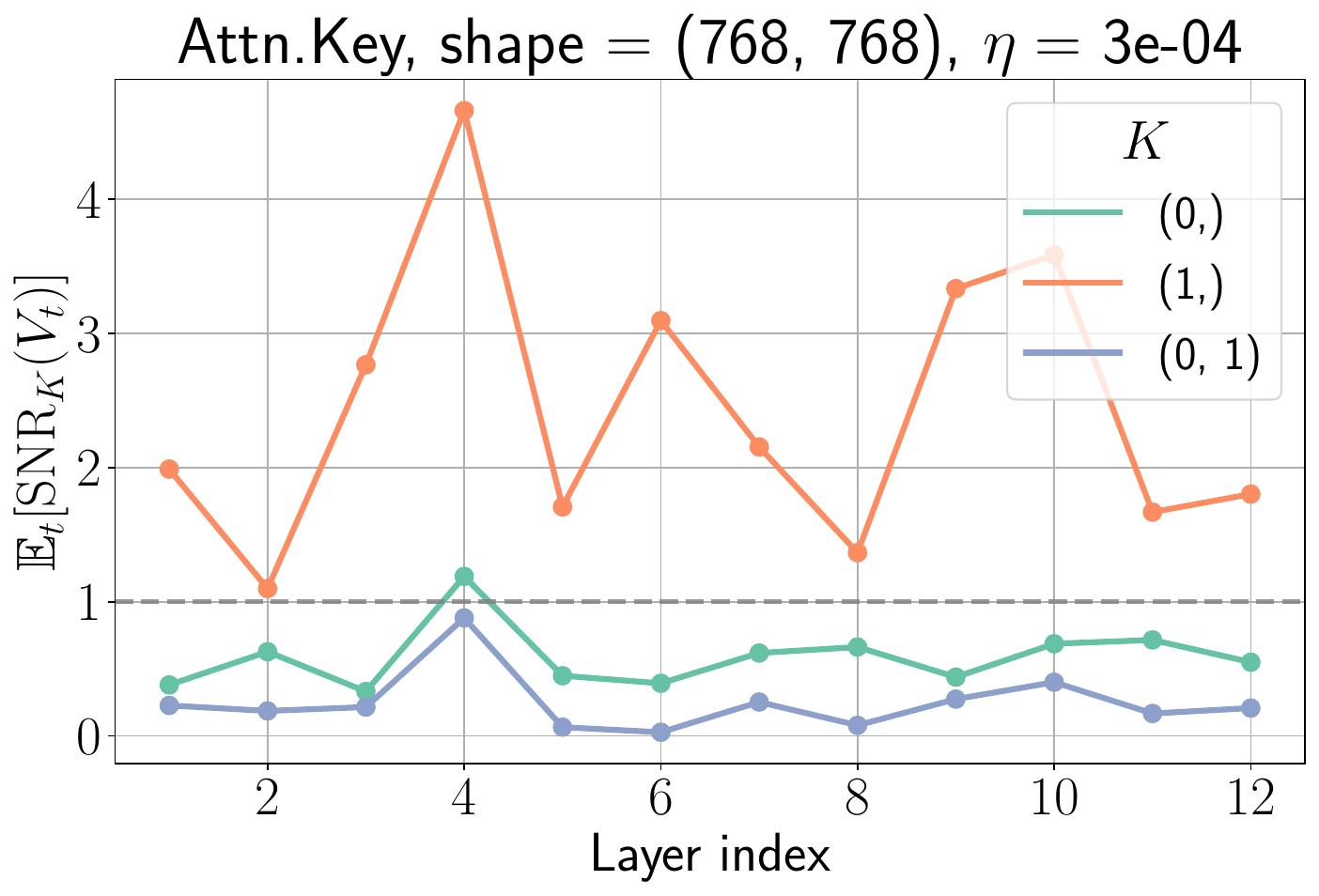}
\end{minipage}
\hfill
\begin{minipage}[b]{0.245\textwidth}
    \centering
    \includegraphics[width=\textwidth]{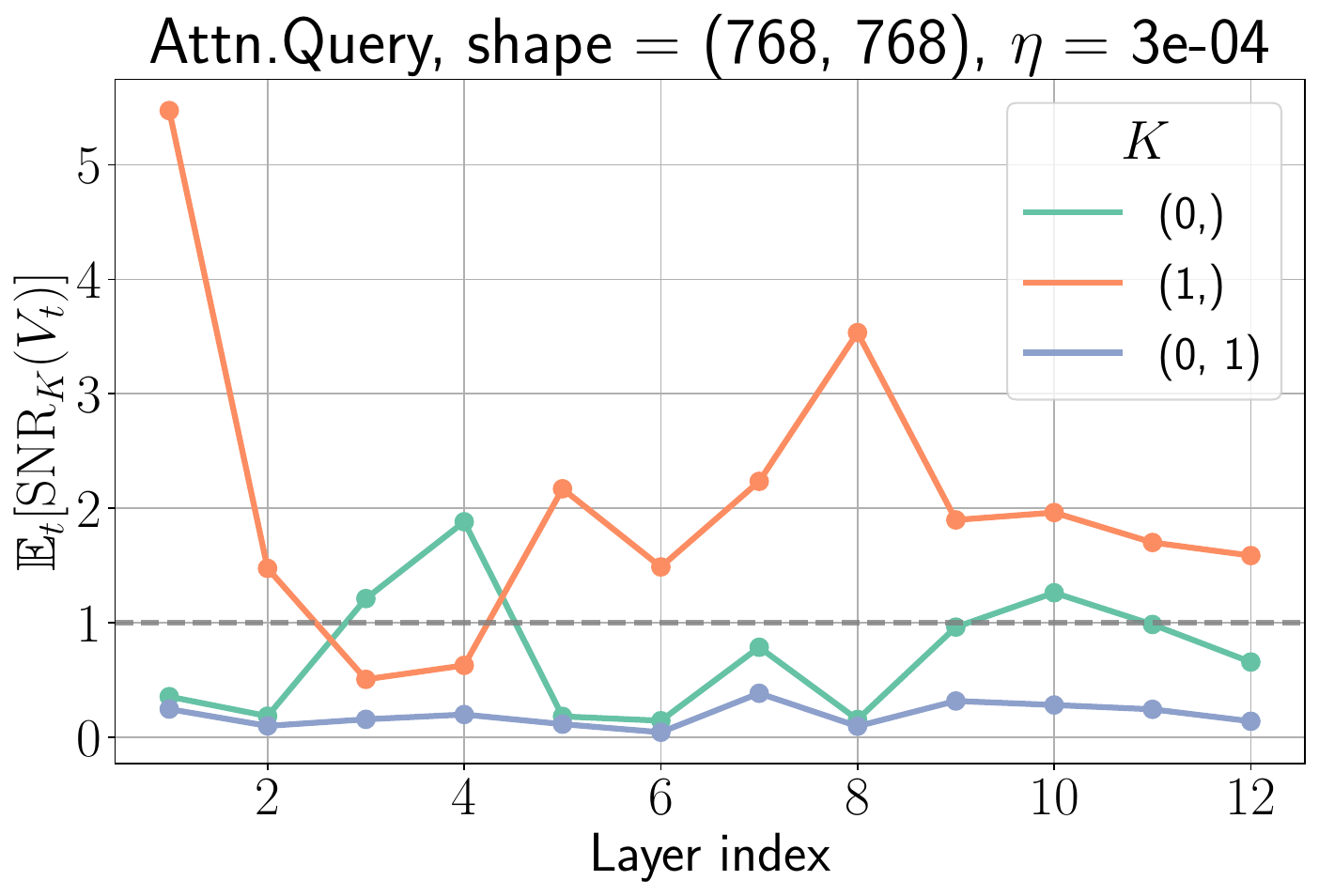}
\end{minipage}
\hfill
\begin{minipage}[b]{0.245\textwidth}
    \centering
    \includegraphics[width=\textwidth]{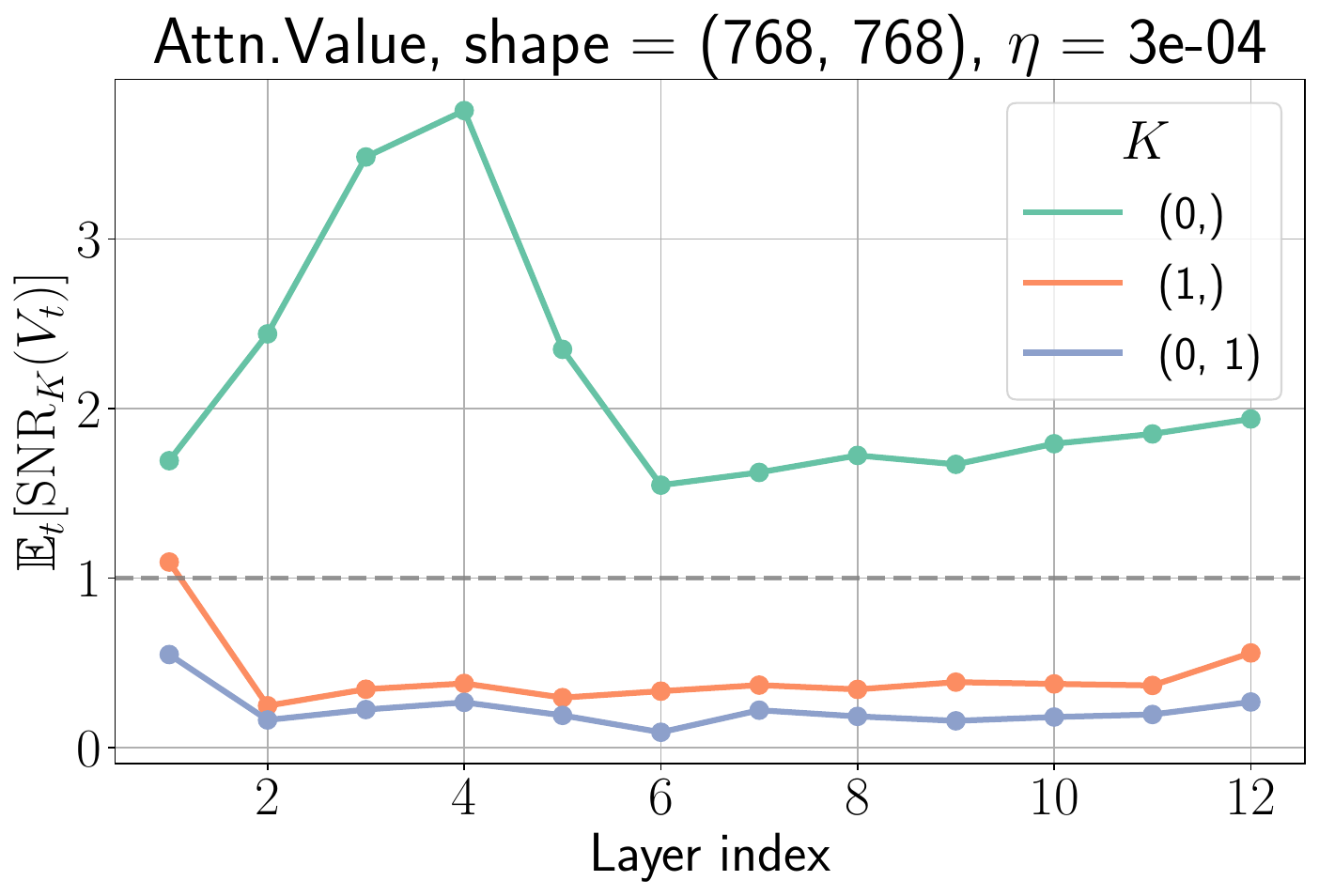}
\end{minipage}
\hfill
\begin{minipage}[b]{0.245\textwidth}
    \centering
    \includegraphics[width=\textwidth]{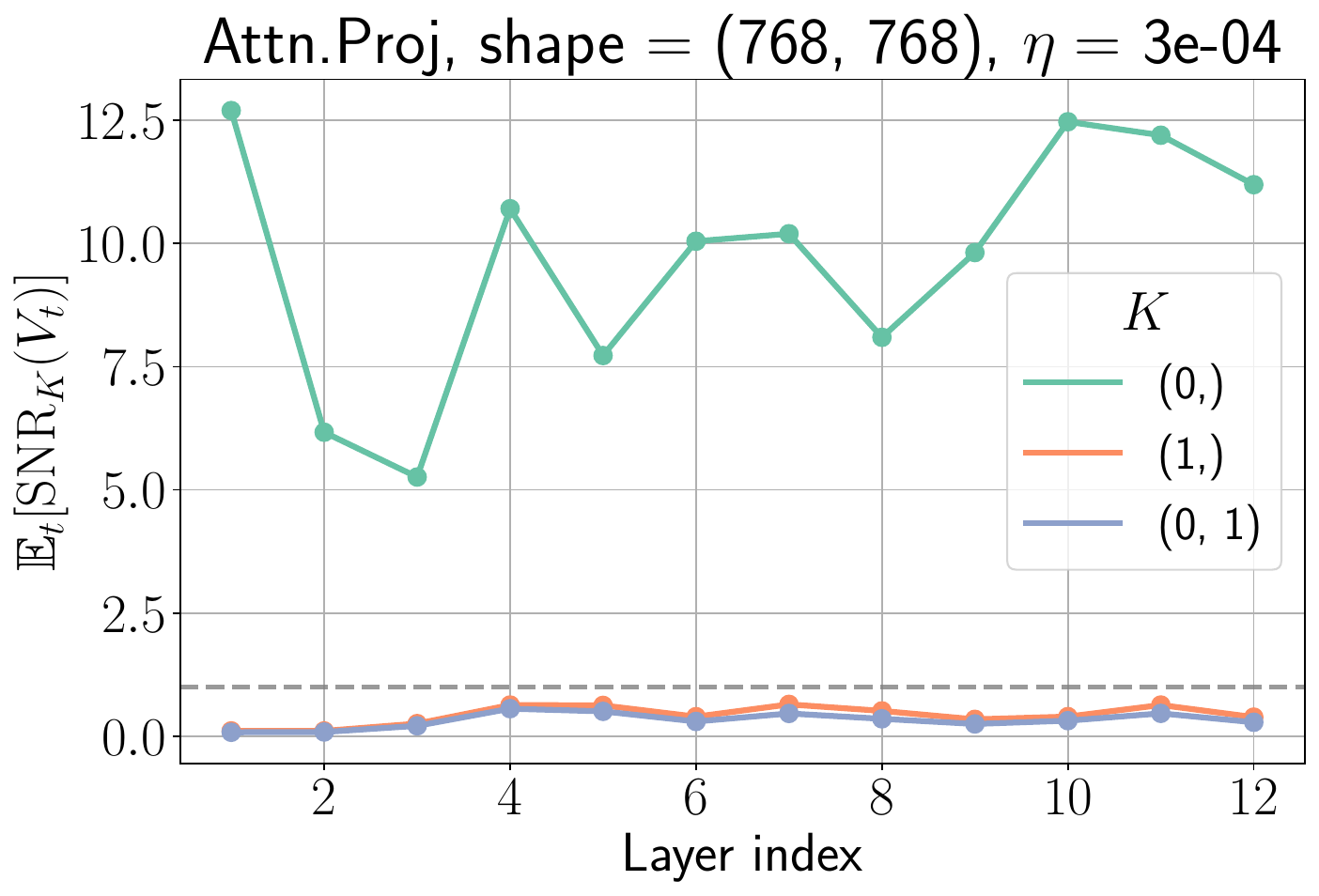}
\end{minipage}

\begin{minipage}[b]{0.245\textwidth}
    \centering
    \includegraphics[width=\textwidth]{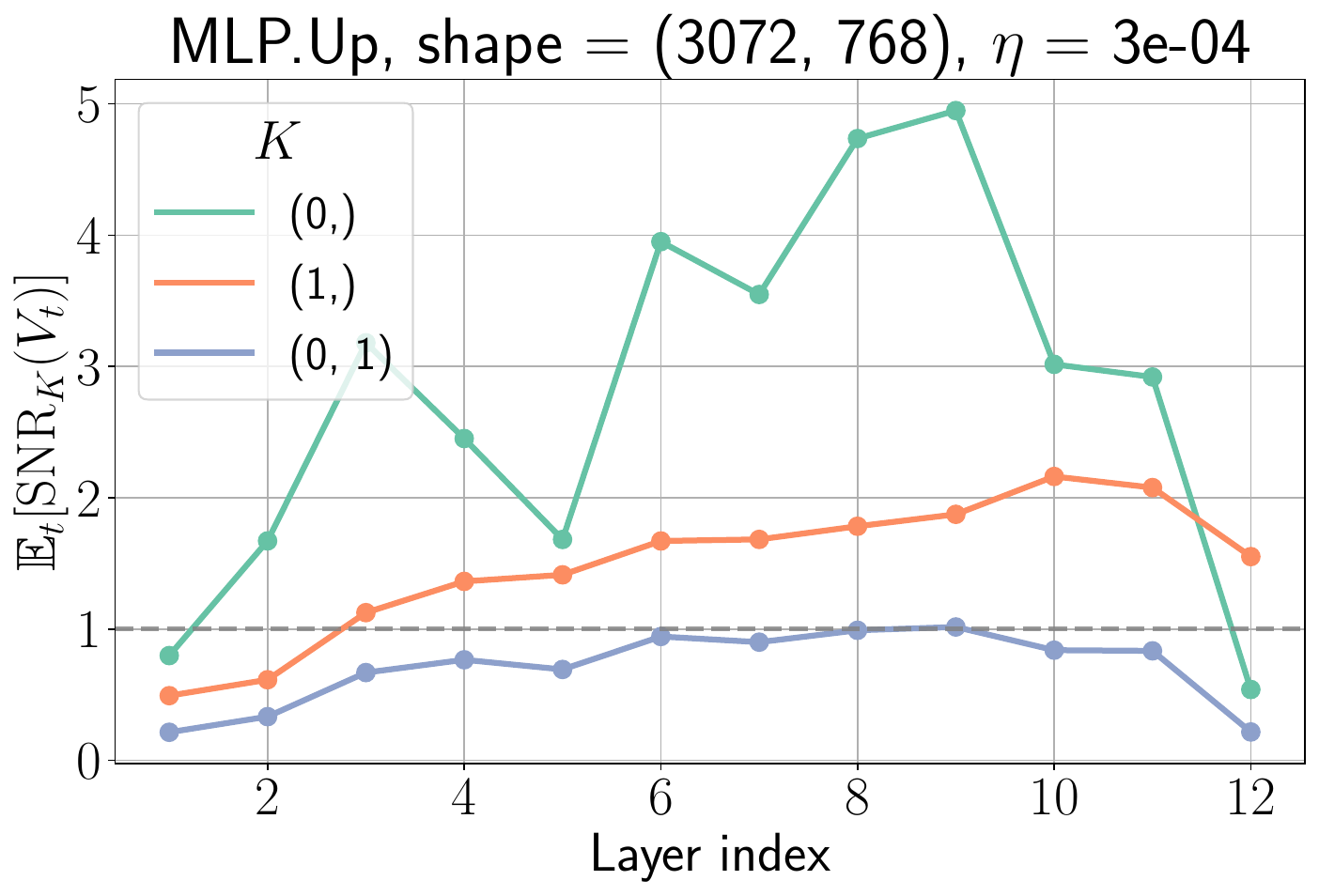}
\end{minipage}
\hfill
\begin{minipage}[b]{0.245\textwidth}
    \centering
    \includegraphics[width=\textwidth]{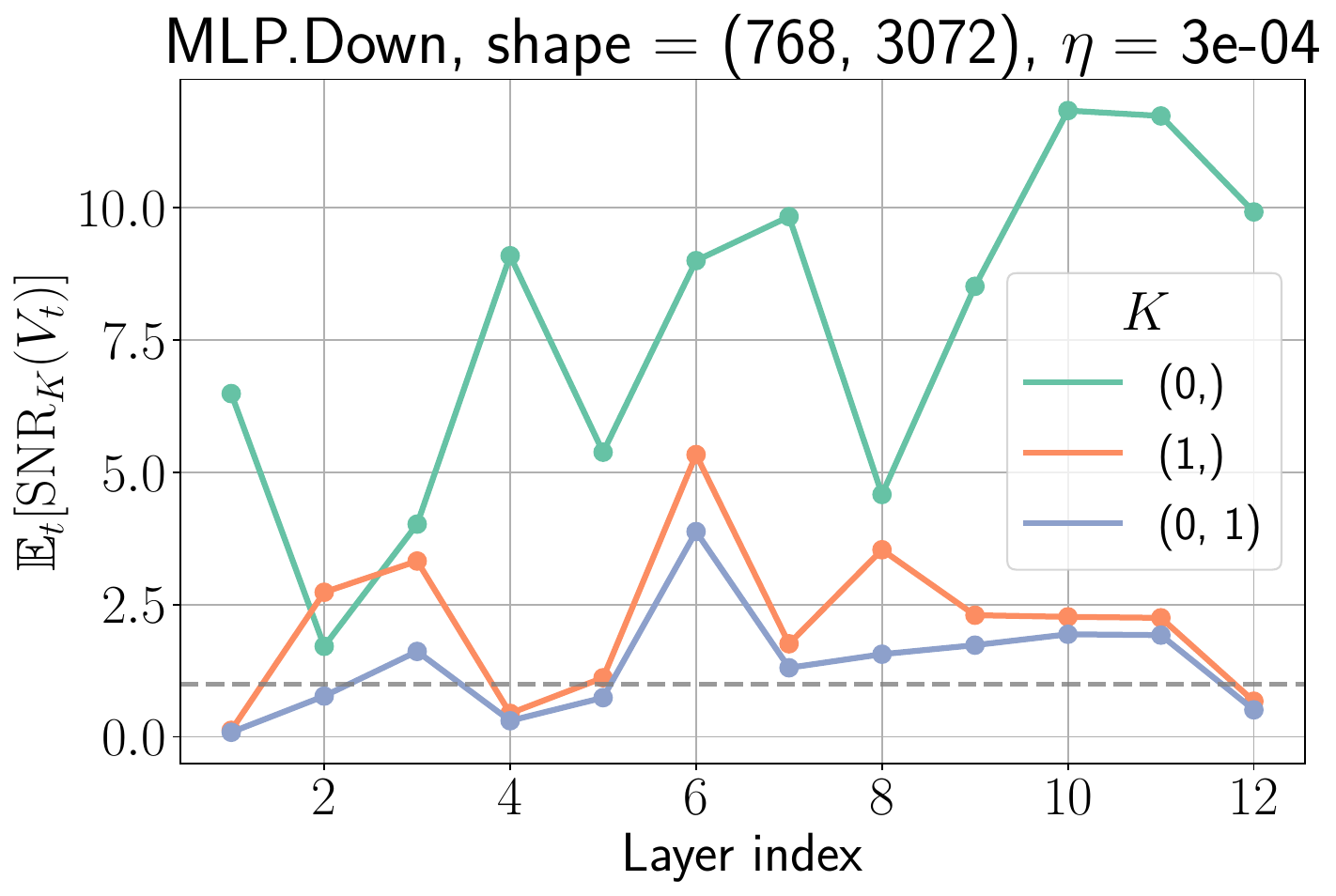}
\end{minipage}
\hfill
\begin{minipage}[b]{0.245\textwidth}
    \centering
    \includegraphics[width=\textwidth]{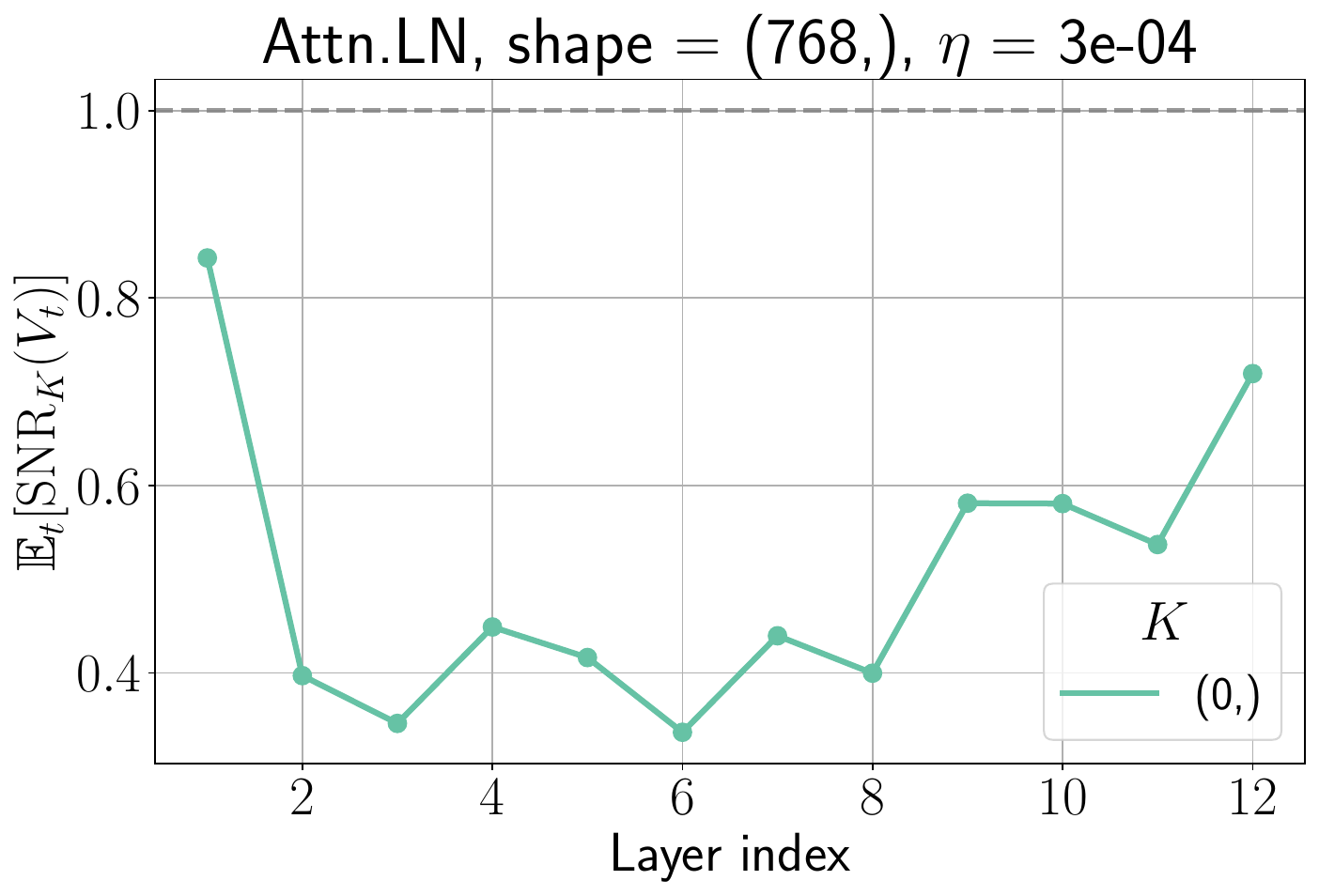}
\end{minipage}
\hfill
\begin{minipage}[b]{0.245\textwidth}
    \centering
    \includegraphics[width=\textwidth]{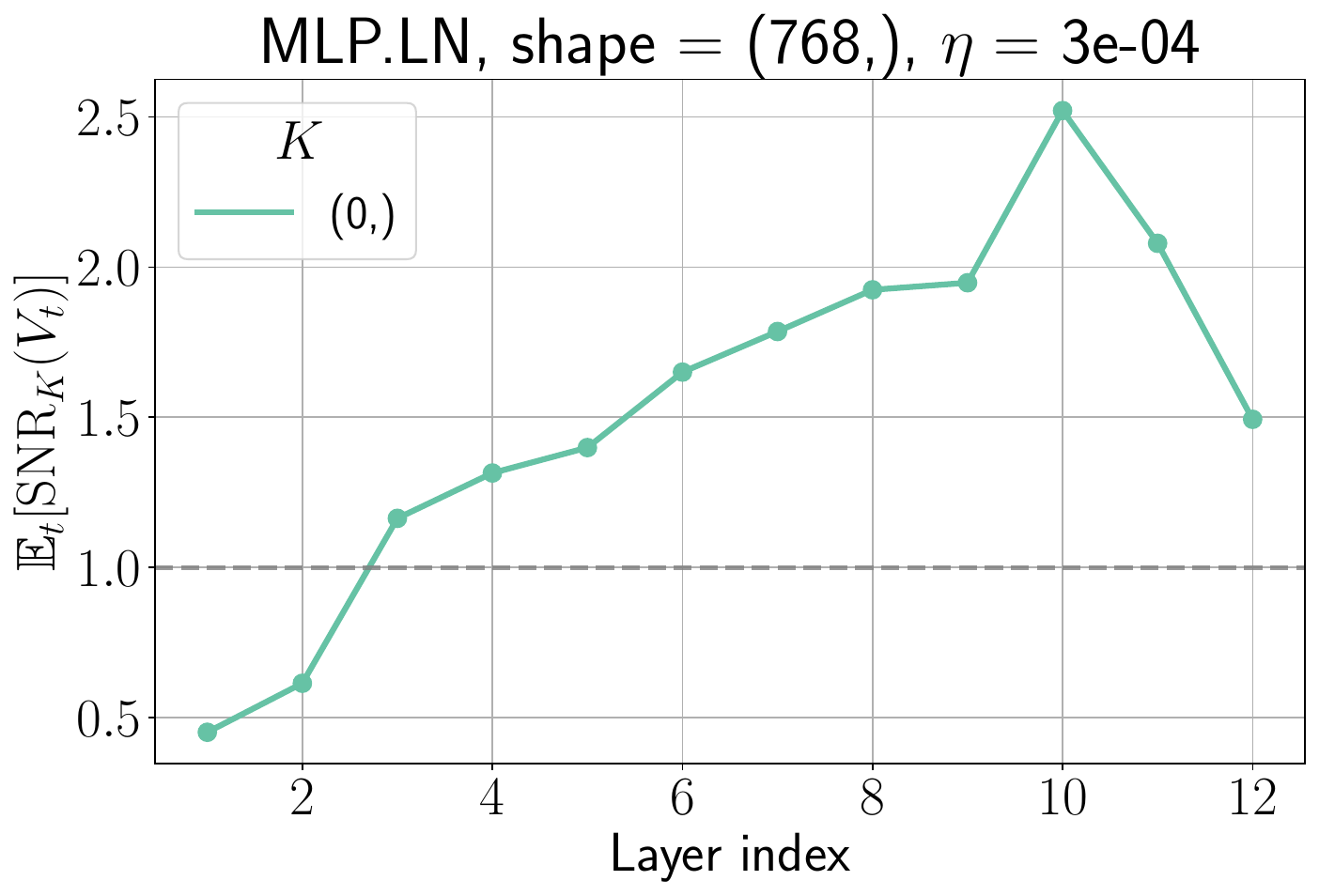}
\end{minipage}
\caption{Layer dependence of averaged SNR values of GPT-small trained on $10$B token subset of FineWeb-Edu.}
\label{fig:snr-layer-gpt-small-fineweb-full}
\end{figure*}

\begin{figure*}[!htbt]
\centering
\begin{minipage}[b]{0.245\textwidth}
    \centering
    \includegraphics[width=\textwidth]{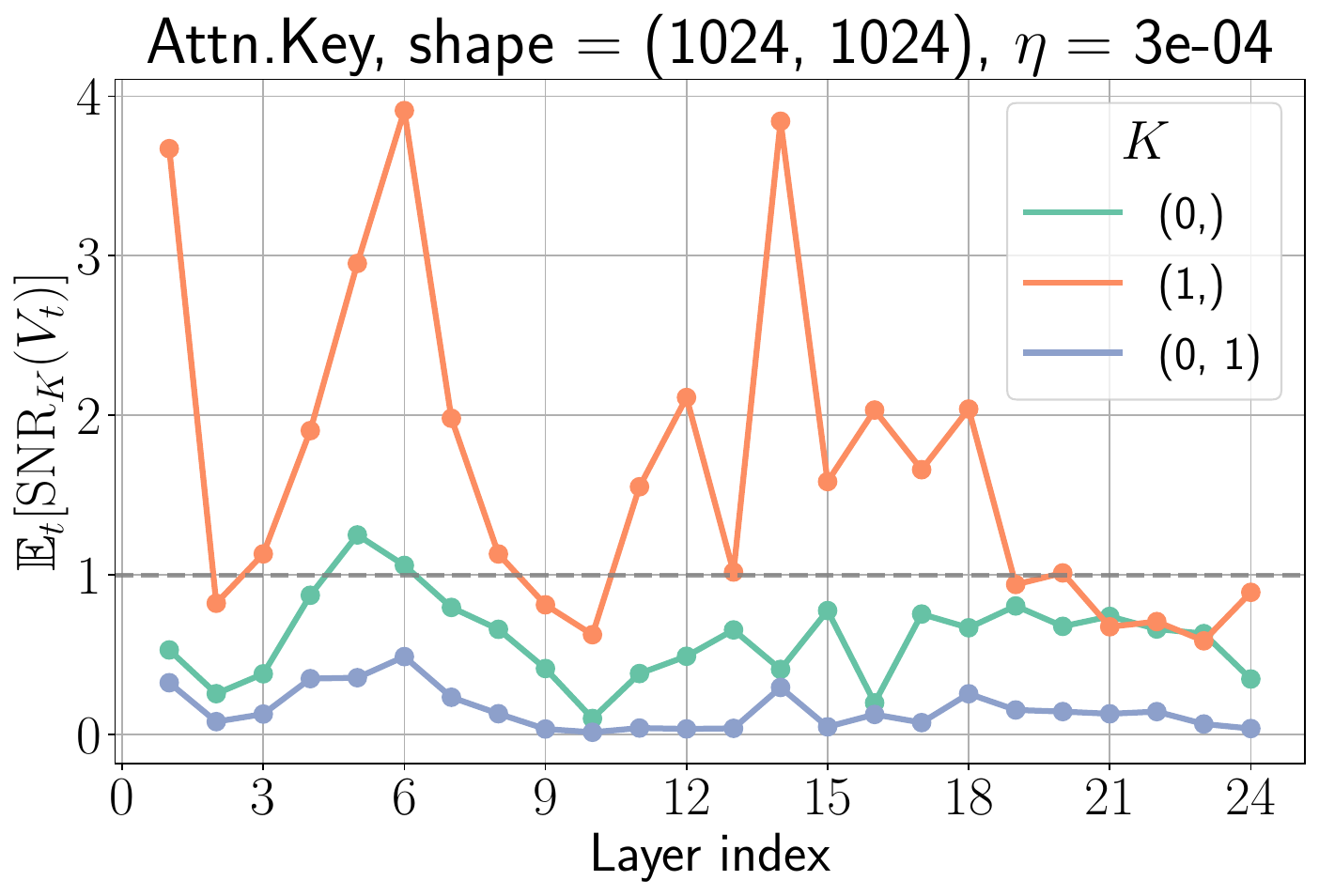}
\end{minipage}
\hfill
\begin{minipage}[b]{0.245\textwidth}
    \centering
    \includegraphics[width=\textwidth]{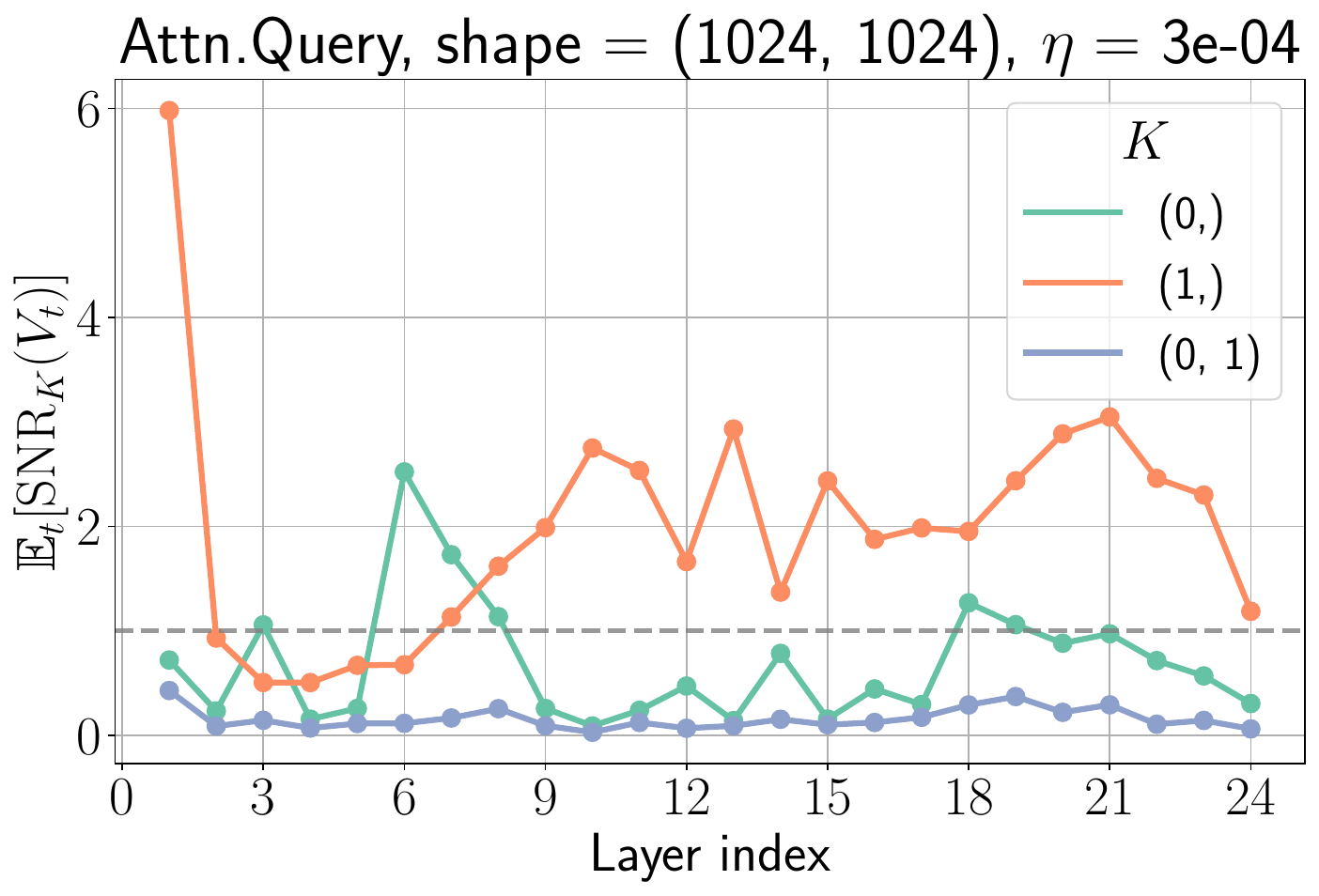}
\end{minipage}
\hfill
\begin{minipage}[b]{0.245\textwidth}
    \centering
    \includegraphics[width=\textwidth]{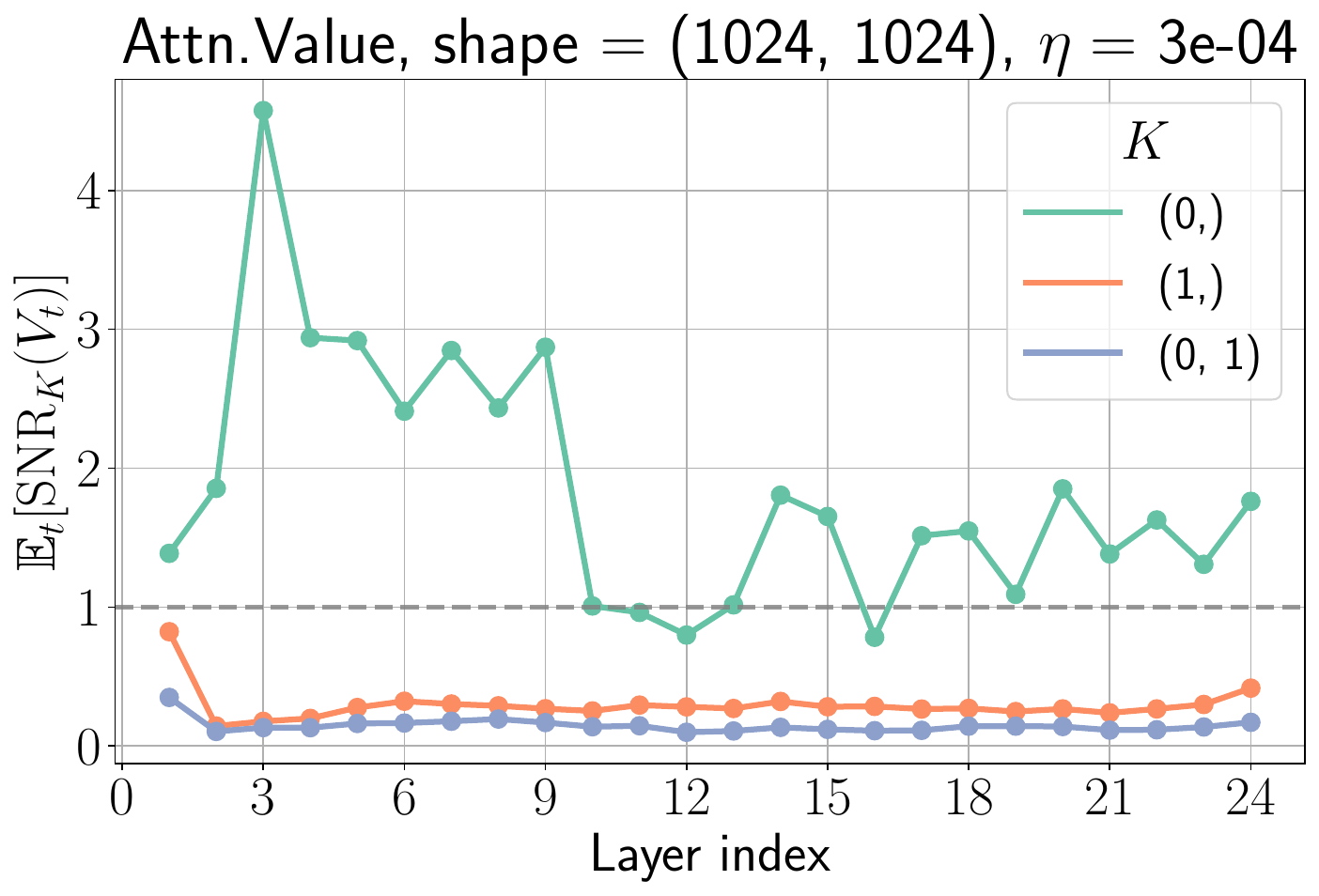}
\end{minipage}
\hfill
\begin{minipage}[b]{0.245\textwidth}
    \centering
    \includegraphics[width=\textwidth]{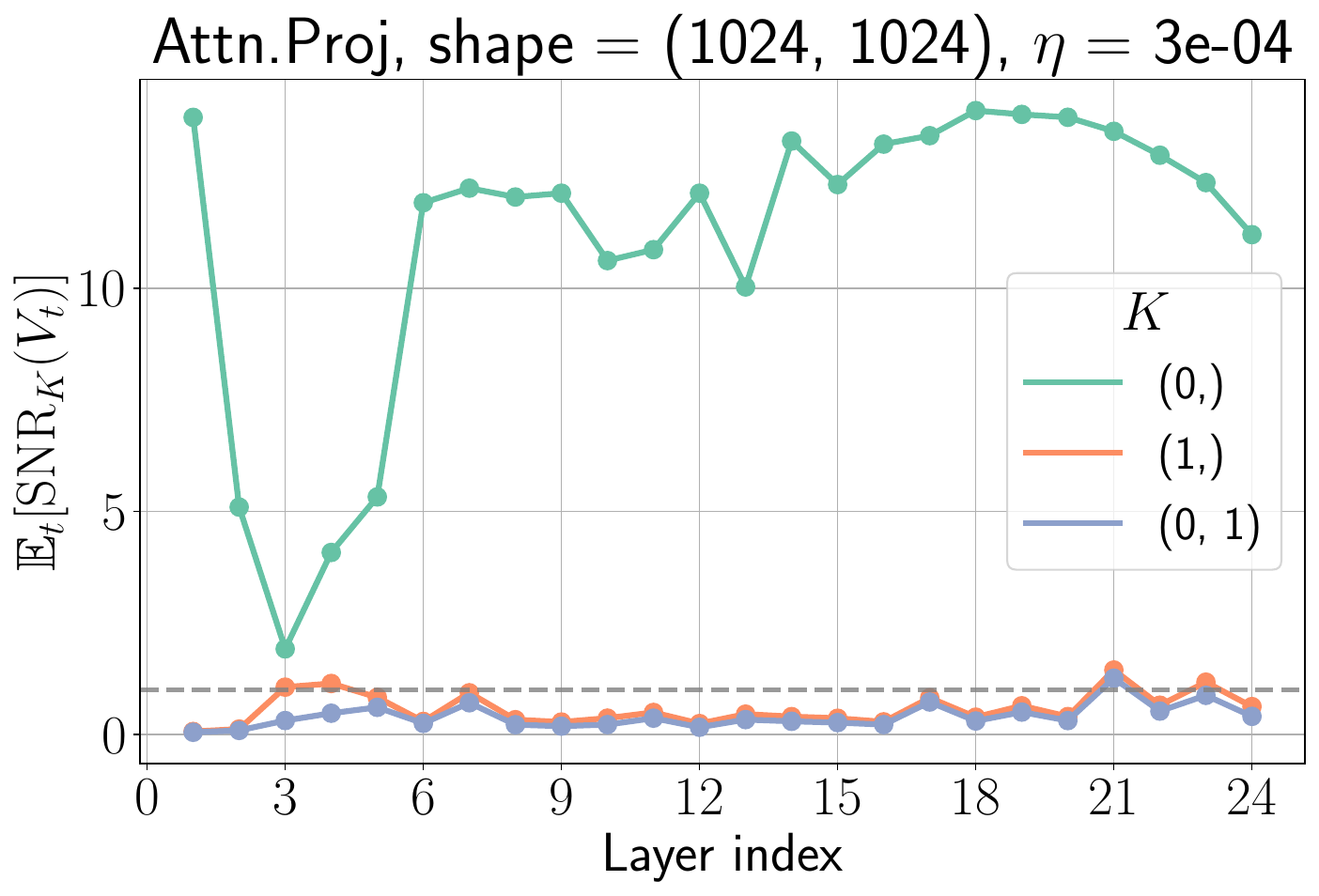}
\end{minipage}

\begin{minipage}[b]{0.245\textwidth}
    \centering
    \includegraphics[width=\textwidth]{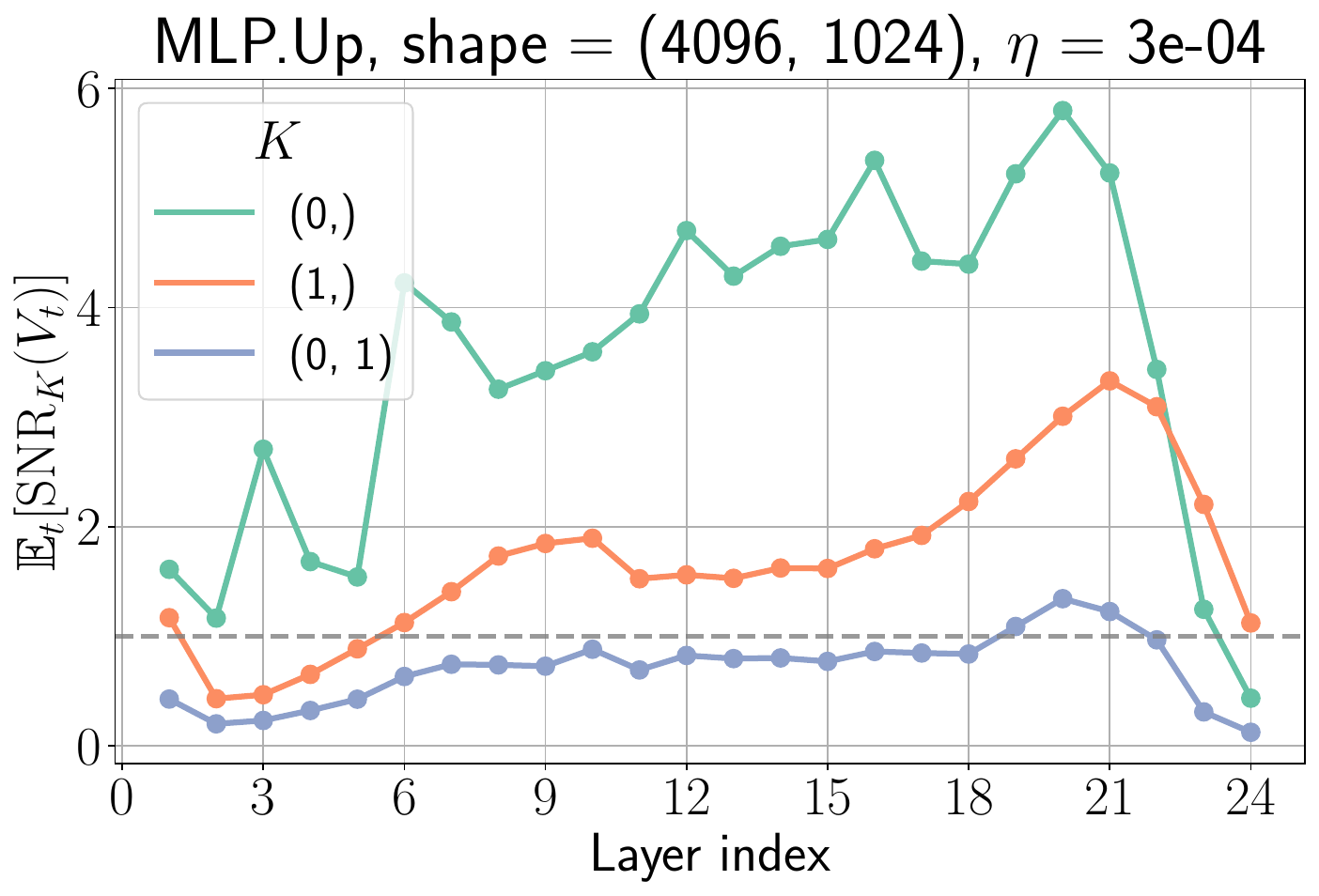}
\end{minipage}
\hfill
\begin{minipage}[b]{0.245\textwidth}
    \centering
    \includegraphics[width=\textwidth]{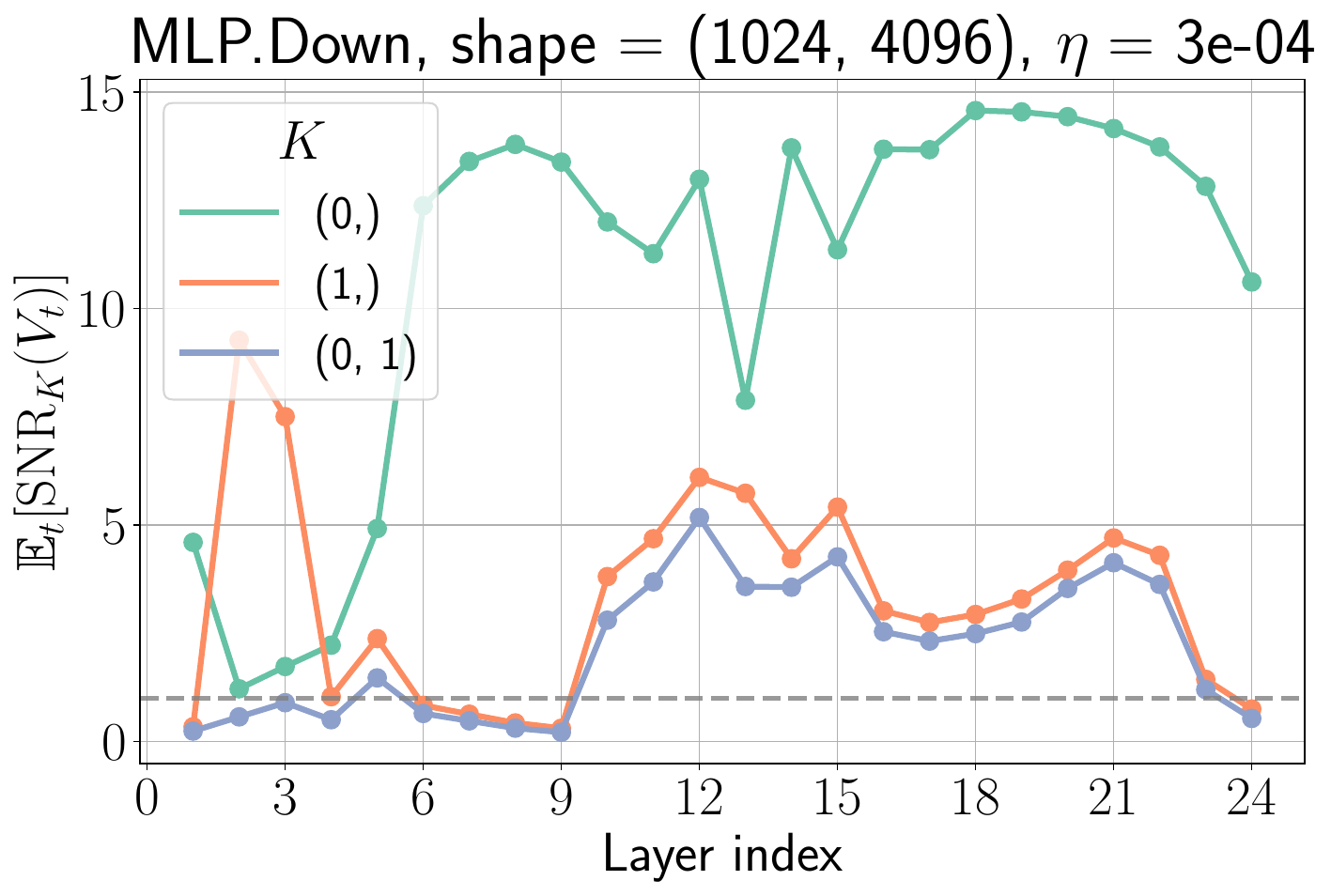}
\end{minipage}
\hfill
\begin{minipage}[b]{0.245\textwidth}
    \centering
    \includegraphics[width=\textwidth]{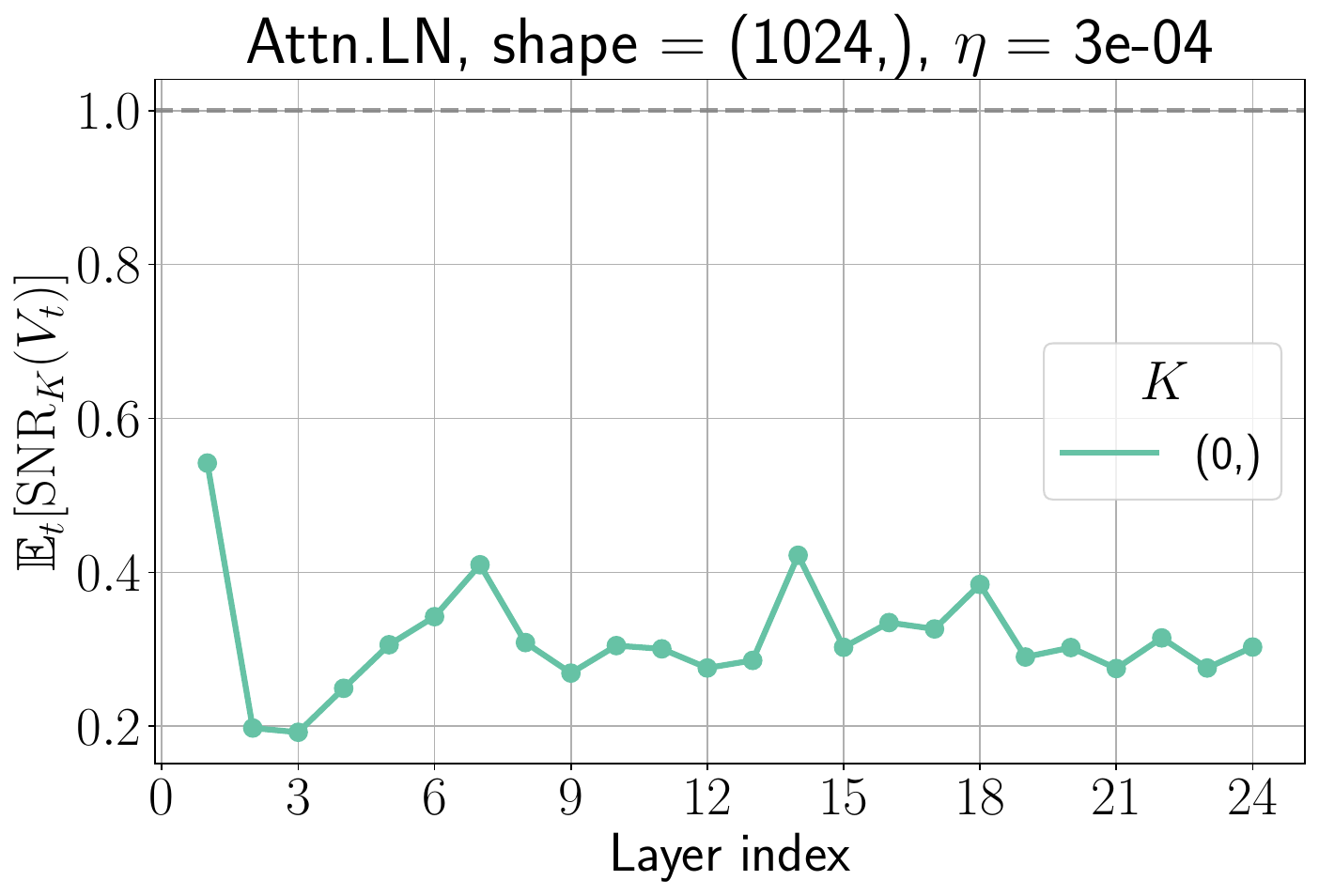}
\end{minipage}
\hfill
\begin{minipage}[b]{0.245\textwidth}
    \centering
    \includegraphics[width=\textwidth]{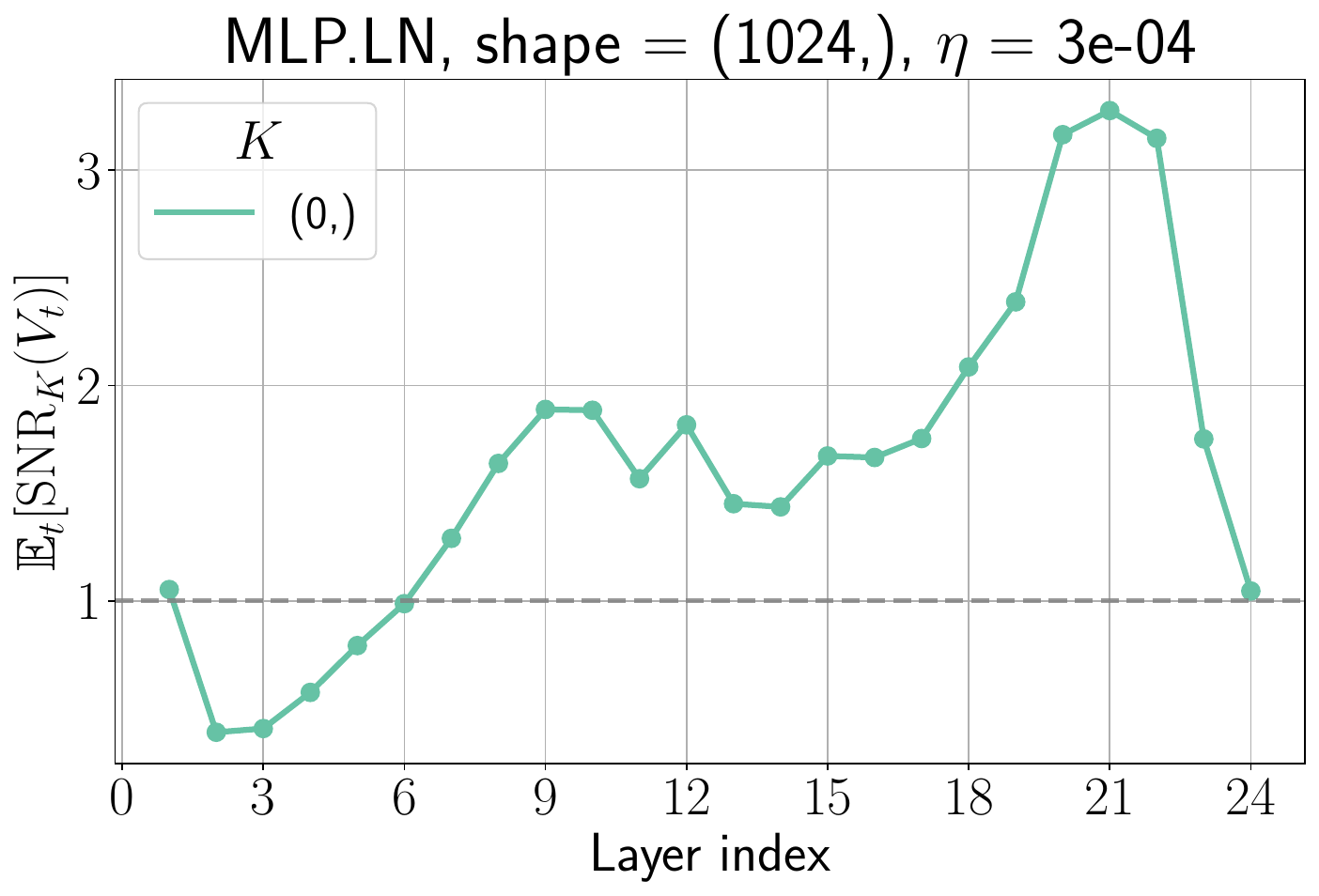}
\end{minipage}
\caption{Layer dependence of average SNR values of the GPT-medium trained on FineWeb-Edu.}
\label{fig:snr-layer-gpt-medium-fineweb-full}
\end{figure*}

This section provides supporting results for the SNR analysis of language pre-training performed in \Cref{section:pre-training}. We considered three experiments to explore the model size and dataset dependency on the SNR results:

\begin{enumerate}
    \item GPT-small trained on OpenWebText (\Cref{fig:snr-curves-gpt-small-openweb-full,fig:snr-layer-gpt-small-openweb-full})
    \item GPT-small trained on FineWeb-Edu (\Cref{fig:snr-curves-gpt-small-fineweb-full,fig:snr-layer-gpt-small-fineweb-full})
    \item GPT-medium trained on FineWeb-Edu (\Cref{fig:snr-layer-gpt-medium-fineweb-full})
\end{enumerate}

\Cref{fig:snr-curves-gpt-small-openweb-full,fig:snr-curves-gpt-small-fineweb-full} show that similar SNR trajectories are observed across different web text datasets. The layerwise trends shown in  \Cref{fig:snr-layer-gpt-small-openweb-full,fig:snr-layer-gpt-small-fineweb-full} further support this claim. 
Furthermore, \Cref{fig:snr-layer-gpt-medium-fineweb-full} shows that similar SNR trends for a GPT-medium model.

\subsection{Language Fine-tuning}
\label{appendix:snr-finetuning}

\begin{figure*}[!htb]
\centering
\begin{minipage}[b]{0.245\textwidth}
    \centering
    \includegraphics[width=\textwidth]{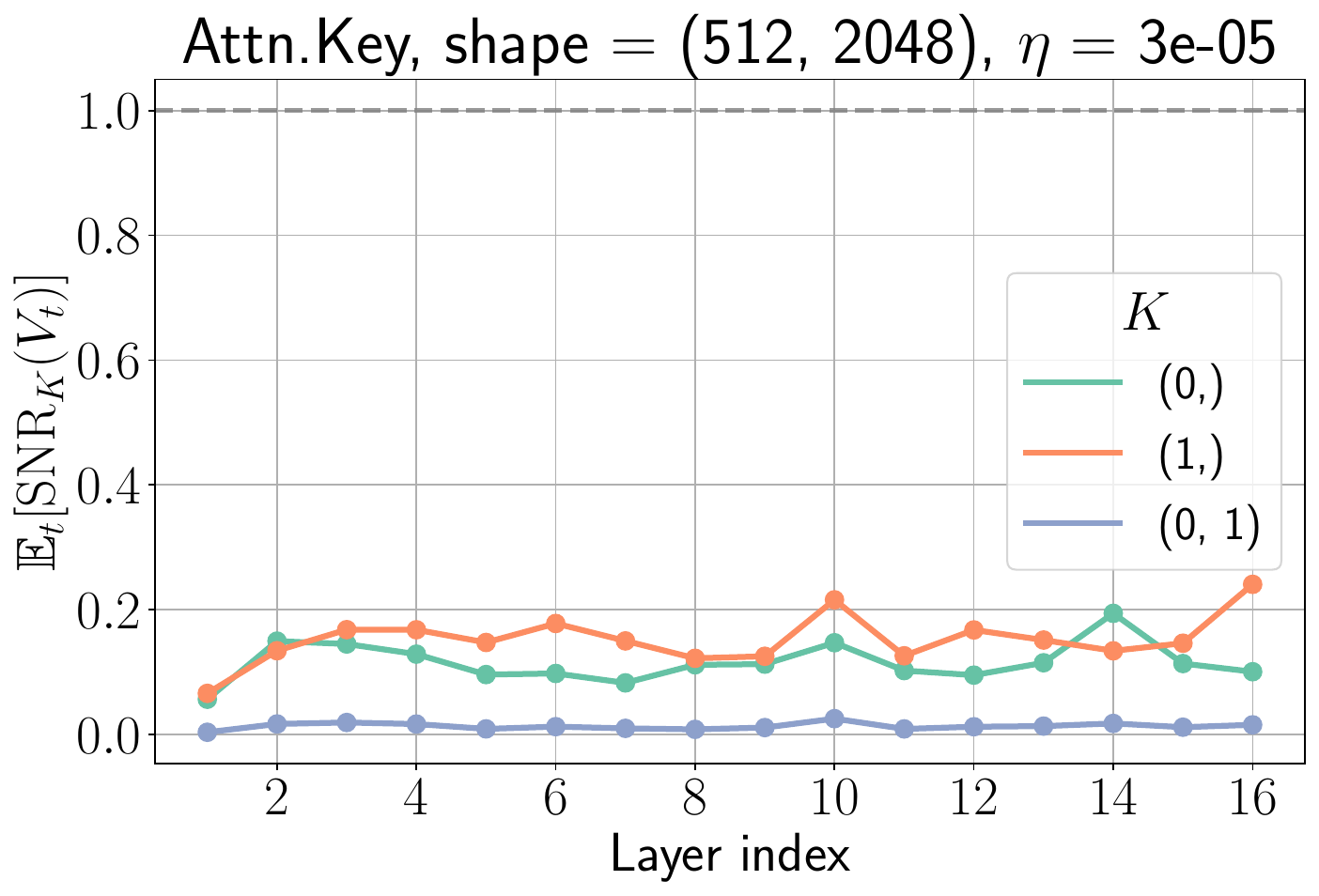}
\end{minipage}
\hfill
\begin{minipage}[b]{0.245\textwidth}
    \centering
    \includegraphics[width=\textwidth]{figures/snr-analysis/snr-layer/Llama-finetuning/snr_layer_Attn.Query_llama-3.2-1b_alpaca_42123_bs16_E3_Adam_lr3e-05_b0.9_b0.999_wd0.1_ga_1.pdf}
\end{minipage}
\hfill
\begin{minipage}[b]{0.245\textwidth}
    \centering
    \includegraphics[width=\textwidth]{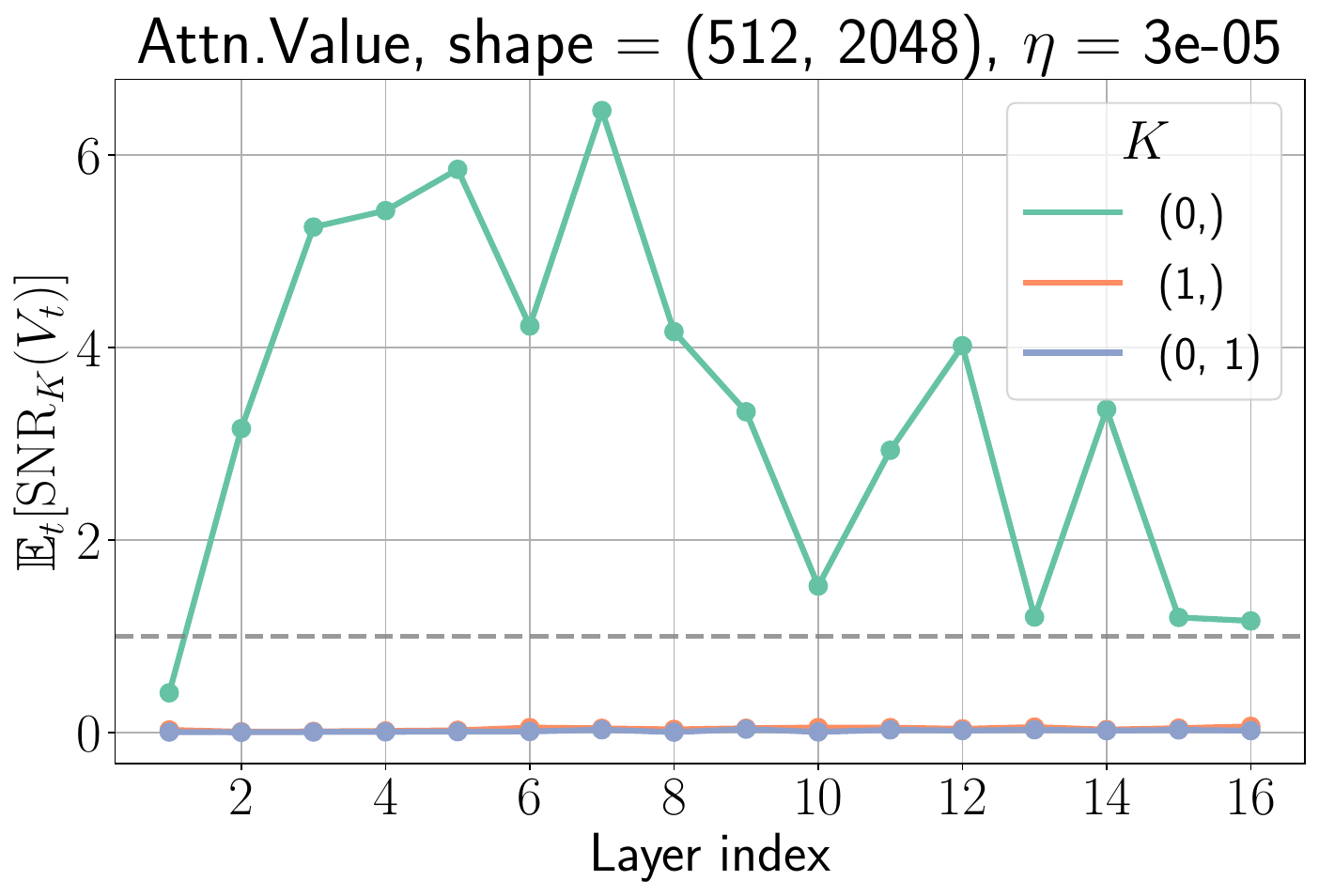}
\end{minipage}
\hfill
\begin{minipage}[b]{0.245\textwidth}
    \centering
    \includegraphics[width=\textwidth]{figures/snr-analysis/snr-layer/Llama-finetuning/snr_layer_Attn.Proj_llama-3.2-1b_alpaca_42123_bs16_E3_Adam_lr3e-05_b0.9_b0.999_wd0.1_ga_1.pdf}
\end{minipage}

\begin{minipage}[b]{0.245\textwidth}
    \centering
    \includegraphics[width=\textwidth]{figures/snr-analysis/snr-layer/Llama-finetuning/snr_layer_MLP.Up_llama-3.2-1b_alpaca_42123_bs16_E3_Adam_lr3e-05_b0.9_b0.999_wd0.1_ga_1.pdf}
\end{minipage}
\begin{minipage}[b]{0.245\textwidth}
    \centering
    \includegraphics[width=\textwidth]{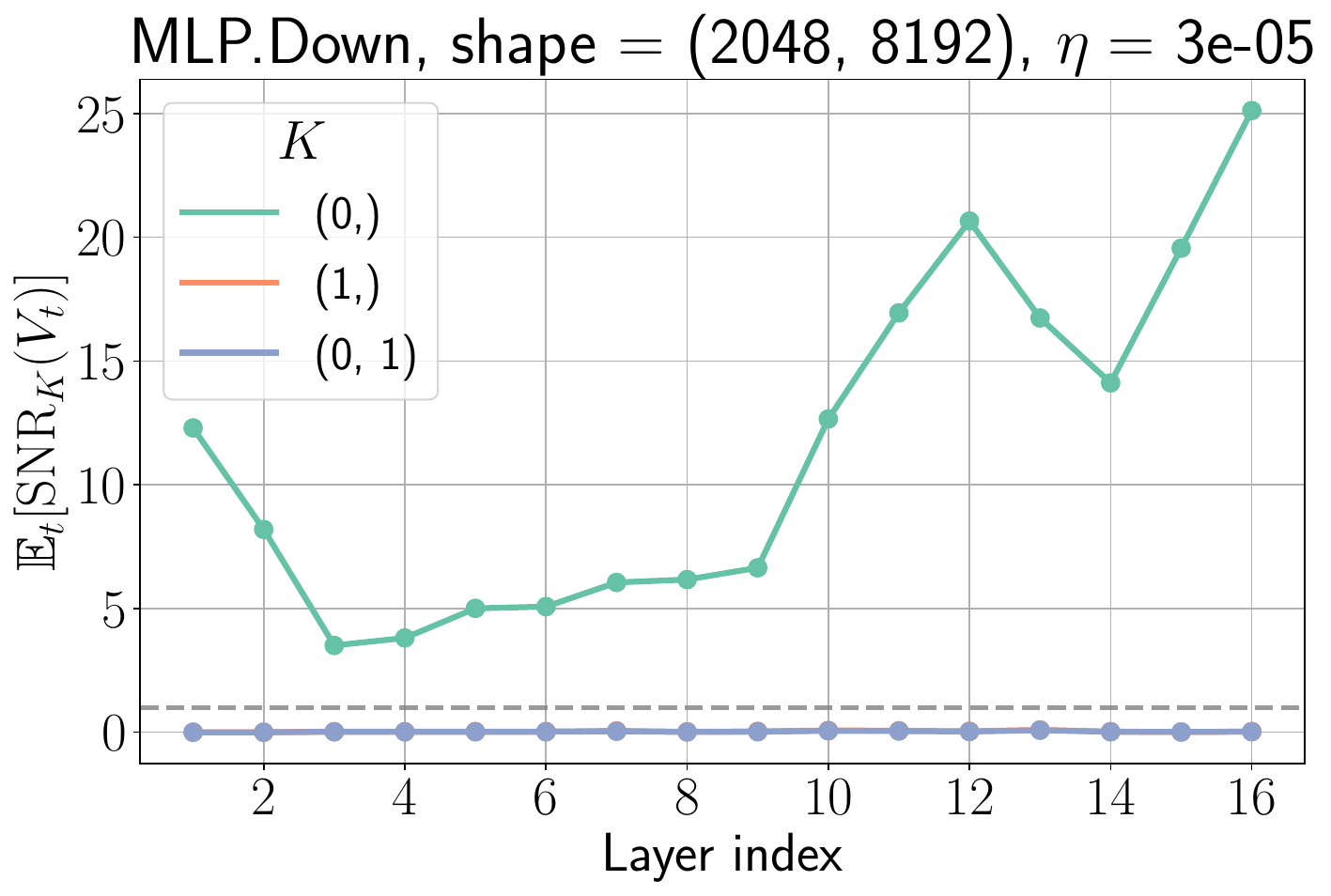}
\end{minipage}
\begin{minipage}[b]{0.245\textwidth}
    \centering
    \includegraphics[width=\textwidth]{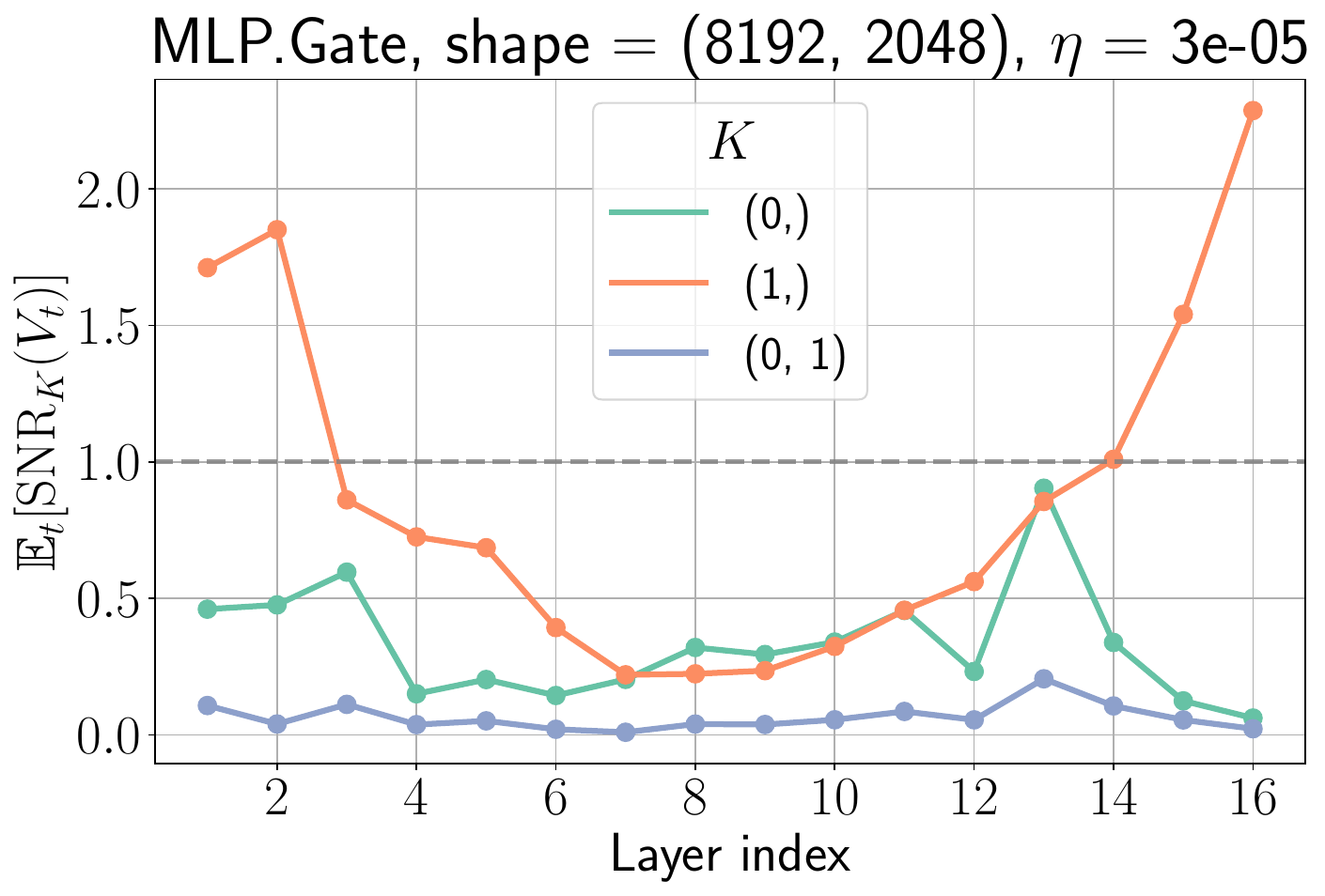}
\end{minipage}

\begin{minipage}[b]{0.245\textwidth}
    \centering
    \includegraphics[width=\textwidth]{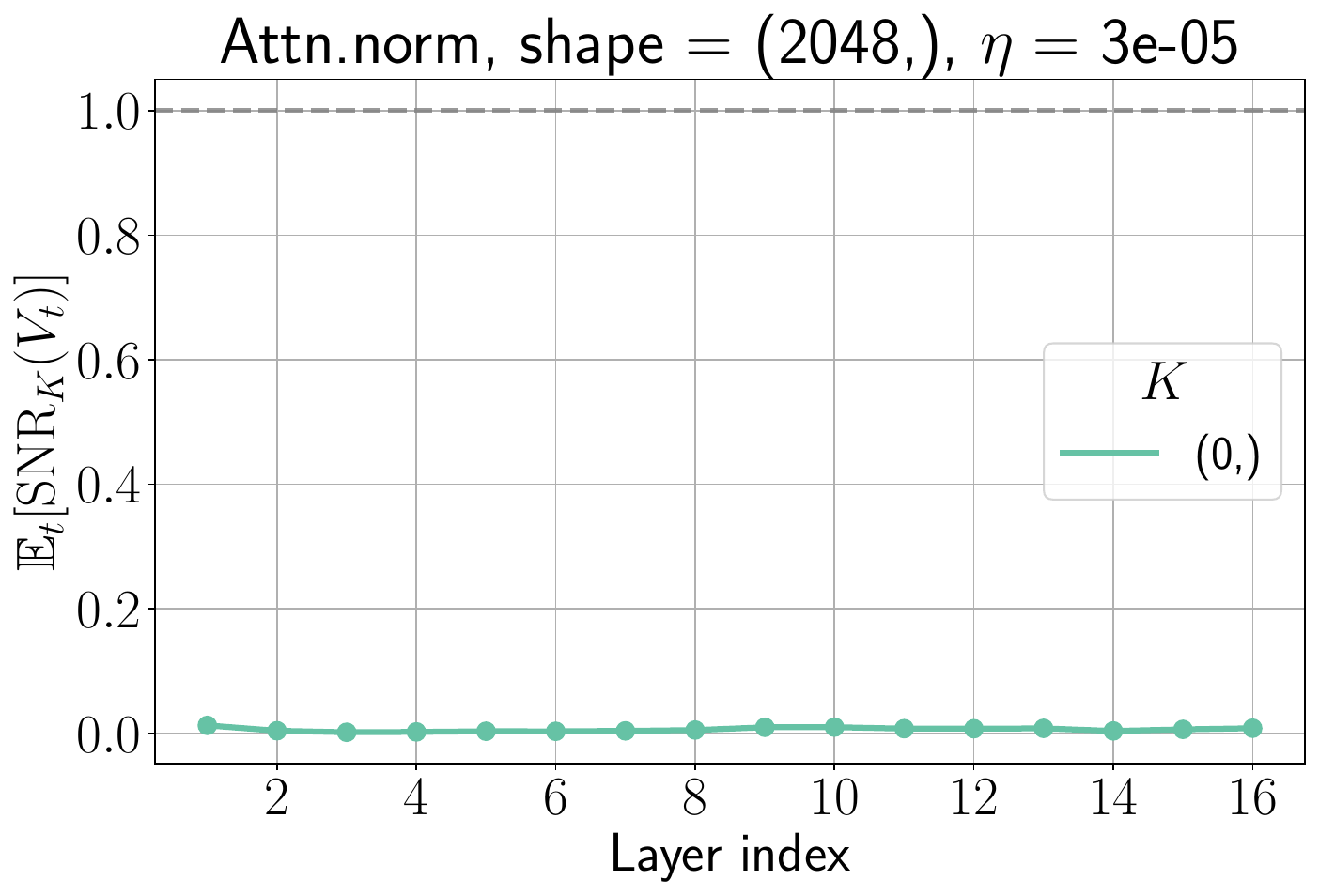}
\end{minipage}
\hfill
\begin{minipage}[b]{0.245\textwidth}
    \centering
    \includegraphics[width=\textwidth]{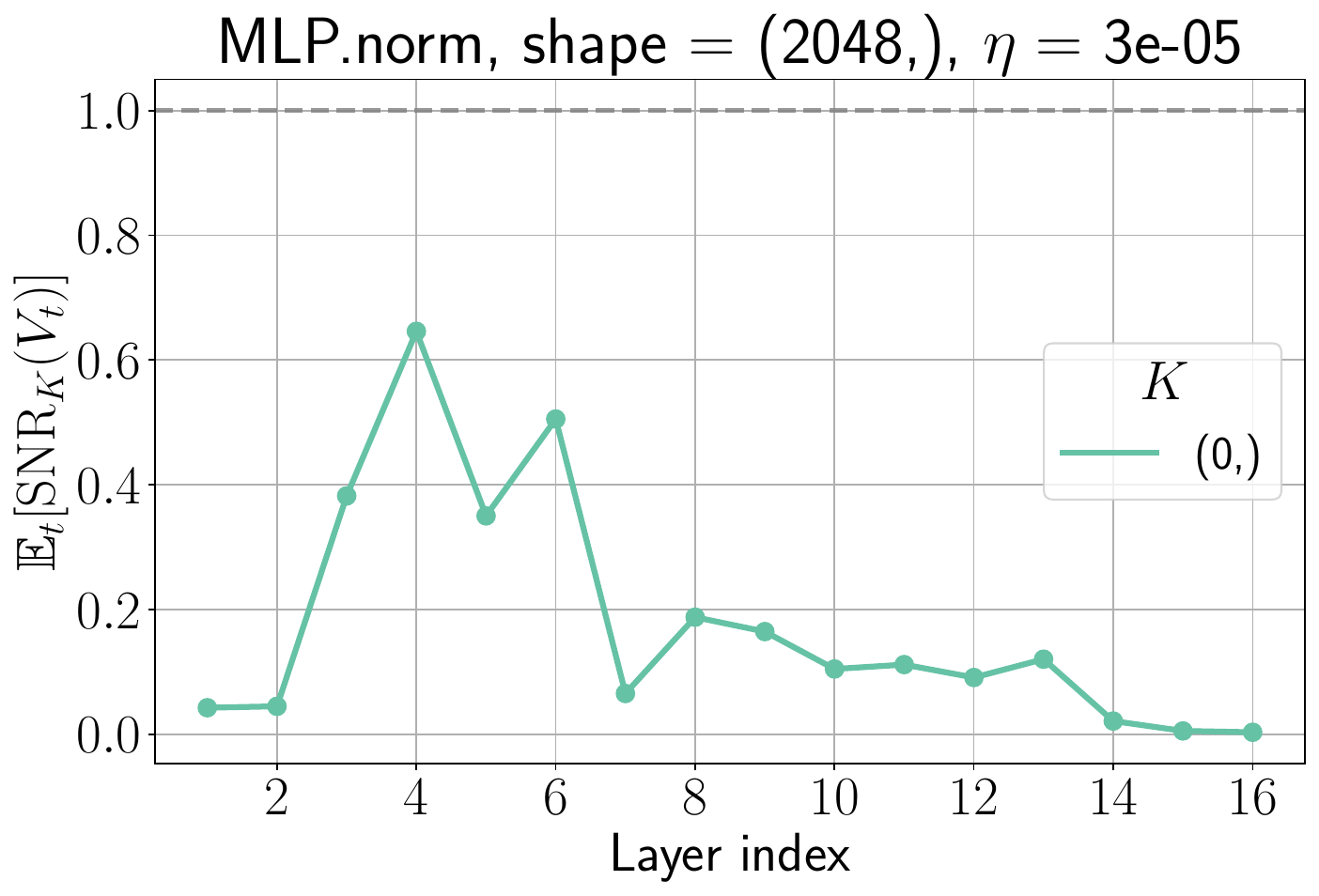}
\end{minipage}
\hfill
\begin{minipage}[b]{0.245\textwidth}
    \centering
    \includegraphics[width=\textwidth]{figures/snr-analysis/snr-trajectories/Llama-finetuning/snr_Tok.Embd_LNone_llama-3.2-1b_alpaca_42123_bs16_E3_Adam_lr3e-05_b0.9_b0.999_wd0.1_ga_1.pdf}
\end{minipage}
\hfill
\begin{minipage}[b]{0.245\textwidth}
    \centering
    \includegraphics[width=\textwidth]{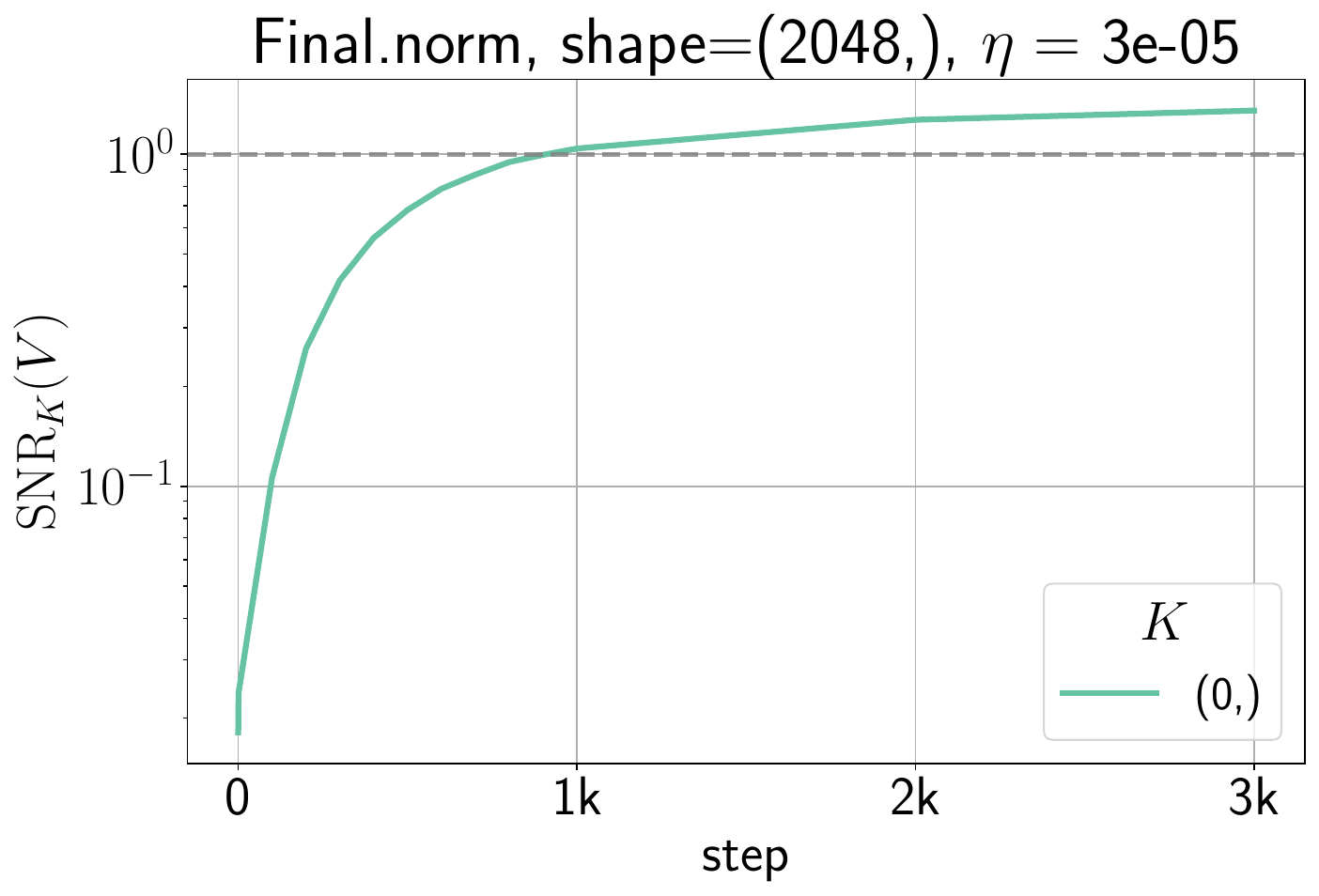}
\end{minipage}
\caption{SNR analysis of pre-trained Llama 3.2 1B fine-tuned on Alpaca dataset.}
\label{fig:snr-layer-llama-1b-alpaca-full}
\end{figure*}

\Cref{fig:snr-layer-llama-1b-alpaca-full} shows the SNR trends for pre-trained Llama 3.2 1B, fine-tuned on the Alpaca dataset. In comparison to the GPT pre-training experiments, we observe that the SNR values of attention key and query second moments are significantly lower than $1.0$. More generally, we observe lower SNR values, suggesting less compressibility.

\subsection{Image Classification}
\label{appendix:image-classification}

\begin{figure*}[!htb]
\centering
\begin{minipage}[b]{0.245\textwidth}
    \centering
    \includegraphics[width=\textwidth]{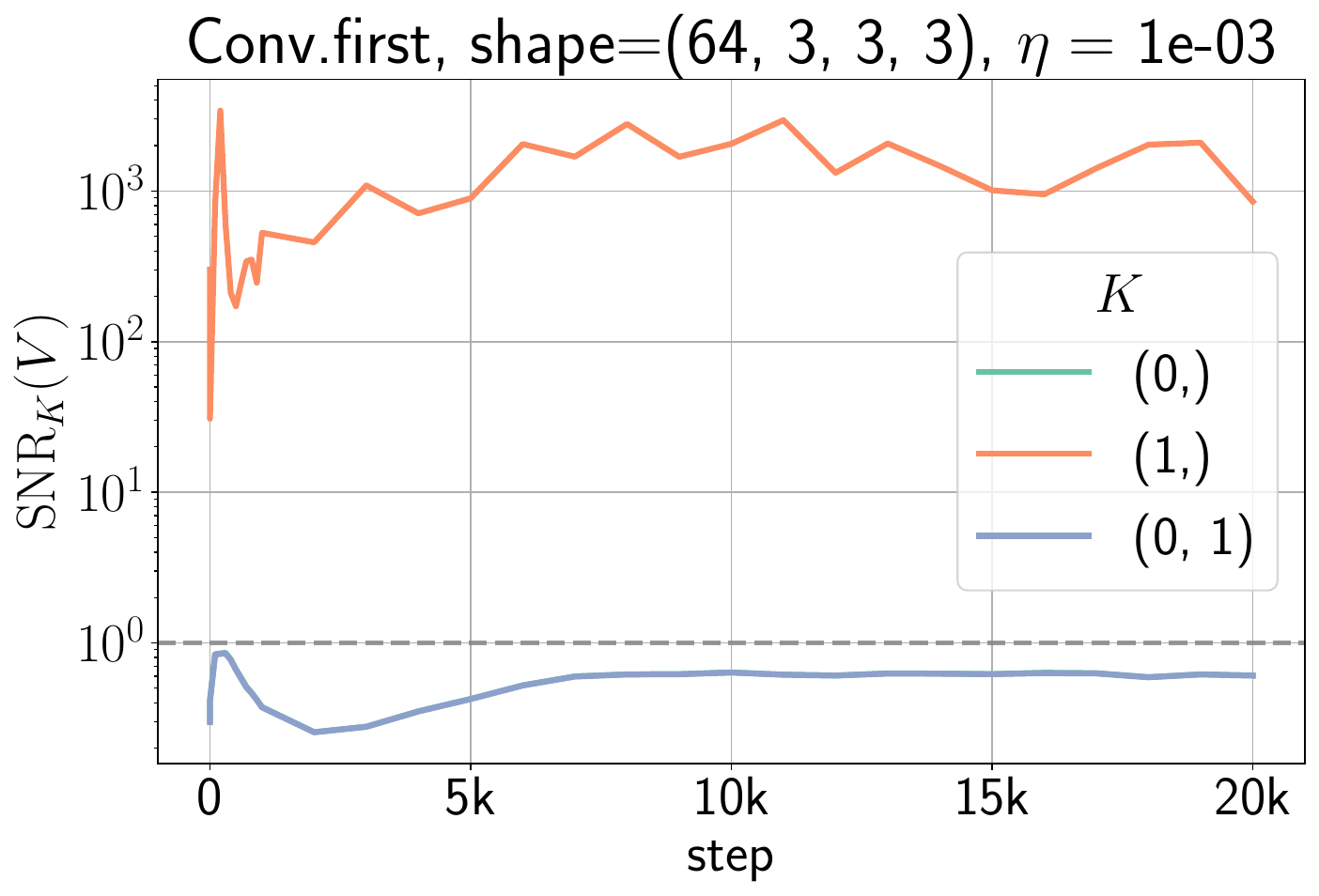}
\end{minipage}
\hfill
\begin{minipage}[b]{0.245\textwidth}
    \centering
    \includegraphics[width=\textwidth]{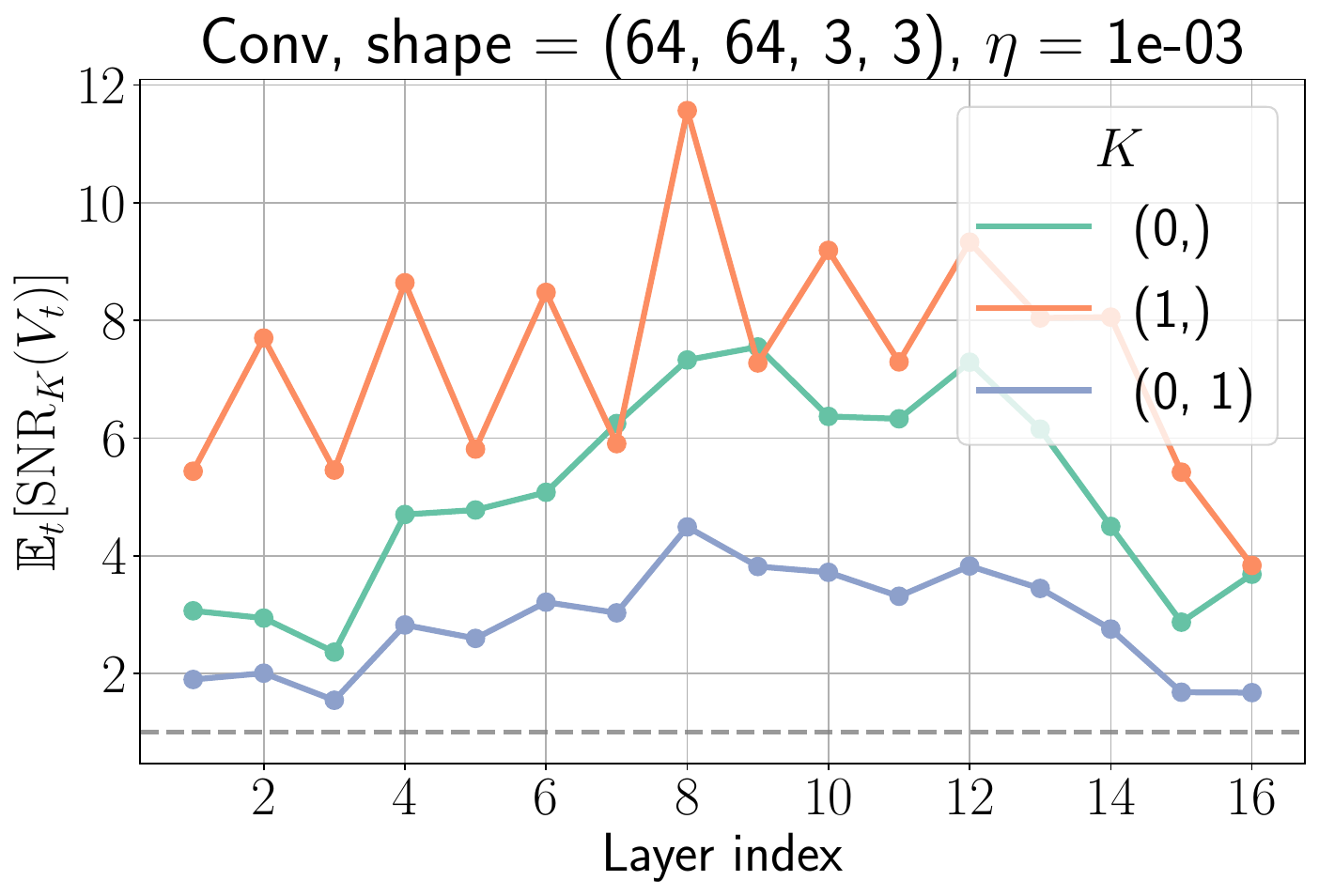}
\end{minipage}
\hfill
\begin{minipage}[b]{0.245\textwidth}
    \centering
    \includegraphics[width=\textwidth]{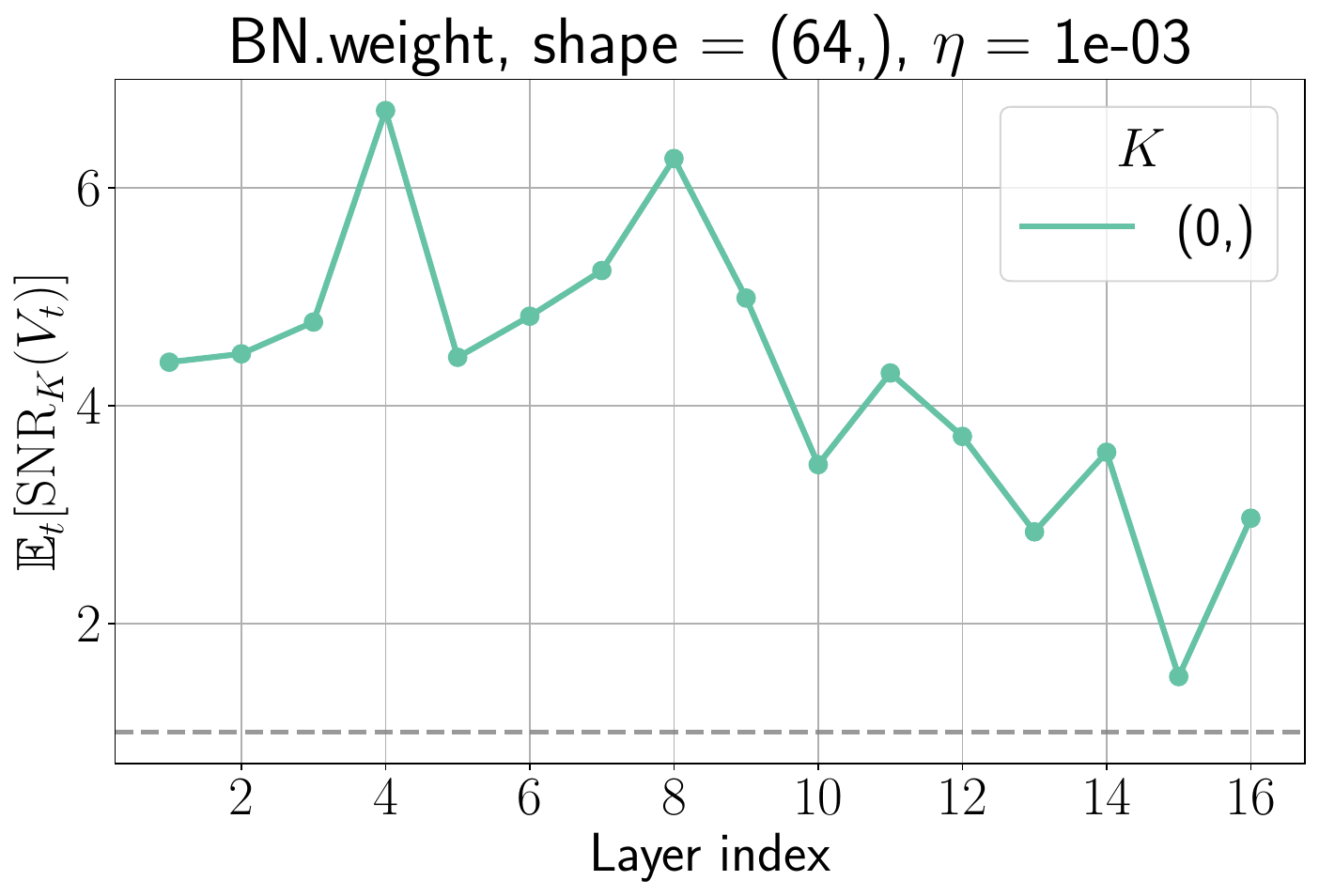}
\end{minipage}
\hfill
\begin{minipage}[b]{0.245\textwidth}
    \centering
    \includegraphics[width=\textwidth]{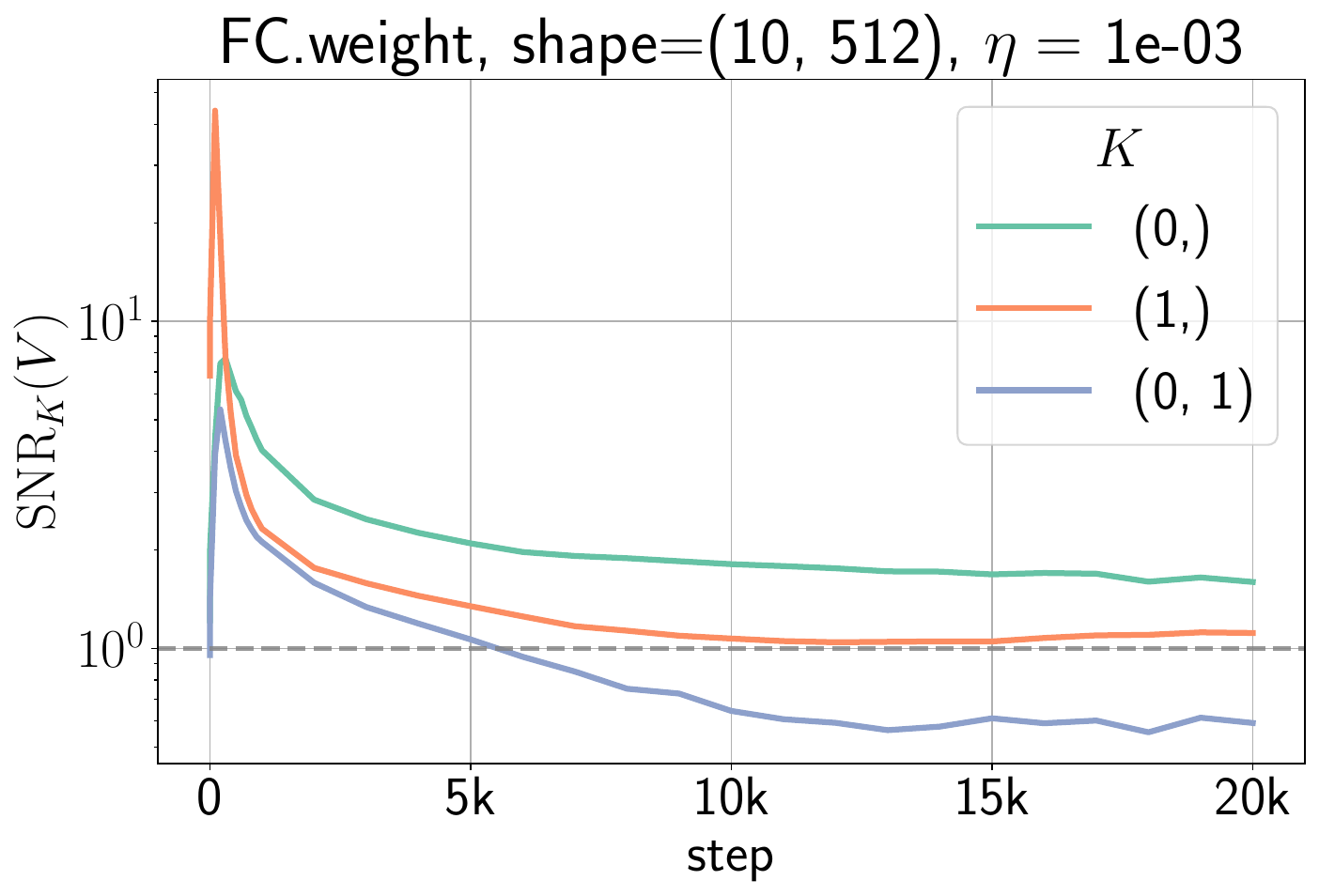}
\end{minipage}
\caption{SNR trends of different layers of ResNet-18 trained on CIFAR-10.}
\label{fig:snr-resnet-cifar-10}
\end{figure*}

\begin{figure*}[!htb]
\centering
\begin{minipage}[b]{0.245\textwidth}
    \centering
    \includegraphics[width=\textwidth]{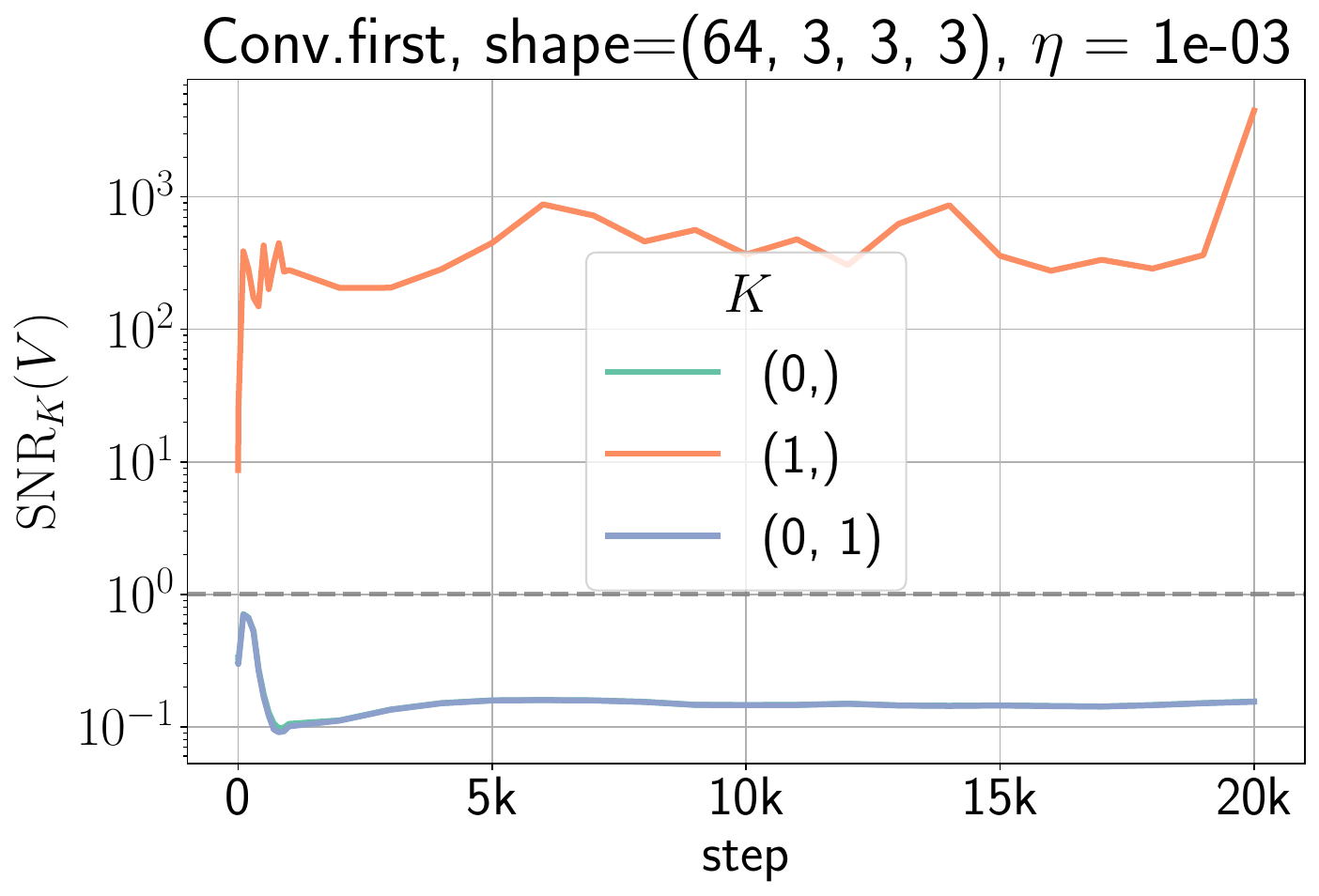}
\end{minipage}
\hfill
\begin{minipage}[b]{0.245\textwidth}
    \centering
    \includegraphics[width=\textwidth]{figures/snr-analysis/snr-layer/Resnet/snr_layer_Conv_cifar-100_Resnet18_n16_d18_relu_varw1.0_B128_T20000_Tw2048_SlimAdamw_None_lr1e-03_b0.9_b0.999_eps1e-08_wd0.01_True.pdf}
\end{minipage}
\hfill
\begin{minipage}[b]{0.245\textwidth}
    \centering
    \includegraphics[width=\textwidth]{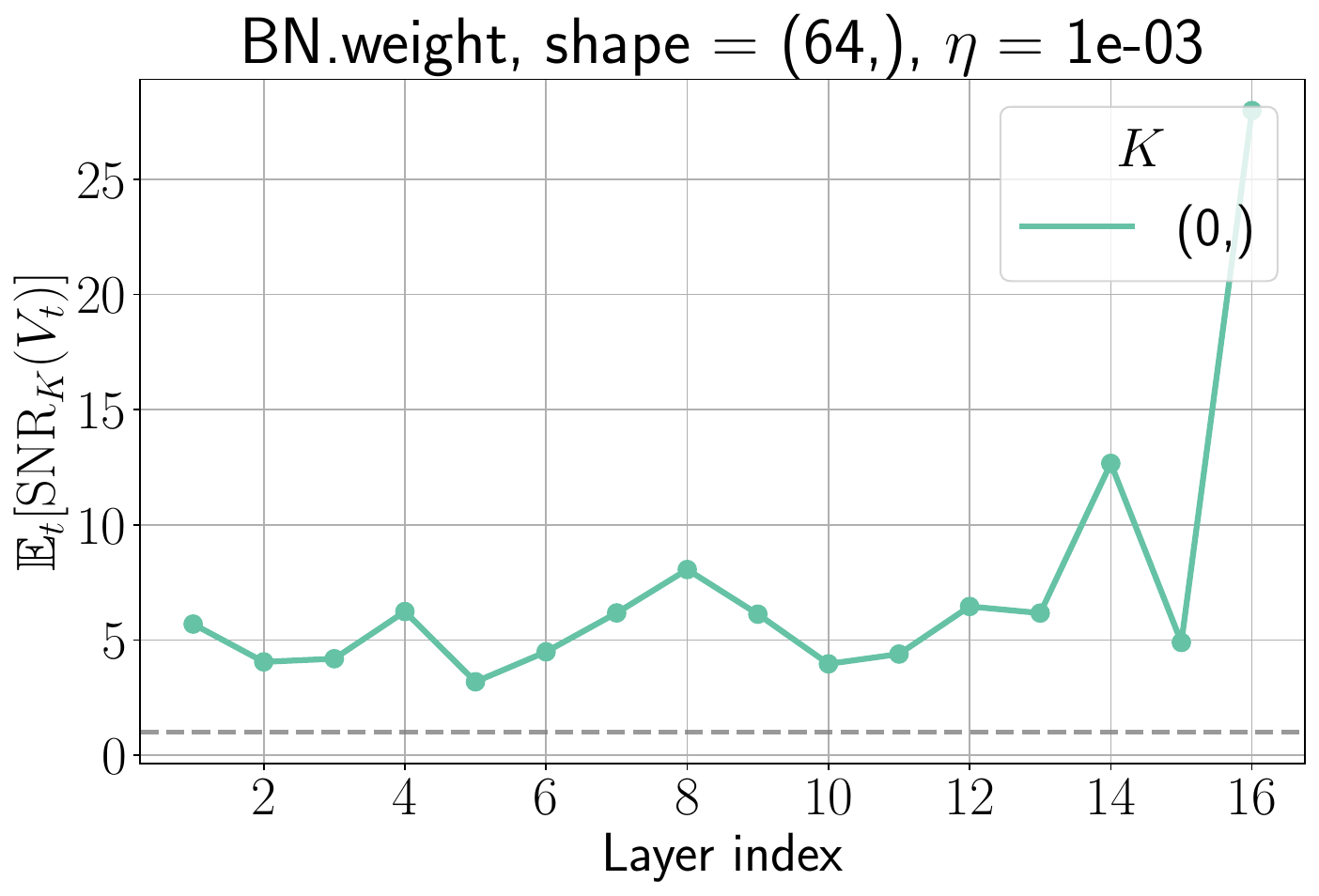}
\end{minipage}
\hfill
\begin{minipage}[b]{0.245\textwidth}
    \centering
    \includegraphics[width=\textwidth]{figures/snr-analysis/snr-trajectories/Resnet/snr_FC.weight_LNone_cifar-100_Resnet18_n16_d18_relu_varw1.0_B128_T20000_Tw2048_SlimAdamw_None_lr1e-03_b0.9_b0.999_eps1e-08_wd0.01_True.pdf}
\end{minipage}
\caption{SNR trends of different layers of ResNet-18 trained on CIFAR-100.}
\label{fig:snr-resnet-cifar-100-full}
\end{figure*}

\begin{figure*}[!htb]
\centering
\begin{minipage}[b]{0.245\textwidth}
    \centering
    \includegraphics[width=\textwidth]{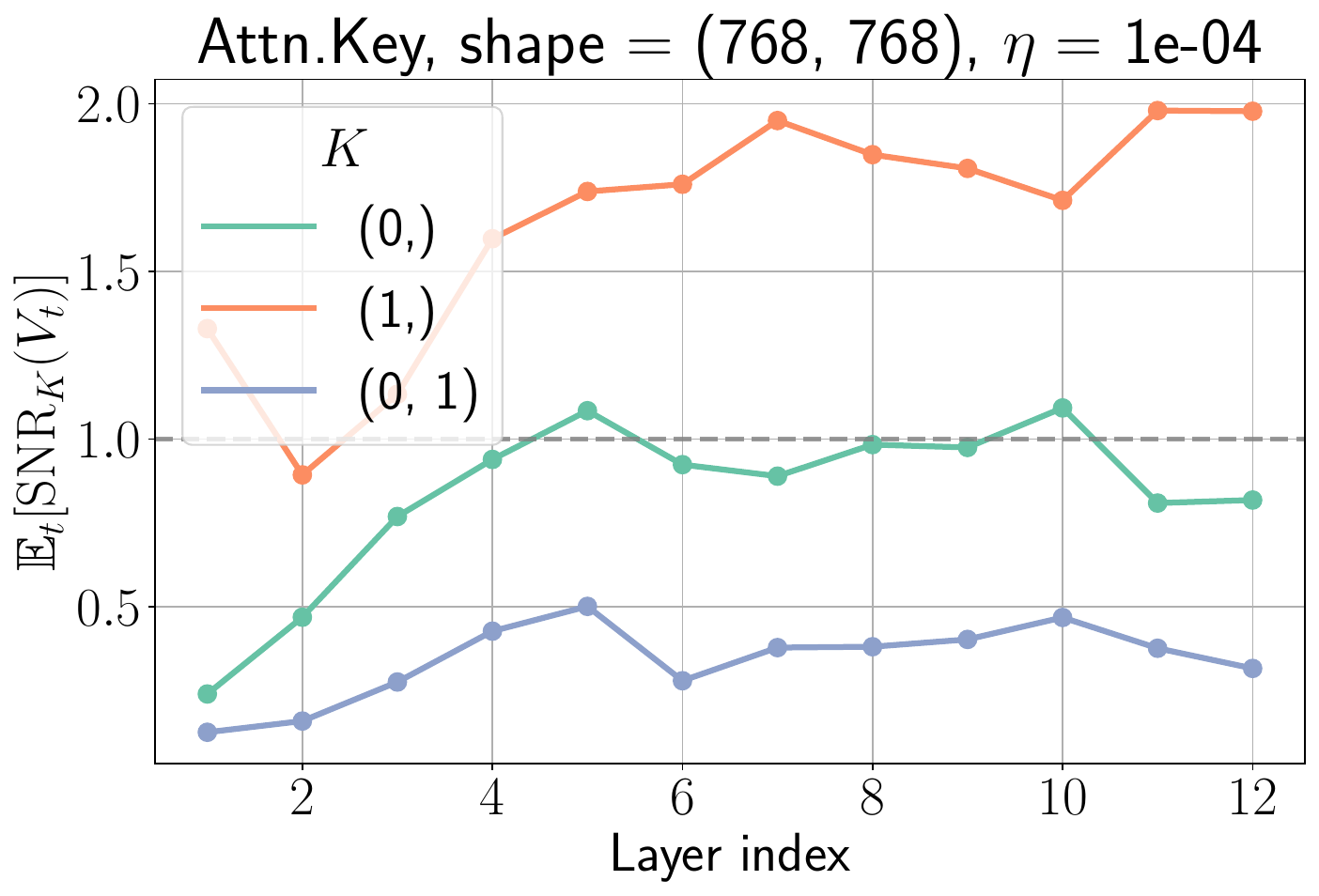}
\end{minipage}
\hfill
\begin{minipage}[b]{0.245\textwidth}
    \centering
    \includegraphics[width=\textwidth]{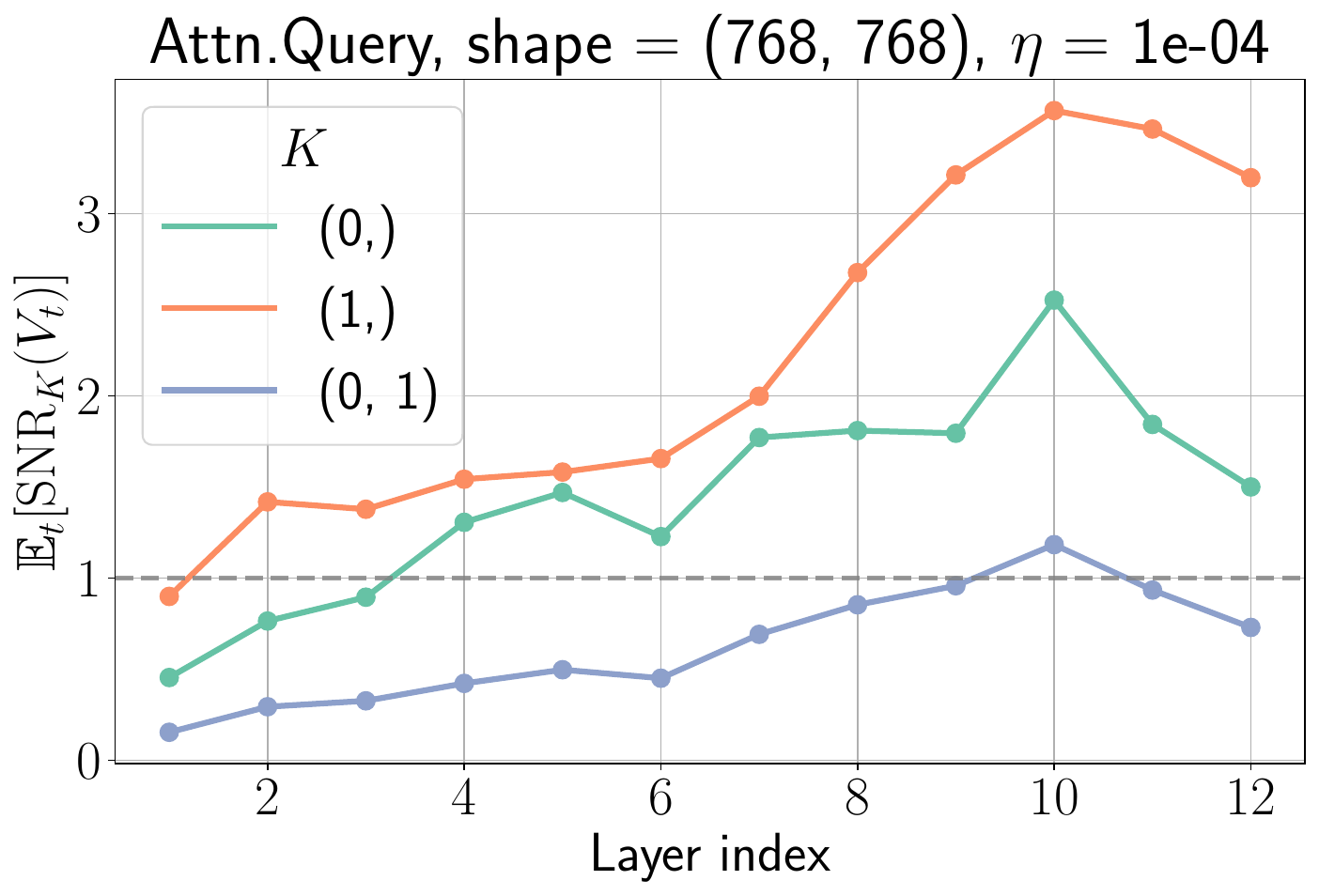}
\end{minipage}
\hfill
\begin{minipage}[b]{0.245\textwidth}
    \centering
    \includegraphics[width=\textwidth]{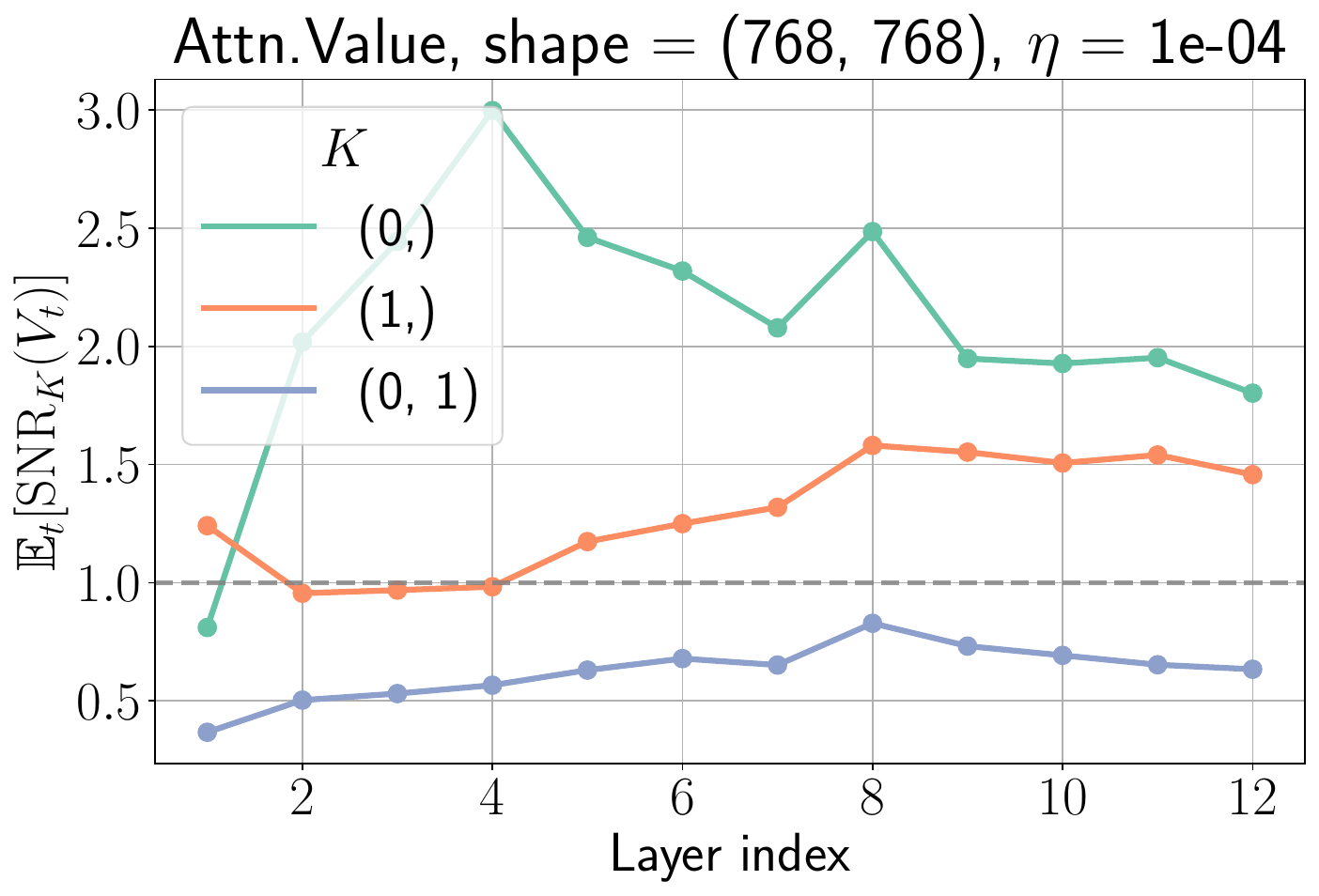}
\end{minipage}
\hfill
\begin{minipage}[b]{0.245\textwidth}
    \centering
    \includegraphics[width=\textwidth]{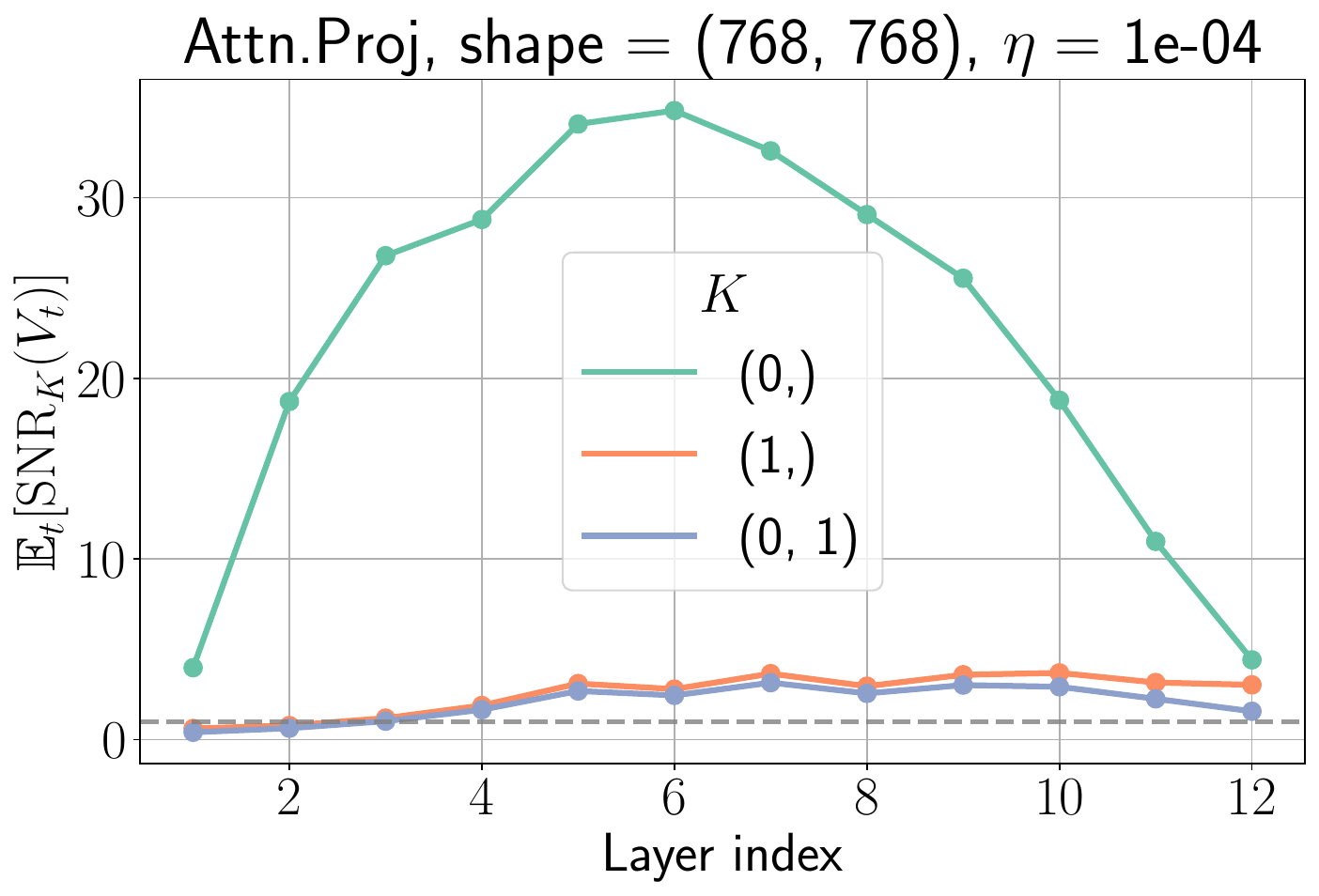}
\end{minipage}

\begin{minipage}[b]{0.245\textwidth}
    \centering
    \includegraphics[width=\textwidth]{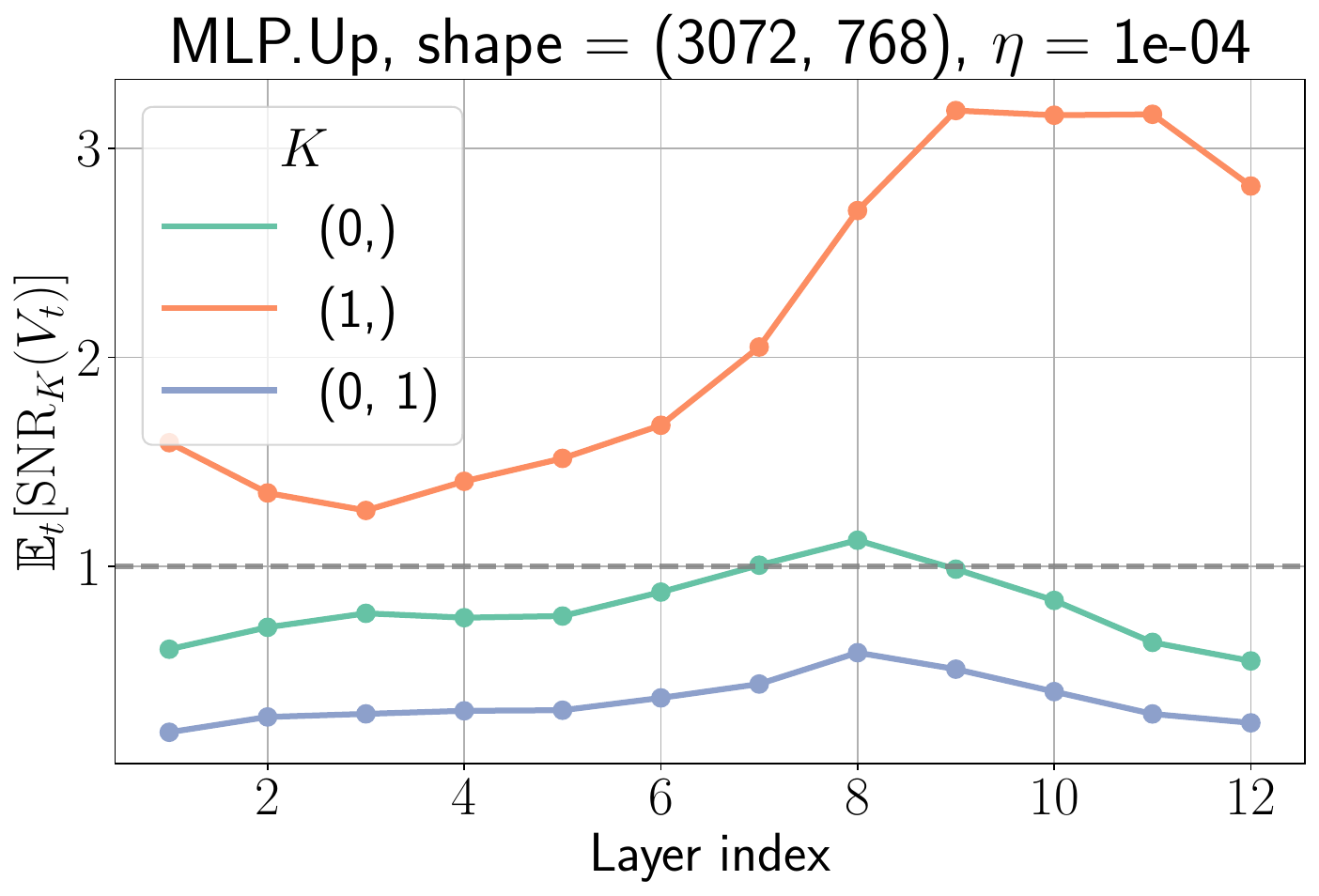}
\end{minipage}
\hfill
\begin{minipage}[b]{0.245\textwidth}
    \centering
    \includegraphics[width=\textwidth]{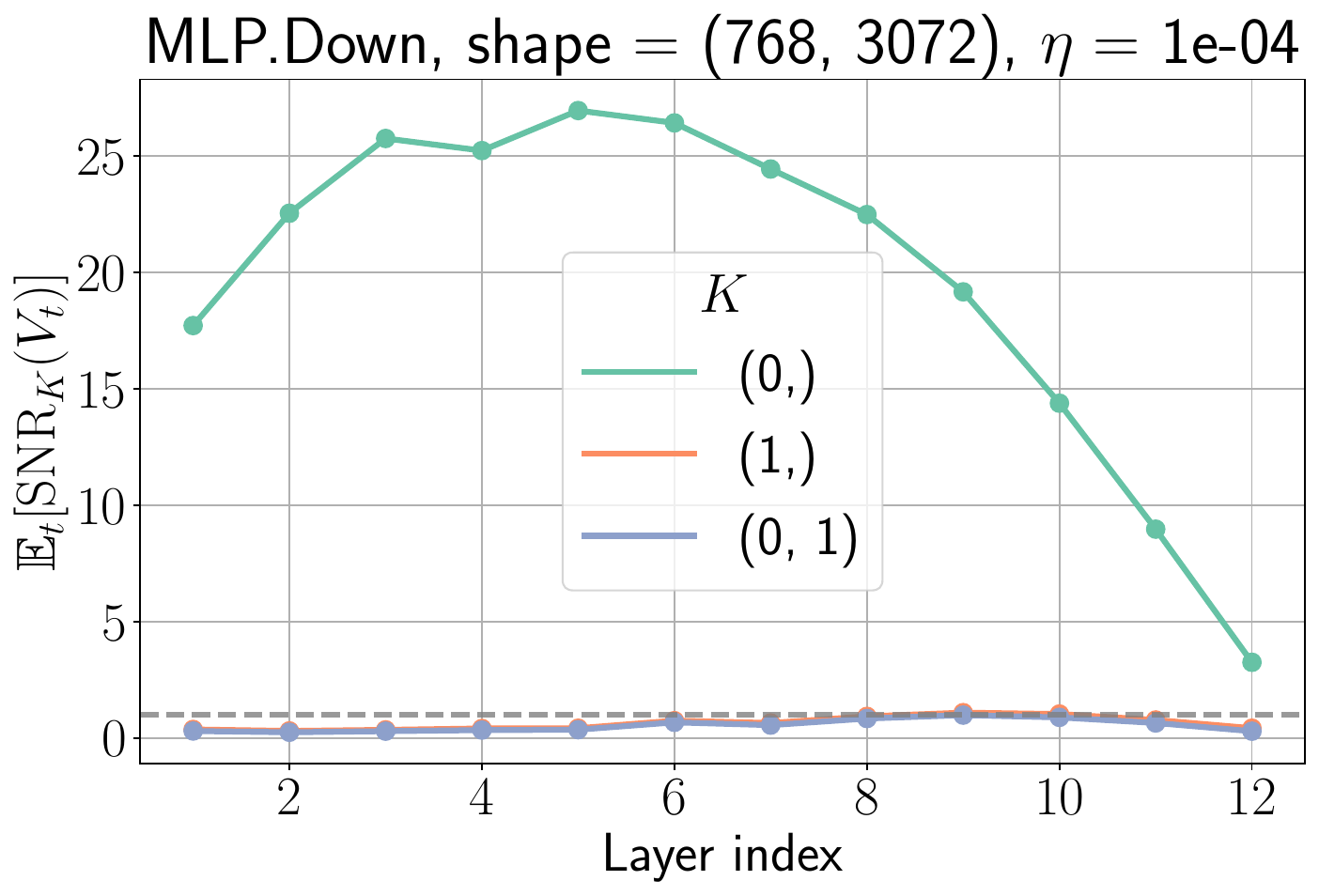}
\end{minipage}
\hfill
\begin{minipage}[b]{0.245\textwidth}
    \centering
    \includegraphics[width=\textwidth]{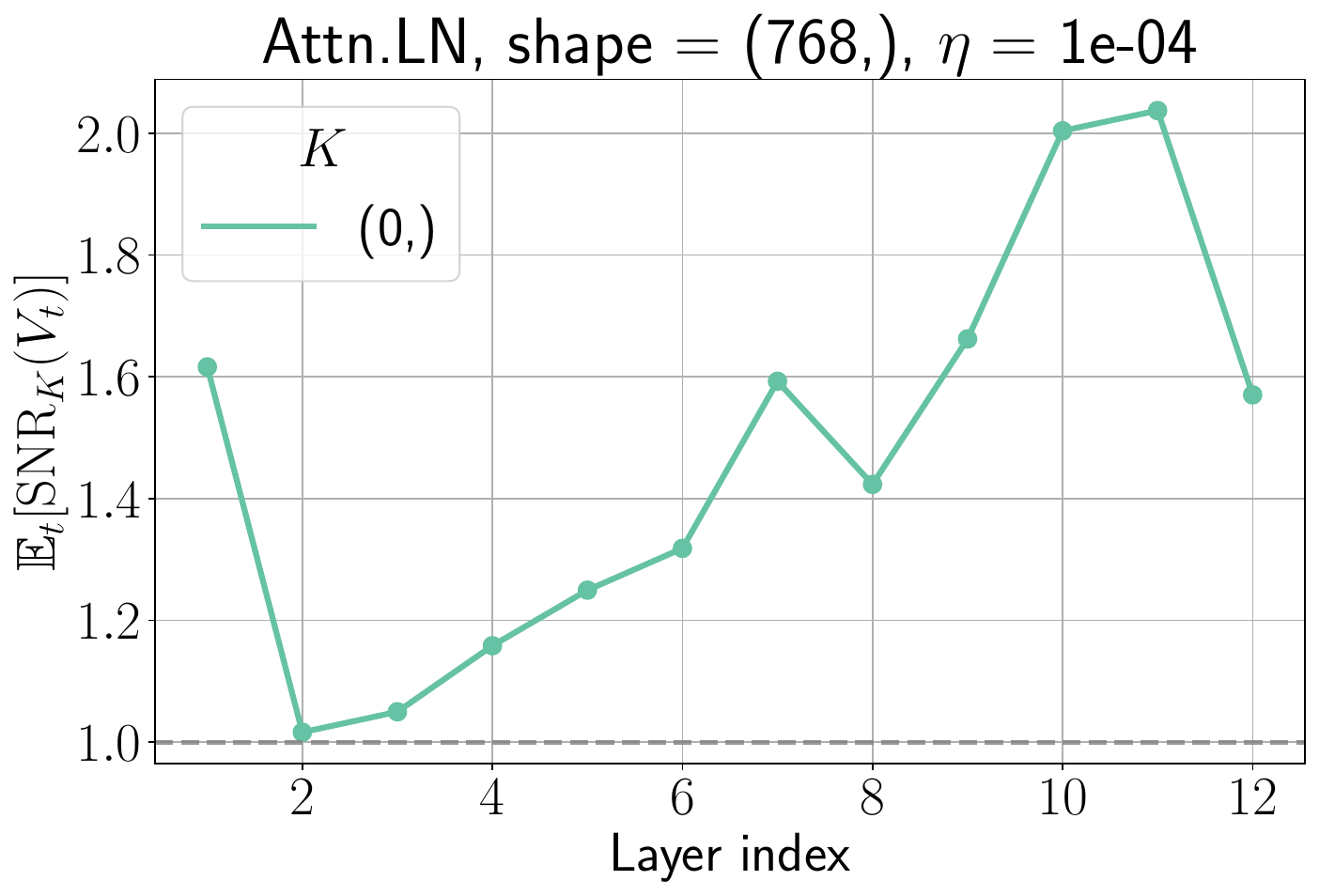}
\end{minipage}
\hfill
\begin{minipage}[b]{0.245\textwidth}
    \centering
    \includegraphics[width=\textwidth]{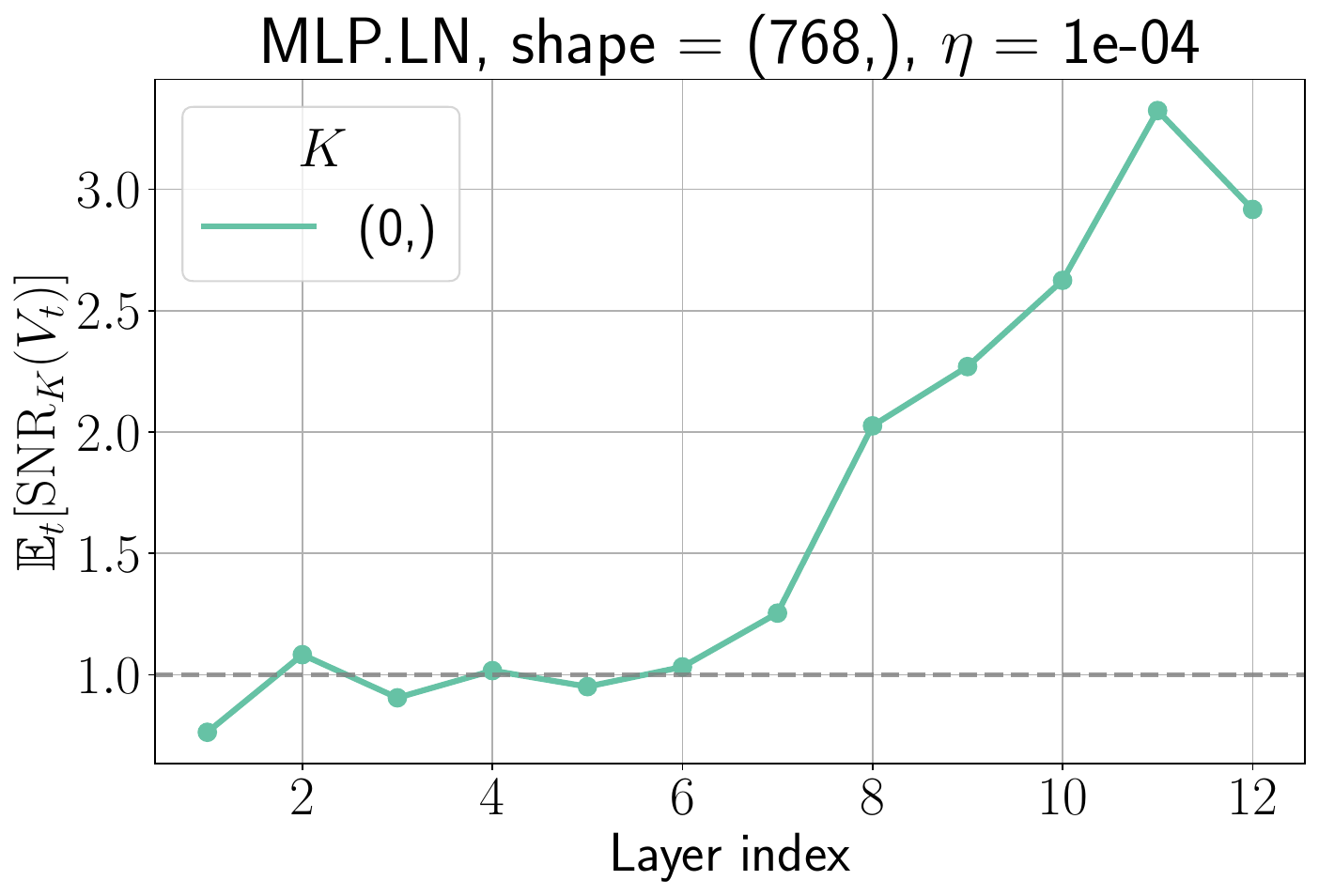}
\end{minipage}

\begin{minipage}[b]{0.225\textwidth}
    \centering
    \includegraphics[width=\textwidth]{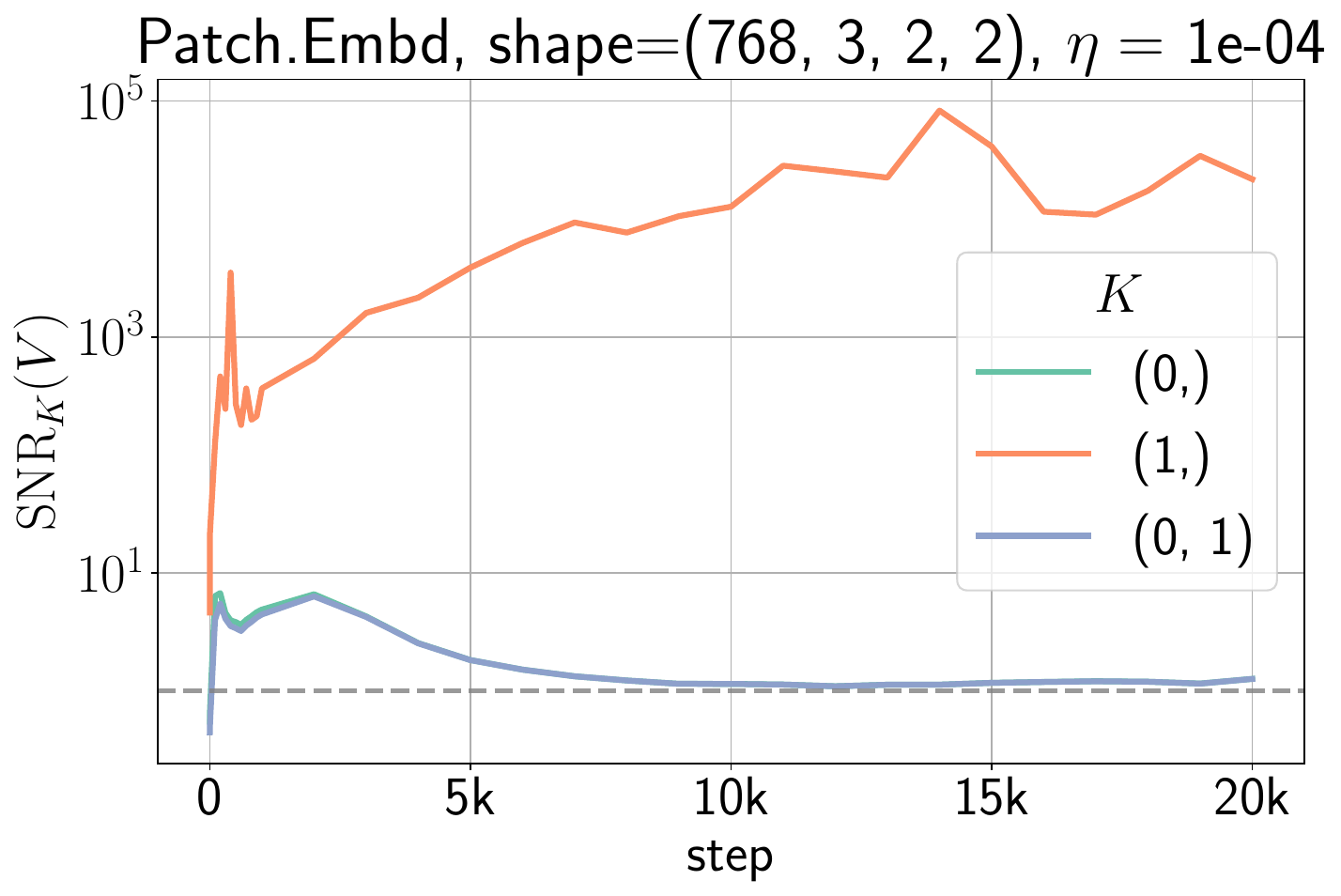}
\end{minipage}
\hfill
\begin{minipage}[b]{0.225\textwidth}
    \centering
    \includegraphics[width=\textwidth]{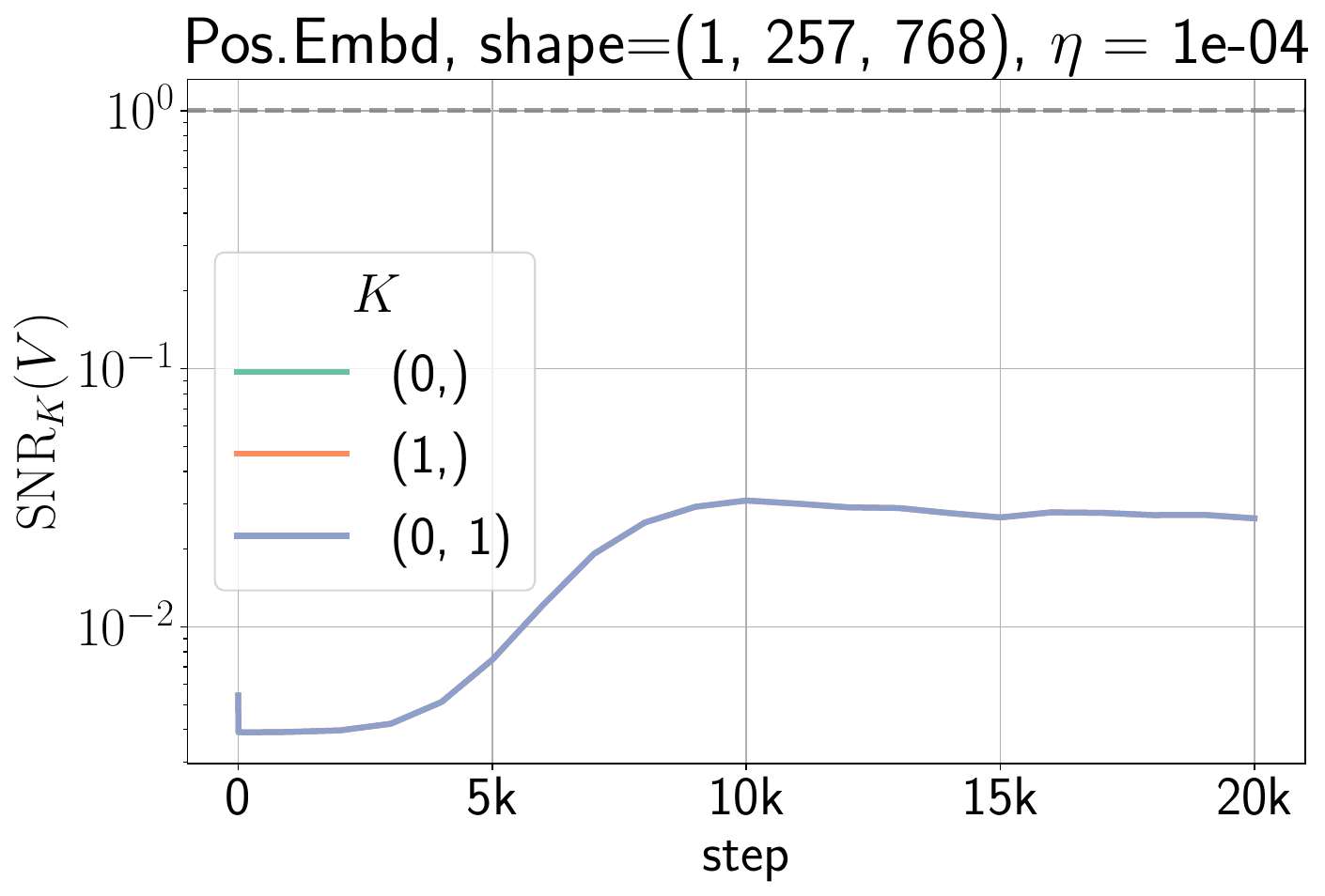}
\end{minipage}
\hfill
\begin{minipage}[b]{0.225\textwidth}
    \centering
    \includegraphics[width=\textwidth]{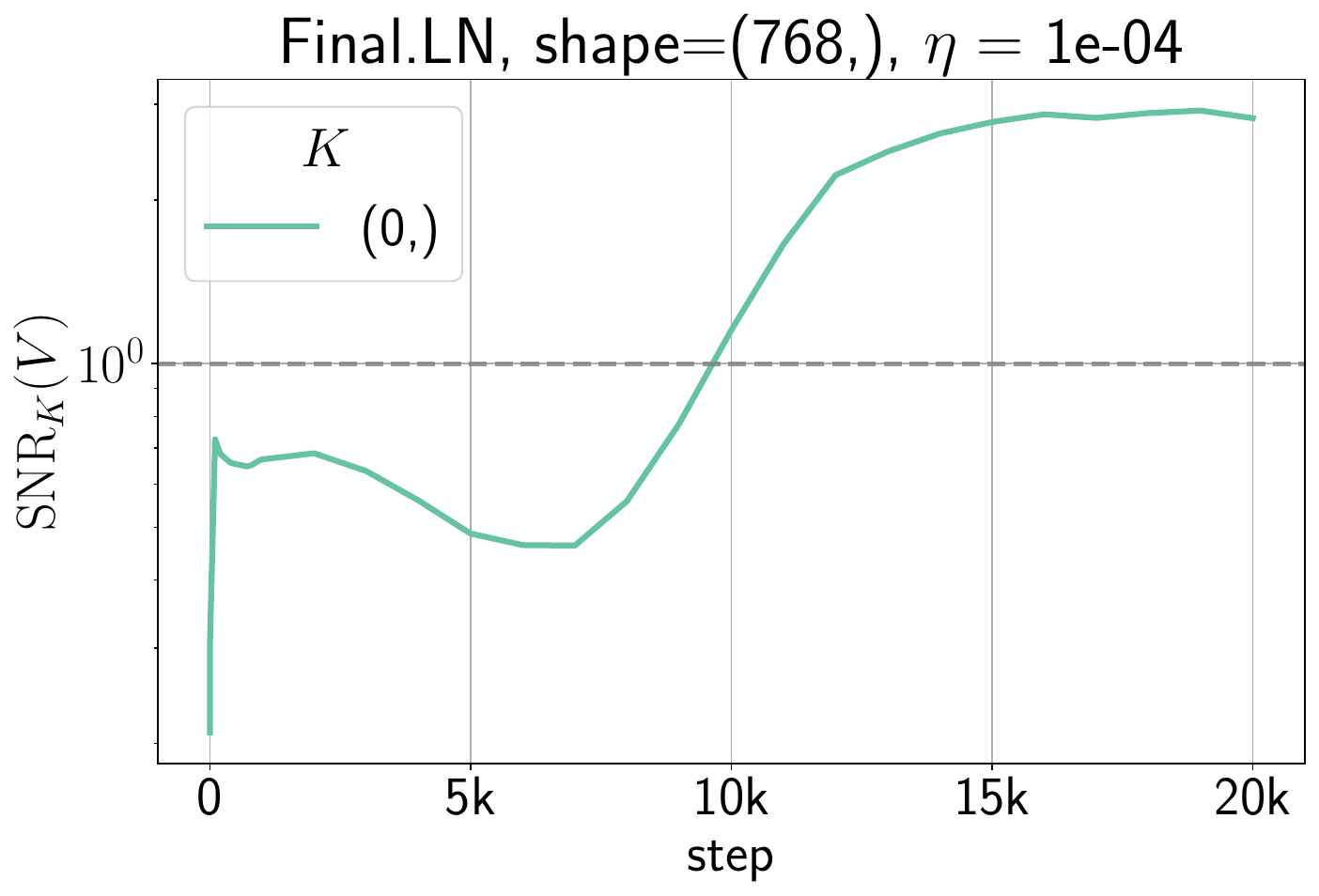}
\end{minipage}
\hfill
\begin{minipage}[b]{0.225\textwidth}
    \centering
    \includegraphics[width=\textwidth]{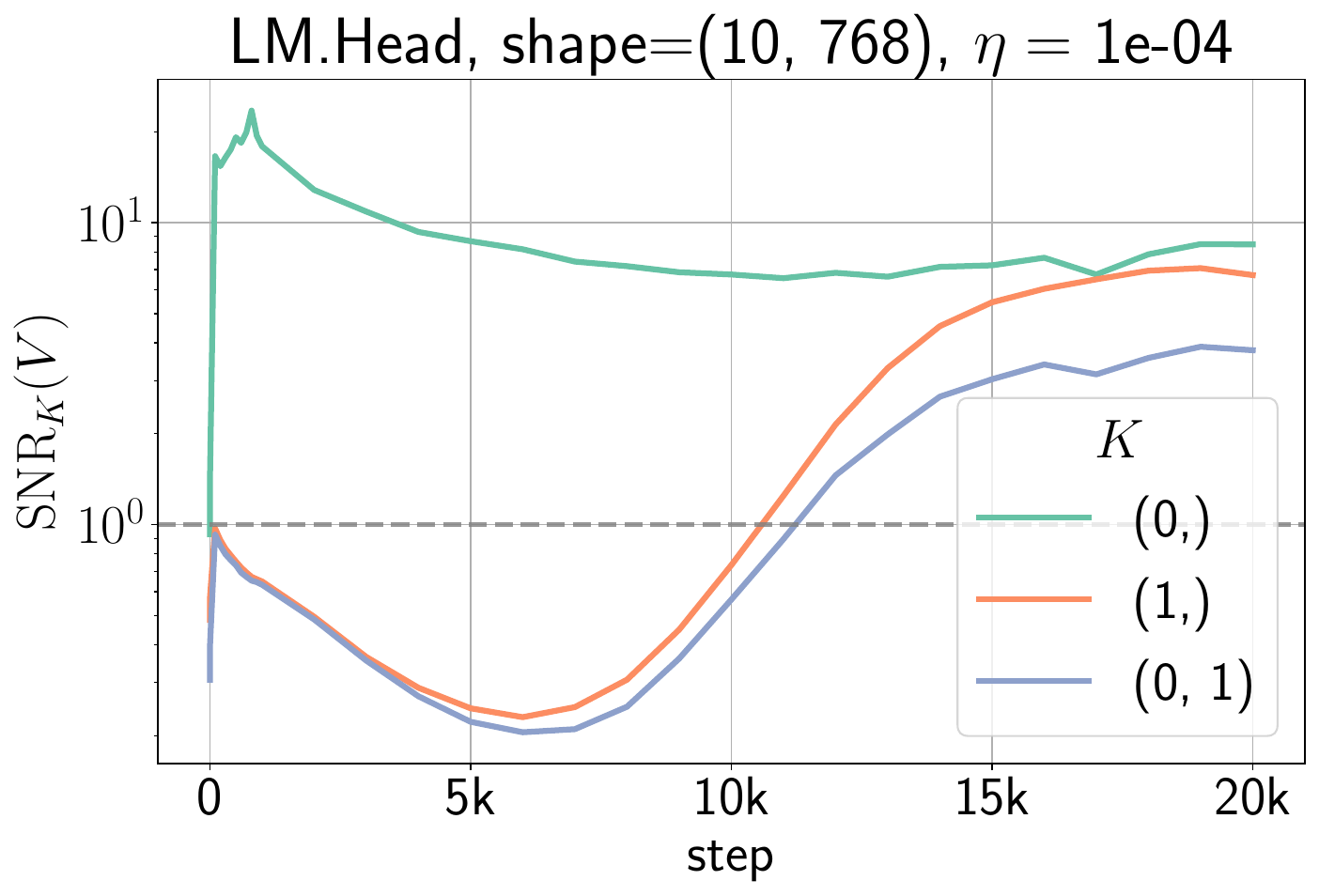}
\end{minipage}
\caption{SNR trends of different layers of ViT-small trained on CIFAR-10.}
\label{fig:snr-vit-cifar-10-full}
\end{figure*}

\begin{figure*}[!htb]
\centering
\begin{minipage}[b]{0.245\textwidth}
    \centering
    \includegraphics[width=\textwidth]{figures/snr-analysis/snr-layer/ViT/snr_layer_Attn.Key_cifar-100_ViT_n768_d12_12_p2_B128_T20000_Tw2048_SlimAdamw_None_lr1e-04_b0.9_b0.999_eps1e-08_wd0.01_True.pdf}
\end{minipage}
\hfill
\begin{minipage}[b]{0.245\textwidth}
    \centering
    \includegraphics[width=\textwidth]{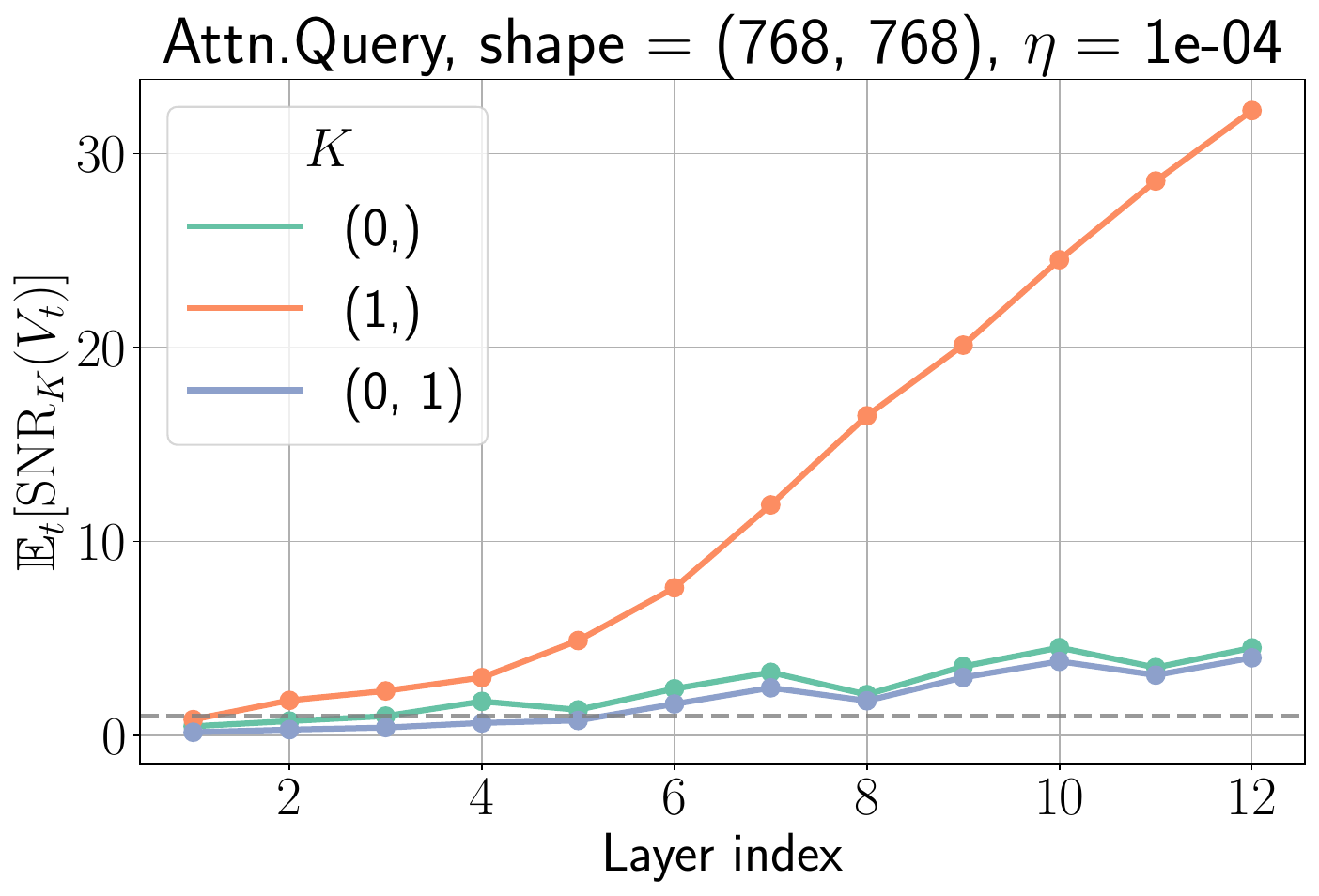}
\end{minipage}
\hfill
\begin{minipage}[b]{0.245\textwidth}
    \centering
    \includegraphics[width=\textwidth]{figures/snr-analysis/snr-layer/ViT/snr_layer_Attn.Value_cifar-100_ViT_n768_d12_12_p2_B128_T20000_Tw2048_SlimAdamw_None_lr1e-04_b0.9_b0.999_eps1e-08_wd0.01_True.pdf}
\end{minipage}
\hfill
\begin{minipage}[b]{0.245\textwidth}
    \centering
    \includegraphics[width=\textwidth]{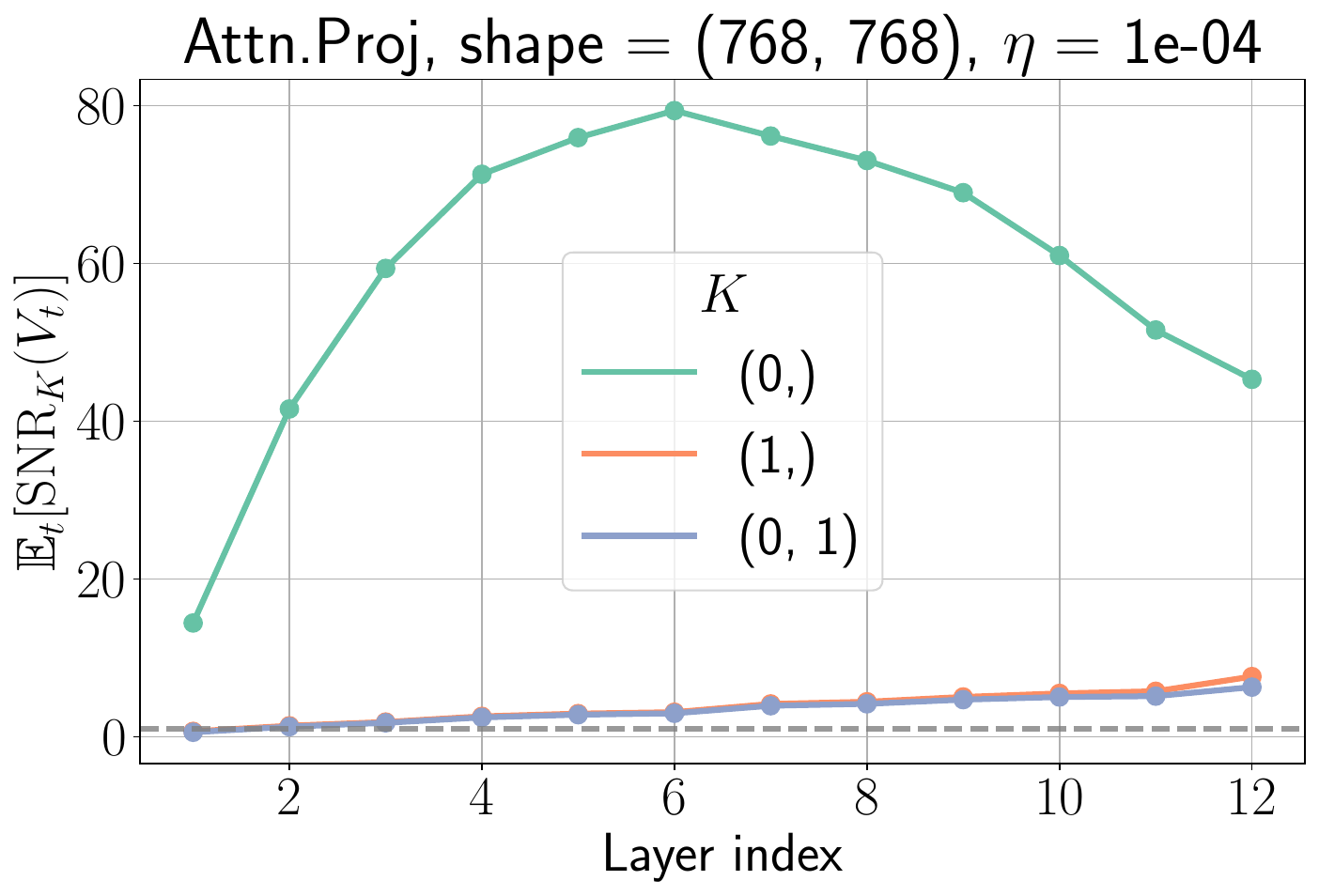}
\end{minipage}

\begin{minipage}[b]{0.245\textwidth}
    \centering
    \includegraphics[width=\textwidth]{figures/snr-analysis/snr-layer/ViT/snr_layer_MLP.Up_cifar-100_ViT_n768_d12_12_p2_B128_T20000_Tw2048_SlimAdamw_None_lr1e-04_b0.9_b0.999_eps1e-08_wd0.01_True.pdf}
\end{minipage}
\hfill
\begin{minipage}[b]{0.245\textwidth}
    \centering
    \includegraphics[width=\textwidth]{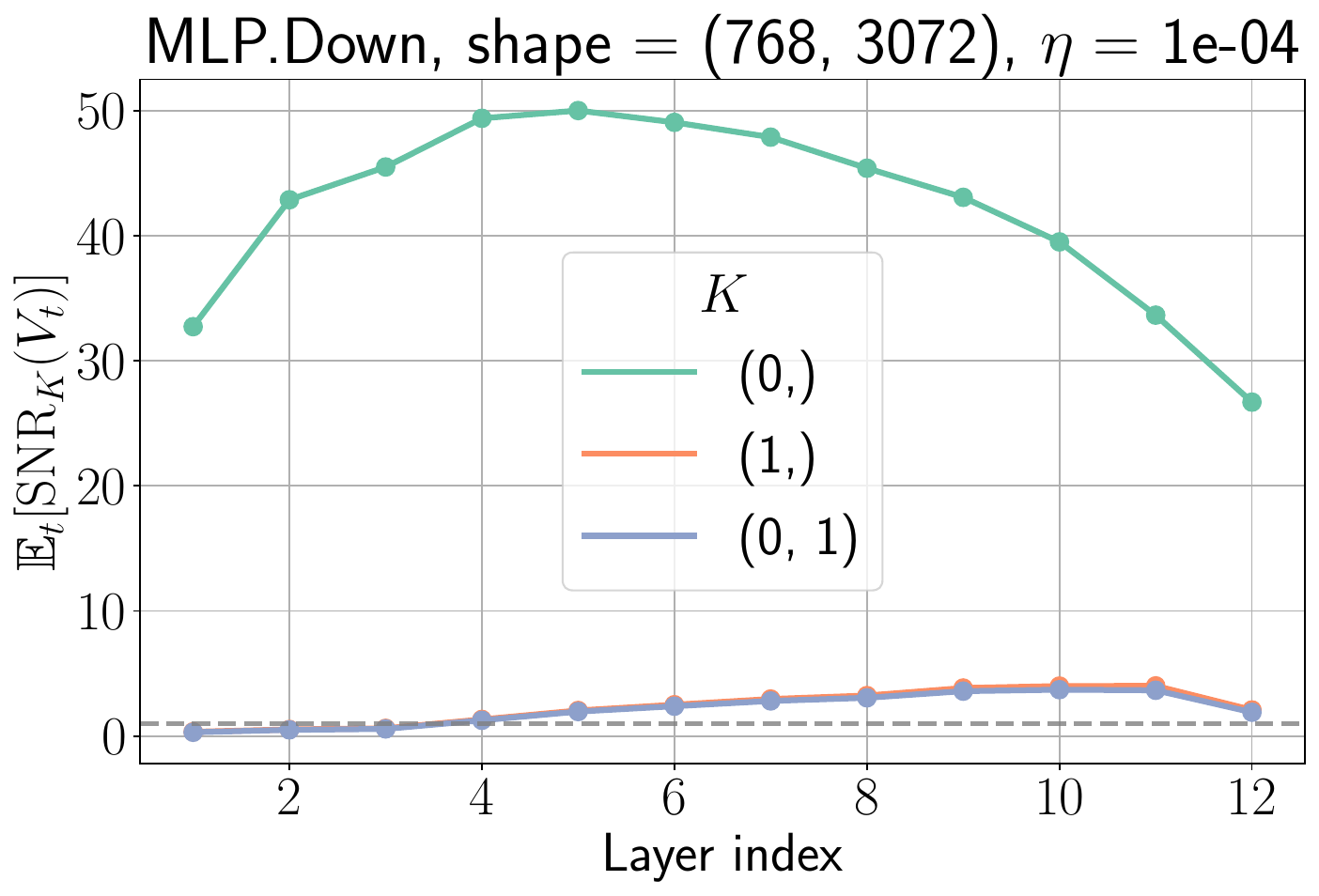}
\end{minipage}
\hfill
\begin{minipage}[b]{0.245\textwidth}
    \centering
    \includegraphics[width=\textwidth]{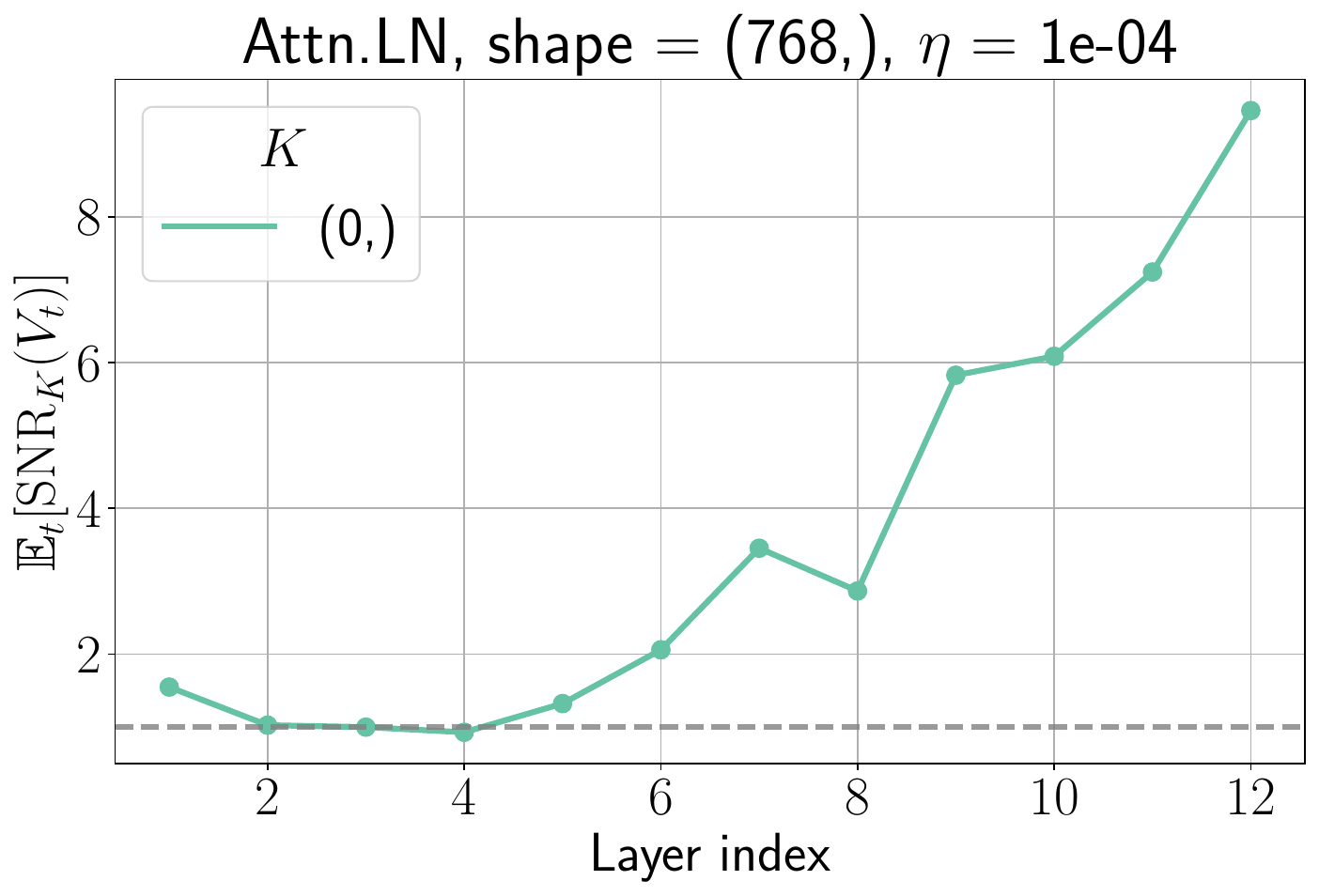}
\end{minipage}
\hfill
\begin{minipage}[b]{0.245\textwidth}
    \centering
    \includegraphics[width=\textwidth]{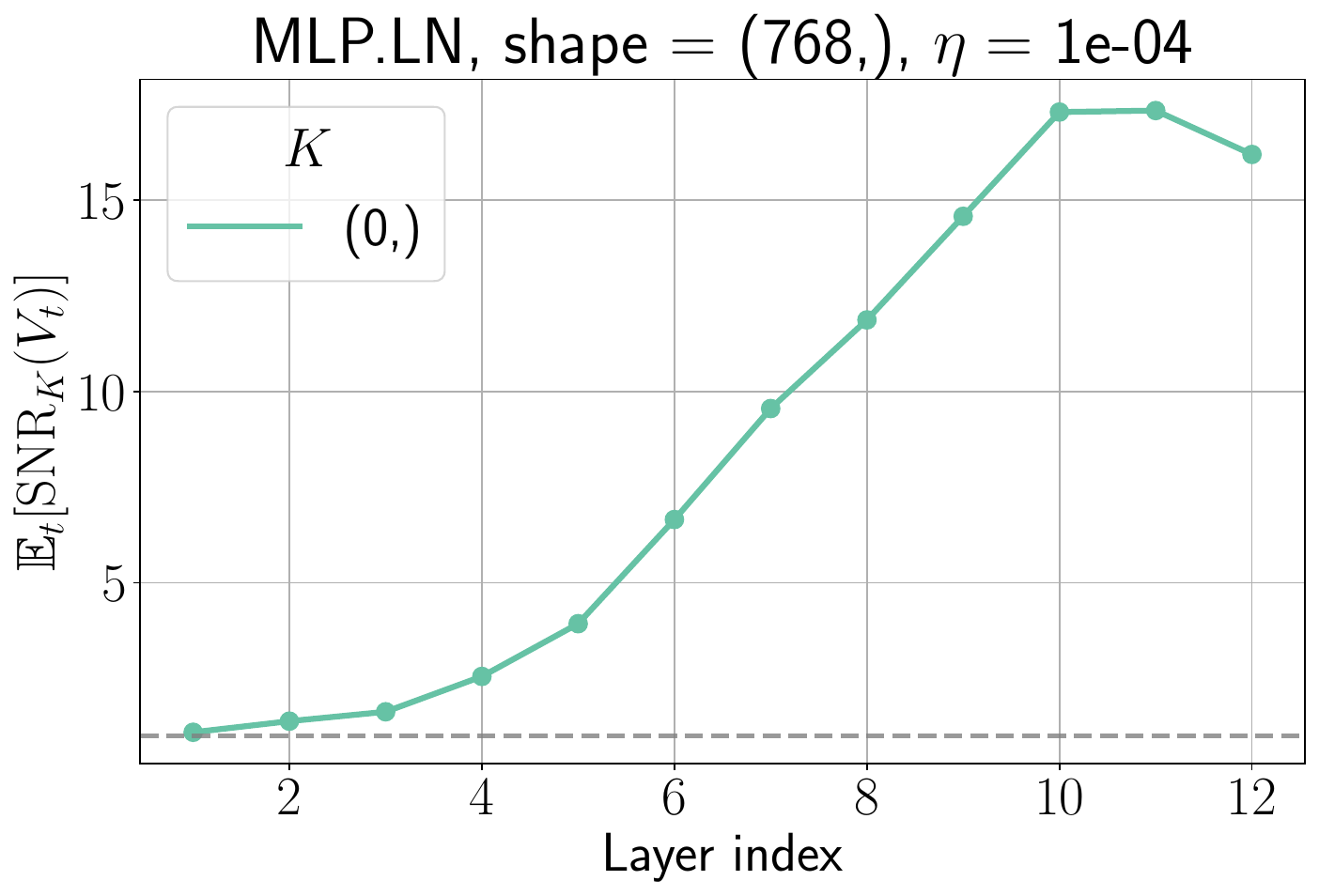}
\end{minipage}

\begin{minipage}[b]{0.245\textwidth}
    \centering
    \includegraphics[width=\textwidth]{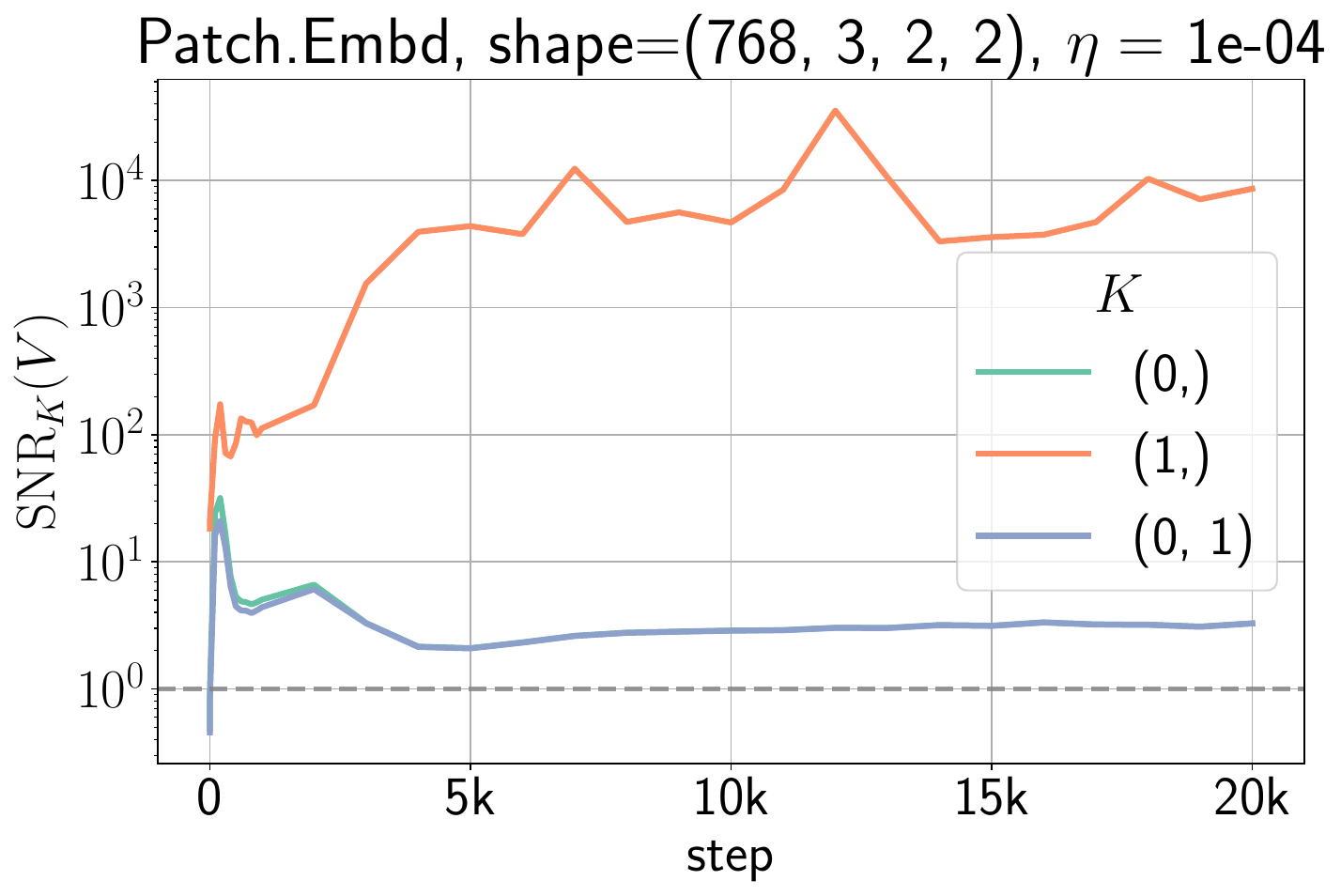}
\end{minipage}
\hfill
\begin{minipage}[b]{0.245\textwidth}
    \centering
    \includegraphics[width=\textwidth]{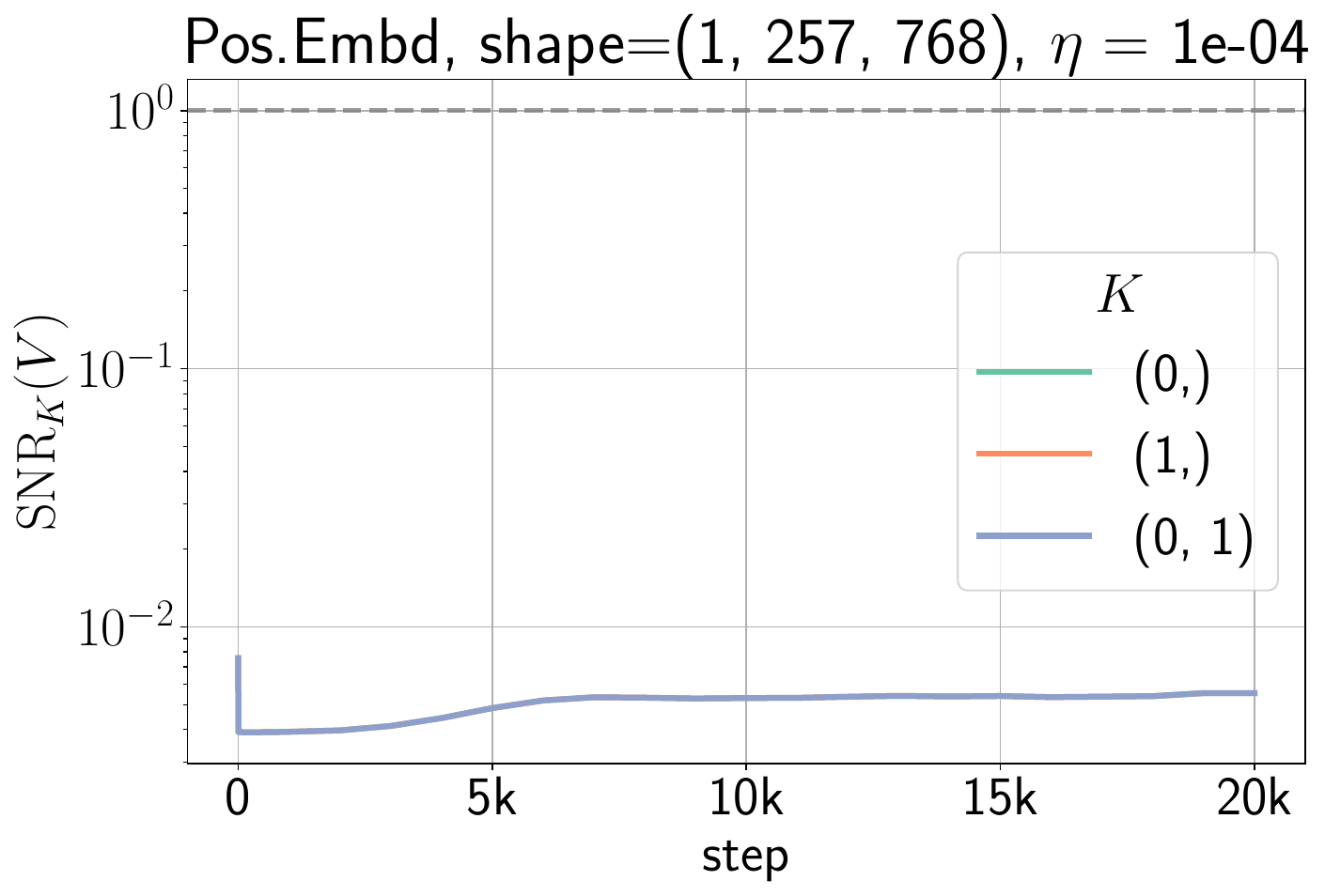}
\end{minipage}
\hfill
\begin{minipage}[b]{0.245\textwidth}
    \centering
    \includegraphics[width=\textwidth]{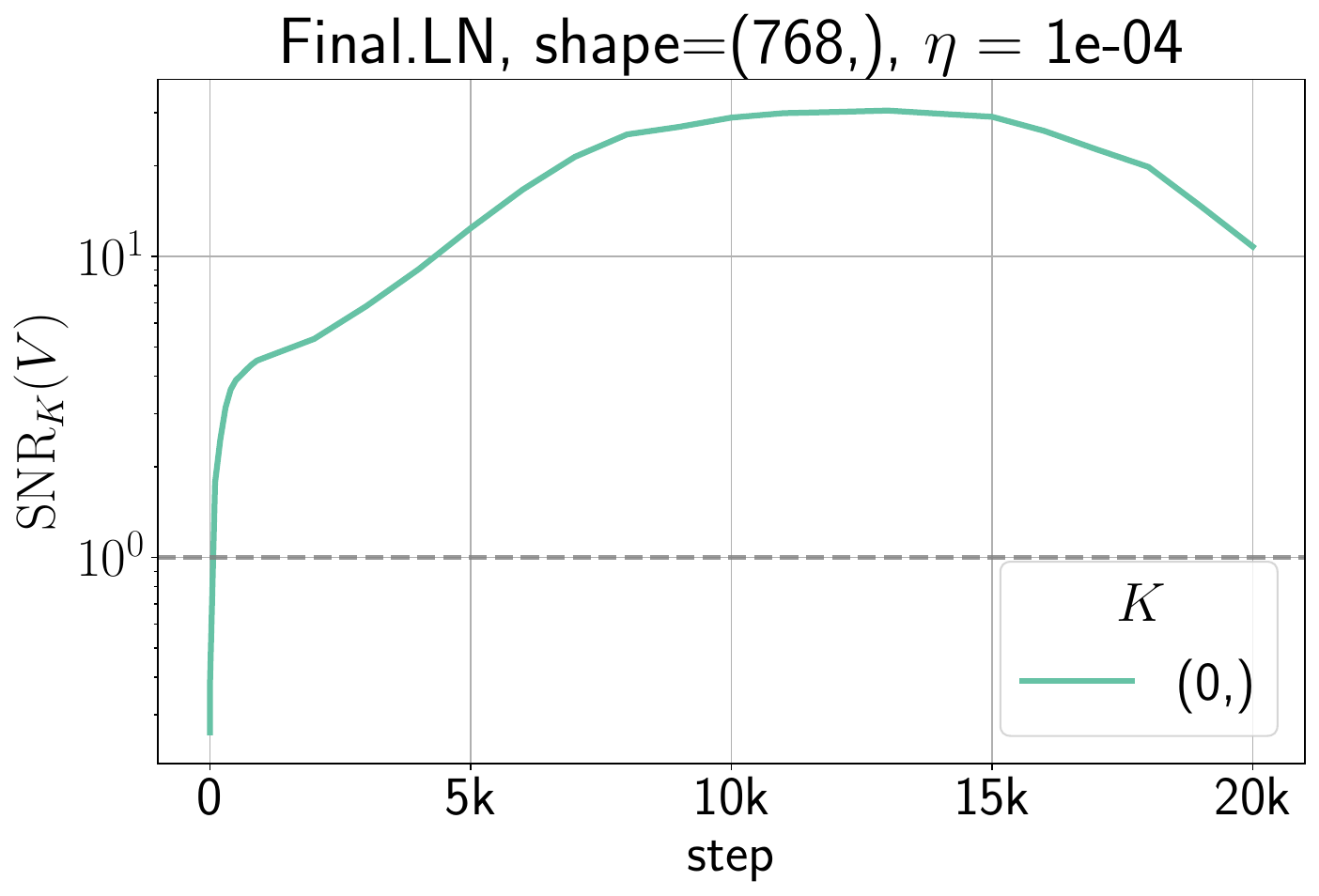}
\end{minipage}
\hfill
\begin{minipage}[b]{0.245\textwidth}
    \centering
    \includegraphics[width=\textwidth]{figures/snr-analysis/snr-trajectories/ViT/snr_LM.Head_LNone_cifar-100_ViT_n768_d12_12_p2_B128_T20000_Tw2048_SlimAdamw_None_lr1e-04_b0.9_b0.999_eps1e-08_wd0.01_True.pdf}
\end{minipage}
\caption{SNR trends of different layers of ViT-small trained on CIFAR-100.}
\label{fig:snr-vit-cifar-100-full}
\end{figure*}

Next, we examine the SNR trends of ResNets and ViTs trained on image classification tasks. 
As shown in \Cref{fig:snr-resnet-cifar-10,fig:snr-resnet-cifar-100-full}, ResNets trained on both CIFAR-10 and CIFAR-100 exhibit consistently high SNR values, suggesting compressibility. Most layers maintain high SNR values throughout training, with notable exceptions at the network boundaries. The first convolutional layer averses compressibility along the $\text{fan}_{\text{out}}$ dimension, while the final layer exhibits declining SNR values during later training stages when both dimensions are compressed. Unlike LayerNorm in Transformers, BatchNorm layers demonstrate SNR values around $1.0$ throughout training.

\begin{figure*}[!htb]
\centering
\begin{minipage}[b]{0.245\textwidth}
    \centering
    \includegraphics[width=\textwidth]{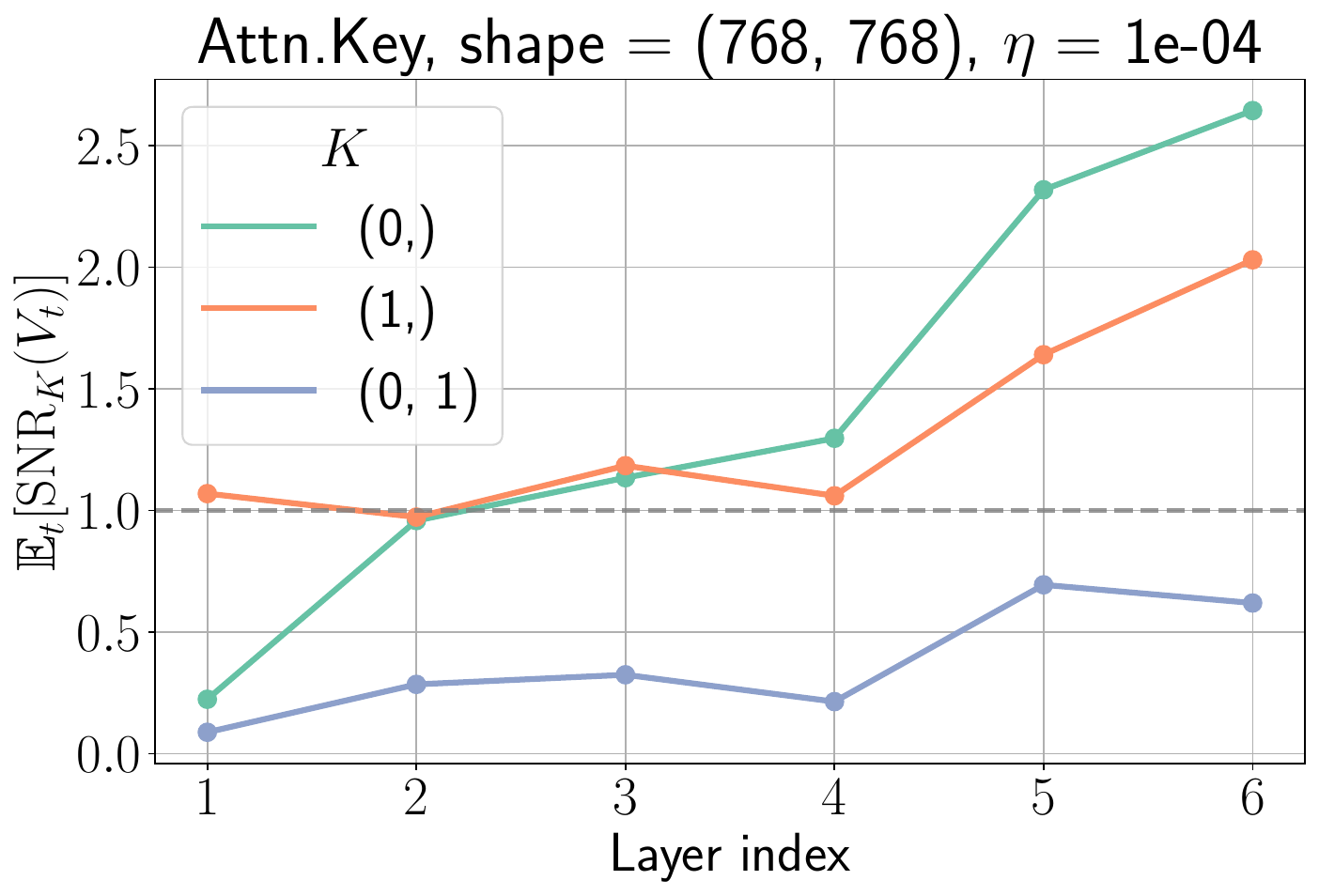}
\end{minipage}
\hfill
\begin{minipage}[b]{0.245\textwidth}
    \centering
    \includegraphics[width=\textwidth]{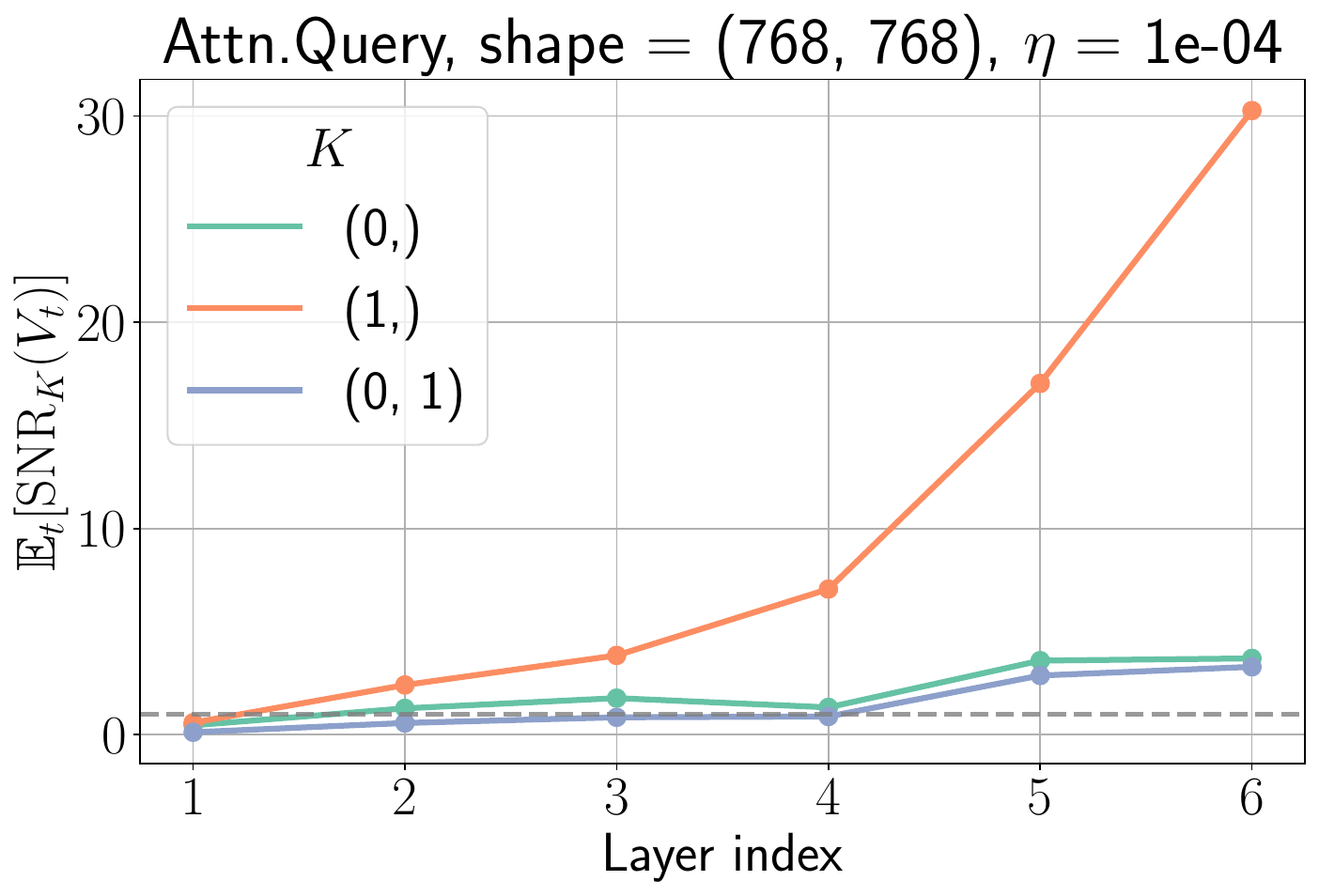}
\end{minipage}
\hfill
\begin{minipage}[b]{0.245\textwidth}
    \centering
    \includegraphics[width=\textwidth]{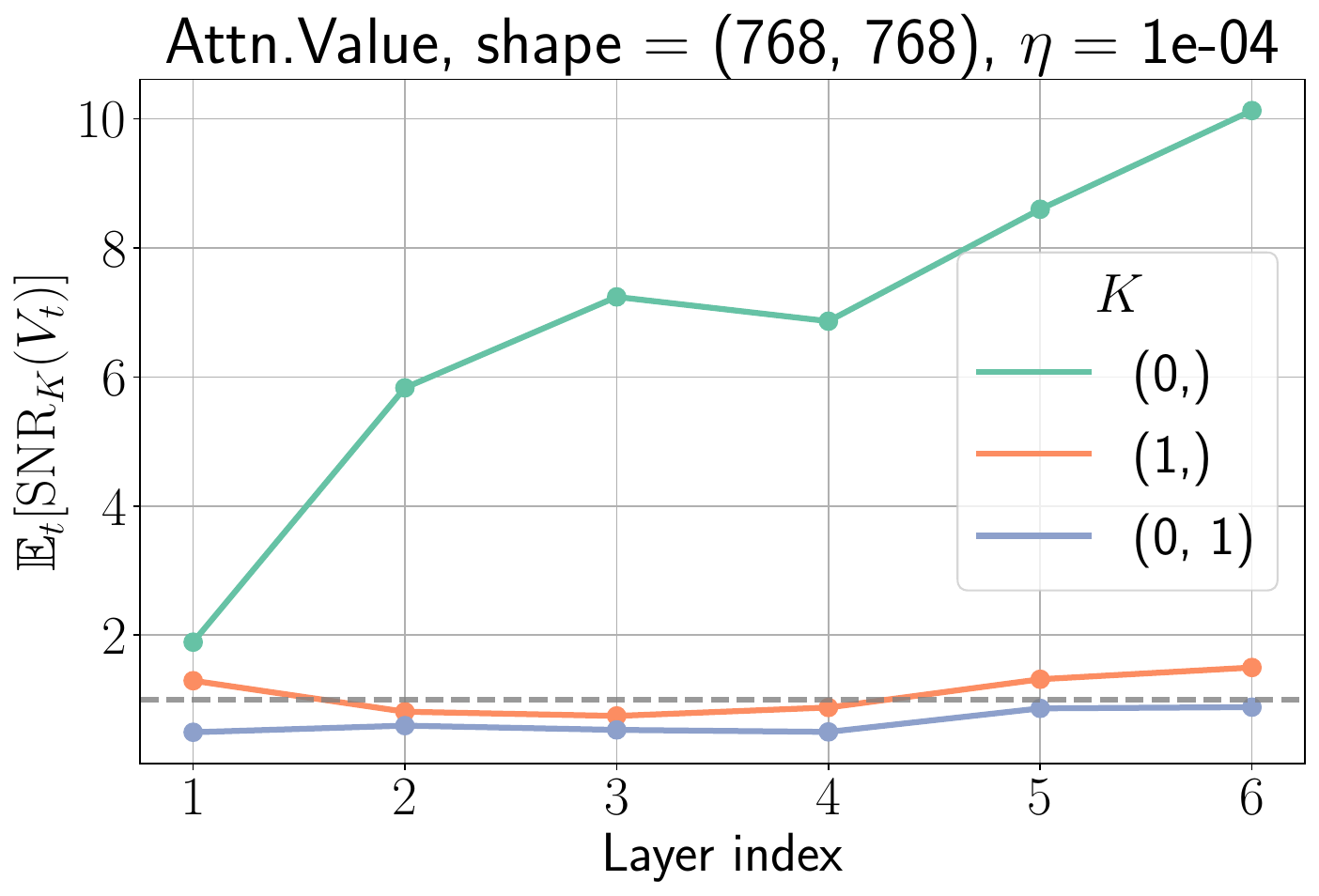}
\end{minipage}
\hfill
\begin{minipage}[b]{0.245\textwidth}
    \centering
    \includegraphics[width=\textwidth]{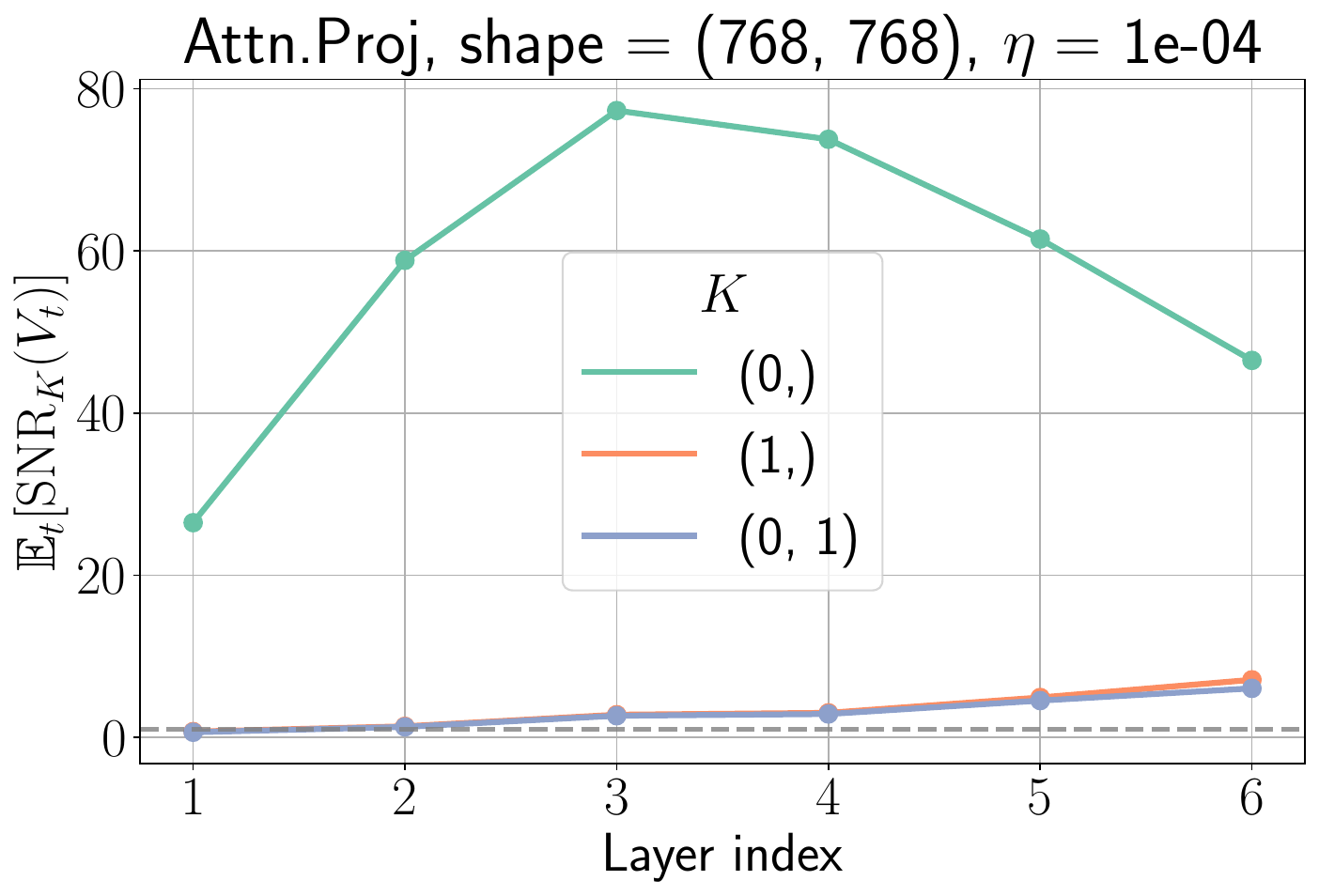}
\end{minipage}

\begin{minipage}[b]{0.245\textwidth}
    \centering
    \includegraphics[width=\textwidth]{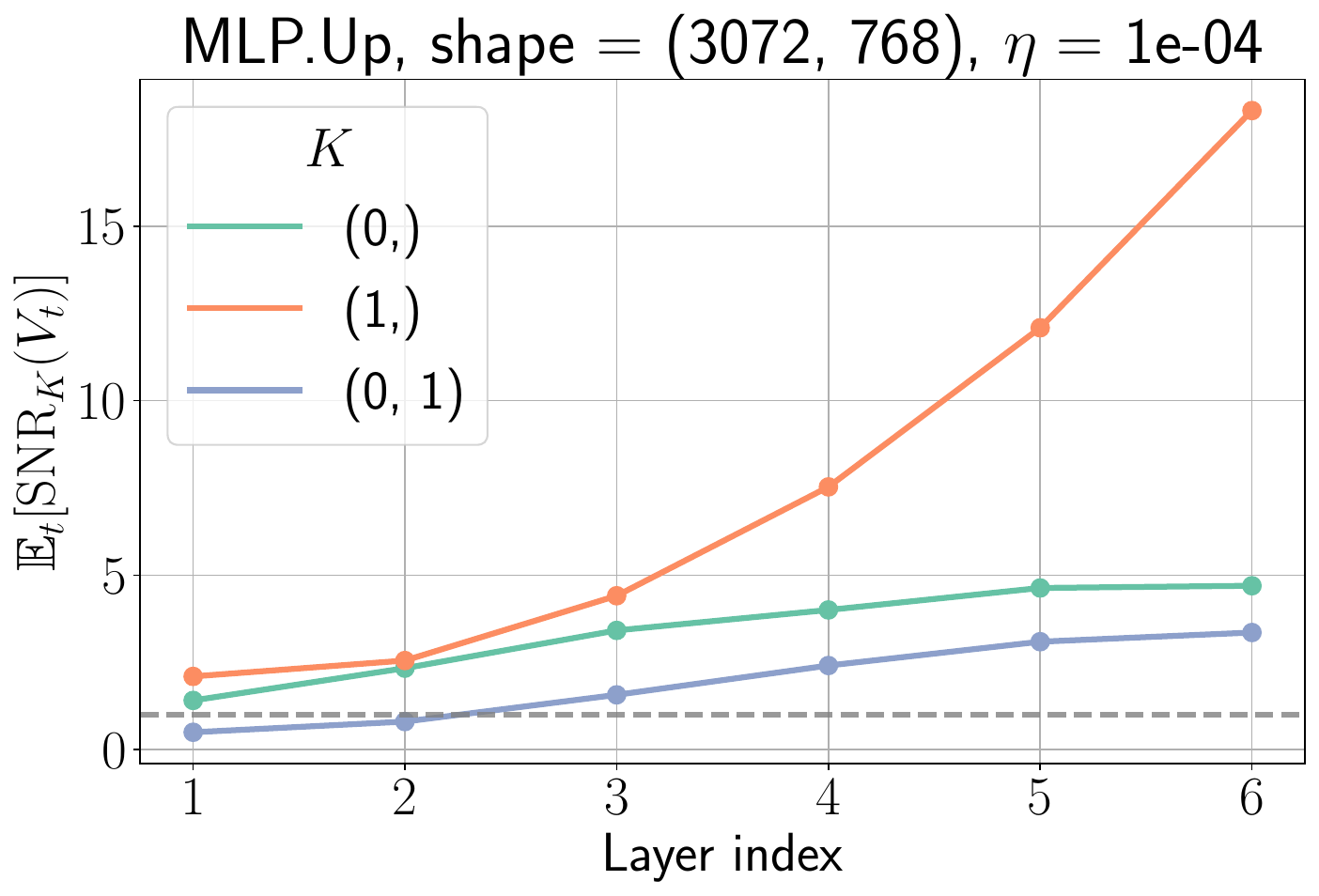}
\end{minipage}
\hfill
\begin{minipage}[b]{0.245\textwidth}
    \centering
    \includegraphics[width=\textwidth]{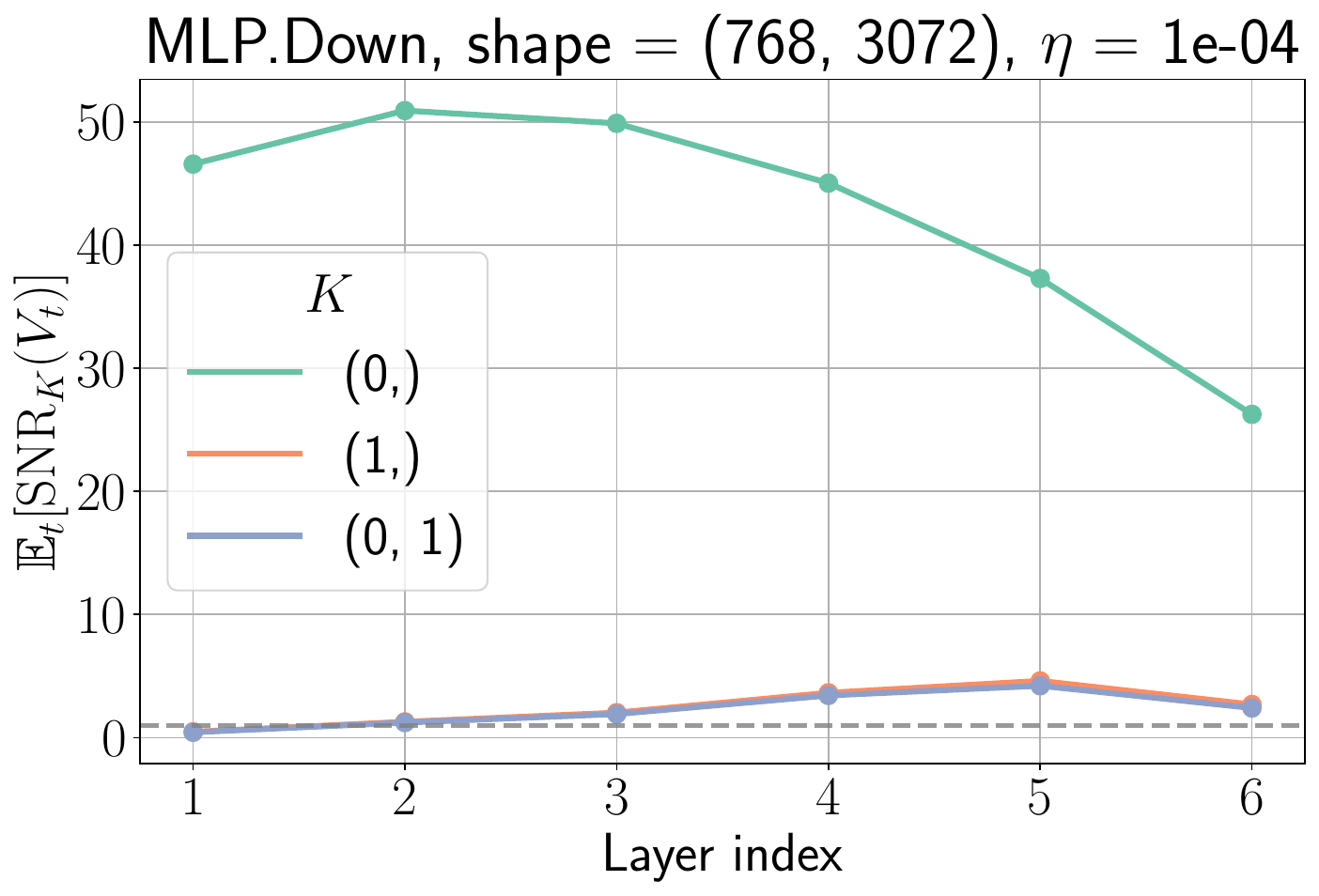}
\end{minipage}
\hfill
\begin{minipage}[b]{0.245\textwidth}
    \centering
    \includegraphics[width=\textwidth]{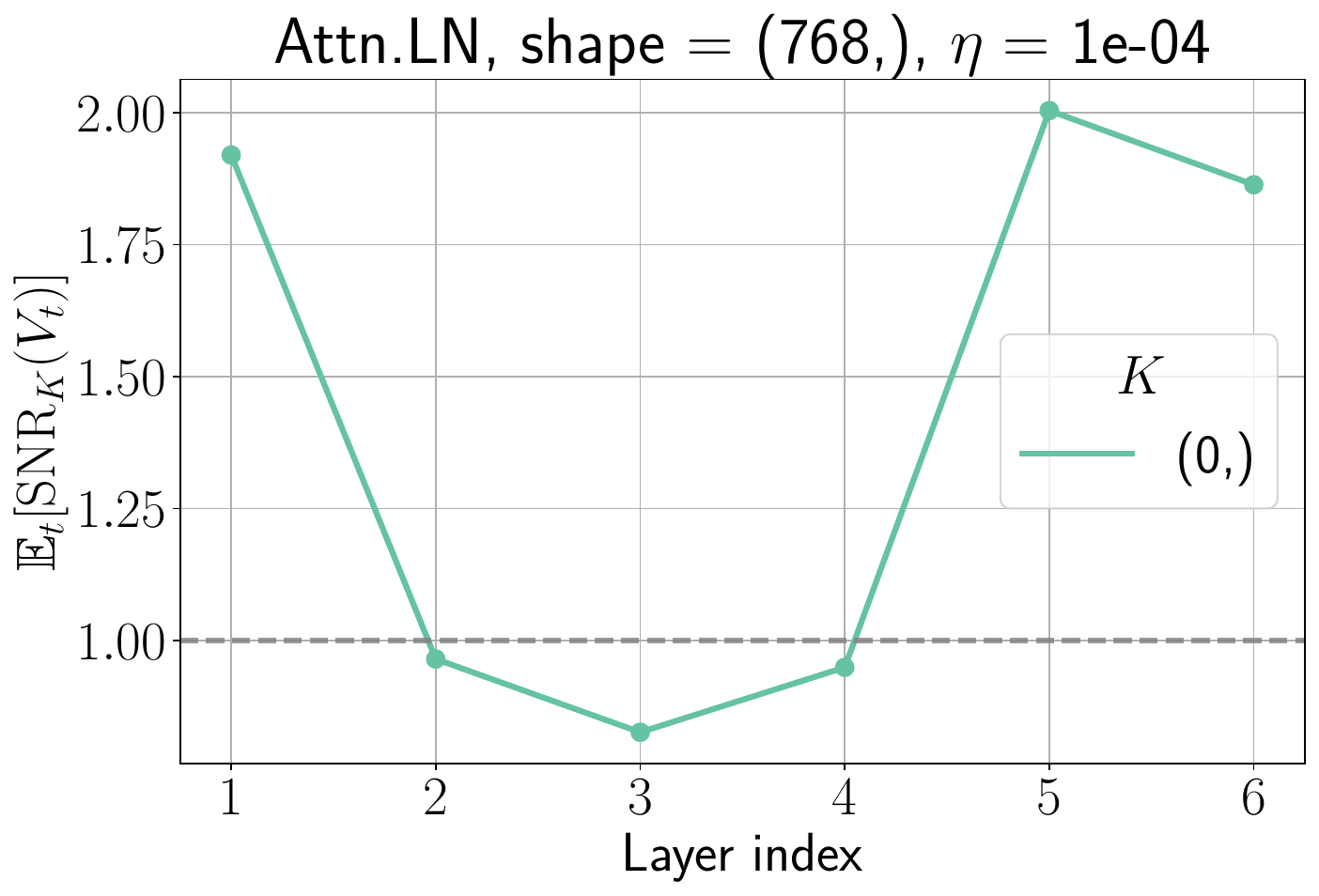}
\end{minipage}
\hfill
\begin{minipage}[b]{0.245\textwidth}
    \centering
    \includegraphics[width=\textwidth]{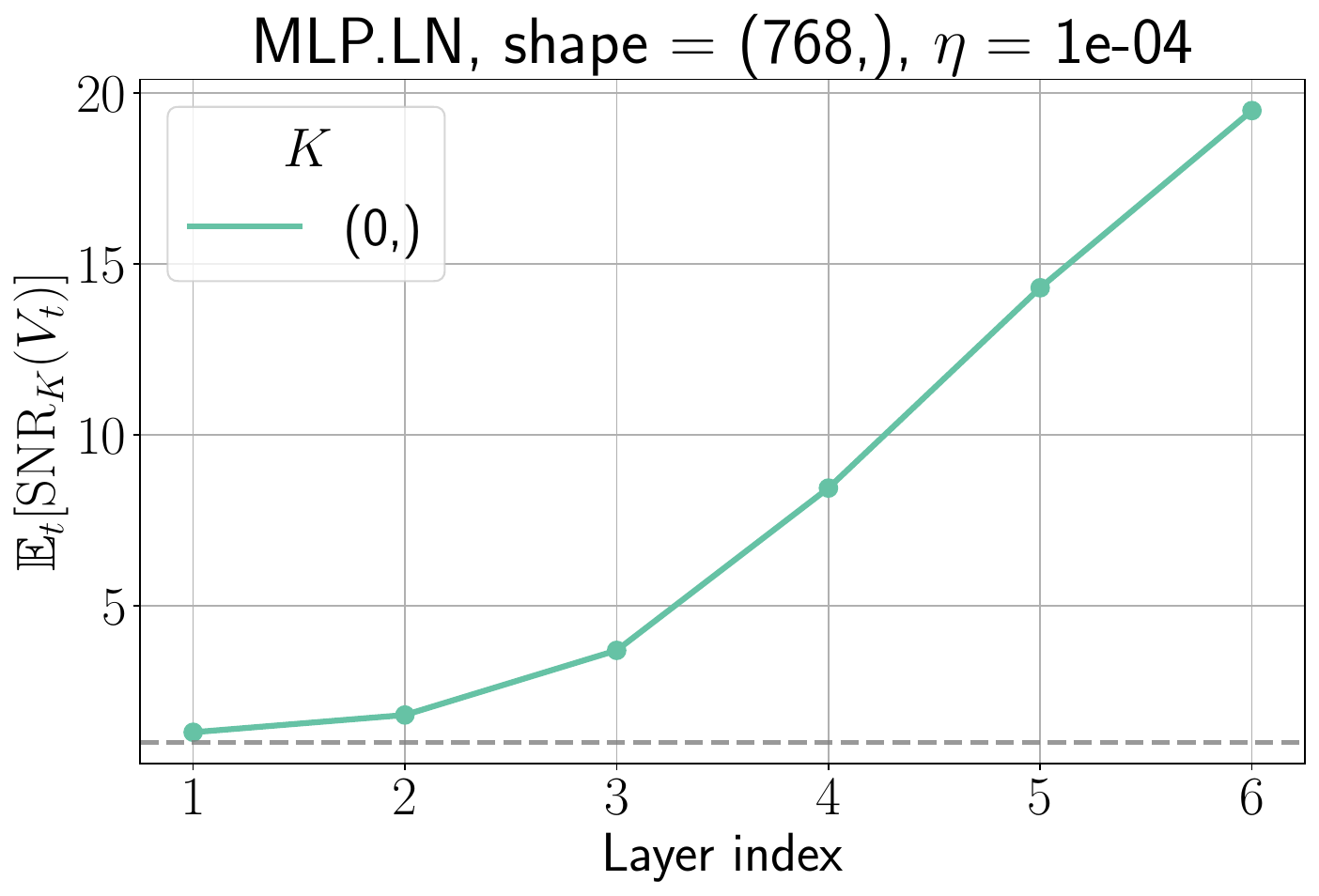}
\end{minipage}

\begin{minipage}[b]{0.245\textwidth}
    \centering
    \includegraphics[width=\textwidth]{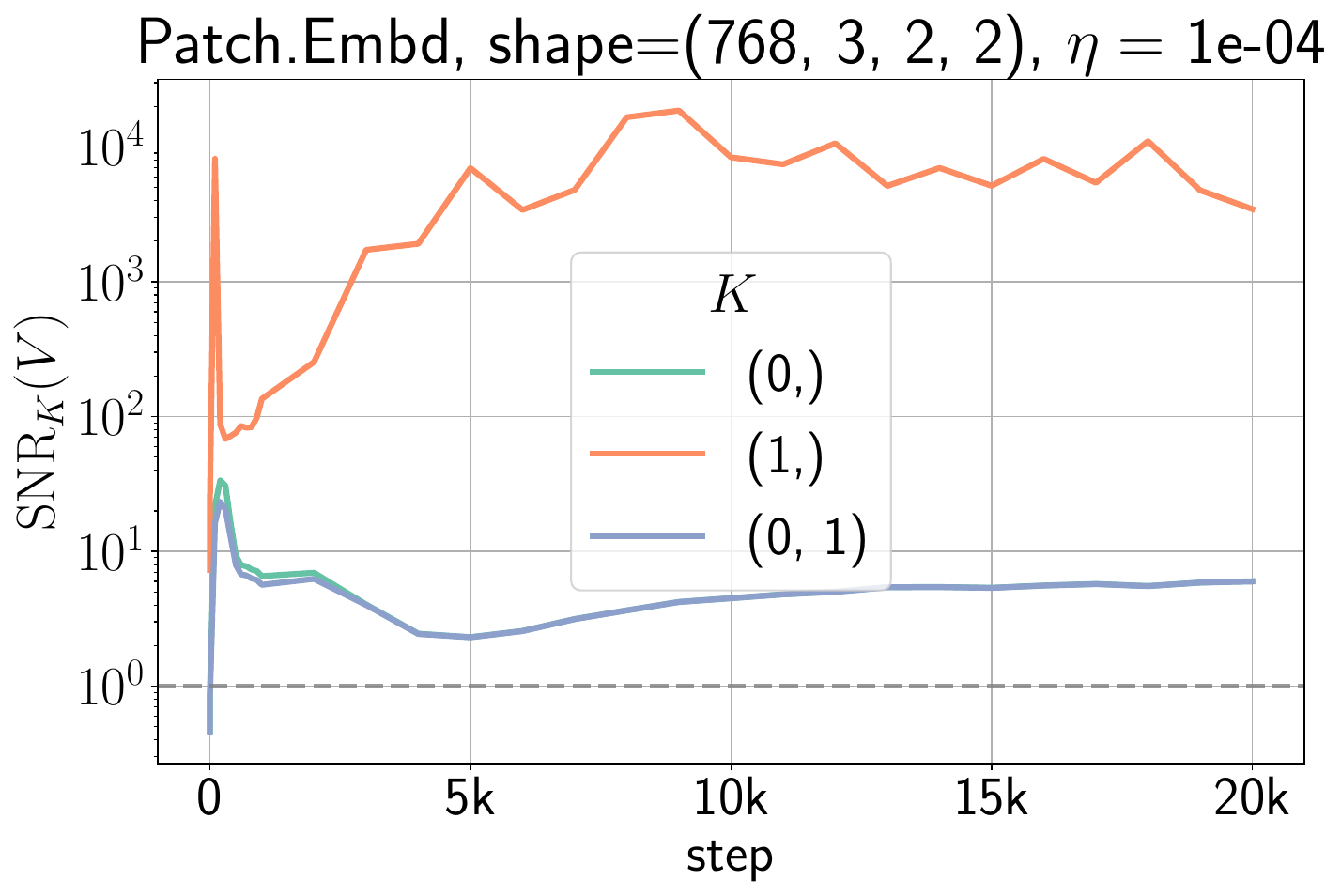}
\end{minipage}
\hfill
\begin{minipage}[b]{0.245\textwidth}
    \centering
    \includegraphics[width=\textwidth]{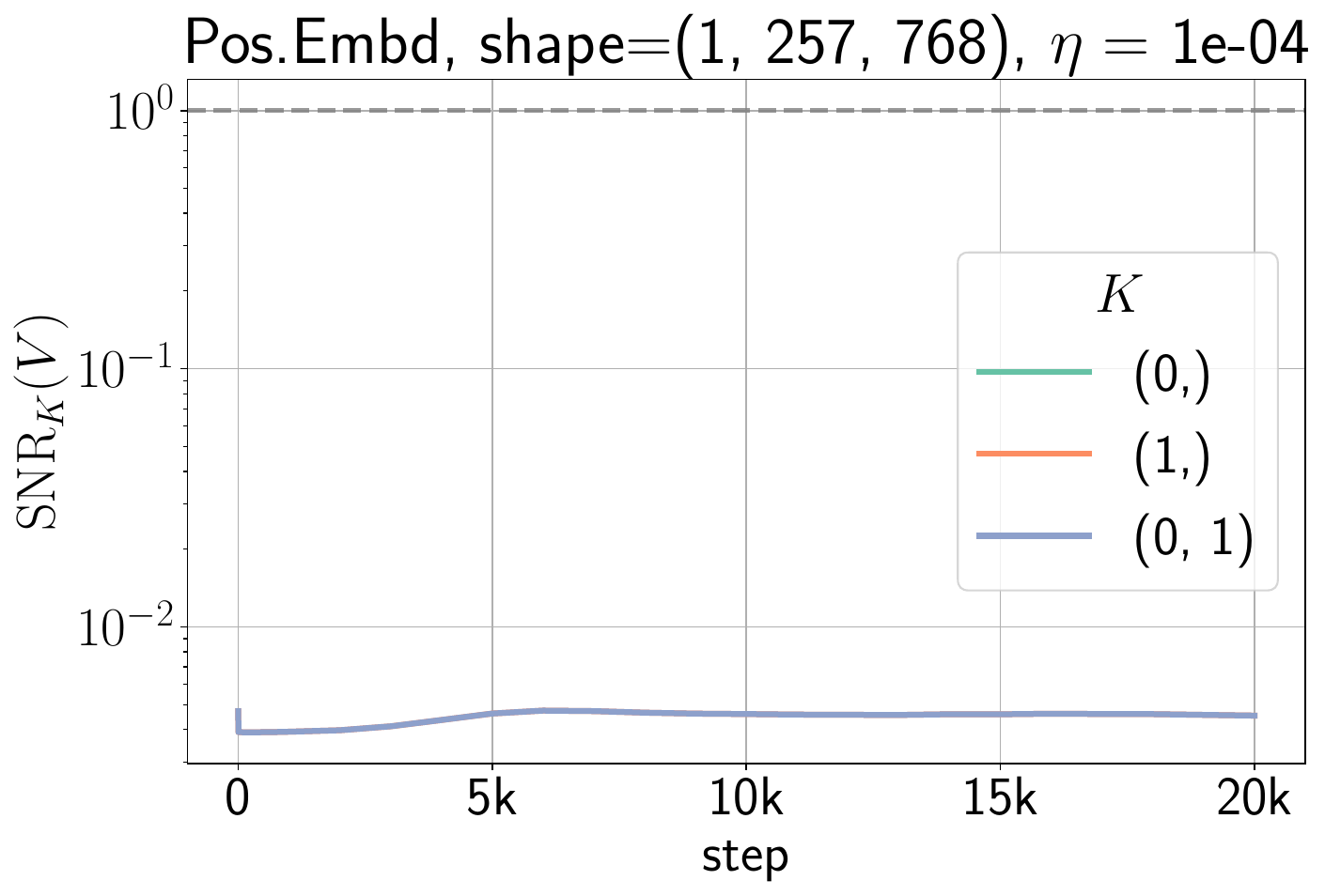}
\end{minipage}
\hfill
\begin{minipage}[b]{0.245\textwidth}
    \centering
    \includegraphics[width=\textwidth]{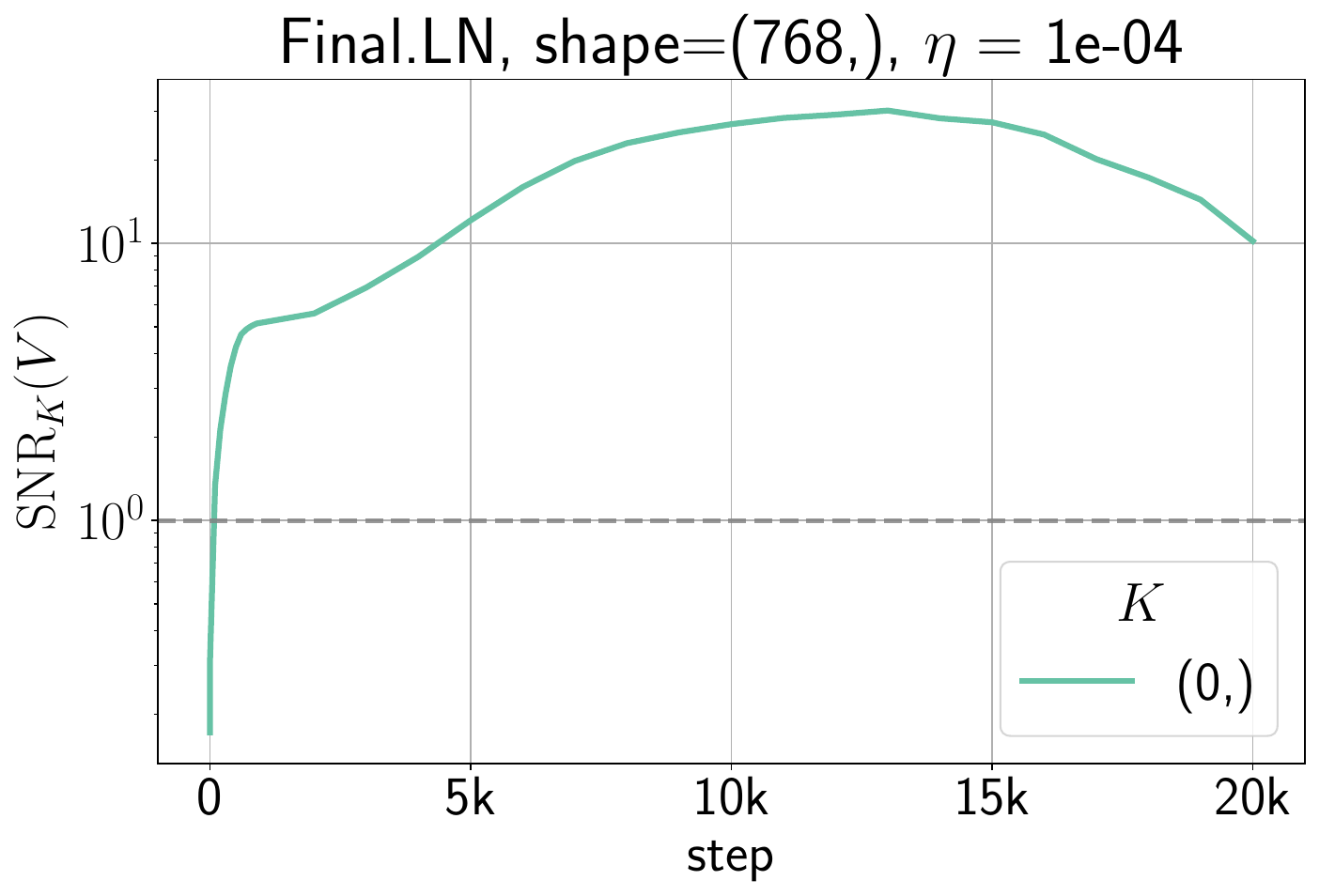}
\end{minipage}
\hfill
\begin{minipage}[b]{0.245\textwidth}
    \centering
    \includegraphics[width=\textwidth]{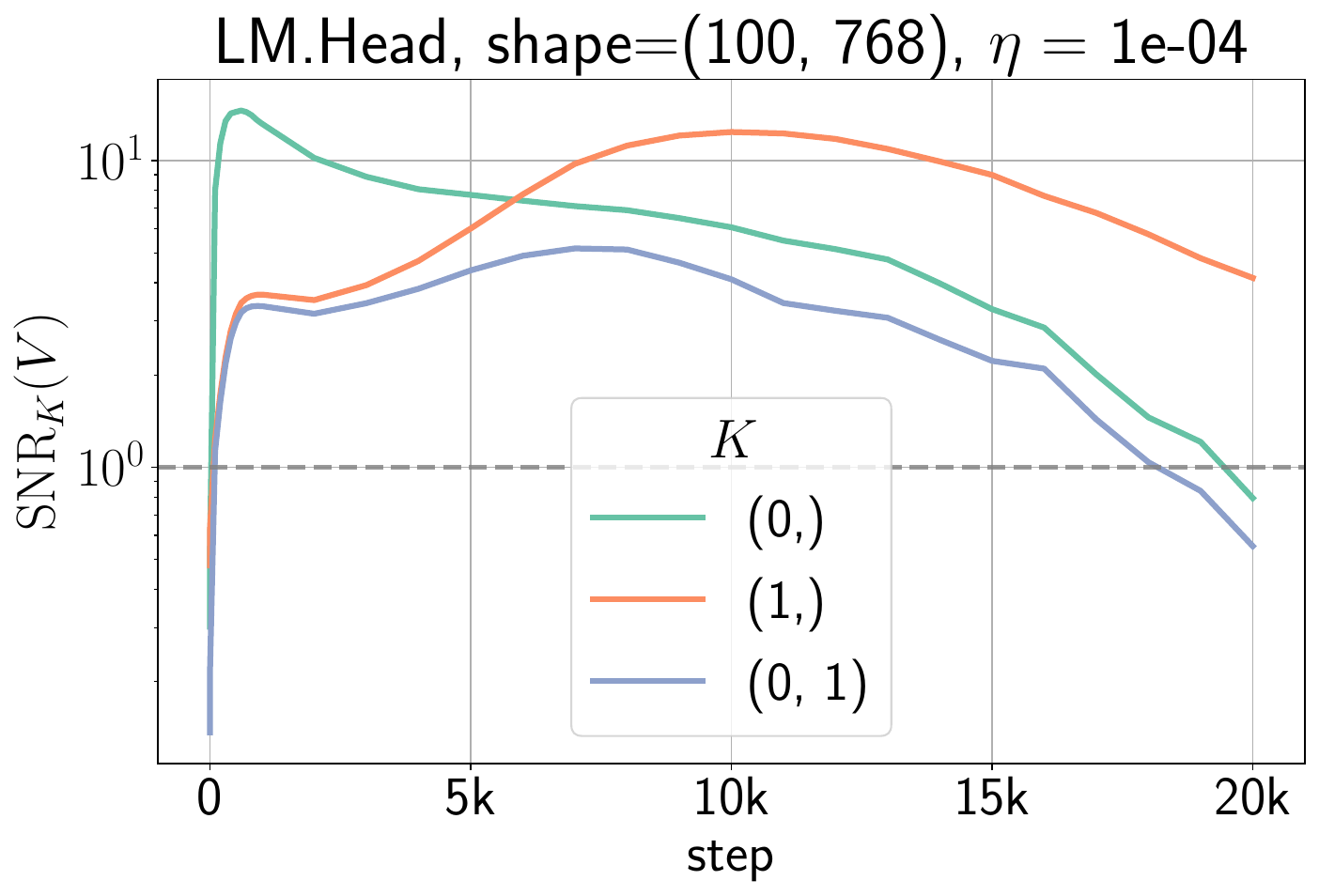}
\end{minipage}
\caption{SNR trends of different layers of ViT-mini trained on CIFAR-100.}
\label{fig:snr-vit-mini-cifar-100}
\end{figure*}

\clearpage
\newpage

\section{Effect of Large Learning Rates on Compressibility}
\label{appendix:large-learning-rates}

\begin{figure*}[!htb]
\centering
\begin{minipage}[b]{0.225\textwidth}
    \centering
    \includegraphics[width=\textwidth]{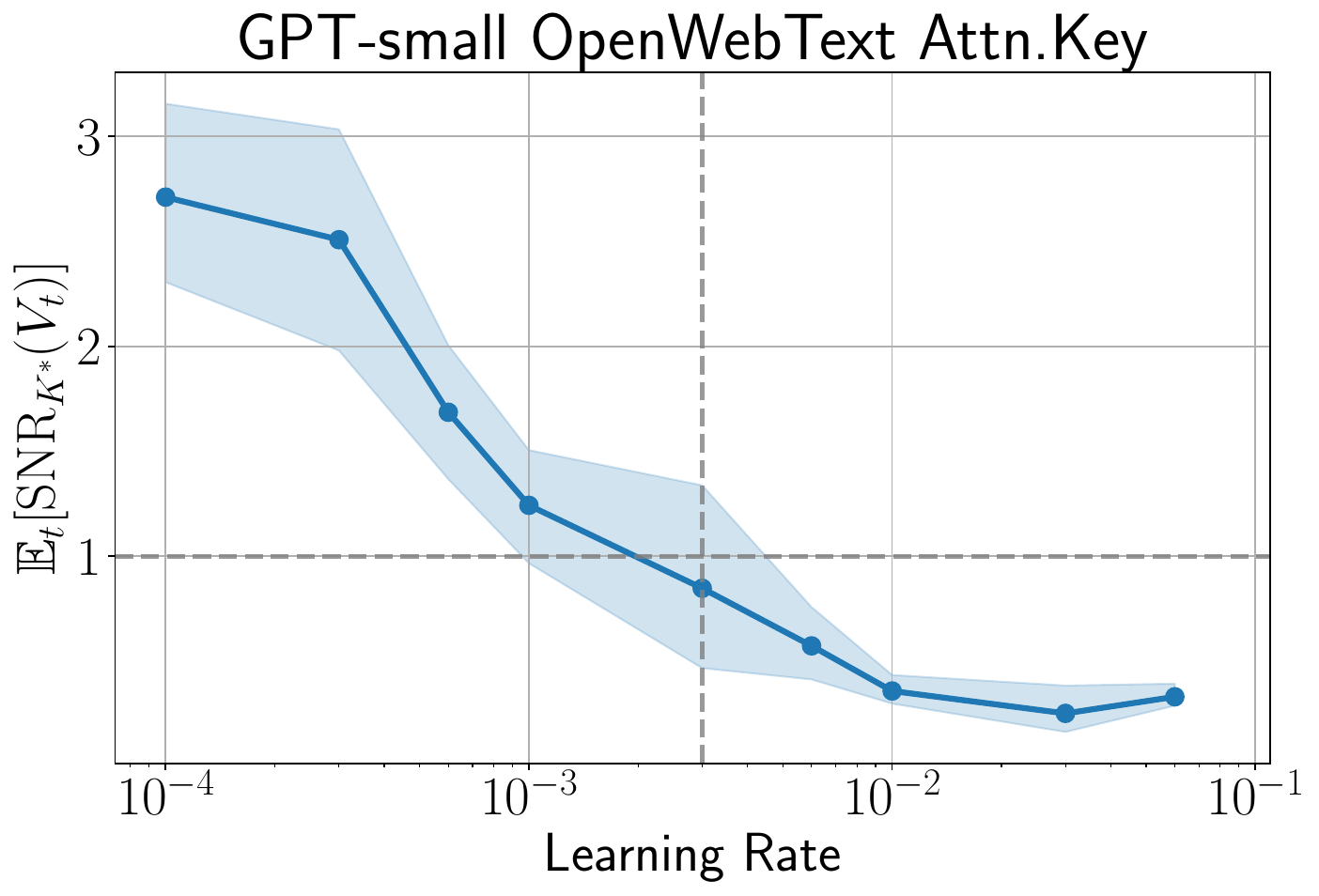}
\end{minipage}
\hfill
\begin{minipage}[b]{0.225\textwidth}
    \centering
    \includegraphics[width=\textwidth]{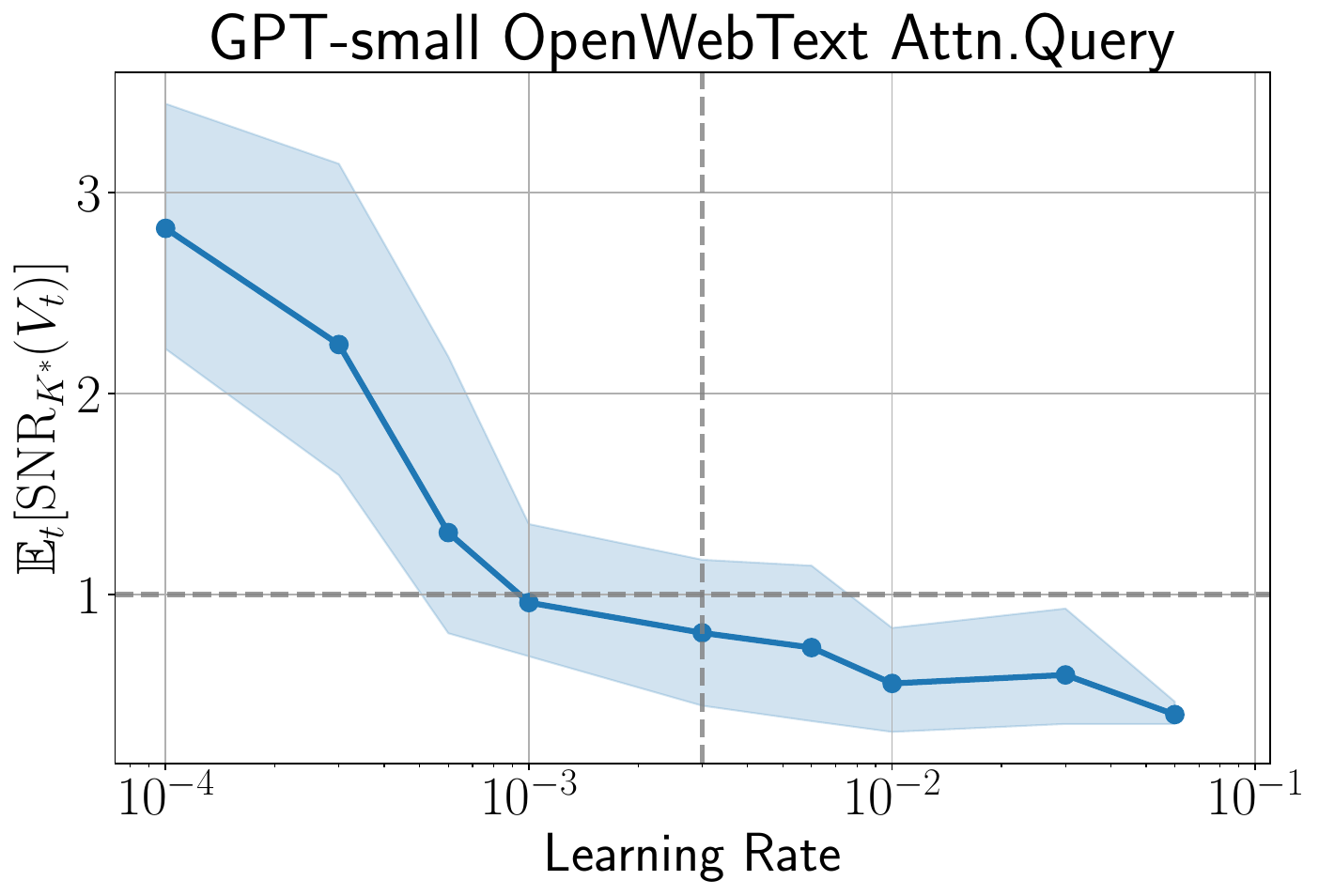}
\end{minipage}
\hfill
\begin{minipage}[b]{0.225\textwidth}
    \centering
    \includegraphics[width=\textwidth]{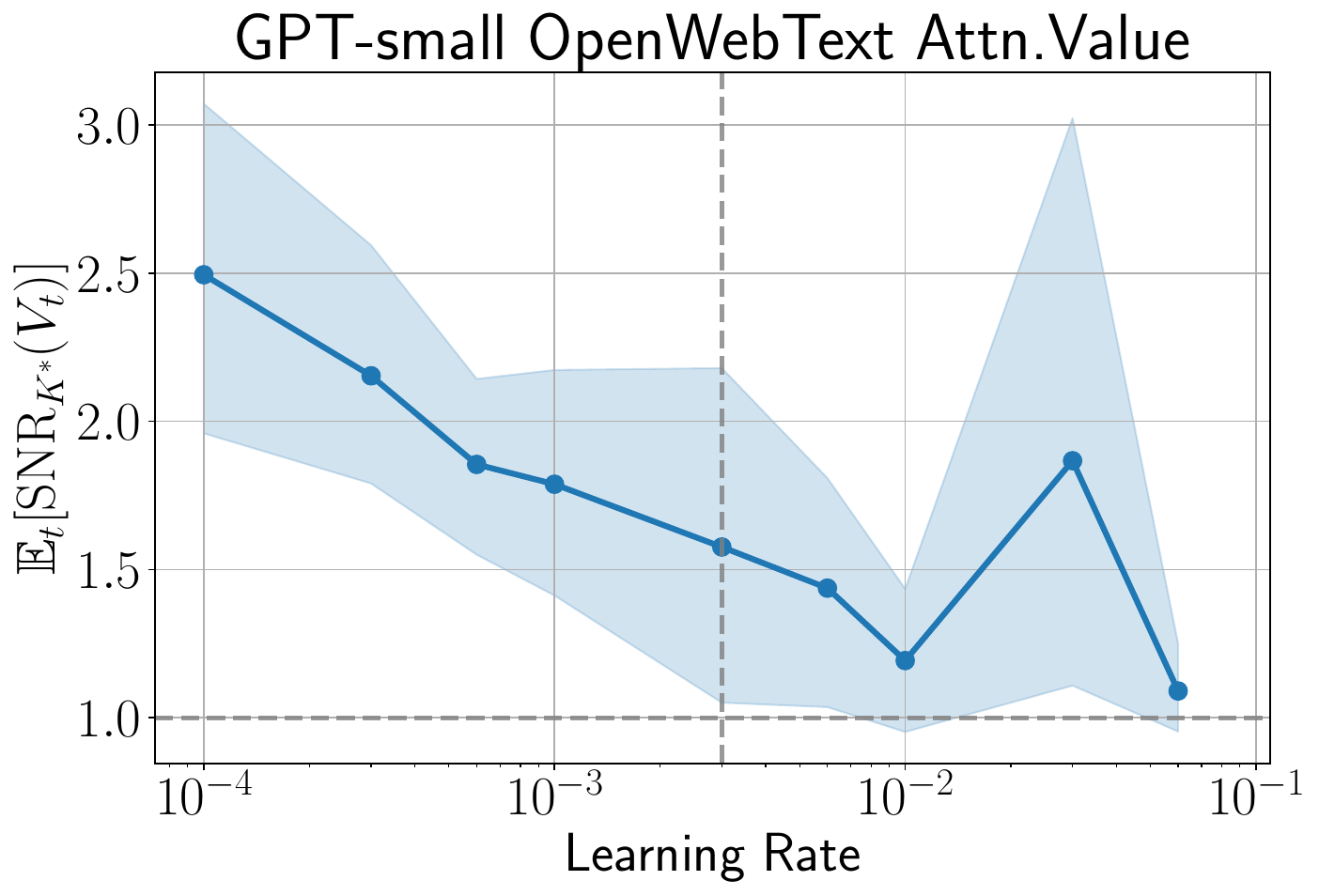}
\end{minipage}
\hfill
\begin{minipage}[b]{0.225\textwidth}
    \centering
    \includegraphics[width=\textwidth]{figures/snr-lr/snr_Attn.Proj_gpt2_openwebtext_SlimAdamW_T10000_ga40_d12_h12_n768_wd0.1_bs32_b0.9_b0.95.pdf}
\end{minipage}

\begin{minipage}[b]{0.225\textwidth}
    \centering
    \includegraphics[width=\textwidth]{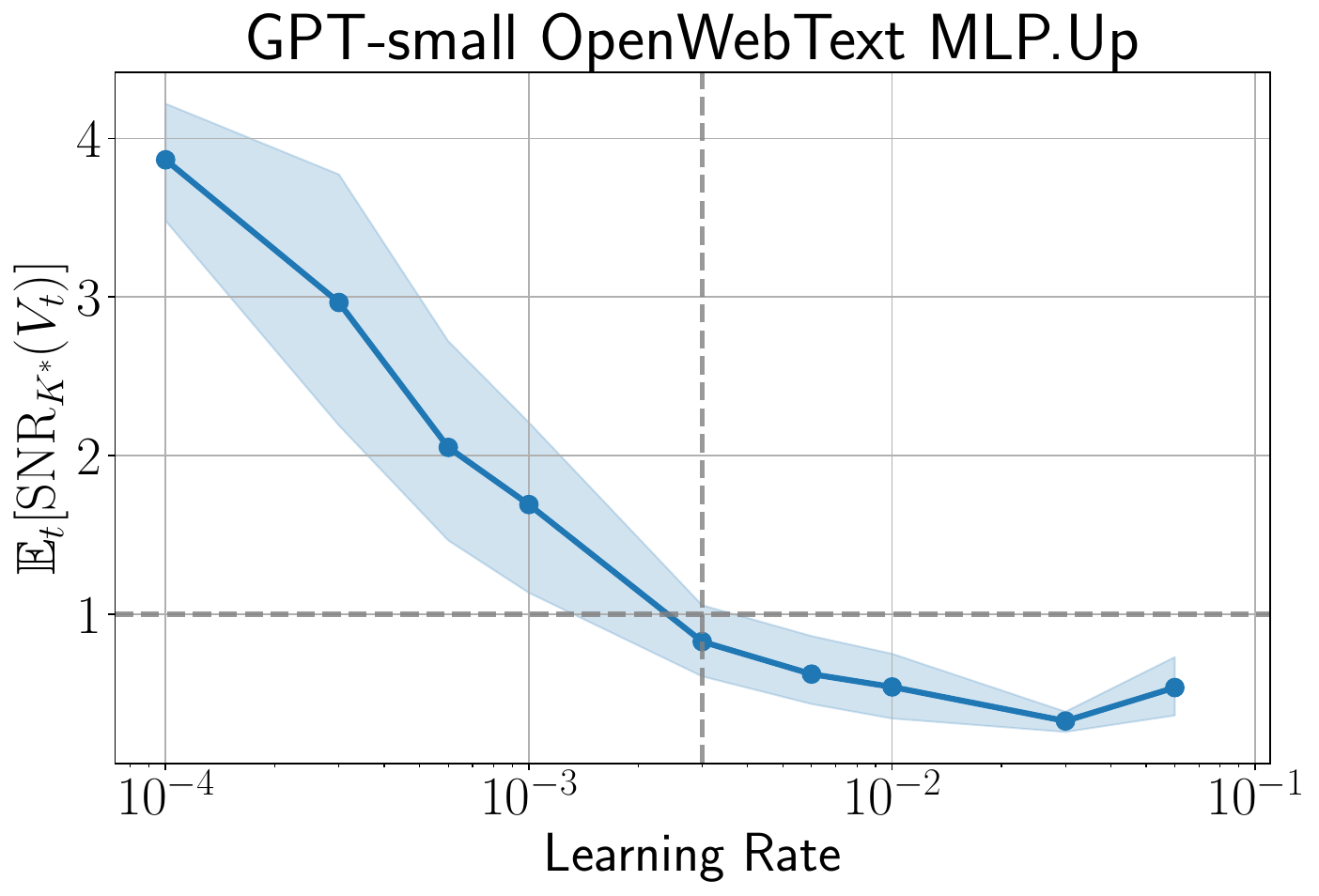}
\end{minipage}
\hfill
\begin{minipage}[b]{0.225\textwidth}
    \centering
    \includegraphics[width=\textwidth]{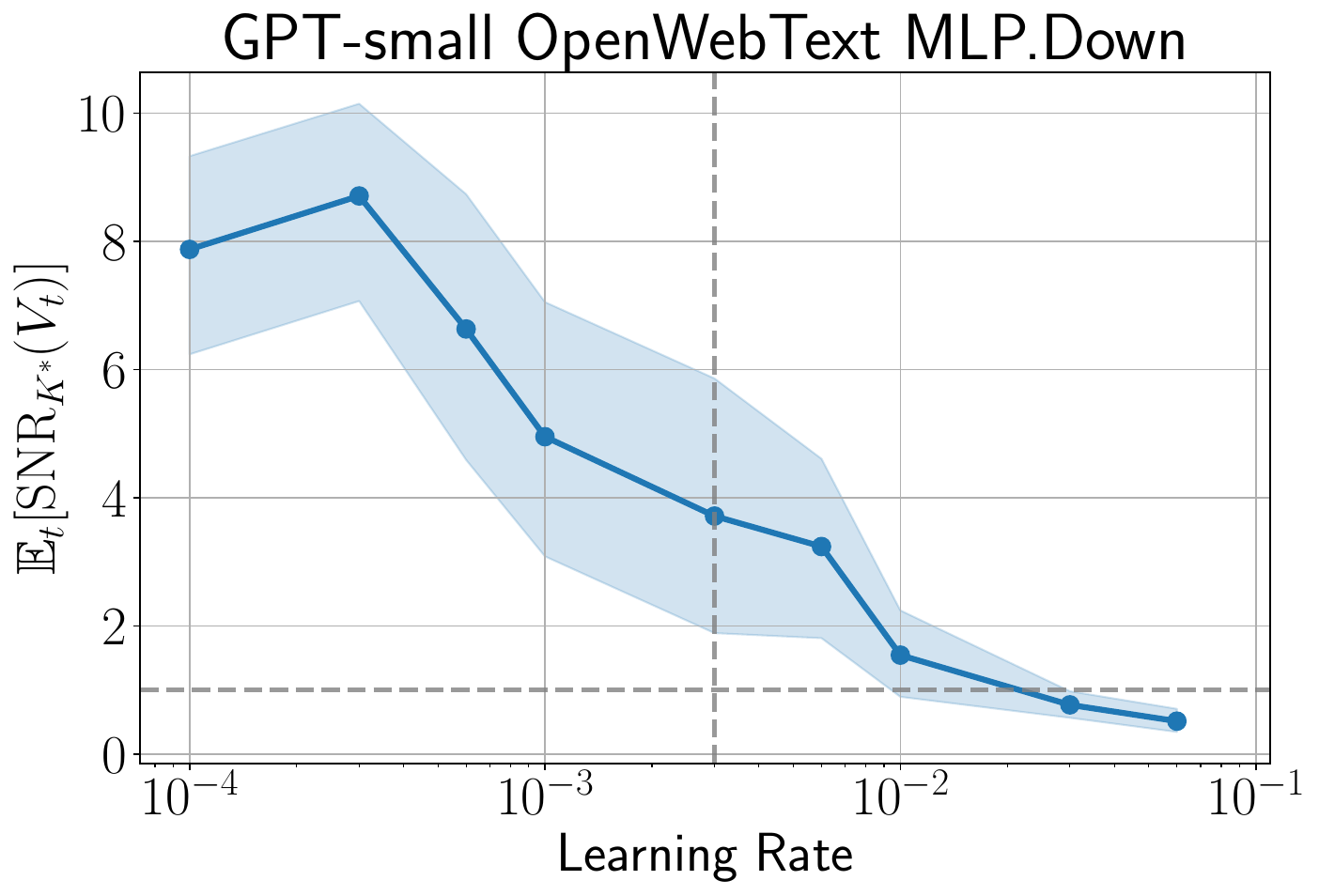}
\end{minipage}
\hfill
\begin{minipage}[b]{0.225\textwidth}
    \centering
    \includegraphics[width=\textwidth]{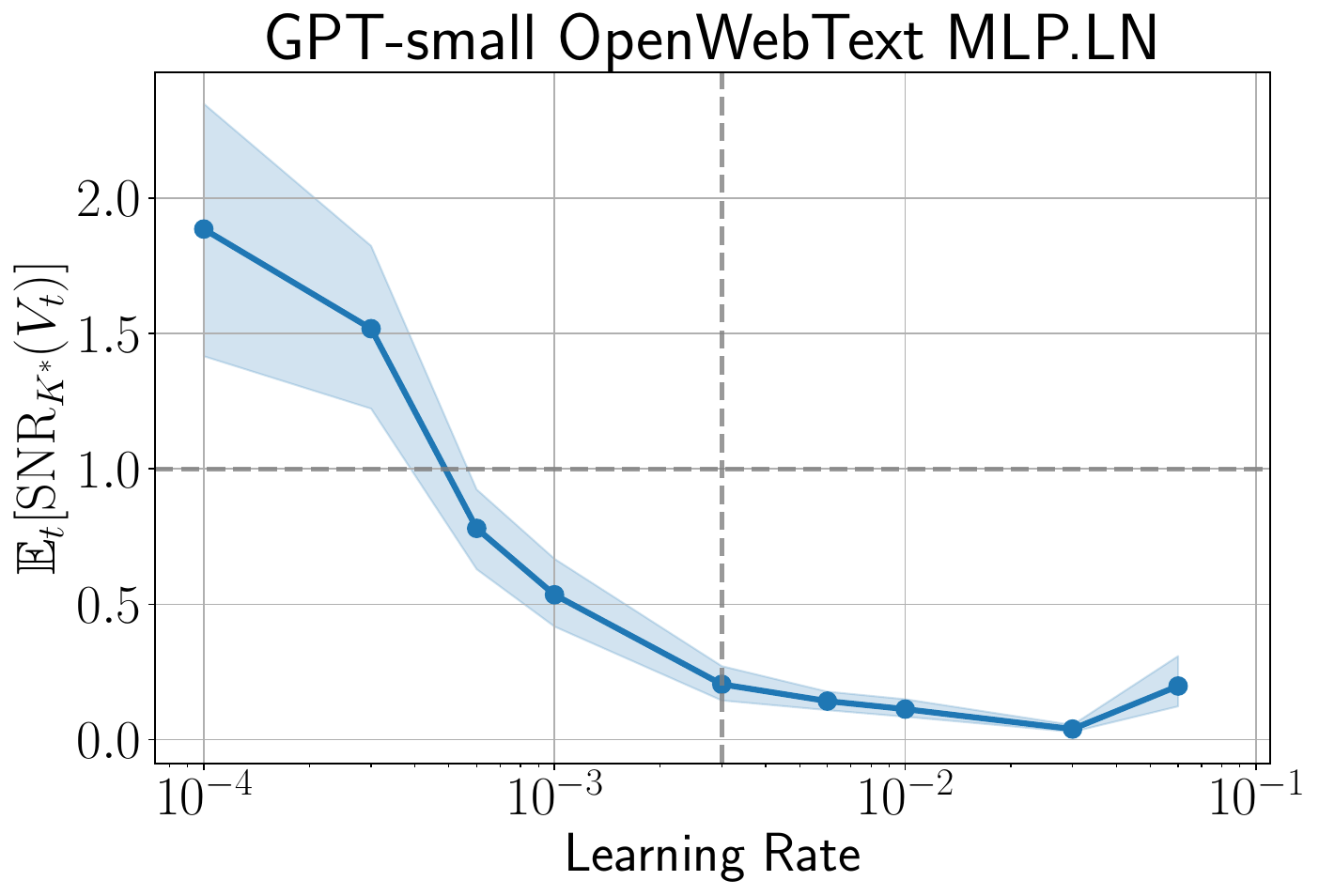}
\end{minipage}
\hfill
\begin{minipage}[b]{0.225\textwidth}
    \centering
    \includegraphics[width=\textwidth]{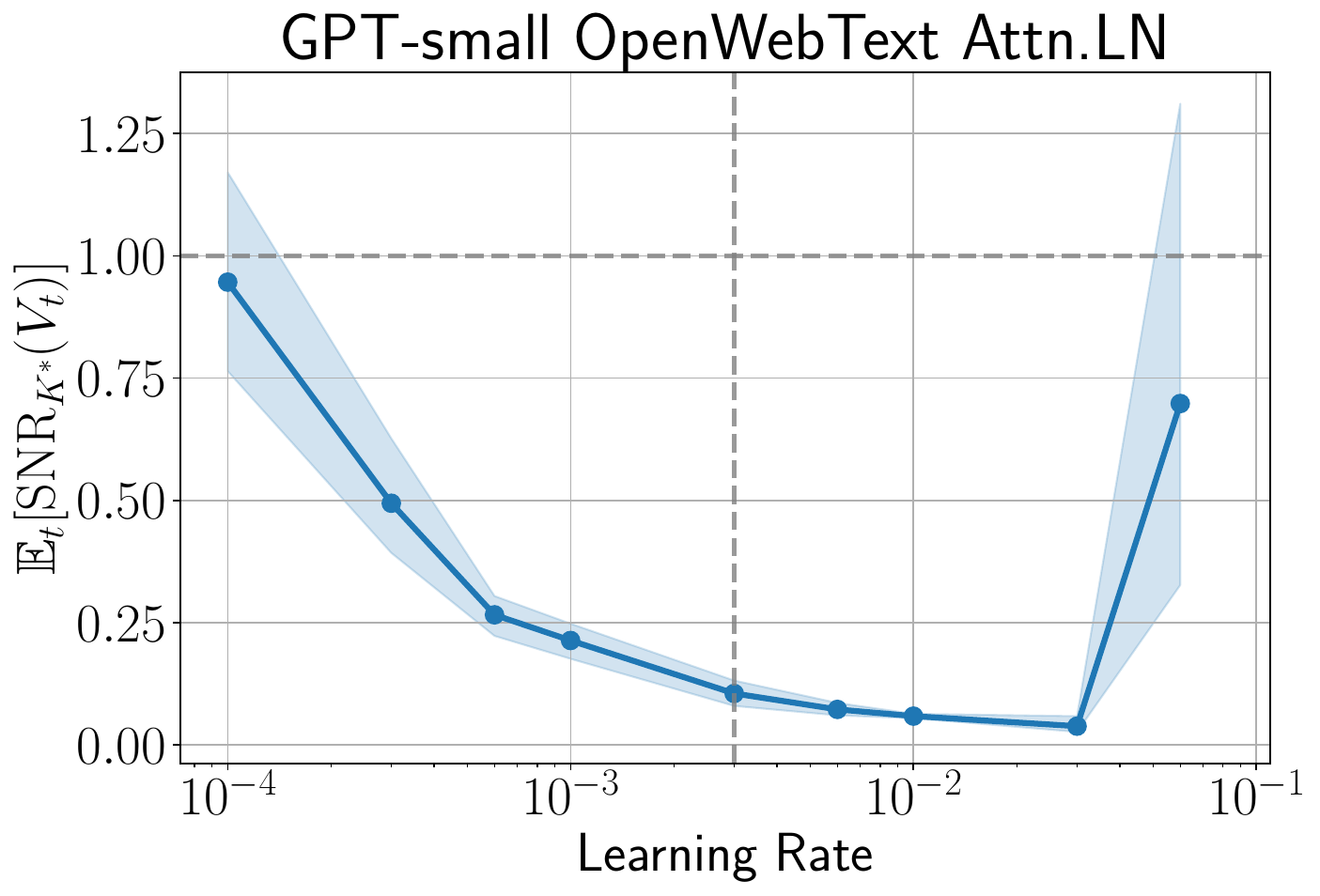}
\end{minipage}

\begin{minipage}[b]{0.225\textwidth}
    \centering
    \includegraphics[width=\textwidth]{figures/snr-lr/snr_Tok.Embd_gpt2_openwebtext_SlimAdamW_T10000_ga40_d12_h12_n768_wd0.1_bs32_b0.9_b0.95.pdf}
\end{minipage}
\begin{minipage}[b]{0.225\textwidth}
    \centering
    \includegraphics[width=\textwidth]{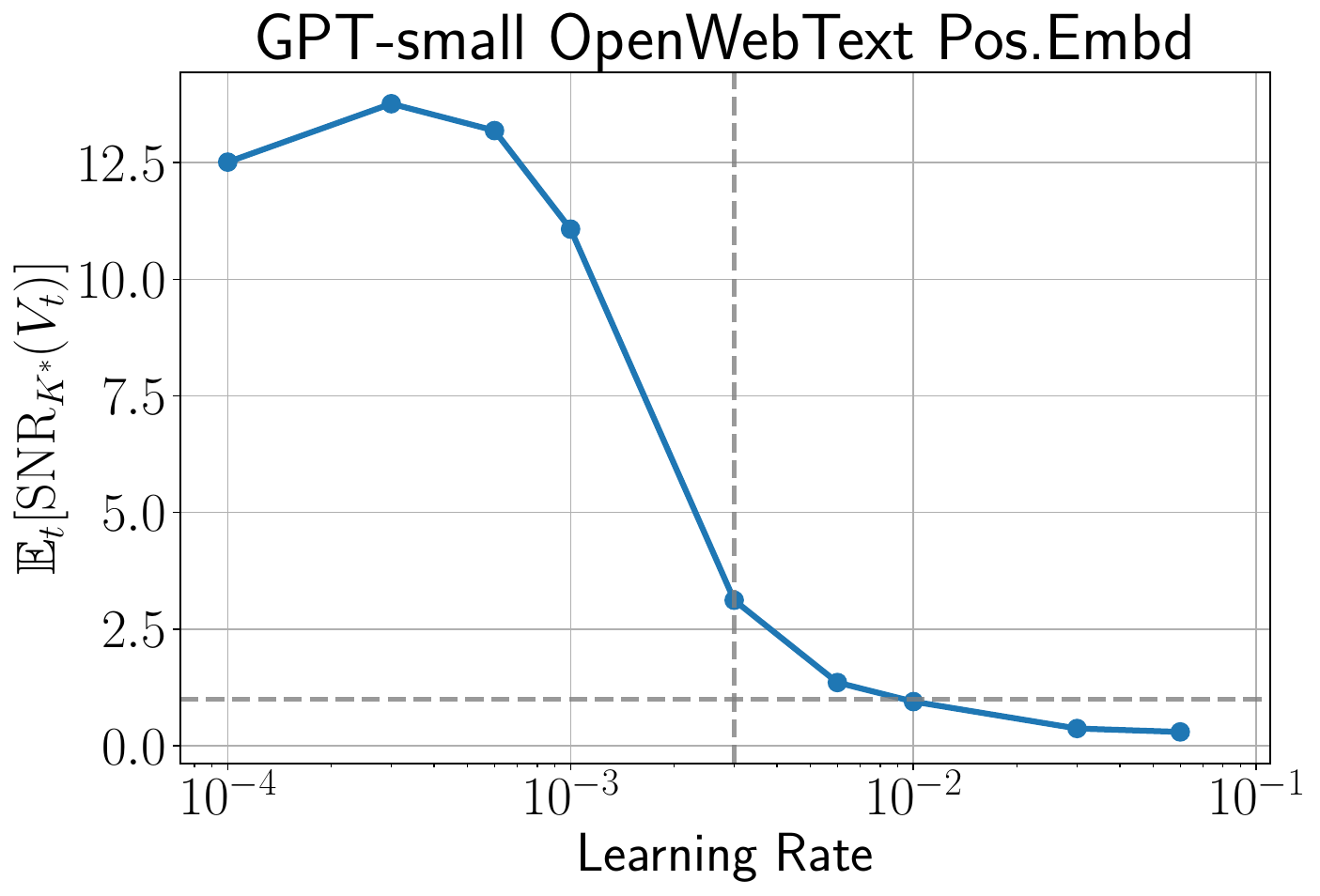}
\end{minipage}
\begin{minipage}[b]{0.225\textwidth}
    \centering
    \includegraphics[width=\textwidth]{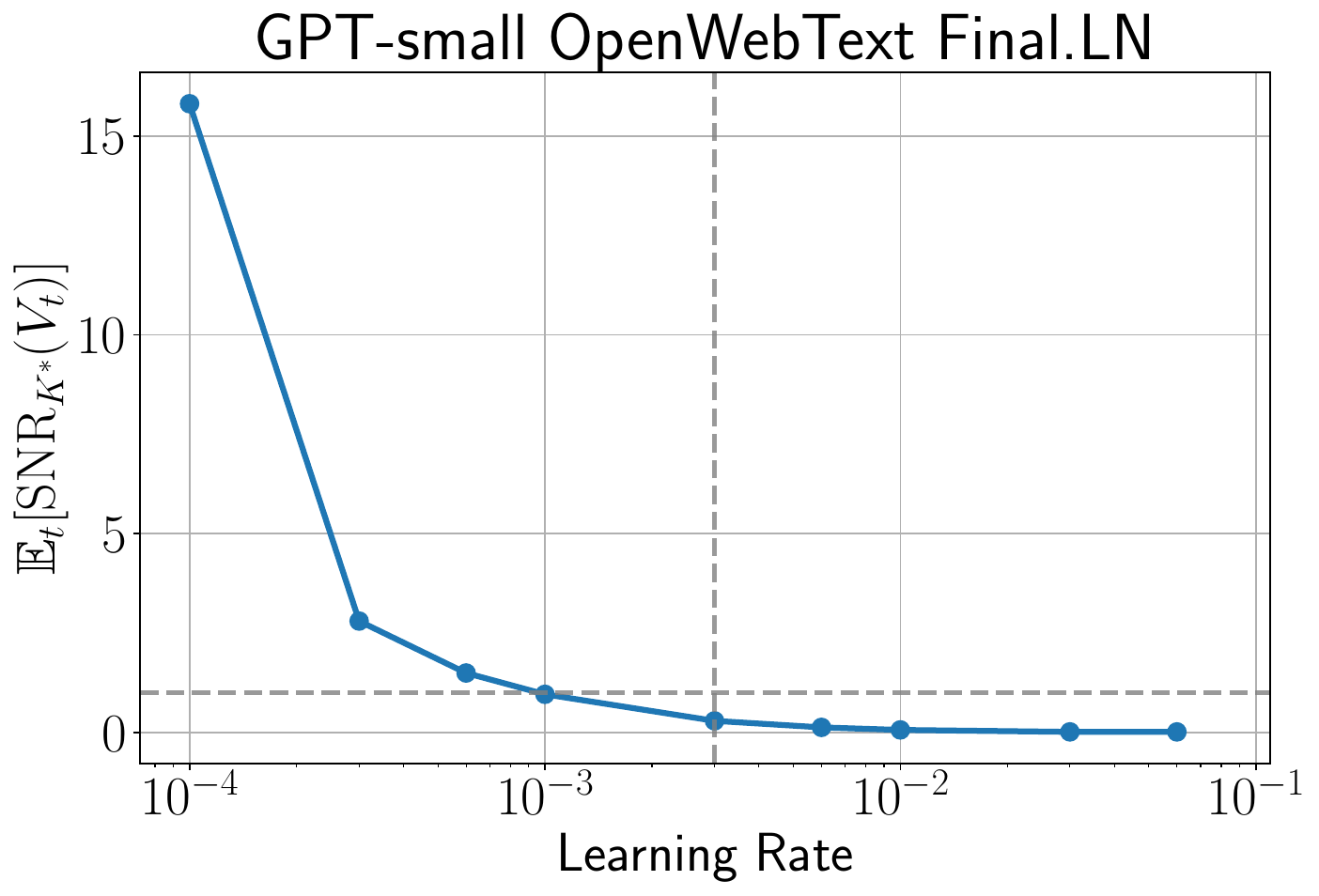}
\end{minipage}

\caption{The effect of learning rate on the averaged SNR values of different layers of a GPT-small model trained on the OpenWebText dataset. For each layer, we have selected the dimension $K^*$ with the highest SNR.
The shaded region around the mean trend shows the variation across depth. The vertical dashed line at $\text{3e-03}$ denotes the optimal learning rate.}
\label{fig:snr-lr-gpt-small-openwebtext-full}
\end{figure*}

This section provides supporting results for \Cref{section:large-learning-rates} on the effect of learning rates on averaged SNR values $\mathbb{E}_t[\mathrm{SNR}_{K}(V_t)]$. For each layer, we analyze the effect of the learning rate on the dimension $K^*$ with the highest SNR.
\Cref{fig:snr-lr-gpt-small-openwebtext-full} shows that the averaged SNR values consistently decrease with the learning rate.
This decline suggests that higher learning rates cause training to explore regions of parameter space where gradients contain more outliers, thereby reducing compression feasibility across all layers.
Based on the effect of increasing the learning rate on SNR values, we classify layer types into two categories:

\begin{enumerate}
    \item \emph{Layers that exhibit low SNR values ($\lesssim 1 $) at the optimal learning rate:} Token Embedding/LM Head, LayerNorm, attention keys, queries and MLp.Up.
    \item \emph{Layers that exhibit high SNR values ($\gtrsim 1$) even at the optimal learning rate:} Attention values, projections and MLP.Down.

\end{enumerate}

\clearpage \newpage

\section{Effect of Initialization on Compressibility}
\label{appendix:initialization}

\begin{figure*}[!htb]
\centering
\begin{minipage}[b]{0.225\textwidth}
    \centering
    \includegraphics[width=\textwidth]{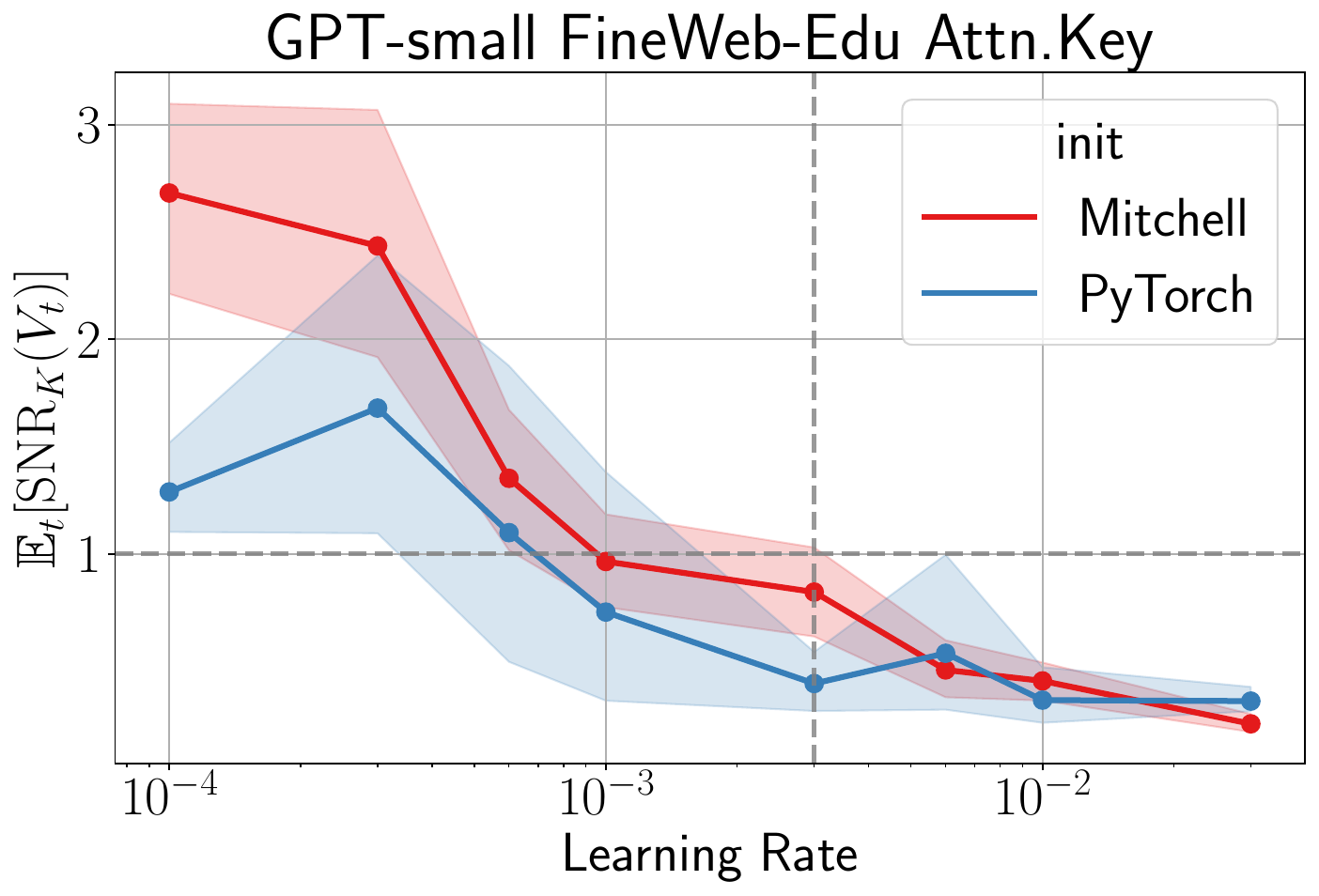}
\end{minipage}
\hfill
\begin{minipage}[b]{0.225\textwidth}
    \centering
    \includegraphics[width=\textwidth]{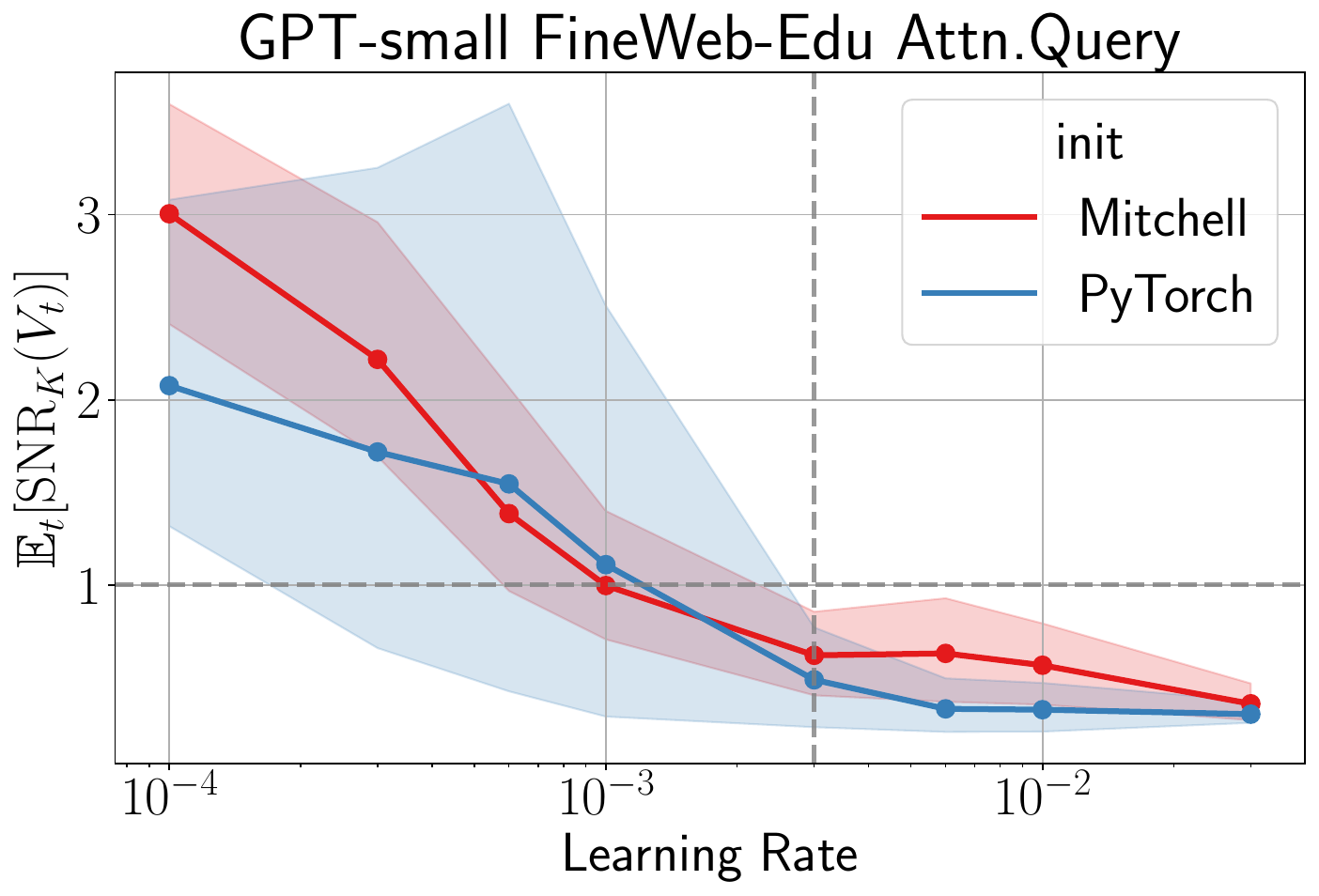}
\end{minipage}
\hfill
\begin{minipage}[b]{0.225\textwidth}
    \centering
    \includegraphics[width=\textwidth]{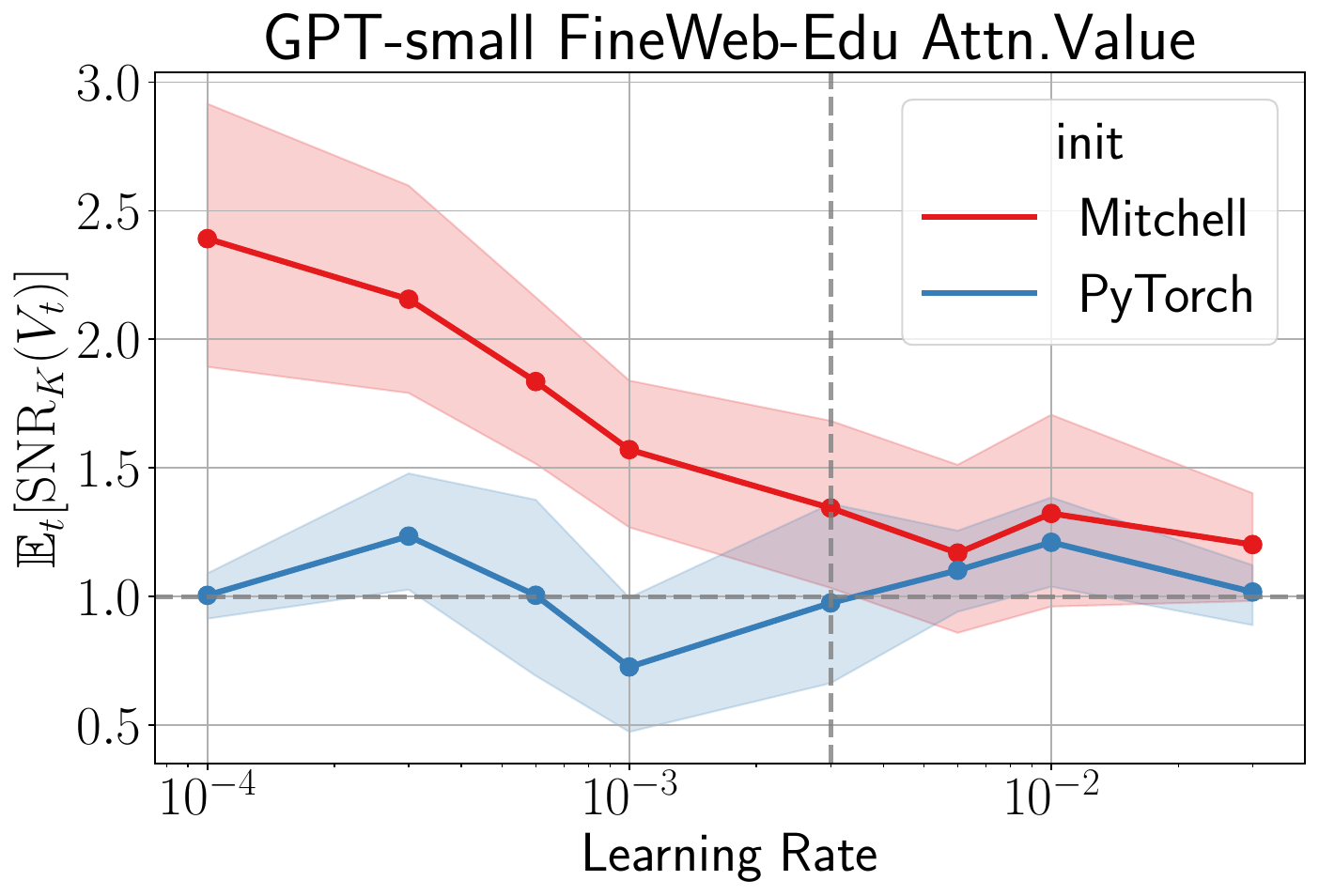}
\end{minipage}
\hfill
\begin{minipage}[b]{0.225\textwidth}
    \centering
    \includegraphics[width=\textwidth]{figures/snr-init/snr_Attn.Proj_gpt2_dinit_fineweb_SlimAdamW_T10000_ga40_d12_h12_n768_wd0.1_bs32_b0.9_b0.95.pdf}
\end{minipage}

\begin{minipage}[b]{0.225\textwidth}
    \centering
    \includegraphics[width=\textwidth]{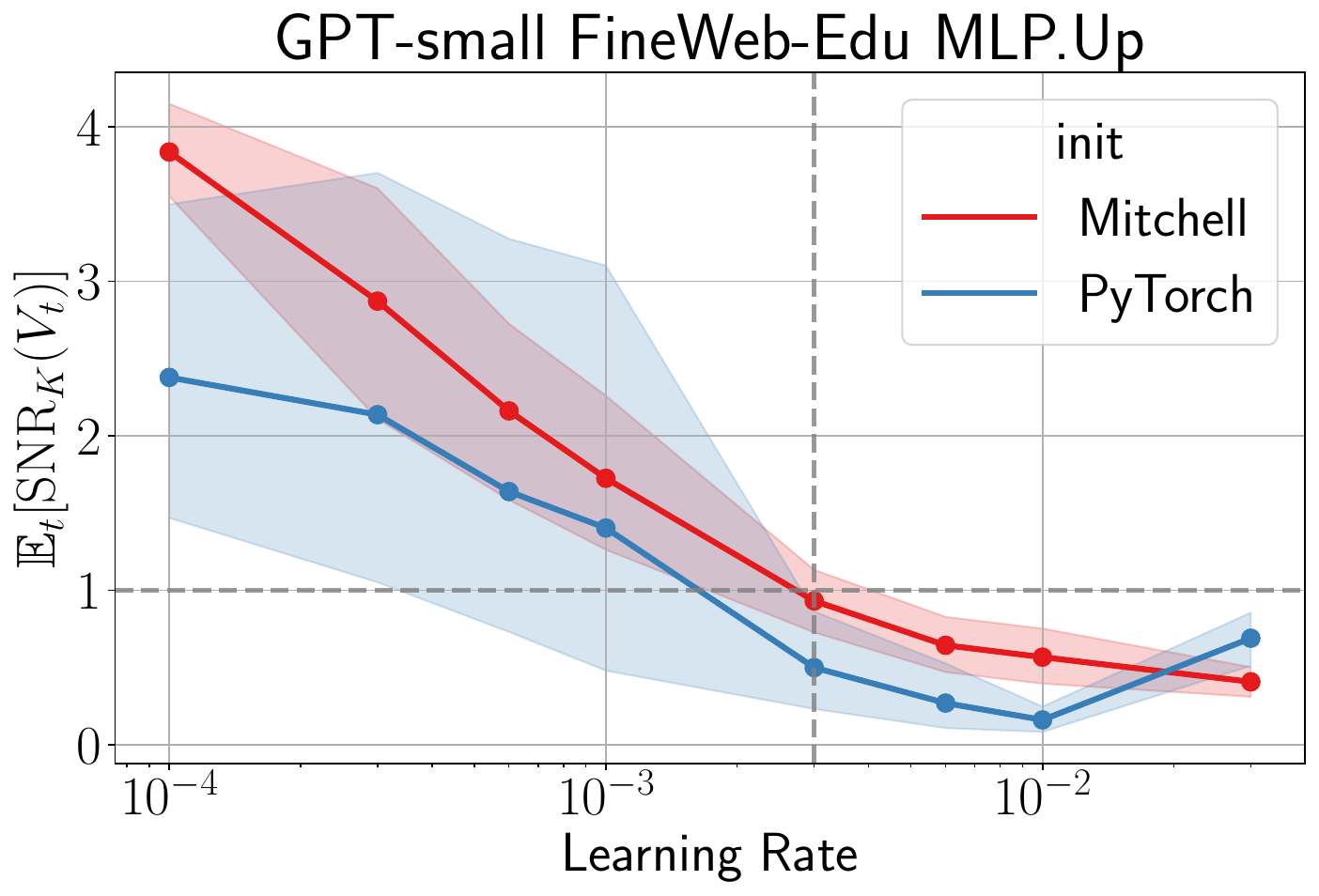}
\end{minipage}
\hfill
\begin{minipage}[b]{0.225\textwidth}
    \centering
    \includegraphics[width=\textwidth]{figures/snr-init/snr_MLP.Down_gpt2_dinit_fineweb_SlimAdamW_T10000_ga40_d12_h12_n768_wd0.1_bs32_b0.9_b0.95.pdf}
\end{minipage}
\hfill
\begin{minipage}[b]{0.225\textwidth}
    \centering
    \includegraphics[width=\textwidth]{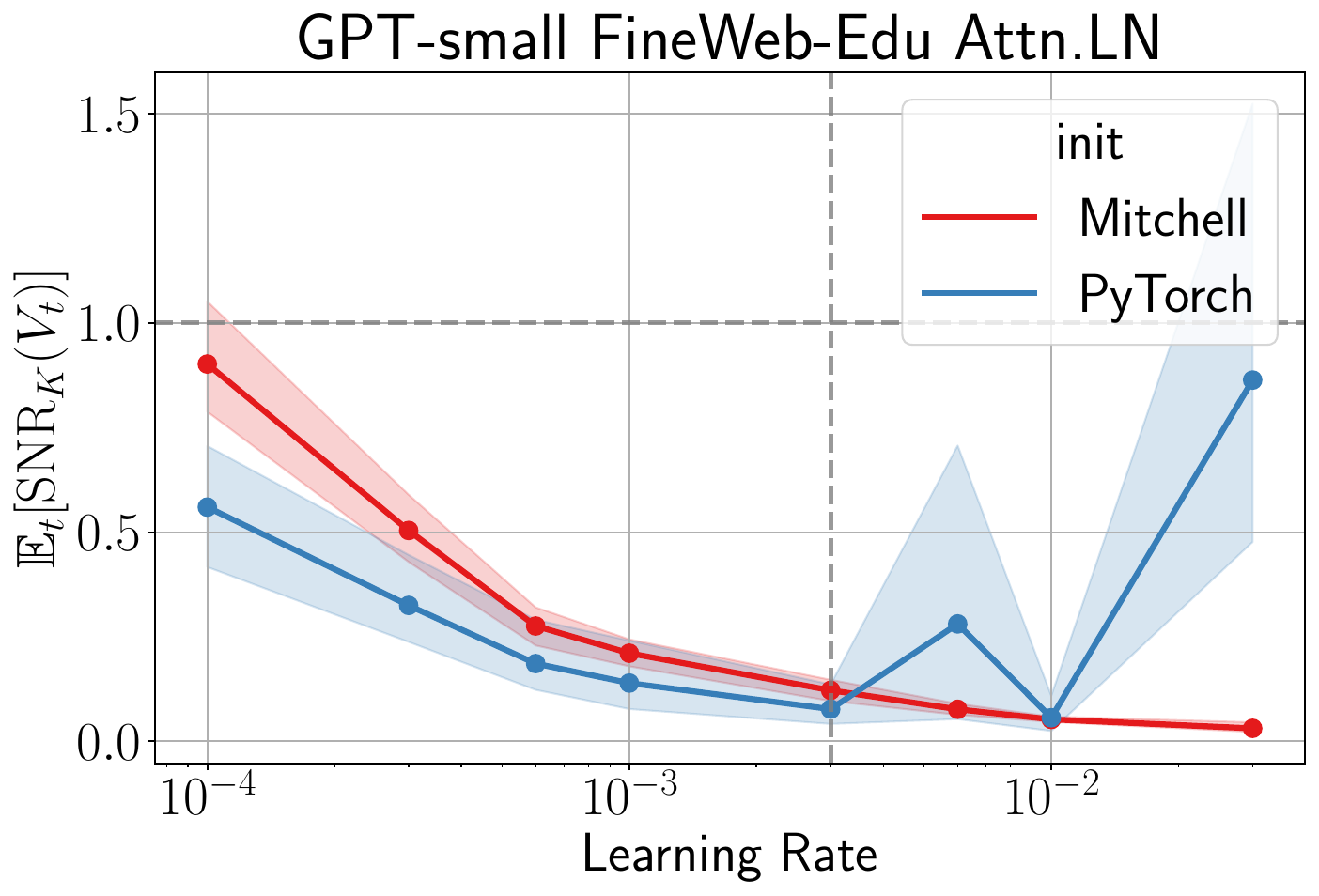}
\end{minipage}
\hfill
\begin{minipage}[b]{0.225\textwidth}
    \centering
    \includegraphics[width=\textwidth]{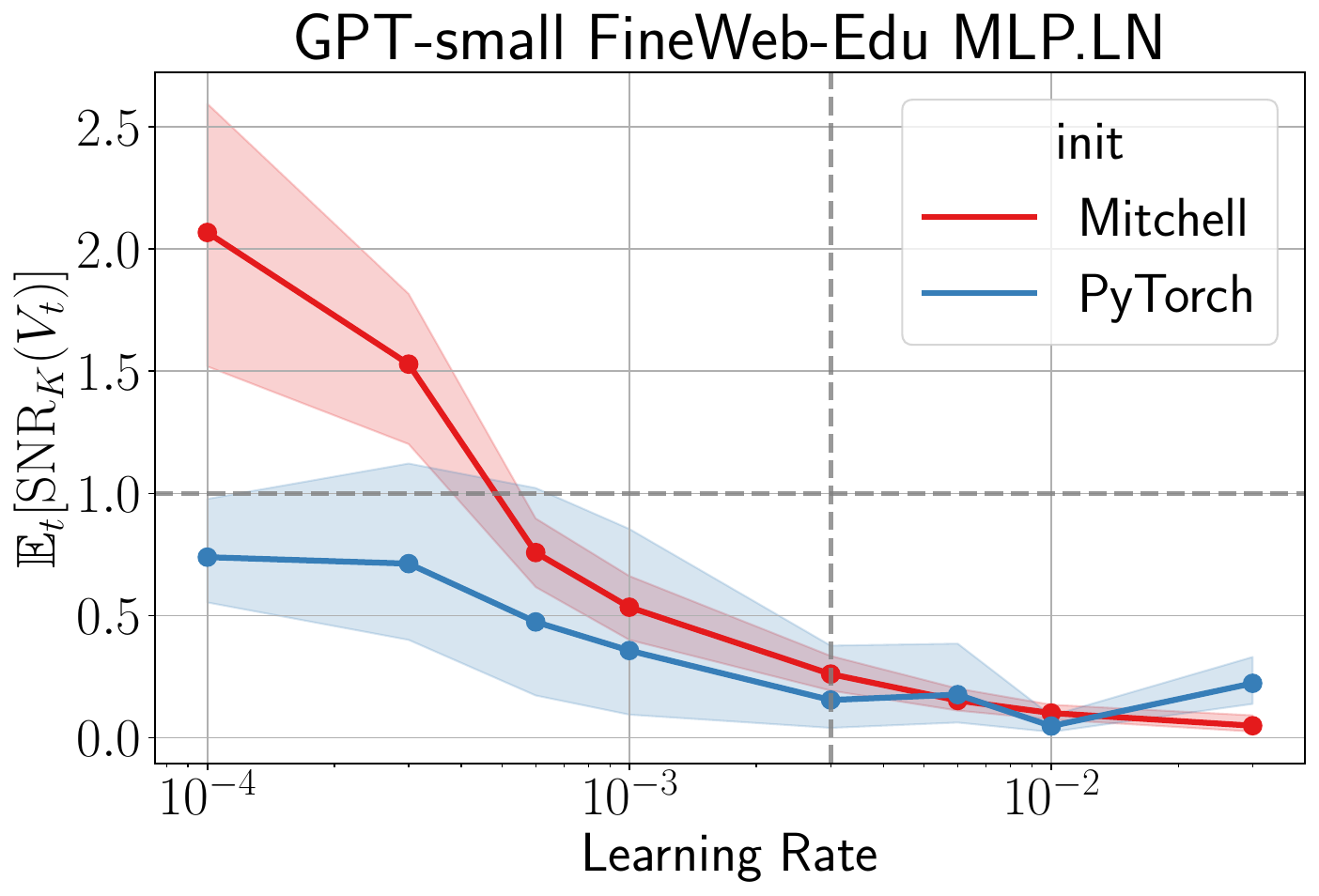}
\end{minipage}

\begin{minipage}[b]{0.225\textwidth}
    \centering
    \includegraphics[width=\textwidth]{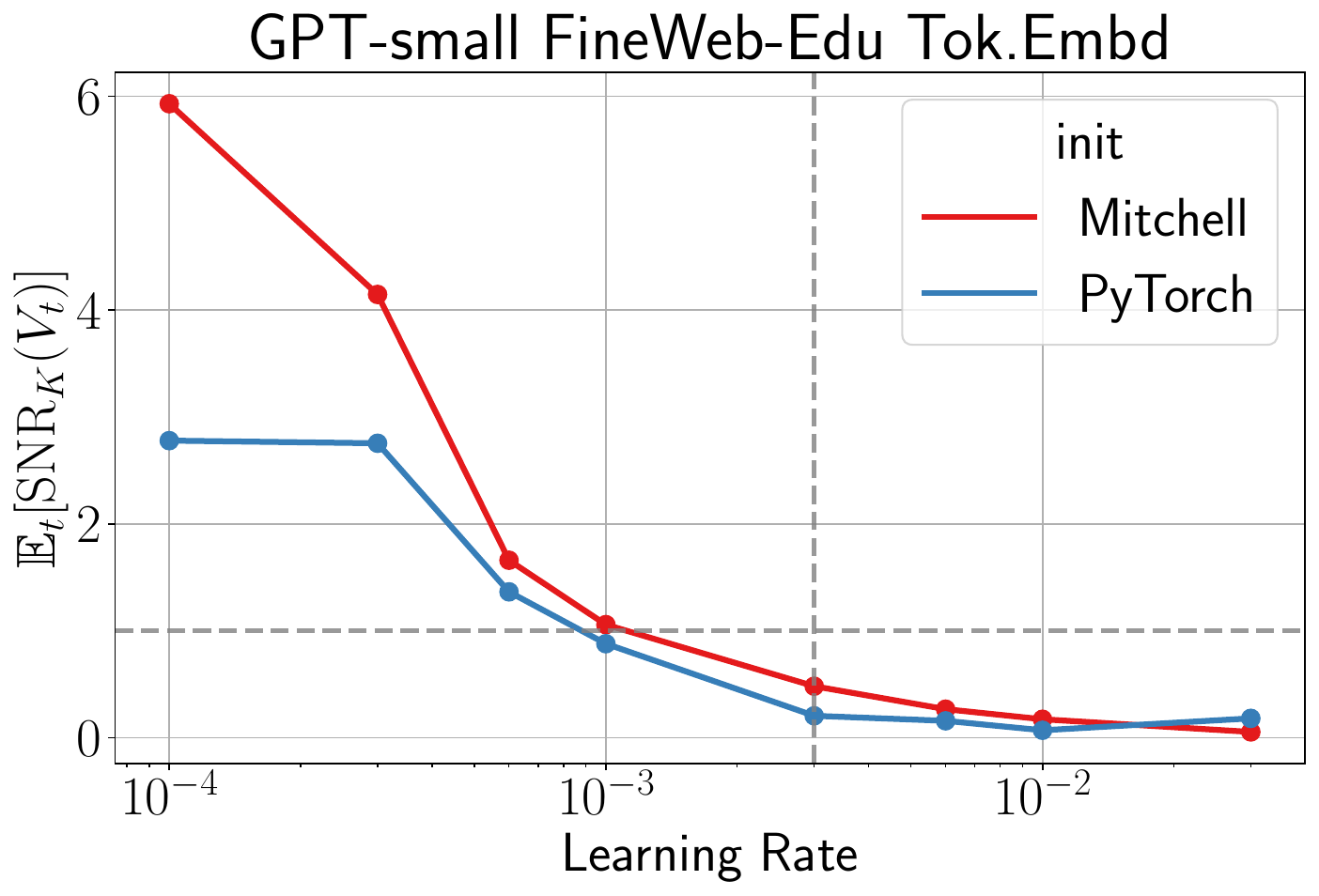}
\end{minipage}
\begin{minipage}[b]{0.225\textwidth}
    \centering
    \includegraphics[width=\textwidth]{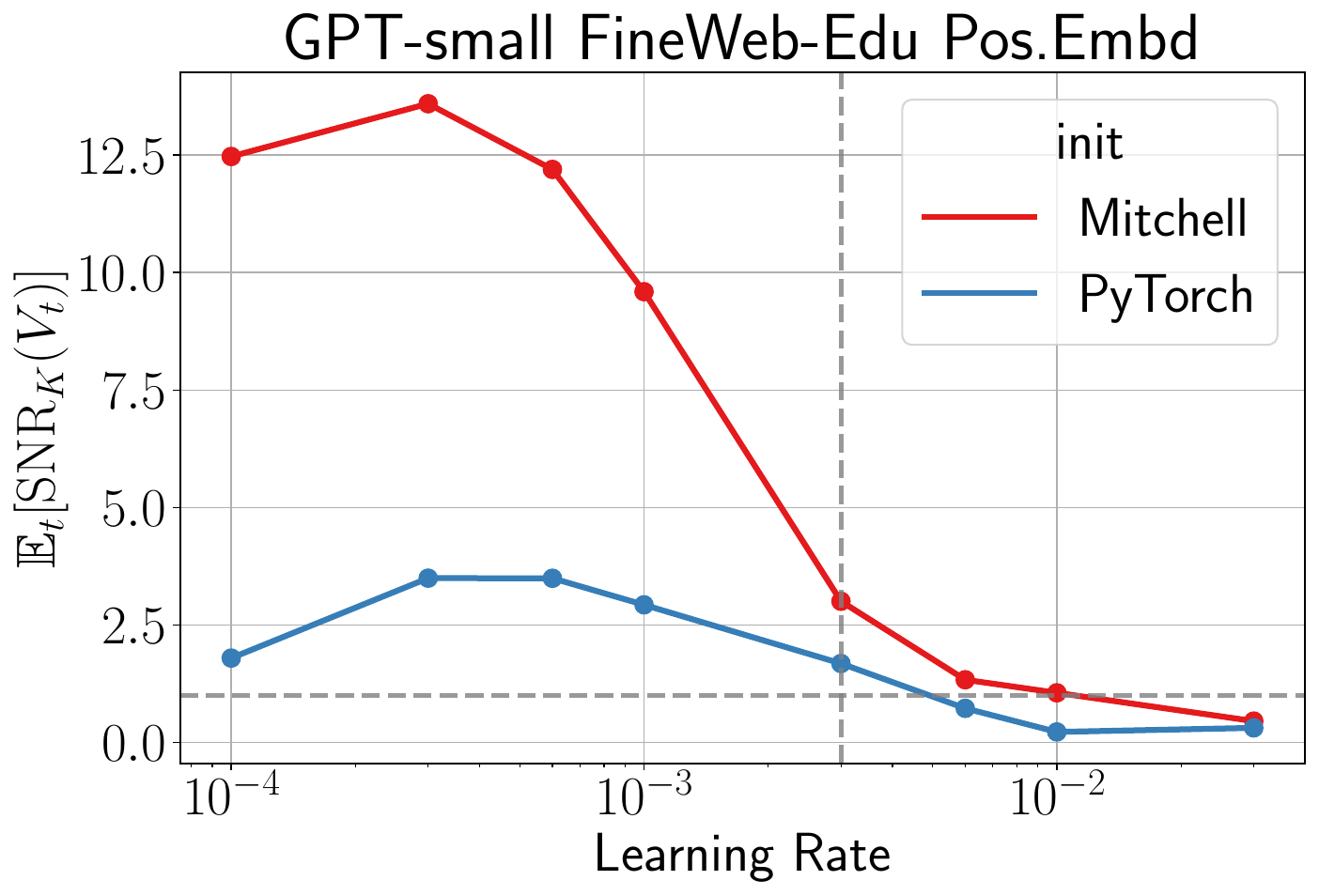}
\end{minipage}
\begin{minipage}[b]{0.225\textwidth}
    \centering
    \includegraphics[width=\textwidth]{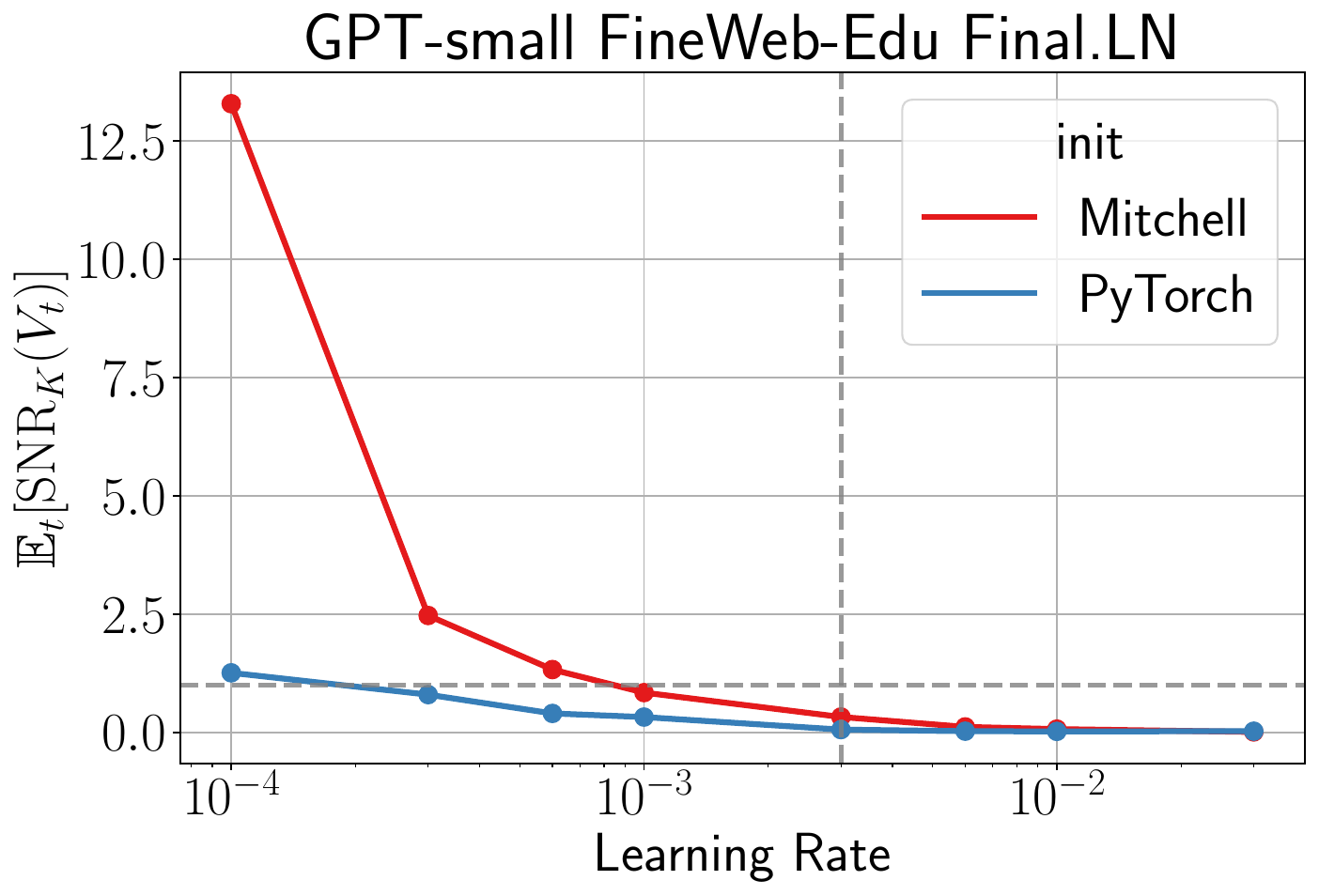}
\end{minipage}
\caption{The effect of initialization on the averaged SNR values of different layers of a GPT-small model trained on the OpenWebText dataset. For each layer, we have selected the dimension $K^*$ with the highest SNR.
The shaded region around the mean trend shows the variation across depth. The vertical dashed line at $\text{3e-03}$ denotes the optimal learning rate for Mitchel initialization.}
\label{fig:snr-layer-gpt-dinit-small-finweb-full}
\end{figure*}

This section provides supporting results for \Cref{section:initialization} on the effect of initialization on averaged SNR values $\mathbb{E}_t[\mathrm{SNR}_{K}(V_t)]$. We analyze how different initialization schemes affect SNR trends by comparing PyTorch's default initialization with the commonly used Mitchell initialization used in GPT models (recall that Mitchell initialization scales down the variance by $1/$depth in layers that add to the residual stream, such as Attn.Proj and MLP.Down).
For simplicity, we select the dimension $K^*$ with the highest SNR for each layer.

\Cref{fig:snr-layer-gpt-dinit-small-finweb-full} shows that PyTorch's default initialization exhibits substantially lower SNR values across layers, especially the layers that add to the residual stream (Attn.Proj and MLP.Down) exhibit substantially lower SNR values.
These results suggest that the compression feasibility depends on initialization choices and architectural details, suggesting that a single compression strategy is unlikely to work universally.

\clearpage \newpage

\section{Additional Results for \emph{SlimAdam}}
\label{appendix:slimadam}

\begin{figure*}[!htb]
\centering
\begin{minipage}[b]{0.225\textwidth}
    \centering
    \includegraphics[width=\textwidth]{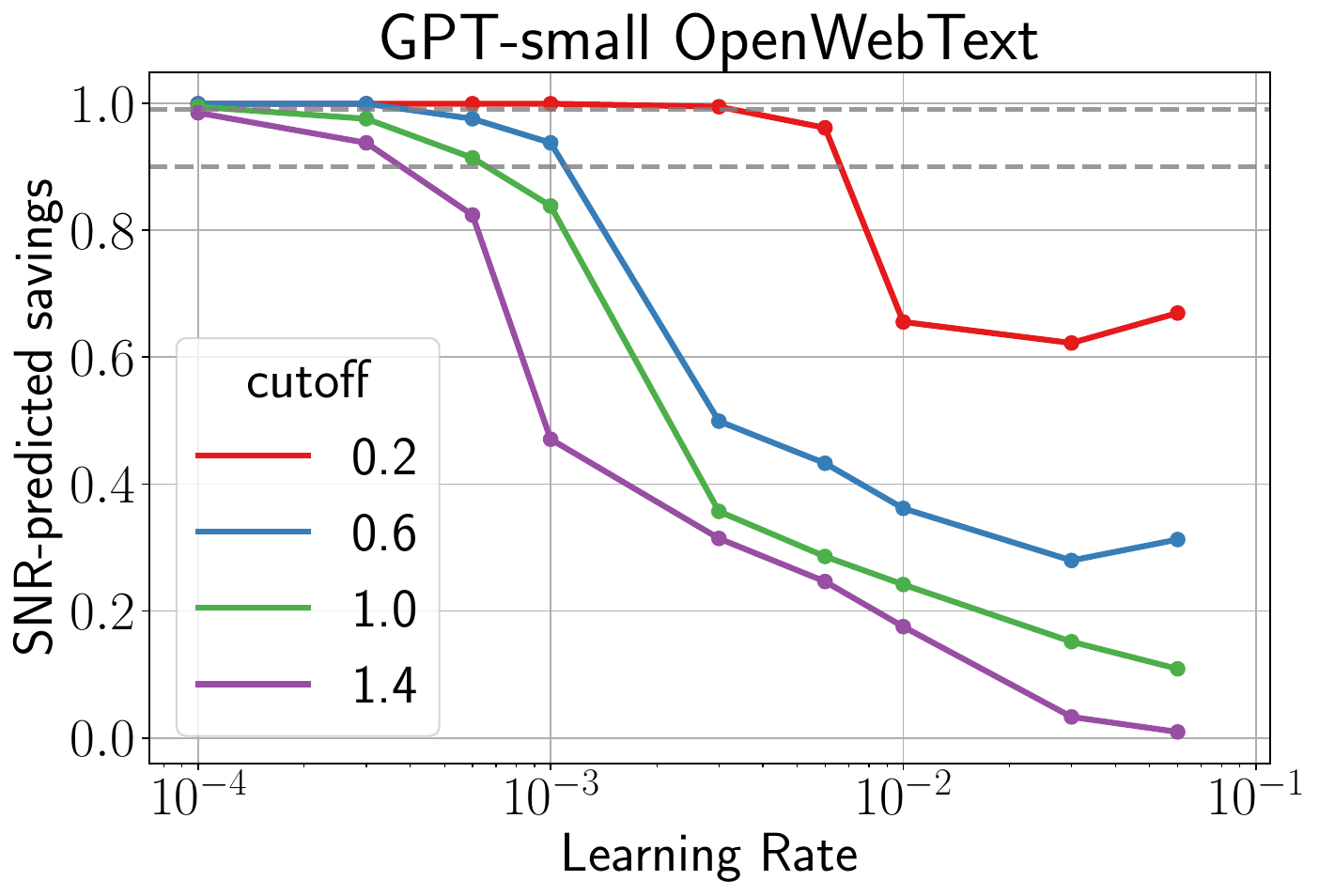}
\end{minipage}
\hfill
\begin{minipage}[b]{0.225\textwidth}
    \centering
    \includegraphics[width=\textwidth]{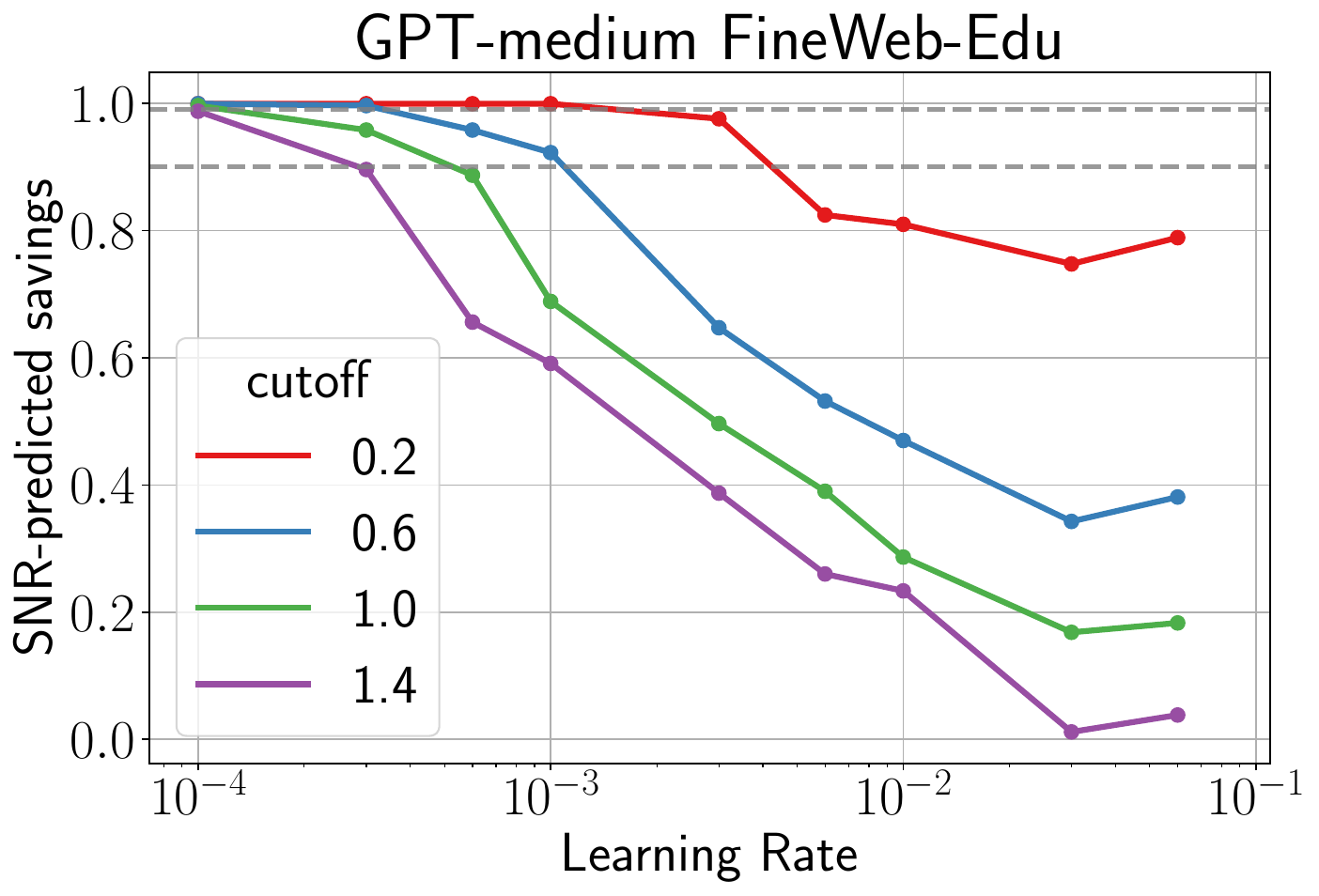}
\end{minipage}
\hfill
\begin{minipage}[b]{0.225\textwidth}
    \centering
    \includegraphics[width=\textwidth]{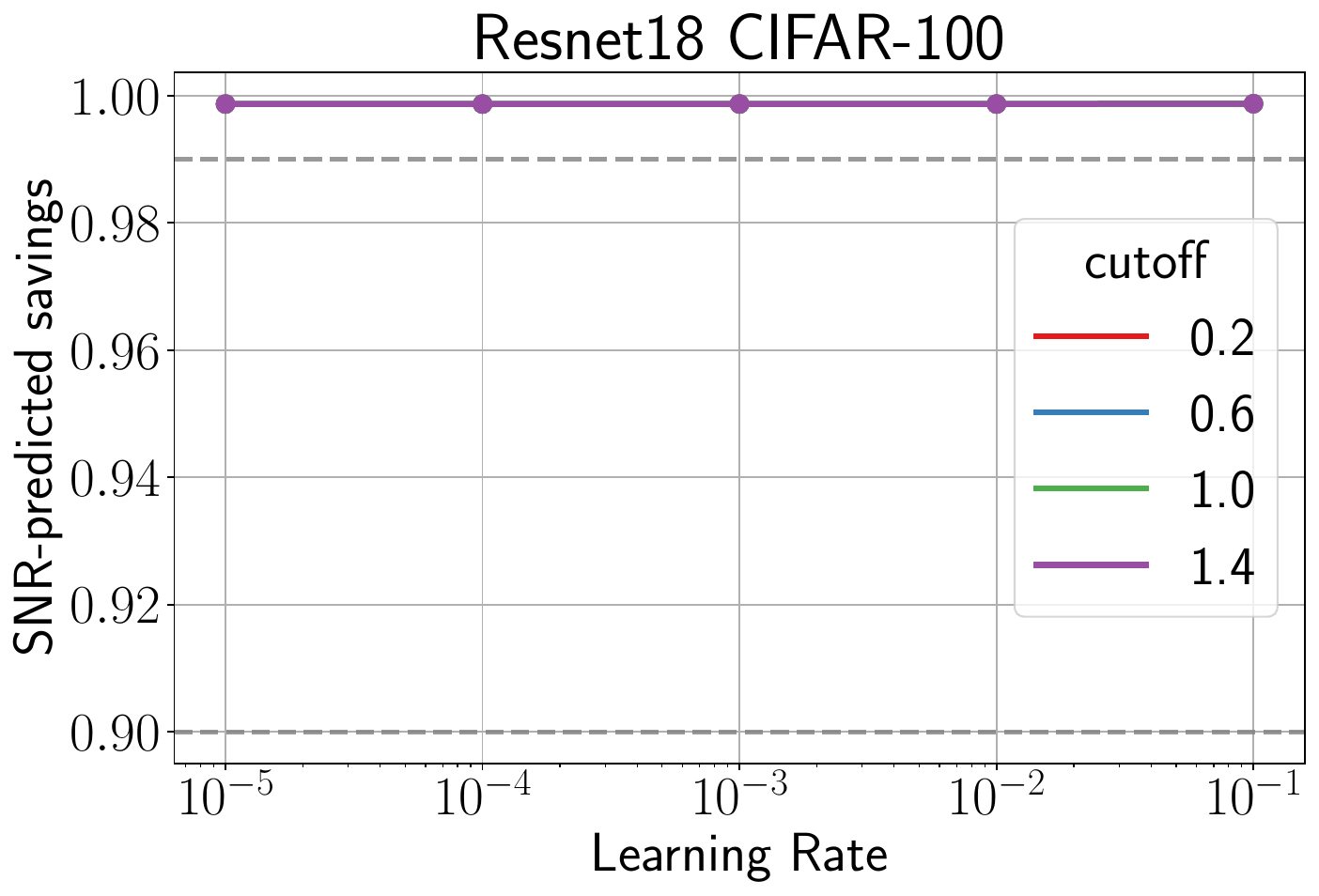}
\end{minipage}
\hfill
\begin{minipage}[b]{0.225\textwidth}
    \centering
    \includegraphics[width=\textwidth]{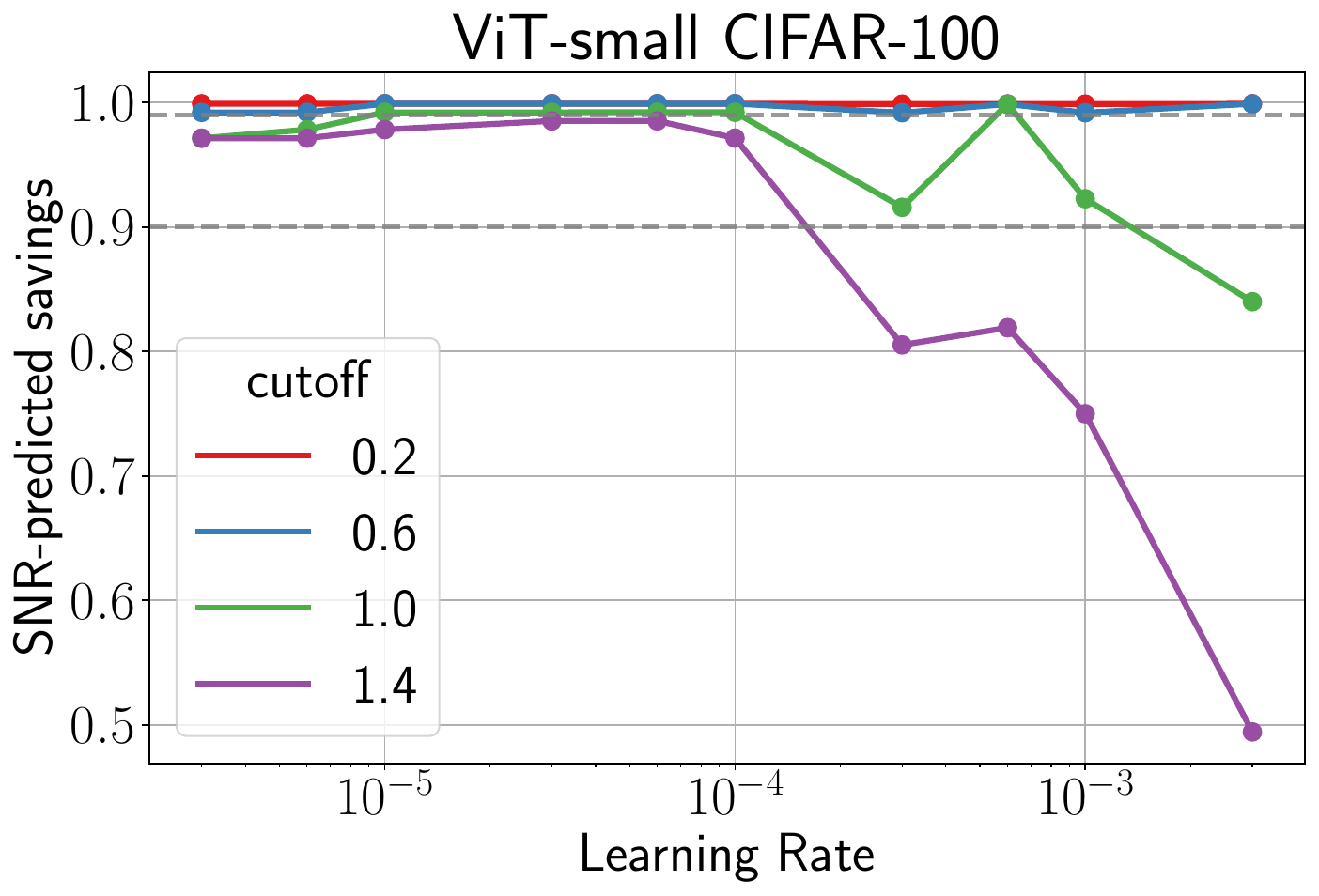}
\end{minipage}

\begin{minipage}[b]{0.225\textwidth}
    \centering
    \includegraphics[width=\textwidth]{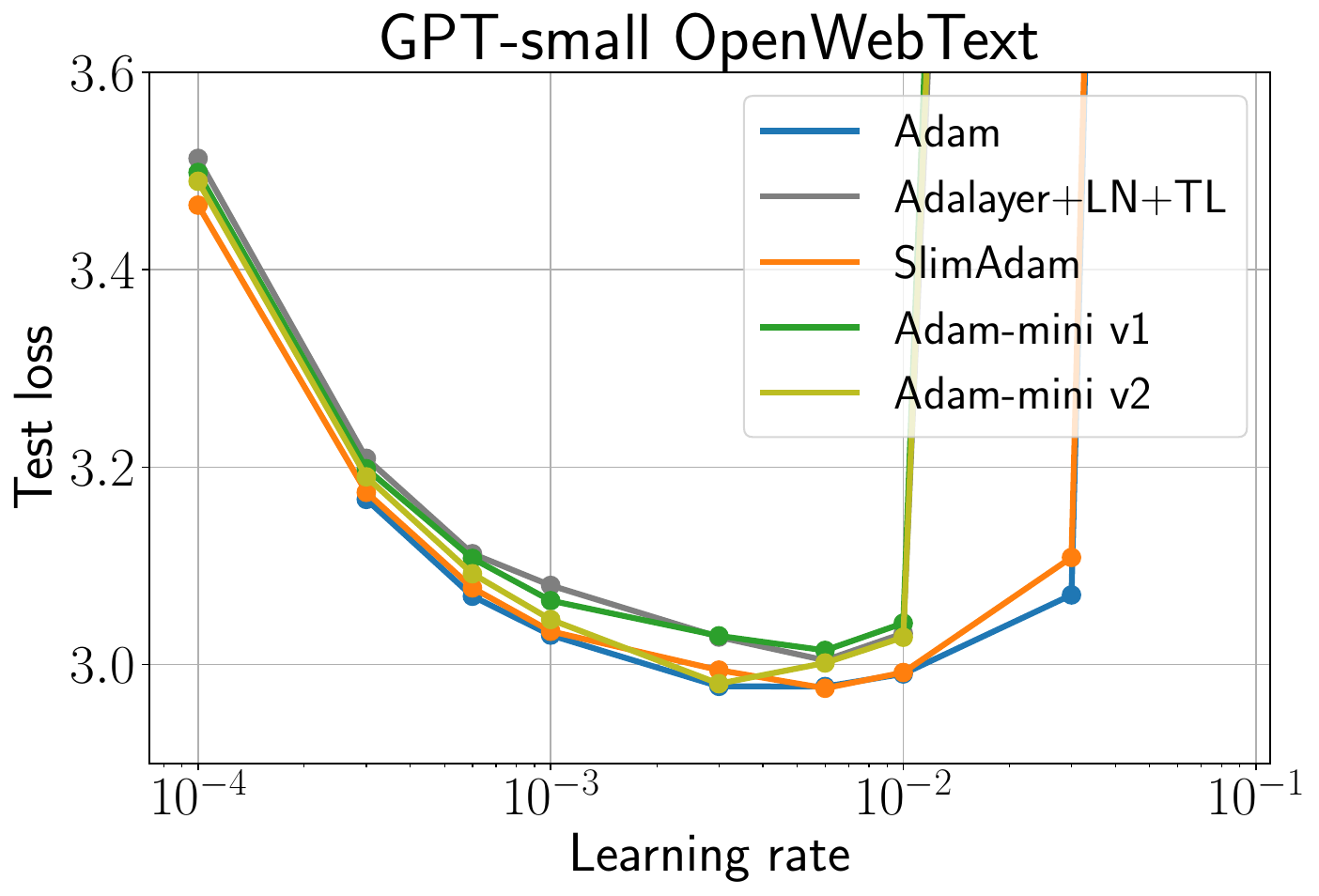}
\end{minipage}
\hfill
\begin{minipage}[b]{0.225\textwidth}
    \centering
    \includegraphics[width=\textwidth]{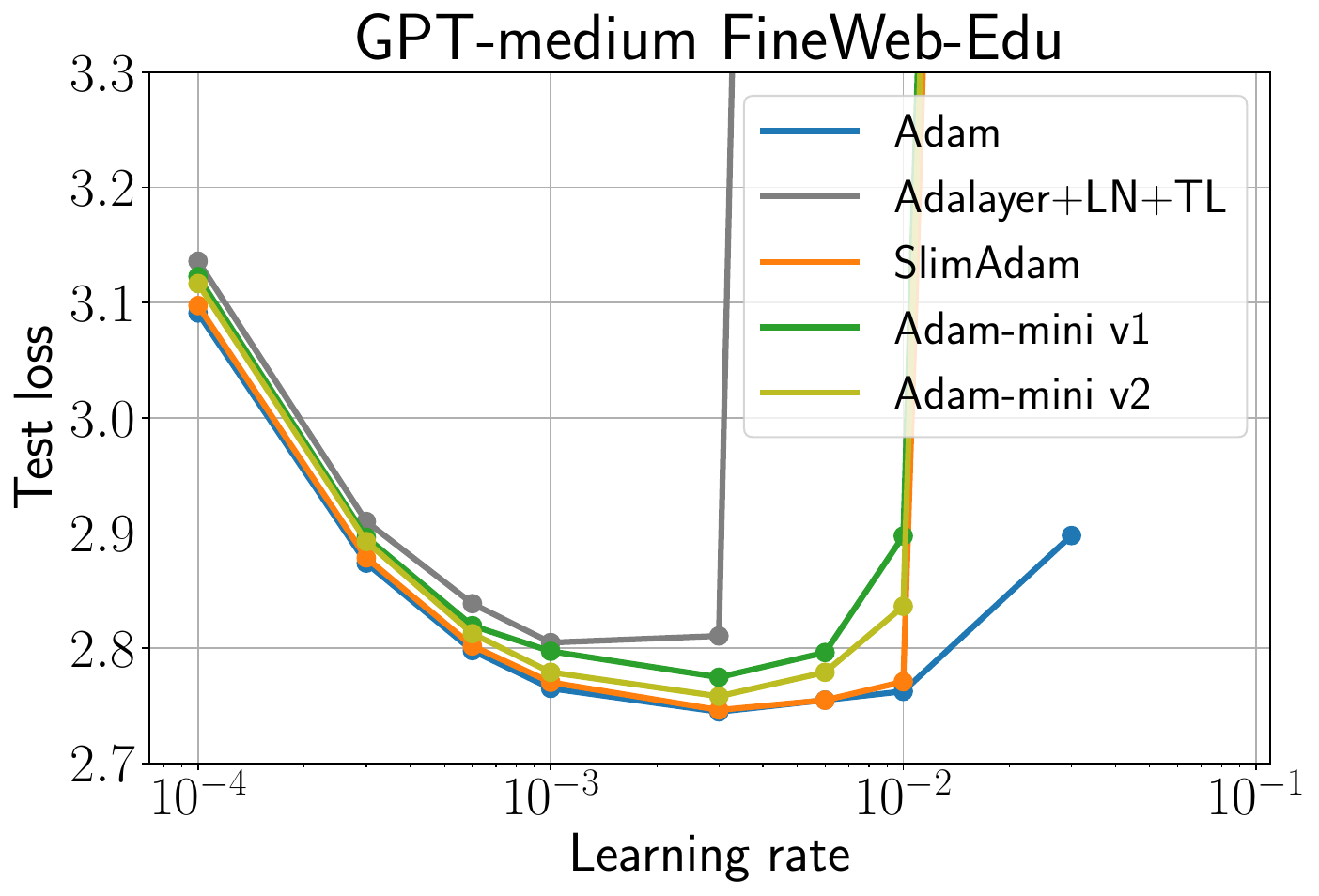}
\end{minipage}
\hfill
\begin{minipage}[b]{0.225\textwidth}
    \centering
    \includegraphics[width=\textwidth]{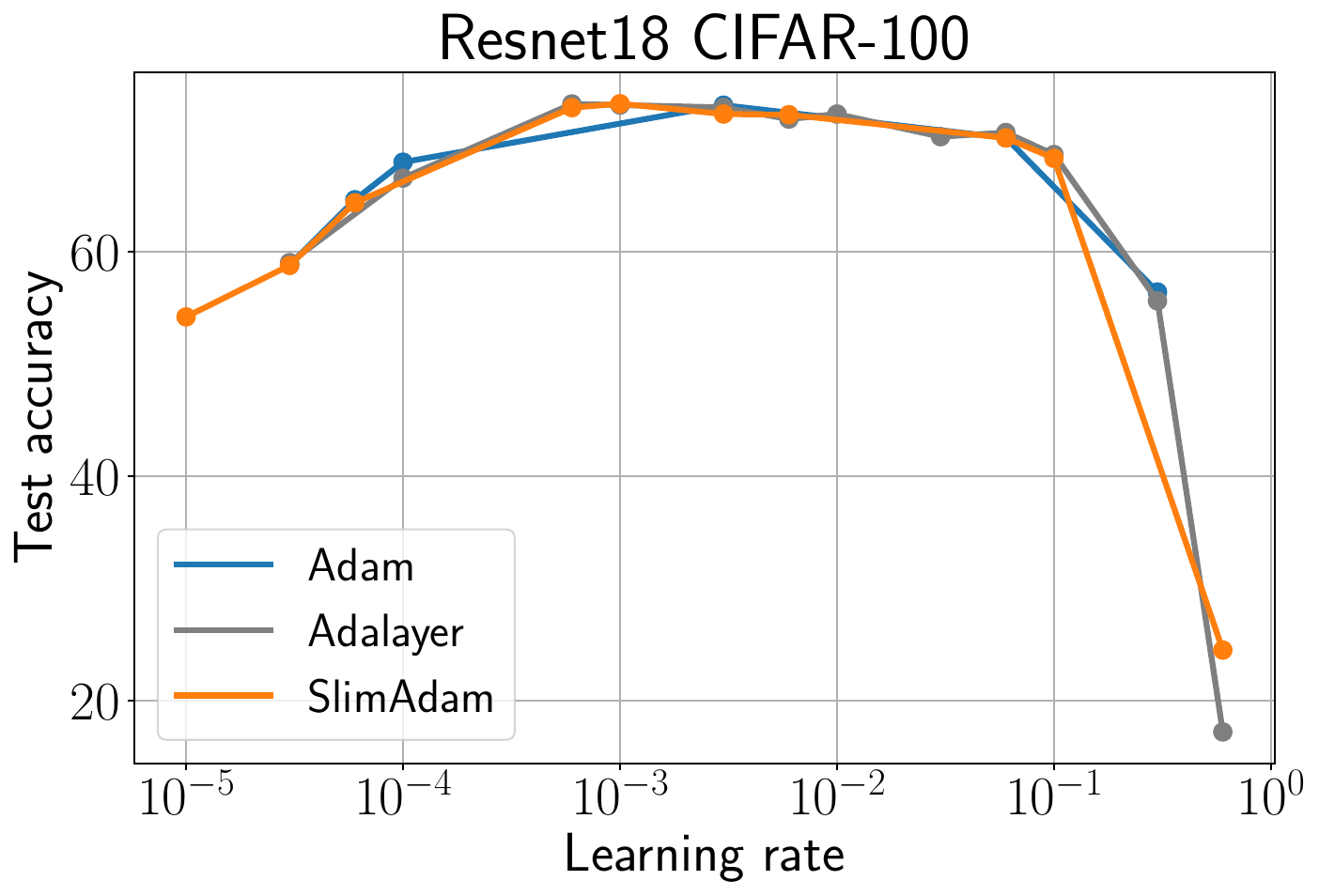}
\end{minipage}
\hfill
\begin{minipage}[b]{0.225\textwidth}
    \centering
    \includegraphics[width=\textwidth]{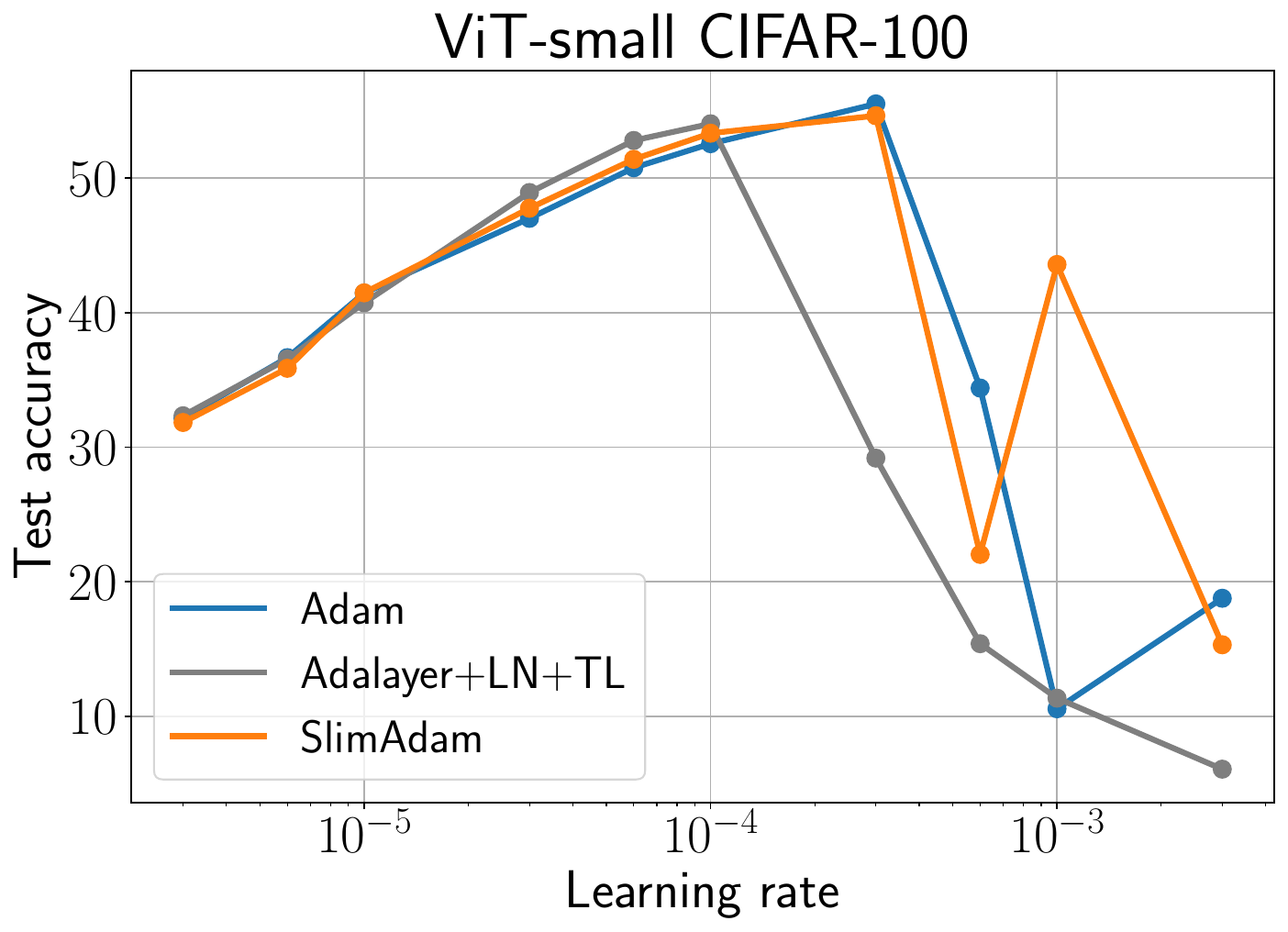}
\end{minipage}

\caption{(Top) Fraction of second moments saved (relative to Adam) as a function of learning rate and SNR cutoff across training configuration, as suggested by the SNR analysis.
(Bottom) Performance comparison across learning rates between SlimAdam and baselines.
}
\label{fig:slim-memory-performance-appendix}
\end{figure*}

\begin{figure*}[!htb]
\centering
\begin{minipage}[b]{0.225\textwidth}
    \centering
    \includegraphics[width=\textwidth]{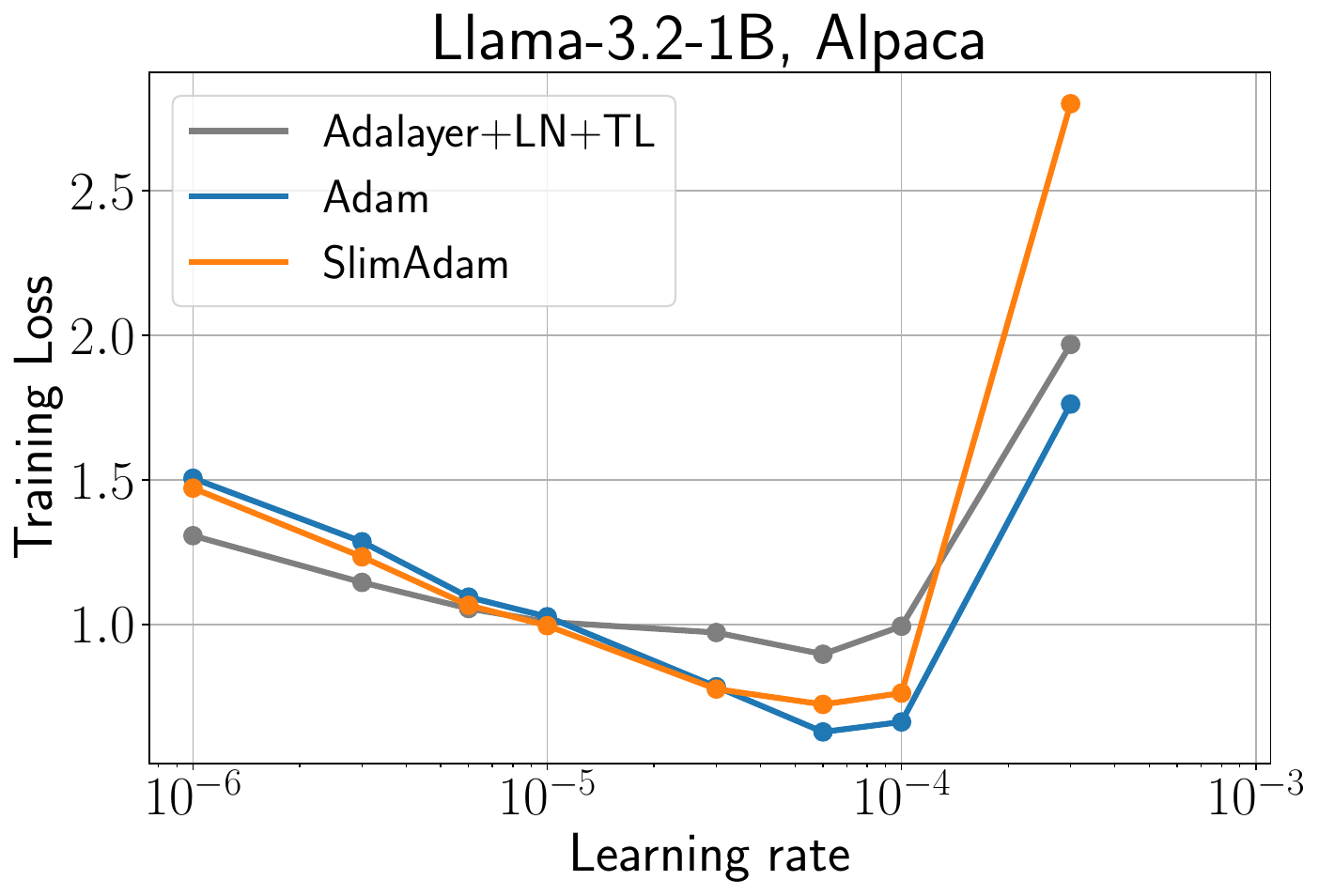}
\end{minipage}
\begin{minipage}[b]{0.225\textwidth}
    \centering
    \includegraphics[width=\textwidth]{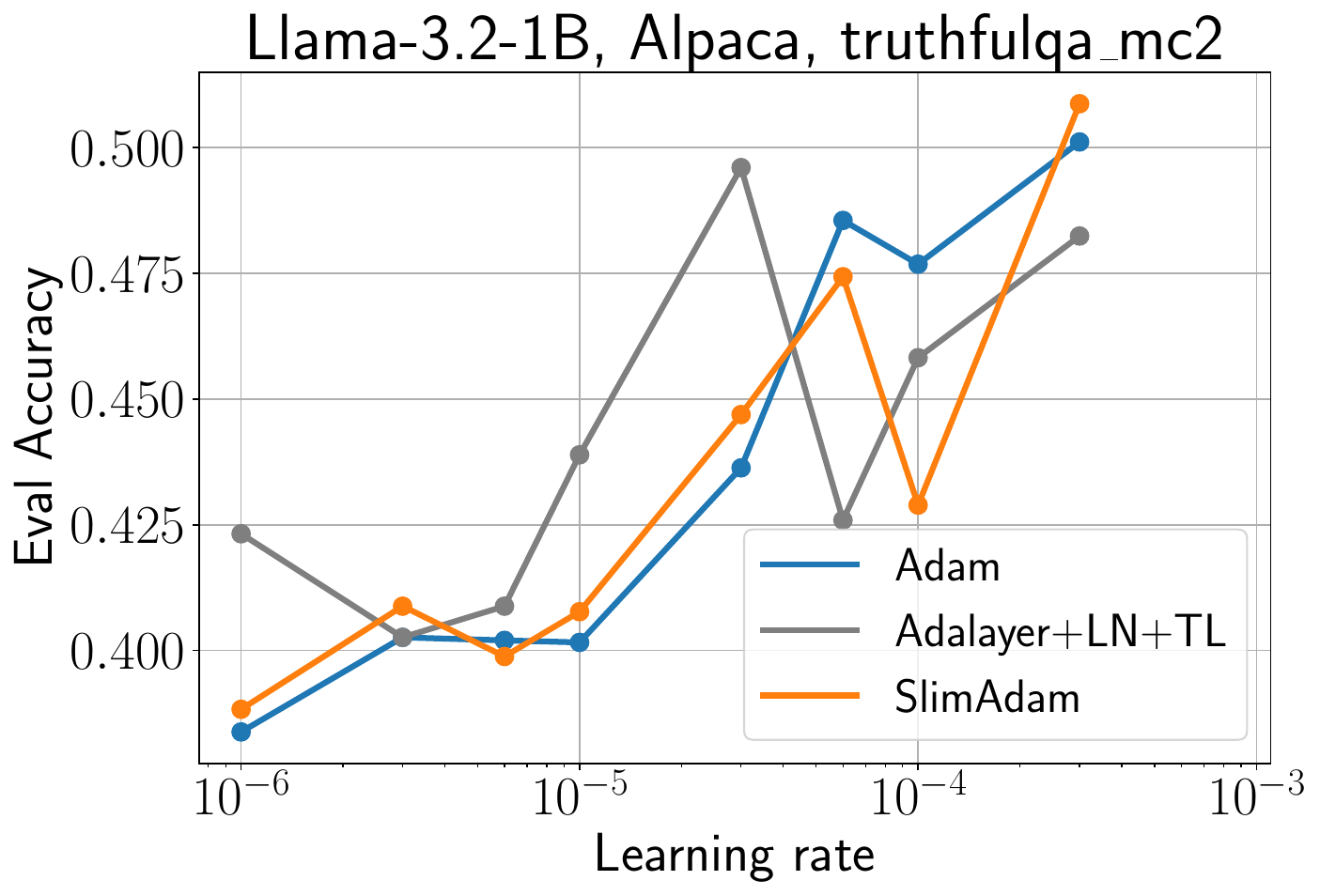}
\end{minipage}
\begin{minipage}[b]{0.225\textwidth}
    \centering
    \includegraphics[width=\textwidth]{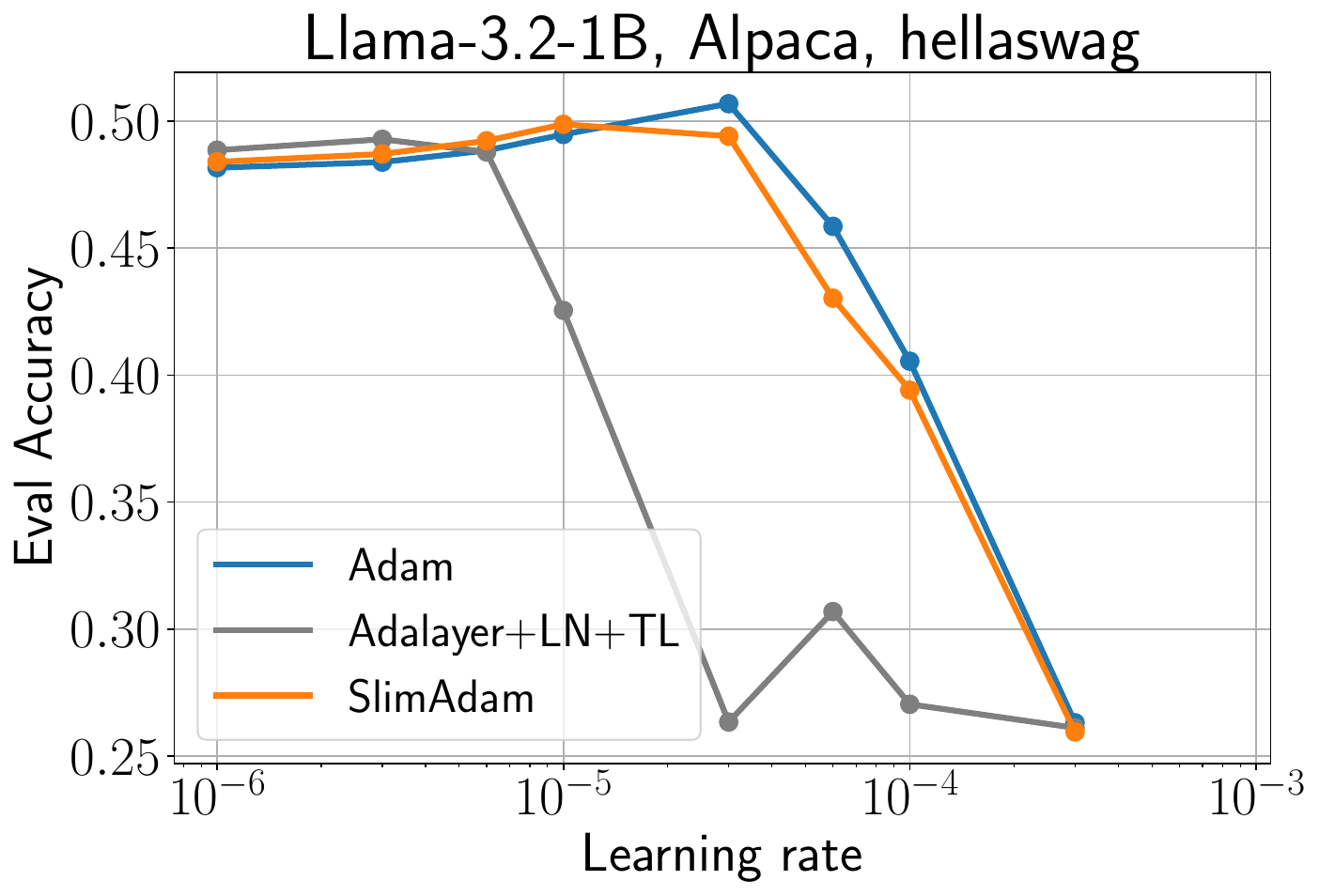}
\end{minipage}

\caption{Training loss and Downstream performance of Llama-3.2 1B finetuned on the Alpaca dataset.}
\label{fig:slim-loss-lr-llama-1b}
\end{figure*}

\begin{figure*}[!htb]
\centering
\begin{minipage}[b]{0.225\textwidth}
    \centering
    \includegraphics[width=\textwidth]{figures/loss-lr-plots/loss_lr_Llama-3.2-3B_Alpaca_bs4_E3_SlimAdamW_b0.9_b0.999_wd0.1_ga_1.pdf}
\end{minipage}
\begin{minipage}[b]{0.225\textwidth}
    \centering
    \includegraphics[width=\textwidth]{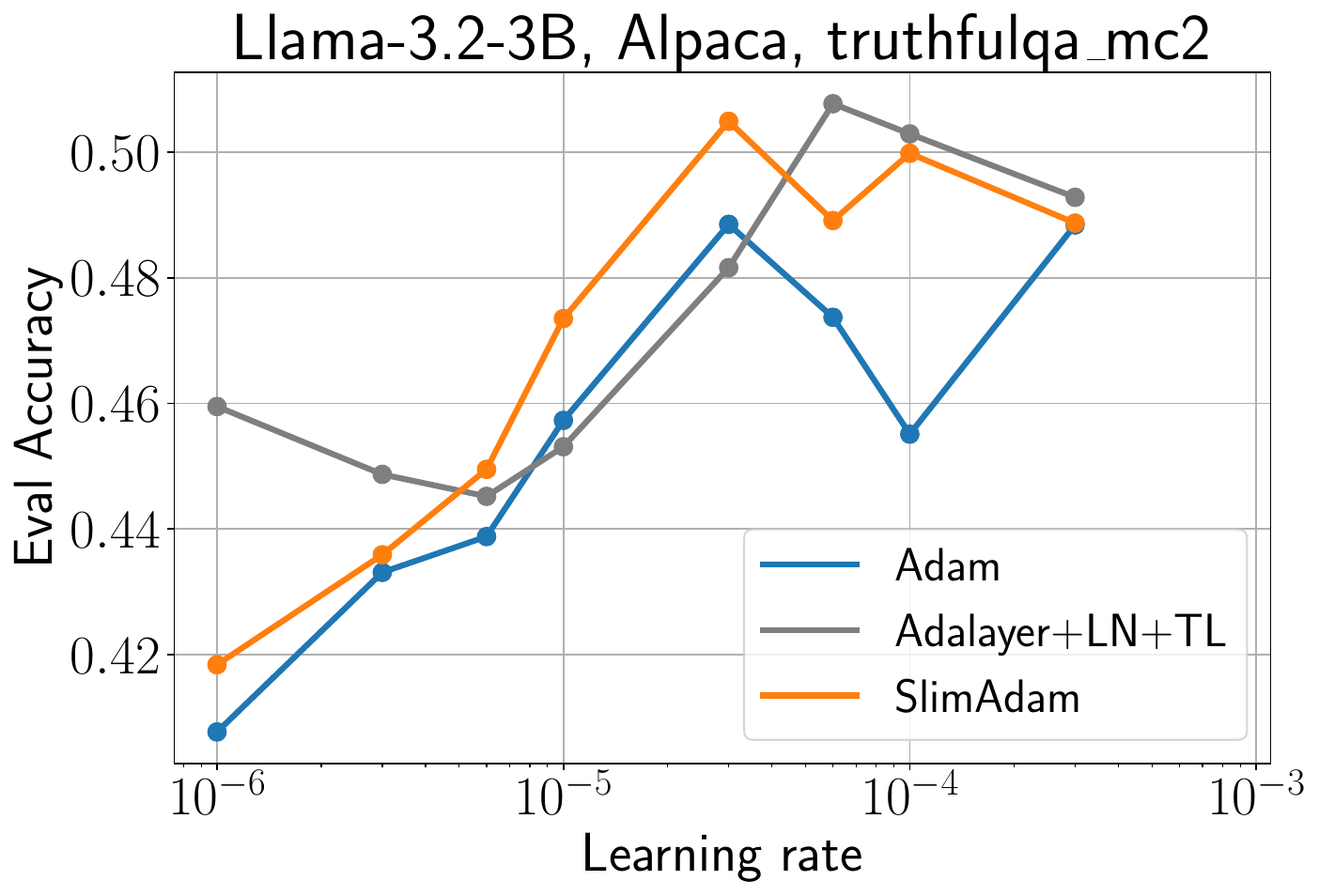}
\end{minipage}
\begin{minipage}[b]{0.225\textwidth}
    \centering
    \includegraphics[width=\textwidth]{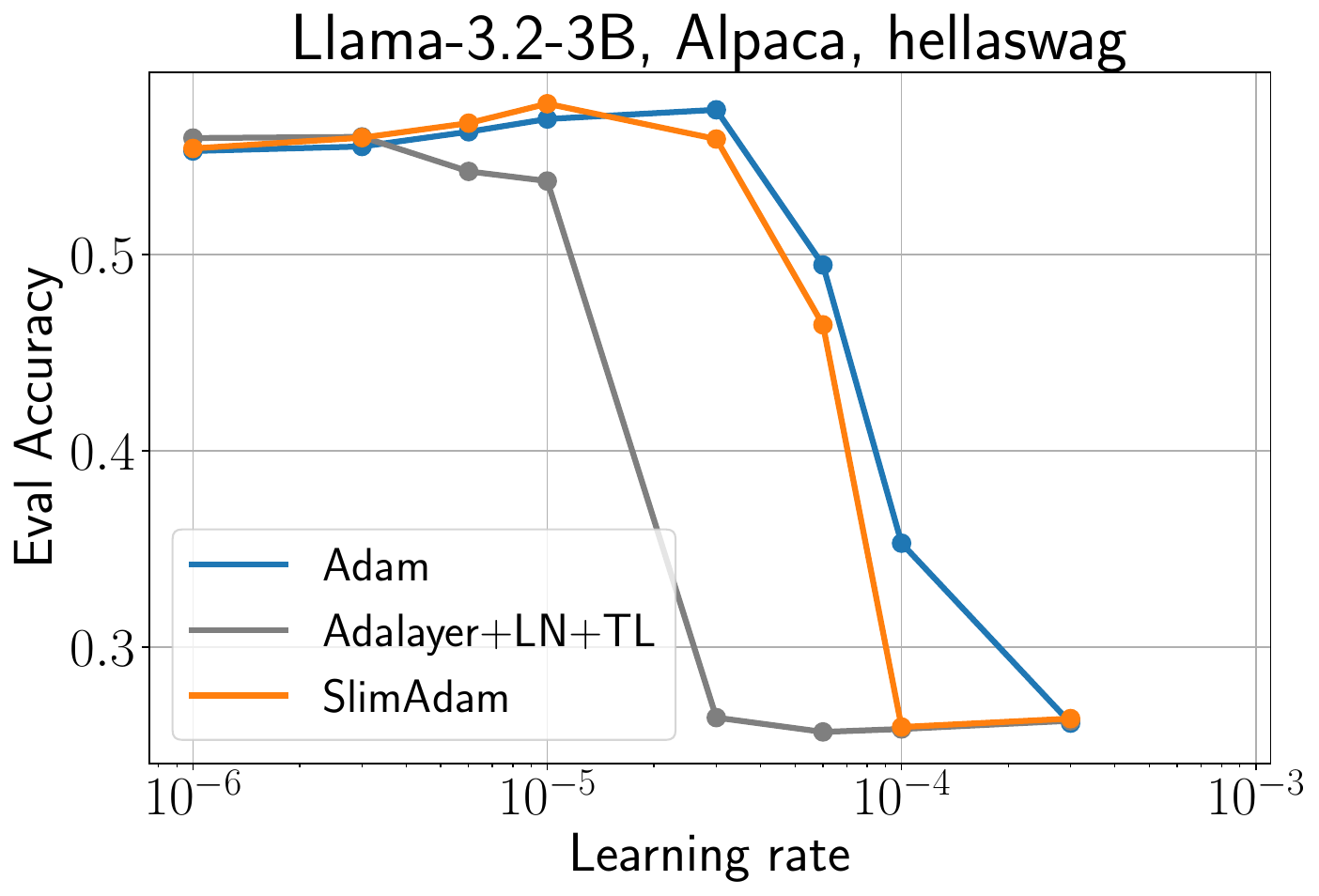}
\end{minipage}

\caption{Training loss and Downstream performance of Llama-3.2 3B finetuned on the Alpaca dataset.}
\label{fig:slim-loss-lr-llama-3b}
\end{figure*}

This section provides additional results for \Cref{section:slimadam}. \Cref{fig:slim-memory-performance-appendix} compares SNR predicted savings and performance of \emph{SlimAdam} with other baselines on additional tasks.
\Cref{fig:slim-loss-lr-llama-1b,fig:slim-loss-lr-llama-3b} shows the training loss and downstream performance (HellaSwag and TruthfulQA) of Llama-3.2 1B and Llama 3.2 3B fine-tuned on the Alpaca dataset.

\clearpage 
\newpage

\section{Tailed Token Distribution Reduce Compressibility}
\label{appendix:vocab-experiment}

\begin{figure*}[!hb]
\centering
\begin{minipage}[b]{0.245\textwidth}
    \centering
    \includegraphics[width=\textwidth]{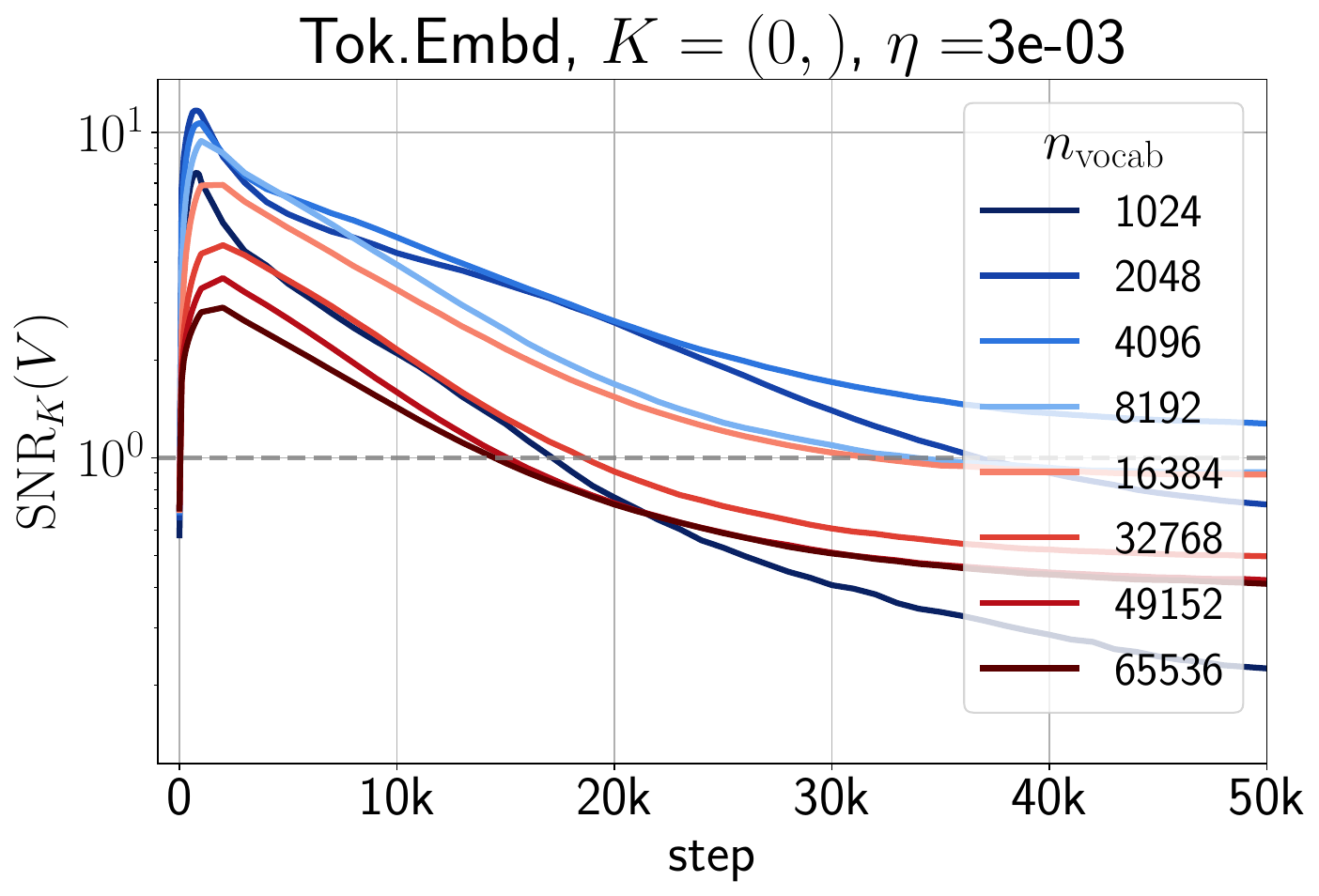}
\end{minipage}
\begin{minipage}[b]{0.245\textwidth}
    \centering
    \includegraphics[width=\textwidth]{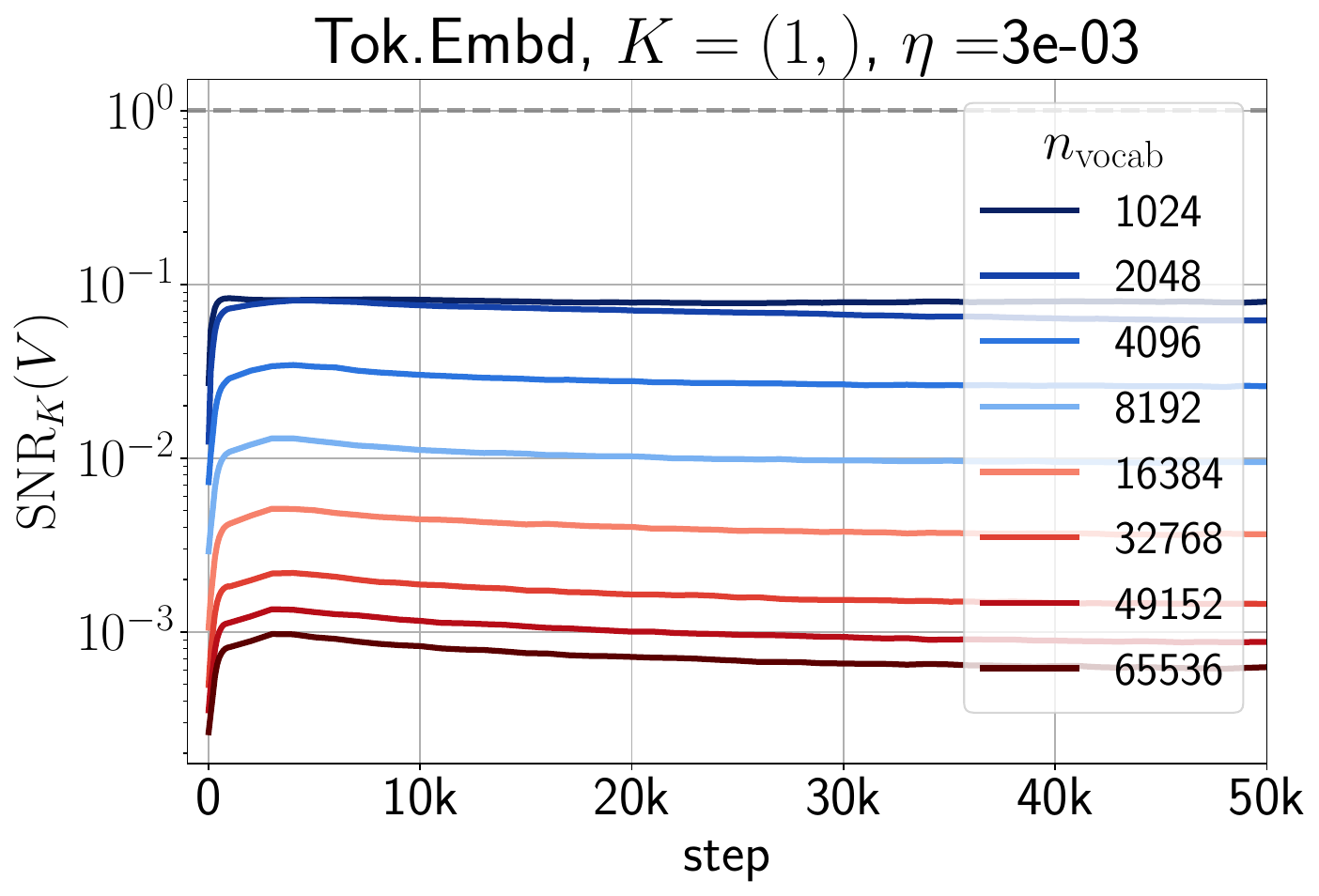}
\end{minipage}
\begin{minipage}[b]{0.245\textwidth}
    \centering
    \includegraphics[width=\textwidth]{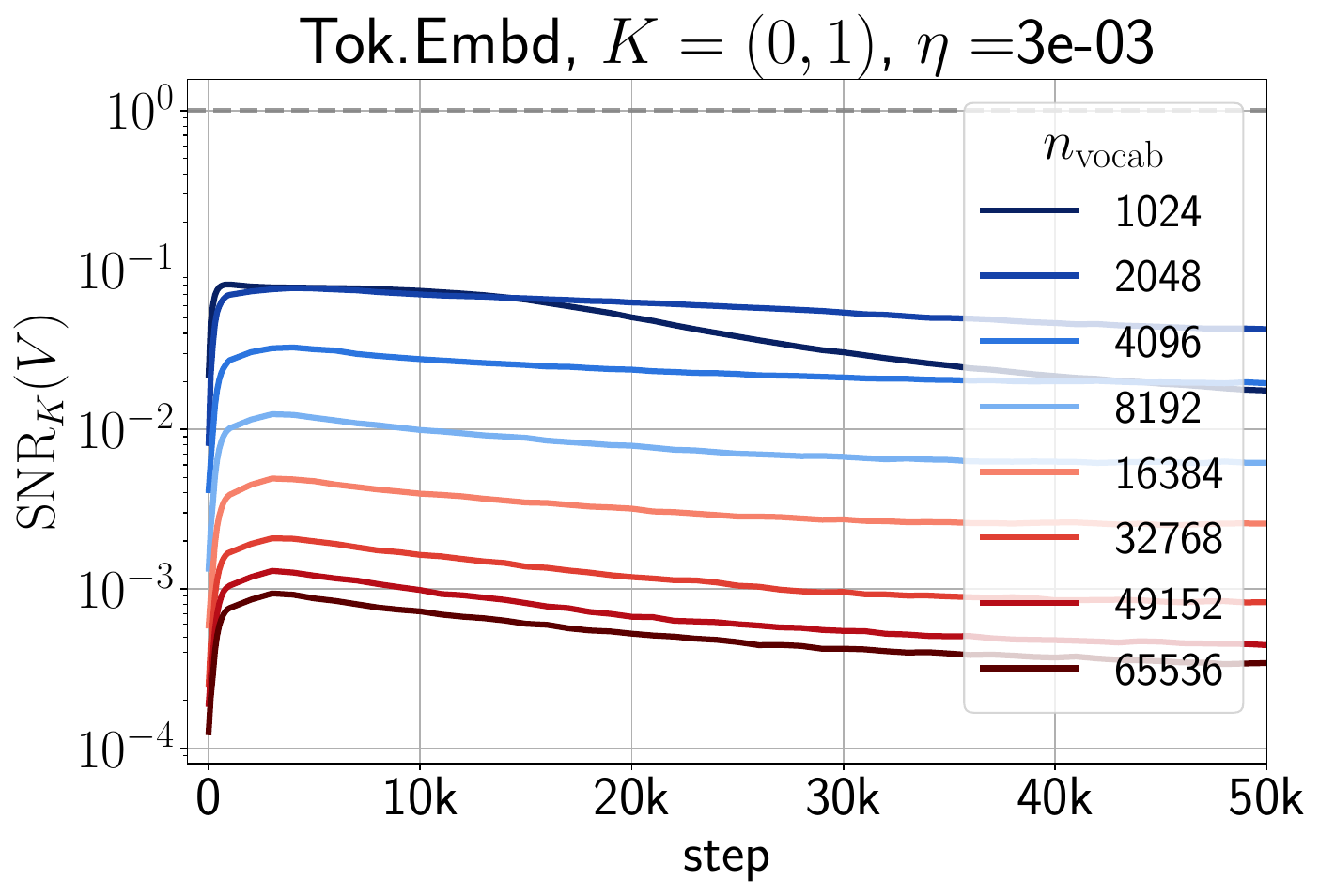}
\end{minipage}

\begin{minipage}[b]{0.245\textwidth}
    \centering
    \includegraphics[width=\textwidth]{figures/analysis/vocab-experiment/snr-trajectories/snr_LM.Head_0_wikitext-103_tokenlm-sp_Tn128_n768_B16_b0.9_b0.999_wd1e-04.pdf}
\end{minipage}
\begin{minipage}[b]{0.245\textwidth}
    \centering
    \includegraphics[width=\textwidth]{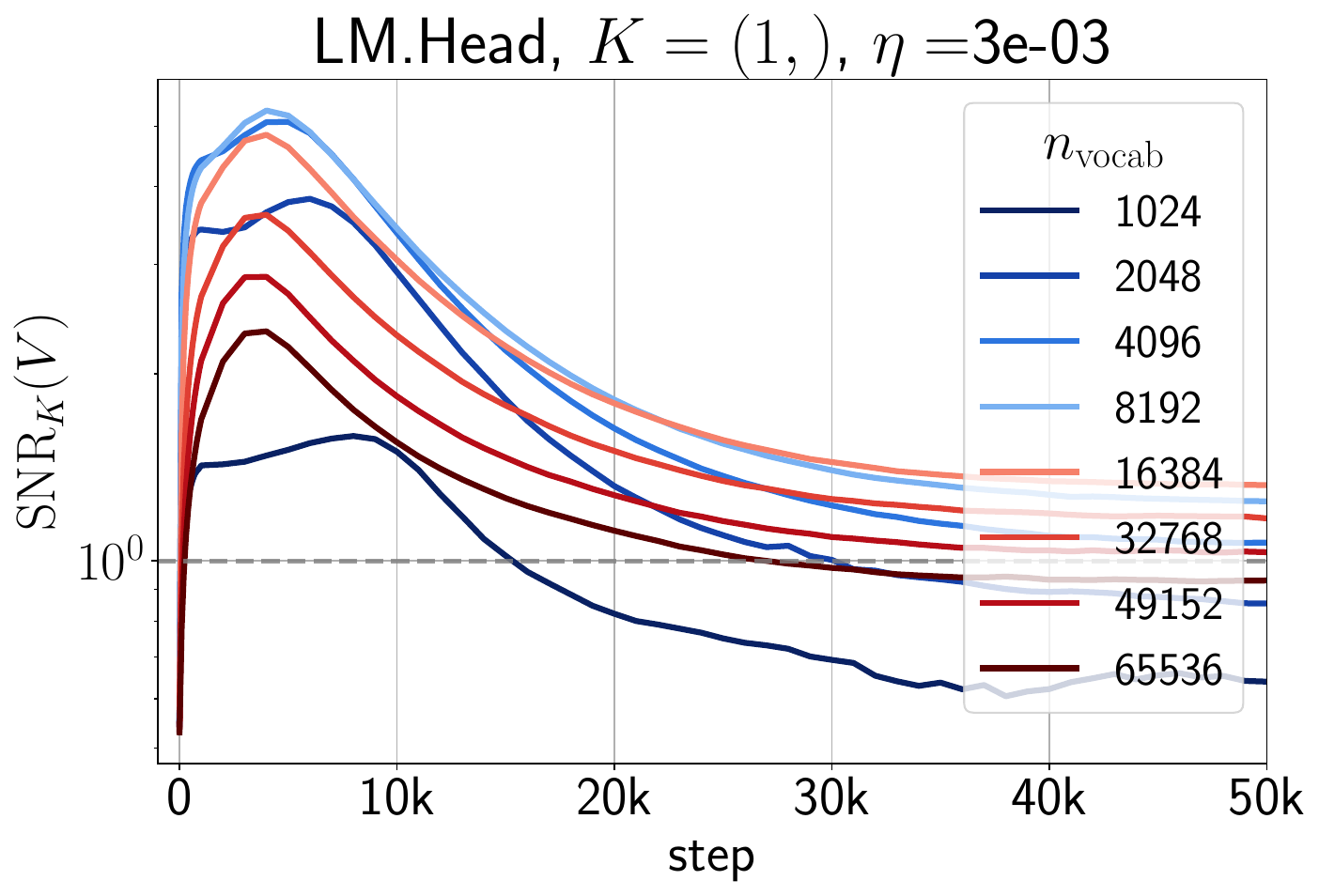}
\end{minipage}
\begin{minipage}[b]{0.245\textwidth}
    \centering
    \includegraphics[width=\textwidth]{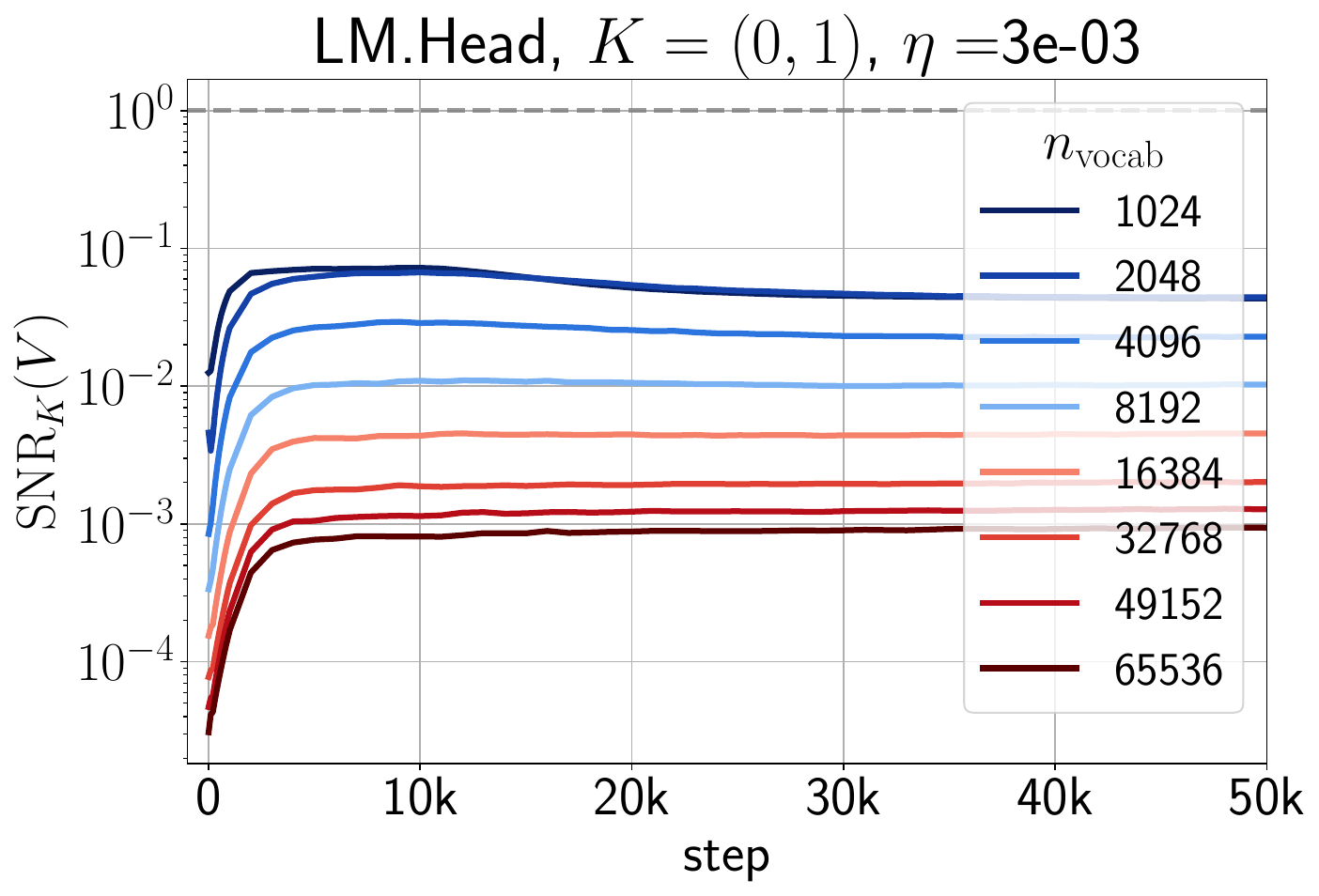}
\end{minipage}

\caption{SNR trajectories of the token embedding and linear head of the simplified two-layer model with varying vocabulary sizes.}
\label{fig:vocab-experiment-full}
\end{figure*}

\Cref{fig:vocab-experiment-full} shows additional SNR trajectories for the token distribution experiment discussed in \Cref{section:heavy-tailed-token-distribution}. For both layers, the SNR values along the token dimension ($K = 0$ for Tok.Embd and $K=1$ for LM.Head) decrease as the vocabulary size is increased. This suggests that at large vocabulary sizes, each token evolves at its own pace and this requires its own effective learning rate.

\section{Robustness of \emph{SlimAdam} Compression Rules}
\label{appendix:rule-transfer}

\begin{figure}[!htb]
    \centering
    \includegraphics[width=0.325\linewidth]{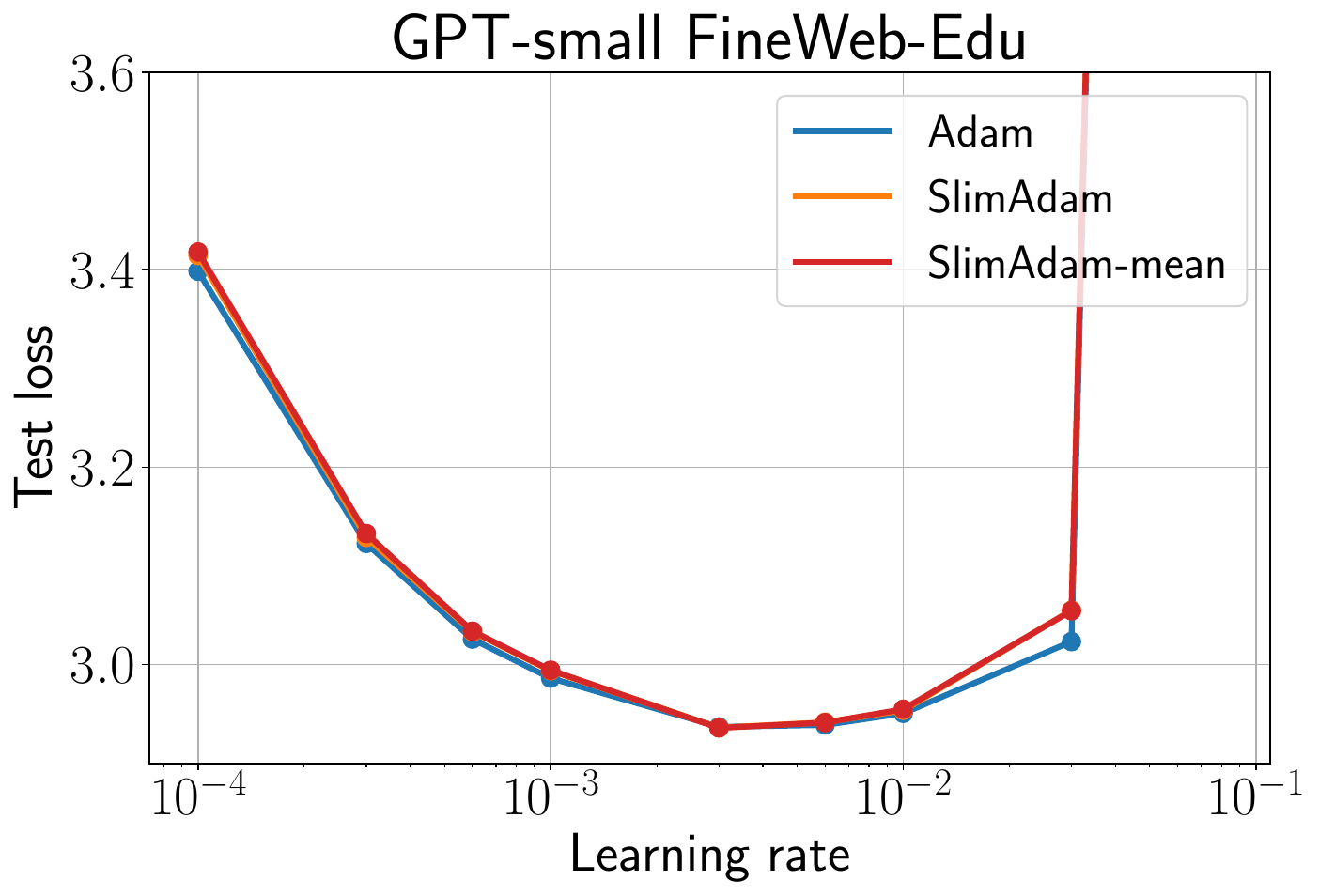}
    \caption{\emph{SlimAdam} with compression rules derived from depth-averaged SNR per layer type (\emph{SlimAdam-mean}) achieves identical performance to SlimAdam with per-layer compression rules.
    }
    \label{fig:snr-mean-rules}
\end{figure}

This section analyzes the robustness of \emph{SlimAdam} rules across datasets and model size.

\subsection{Dataset Dependency of SlimAdam Rules}
\label{appendix:dataset-dependency}
This section analyzes how \emph{SlimAdam}'s compression rules vary across different datasets. We compare rules derived from OpenWebText against FineWeb-Edu using GPT-small. The compression rules remain largely consistent, with differences in only five matrices, primarily in early MLP layers, as summarized in \Cref{table:dataset-compression-rules}.

\begin{table}[!htb]
\caption{Compression rule differences between datasets for GPT-small.}
\label{table:dataset-compression-rules}
\centering
\begin{tabular}{lcc}
\toprule
Layer & OpenWebText & FineWeb-Edu \\
\midrule
\multicolumn{3}{l}{\textit{Attention}} \\
Attn Query (L3) & None & fan-out \\
\midrule
\multicolumn{3}{l}{\textit{MLP}} \\
MLP Up (L0) & fan-out & None \\
MLP Up (L1) & None & fan-out \\
MLP Proj (L1) & fan-out & fan-in \\
MLP Proj (L2) & fan-in & fan-out \\
\bottomrule
\end{tabular}
\end{table}

\subsection{Width Dependency of SlimAdam Rules}
\label{appendix:width-dependency}

 This section analyzes the robustness of \emph{SlimAdam}'s compression rules across model widths ($d_{\text{model}}$). We compare the SNR-derived compression rules for GPT-small with embedding dimension $d_{\text{model}} = 768$ against a narrower model ($d_{\text{model}} = 256$. Out of all layer matrices, we observe differences in compression rules for only 12 matrices, primarily in early to middle layers, as shown in \Cref{table:width-compression-rules}.

\begin{table}[!htb]
\caption{\emph{SlimAdam} compression rule differences between narrow (width 256) and wide (width 768) models.}
\label{table:width-compression-rules}
\centering
\begin{tabular}{lcc}
\toprule
Layer & $d_{\text{model}} = 256$ & $d_{\text{model}} = 768$ \\
\midrule
\multicolumn{3}{l}{\textit{Attention Components}} \\
Attention Value (L0) & fan-in & fan-out \\
Attention Key (L2) & fan-out & fan-in \\
Attention Query (L2) & fan-in & fan-out \\
Attention Query (L3) & fan-in & None \\
\midrule
\multicolumn{3}{l}{\textit{MLP Components}} \\
MLP Up (L0) & fan-in & fan-out \\
MLP Up (L1) & fan-out & None \\
MLP Proj (L2) & fan-out & fan-in \\
MLP Up (L3) & fan-in & fan-out \\
MLP Up (L4) & fan-in & fan-out \\
MLP Proj (L4) & fan-in & fan-out \\
MLP Proj (L5) & fan-in & fan-out \\
MLP Up (L6) & fan-in & fan-out \\
\bottomrule
\end{tabular}
\end{table}

The variations observed in \Cref{table:dataset-compression-rules,table:width-compression-rules} can be eliminated by deriving compression rules using SNR values averaged over depth for each layer type.
 \Cref{fig:snr-mean-rules} shows that compression rules derived from depth-averaged SNR result in identical performance to SlimAdam with per-layer compression rules. \Cref{tab:best-compression-dimensions} shows the typical compression rules we observe across training regimes.

\begin{table}[!htb]
\centering
\caption{Recommended compression dimensions for different layer types. Layers with compression dimension marked with $^\star$ show inconsistent trends across models and tasks.}
\vspace{0.2 in}
\label{tab:best-compression-dimensions}
\begin{tabular}{ll}
\toprule
\textbf{Layer Type} & $\mathbf{K^*}$ \\
\midrule
\multicolumn{2}{l}{\textit{Attention}} \\
Key \& Query & $\text{fan}_{\text{in}}$ \\
Value \& Projection & $\text{fan}_{\text{out}}$ \\
\midrule
\multicolumn{2}{l}{\textit{MLP Layers}} \\
First layer (Up) & $\text{fan}_{\text{out}}^\star$ \\
Middle layer (Gate, Llama only) & $\text{fan}_{\text{out}}^\star$  \\
Last layer (Down) & $\text{fan}_{\text{out}}$ \\
\midrule
\multicolumn{2}{l}{\textit{Special Layers}} \\
Token Embedding & $\text{fan}_{\text{out}}$ \\
Language Modeling Head & $\text{fan}_{\text{in}}$ \\
Vision First Layer & $\text{fan}_{\text{in}}$ \\
Vision Classification Head & $\text{fan}_{\text{in}}^\star$ \\
\midrule
{\textit{Normalization Layers}} & - \\
\bottomrule
\end{tabular}
\vspace{-0.2 in}
\end{table}

\end{document}